\newif\ifarxiv
\arxivtrue
\documentclass[11pt]{article}
\ifarxiv
\usepackage[preprint]{acl}
\else
\usepackage[review]{acl}
\fi
\usepackage{times}
\usepackage{latexsym}
\usepackage[T1]{fontenc}
\usepackage[utf8]{inputenc}
\usepackage{microtype}
\usepackage{inconsolata}
\usepackage{graphicx}
\usepackage{booktabs}
\usepackage{multirow}
\usepackage{amsmath,amssymb}
\usepackage{xcolor}
\usepackage{colortbl}  %
\usepackage{enumitem}
\usepackage{subcaption}
\usepackage{tikz}
\usetikzlibrary{positioning,arrows.meta,fit,backgrounds,calc,shapes.geometric,shapes.misc}
\usetikzlibrary{shadows.blur}  %
\usepackage{fontawesome5}  %
\usepackage{placeins}      %
\usepackage{etoc}          %

\newcommand{\sysone}{System-1}
\newcommand{\systwo}{System-2}
\newcommand{\jev}{\textsc{Jev}}

\newcommand{\benchtext}{AgentDecisionBench}
\newcommand{\harnesstext}{JevAgentHarness}%

\definecolor{nmBench}{HTML}{1F5FAD}\definecolor{nmHarness}{HTML}{1B7A4C}\definecolor{nmModel}{HTML}{7448AE}
\definecolor{keyB}{HTML}{3A7BC8}\definecolor{keyH}{HTML}{2E8B62}\definecolor{keyM}{HTML}{8A62C2}
\newcommand{\keyB}[1]{{\color{keyB}\textbf{#1}}}\newcommand{\keyH}[1]{{\color{keyH}\textbf{#1}}}
\newcommand{\keyM}[1]{{\color{keyM}\textbf{#1}}}
\definecolor{keyR}{HTML}{C2557A}\definecolor{keyA}{HTML}{C27C2A}
\newcommand{\keyA}[1]{{\color{keyA}\textbf{#1}}}
\providecommand{\texorpdfstring}[2]{#1}
\newcommand{\nmStyle}[2]{\texorpdfstring{{\color{#1}\textbf{\textsc{#2}}}}{#2}}
\newcommand{\benchname}{\nmStyle{nmBench}{\benchtext}}
\newcommand{\harnessname}{\nmStyle{nmHarness}{\harnesstext}}

\newcommand{\gym}{\textsc{DecisionGym}}
\newcommand{\modeltext}{AgentDecision}%
\newcommand{\ours}{\nmStyle{nmModel}{\modeltext-4B}}
\newcommand{\oursV}{\nmStyle{nmModel}{\modeltext-4B v1}}
\newcommand{\oursS}{\nmStyle{nmModel}{\modeltext-0.8B}}%

\newcommand{\Choice}{\textsc{Choice}}
\newcommand{\Score}{\textsc{Score}}
\newcommand{\Noul}{\textsc{Noul}}

\newcommand{\modeRep}{\textsc{Replace}}
\newcommand{\modeCas}{\textsc{Cascade}}
\newcommand{\modeAss}{\textsc{Assist}}
\newcommand{\modeGua}{\textsc{Guard}}
\newcommand{\modeTri}{\textsc{Trigger}}
\definecolor{modeCol}{HTML}{B35900}\newcommand{\modeT}[1]{{\color{modeCol}\textbf{#1}}}

\newcounter{jevplaceholders}
\ifdefined\jevsubmission
  \DeclareRobustCommand{\todo}[1]{\stepcounter{jevplaceholders}%
    \PackageWarning{jevpaper}{\string\todo\space placeholder left in a submission build}%
    \textcolor{red}{[TODO]}}
  \DeclareRobustCommand{\tbd}{\stepcounter{jevplaceholders}%
    \PackageWarning{jevpaper}{\string\tbd\space or unfilled \string\res\space left in a submission build}%
    \textcolor{blue}{\textbf{??}}}
  \AtEndDocument{\ifnum\value{jevplaceholders}>0
    \PackageError{jevpaper}{\arabic{jevplaceholders}\space placeholder(s) left (\string\todo, \string\tbd,
      unfilled \string\res); see the jevpaper warnings in the log}{Fill or remove them before submitting.}\fi}
\else
  \newcommand{\todo}[1]{\textcolor{red}{[TODO: #1]}}
  \newcommand{\tbd}{\textcolor{blue}{\textbf{??}}}
\fi

\makeatletter
\newcommand{\res}[1]{\ifcsname jevres@#1\endcsname\csname jevres@#1\endcsname\else\tbd\fi}
\newcommand{\defres}[2]{\expandafter\def\csname jevres@#1\endcsname{#2}}
\makeatother
\newcommand{\resm}[1]{\ensuremath{\res{#1}}}
\newcommand{\resratio}[3]{\fpeval{round(100*\res{#1}/\res{#2},#3)}}
\newcommand{\resgap}[3]{\fpeval{round(\res{#1}-\res{#2},#3)}}
\makeatletter
\def\jev@absstrip#1#2\relax{\ifx-#1#2\else\ifx$#1\jev@absmath#2\relax\else#1#2\fi\fi}
\def\jev@absmath#1$#2\relax{#2}%
\newcommand{\resabs}[1]{\ifcsname jevres@#1\endcsname
  \expandafter\expandafter\expandafter\jev@absstrip\csname jevres@#1\endcsname\relax\else\tbd\fi}
\makeatother
\makeatletter

\newcommand{\dg}{\ensuremath{^{\dagger}}}
\newcommand{\resna}[1]{\ifcsname jevres@#1\endcsname\res{#1}\else--\fi}
\makeatletter
\newcommand{\rbSetBest}[2]{\def\rb@m{-1e9}%
  \@for\rb@k:=#2\do{\ifcsname jevres@\rb@k\endcsname\edef\rb@m{\fpeval{max(\rb@m,\csname jevres@\rb@k\endcsname)}}\fi}%
  \edef\rbFlag{\fpeval{\csname jevres@#1\endcsname>=\rb@m}}}
\newcommand{\resbest}[2]{\ifcsname jevres@#1\endcsname\rbSetBest{#1}{#2}\ifnum\rbFlag=1 \textbf{\res{#1}}\else\res{#1}\fi\else--\fi}
\newcommand{\boldifbest}[3]{\ifcsname jevres@#1\endcsname\rbSetBest{#1}{#2}\ifnum\rbFlag=1 \textbf{#3}\else#3\fi\else#3\fi}
\def\rb@saved#1#2{\ifcsname jevres@#1\endcsname\ifcsname jevres@#2\endcsname
  \fpeval{100*(\csname jevres@#2\endcsname-\csname jevres@#1\endcsname)/\csname jevres@#2\endcsname}\else-1e9\fi\else-1e9\fi}
\def\rb@pair#1=#2\@nil{\rb@saved{#1}{#2}}
\newcommand{\boldifbesttime}[4]{\def\rb@m{-1e9}%
  \@for\rb@k:=#3\do{\edef\rb@v{\expandafter\rb@pair\rb@k\@nil}\edef\rb@m{\fpeval{max(\rb@m,\rb@v)}}}%
  \edef\rb@v{\rb@saved{#1}{#2}}\ifnum\fpeval{\rb@v>=\rb@m-1e-9}=1 \textbf{#4}\else#4\fi}
\makeatother
\newcommand{\resmna}[1]{\ifcsname jevres@#1\endcsname\ensuremath{\res{#1}}\else--\fi}
\newcommand{\respct}[1]{\ifcsname jevres@#1\endcsname\pgfmathparse{100*\csname jevres@#1\endcsname}%
  \pgfmathprintnumber[fixed,fixed zerofill,precision=1]{\pgfmathresult}\else\tbd\fi}
\makeatother
\makeatletter
\def\jev@decmap#1#2{\ifnum\pdfstrcmp{\csname jevres@\jev@deckey\endcsname}{#1}=0 #2\expandafter\jev@decdone\fi}
\def\jev@decdone#1\jev@decend{}
\let\jev@decend\relax
\newcommand{\resdec}[1]{\ifcsname jevres@#1\endcsname\def\jev@deckey{#1}%
  \jev@decmap{INCOMPLETE (p = 1 entered Holm)}{INCOMPLETE$^\S$}%
  \jev@decmap{not rejected: the registered joint claim was not established}{not established}%
  \jev@decmap{not tested (no frozen JevAgentRL checkpoint: no freeze record, A7.10 (h))}{pending}%
  \jev@decmap{not tested (no decision point qualified in Stage 1; the best-practice profile is LLM-only) -- A7.1 (g) 5}{not tested (empty profile)}%
  \jev@decmap{not run (A7.1 (h): E6 not run in Stage 1)}{not run}%
  \jev@decmap{NI: policy NI with zero coverage for Intern-Decision (A6.11 (c).4, A6.12, A6.16 (m), A6.20 (e), A6.22 (a))}{NI$^\|$}%
  \jev@decmap{not NI as a conjunction: Jev NI; Intern-Decision inconclusive at this n}{not NI (split)}%
  \jev@decmap{not NI as a conjunction: Jev not NI: loss of 7.8 pt (95 \% CI entirely below 0); Intern-Decision not NI: loss of 3.6 pt (95 \% CI entirely below 0)}{not NI (both lose)}%
  \jev@decmap{not NI as a conjunction: Jev inconclusive at this n; Intern-Decision not NI: loss of 14.8 pt (95 \% CI entirely below 0)}{not NI (Intern loses)}%
  \jev@decmap{not NI as a conjunction: Jev inconclusive at this n; Intern-Decision not NI: loss of 6.0 pt (95 \% CI entirely below 0)}{not NI (Intern loses)}%
  \jev@decmap{superiority not established (Holm step-down stopped earlier)}{sup.\ not established}%
  \jev@decmap{transfer NI was not established; inconclusive at this n}{transfer NI not established}%
  \jev@decmap{rejects: the claim holds}{holds}%
  \res{#1}\jev@decend
  \else\tbd\fi}
\newcommand{\resplain}[1]{\ifcsname jevres@#1\endcsname\def\jev@deckey{#1}%
  \jev@decmap{NI}{non-inferior}%
  \jev@decmap{superior}{better}%
  \jev@decmap{rejects: the claim holds}{holds}%
  \jev@decmap{not rejected: the registered joint claim was not established}{not established}%
  \jev@decmap{superiority not established (Holm step-down stopped earlier)}{not established}%
  \jev@decmap{transfer NI was not established; inconclusive at this n}{not established (inconclusive)}%
  \jev@decmap{NI: policy NI with zero coverage for Intern-Decision (A6.11 (c).4, A6.12, A6.16 (m), A6.20 (e), A6.22 (a))}{non-inferior (only \jev{} delegated)}%
  \jev@decmap{not NI as a conjunction: Jev NI; Intern-Decision inconclusive at this n}{not established (Intern-Decision inconclusive)}%
  \jev@decmap{not NI as a conjunction: Jev not NI: loss of 7.8 pt (95 \% CI entirely below 0); Intern-Decision not NI: loss of 3.6 pt (95 \% CI entirely below 0)}{not established (both models lose)}%
  \jev@decmap{not NI as a conjunction: Jev inconclusive at this n; Intern-Decision not NI: loss of 14.8 pt (95 \% CI entirely below 0)}{not established (Intern-Decision loses)}%
  \jev@decmap{not NI as a conjunction: Jev inconclusive at this n; Intern-Decision not NI: loss of 6.0 pt (95 \% CI entirely below 0)}{not established (Intern-Decision loses)}%
  \jev@decmap{not tested (no decision point qualified in Stage 1; the best-practice profile is LLM-only) -- A7.1 (g) 5}{not tested (none qualified)}%
  \jev@decmap{not run (A7.1 (h): E6 not run in Stage 1)}{not run}%
  \jev@decmap{not tested (no frozen JevAgentRL checkpoint: no freeze record, A7.10 (h))}{not tested (no frozen model)}%
  \resdec{#1}\jev@decend
  \else\tbd\fi}
\makeatother

\newcommand{\genfig}[4][42mm]{\IfFileExists{figures/generated/#2}{\includegraphics[width=#3]{figures/generated/#2}}%
  {\fbox{\parbox[c][#1][c]{\dimexpr#3-2\fboxsep-2\fboxrule\relax}{\centering\footnotesize\textit{[#4]}}}}}

\newcommand{\eone}[2]{\res{e1_bfcl_when2call/#1/#2}}
\newcommand{\eonerowk}[2]{#1 & \res{#2/tool_select/bfcl/acc} & \res{#2/tool_select/bfcl/ece}
  & \res{#2/tool_select/bfcl/auroc} & \res{#2/tool_params/bfcl/acc}
  & \res{#2/tool_params/bfcl/acc_gold_in_candidates} & \res{#2/tool_params/bfcl/ece}
  & \res{#2/tool_select/when2call/acc} & \res{#2/tool_select/when2call/ece}
  & \res{#2/tool_select/bfcl/latency_p50_ms} & \res{#2/tool_select/bfcl/cost_per_1k_usd}\\}
\newcommand{\eonerow}[2]{\eonerowk{#1}{e1_bfcl_when2call/#2}}

\makeatletter
\newcommand{\resmark}[1]{\ifcsname jevres@#1\endcsname
    \ifnum\pdfmatch{^not (tested|run)}{\csname jevres@#1\endcsname}=1 --%
    \else\ifnum\pdfmatch{^(NI|superior|rejects)([^a-z]|$)}{\csname jevres@#1\endcsname}=1 \checkmark%
    \else\ifnum\pdfmatch{(^|[^a-z])not([^a-z]|$)}{\csname jevres@#1\endcsname}=1 $\times$%
    \else\textcolor{red}{\textbf{?}}\fi\fi\fi
  \else\tbd\fi}
\makeatletter
\newcommand{\resword}[1]{\ifcsname jevres@#1\endcsname
    \ifnum\pdfmatch{^not tested}{\csname jevres@#1\endcsname}=1 not tested%
    \else\ifnum\pdfmatch{^not run}{\csname jevres@#1\endcsname}=1 not run%
    \else\ifnum\pdfmatch{^(NI|superior|rejects)([^a-z]|$)}{\csname jevres@#1\endcsname}=1 held%
    \else\ifnum\pdfmatch{(^|[^a-z])not([^a-z]|$)}{\csname jevres@#1\endcsname}=1 did not hold%
    \else\textcolor{red}{\textbf{?}}\fi\fi\fi\fi
  \else\tbd\fi}
\makeatother
\makeatletter
\newcommand{\resnum}[1]{\ifcsname jevres@#1\endcsname\pgfmathprintnumber[fixed,precision=0,1000 sep={\mathord{,}}]{%
  \csname jevres@#1\endcsname}\else\tbd\fi}
\makeatother
\newcommand{\resk}[1]{\ifcsname jevres@#1\endcsname\pgfmathparse{1000*\csname jevres@#1\endcsname}%
  \pgfmathprintnumber[fixed,fixed zerofill,precision=1]{\pgfmathresult}\else\tbd\fi}
\makeatother

\newif\ifshowcomments
\ifarxiv\showcommentsfalse\else\showcommentstrue\fi
\ifshowcomments
\NewDocumentCommand{\xs}{ mO{} }{\textcolor{cyan}{\textsuperscript{\textit{Xiusi}}\textsf{\textbf{\small[#1]}}}}
\NewDocumentCommand{\kl}{ mO{} }{\textcolor{red}{\textsuperscript{\textit{Kunlun}}\textsf{\textbf{\small[#1]}}}}
\NewDocumentCommand{\heng}{ mO{} }{\textcolor{blue}{\textsuperscript{\textit{Heng}}\textsf{\textbf{\small[#1]}}}}
\NewDocumentCommand{\tao}{ mO{} }{\textcolor{orange}{\textsuperscript{\textit{Tao}}\textsf{\textbf{\small[#1]}}}}
\NewDocumentCommand{\xu}{ mO{} }{\textcolor{magenta}{\textsuperscript{\textit{Xu}}\textsf{\textbf{\small[#1]}}}}
\NewDocumentCommand{\hw}{ mO{} }{\textcolor{teal}{\textsuperscript{\textit{Hengwang}}\textsf{\textbf{\small[#1]}}}}
\else
\NewDocumentCommand{\xs}{ mO{} }{}
\NewDocumentCommand{\kl}{ mO{} }{}
\NewDocumentCommand{\heng}{ mO{} }{}
\NewDocumentCommand{\tao}{ mO{} }{}
\NewDocumentCommand{\xu}{ mO{} }{}
\NewDocumentCommand{\hw}{ mO{} }{}
\fi
\expandafter\def\expandafter\UrlBreaks\expandafter{\UrlBreaks\do\-}%

\definecolor{sOne}{HTML}{2A78D6}  \definecolor{sOneT}{HTML}{E3EEFA}   %
\definecolor{sTwo}{HTML}{EB6834}  \definecolor{sTwoT}{HTML}{FCE8DF}   %
\definecolor{envC}{HTML}{6B6A65}  \definecolor{envT}{HTML}{F4F4F1}    %
\definecolor{ink}{HTML}{0B0B0B}   \definecolor{ink2}{HTML}{52514E}    %

\newcommand{\figfs}{\fontsize{7}{8.2}\selectfont}            %
\newcommand{\figts}{\fontsize{8}{9.6}\selectfont\bfseries}   %
\newcommand{\figrm}[1]{{\rmfamily #1}}                       %

\tikzset{
  jevfig/.style={x=1mm, y=1mm, every node/.style={inner sep=0pt, outer sep=0pt, text=ink}},
  flbl/.style={font=\figfs\sffamily, align=center},
  fnote/.style={font=\figfs\sffamily, text=ink2},
  fttl/.style={font=\figts\sffamily, anchor=north west},
  fbox/.style={draw=#1, line width=0.5pt, rounded corners=1.5pt},
  farr/.style={-{Stealth[length=1.3mm,width=1mm]}, line width=0.5pt, draw=ink2},
  fdefer/.style={-{Stealth[length=1.3mm,width=1mm]}, line width=0.6pt, draw=sTwo, dash pattern=on 1.4pt off 1pt},
  frail/.style={line width=1.4pt, draw=envC!45, line cap=round},
  ftap/.style={rectangle, rounded corners=0.75mm, minimum width=1.6mm, minimum height=3.2mm, line width=0.6pt,
    draw=envC},
  fobjtap/.style={ftap, fill=envC},
  fmetatap/.style={ftap, fill=white},
  fmk/.style={fill=#1, draw=none},
  fmRep/.style={fmk=#1, circle, minimum size=1.8mm},
  fmCas/.style={fmk=#1, regular polygon, regular polygon sides=3, minimum size=2.45mm},
  fmAss/.style={fmk=#1, rectangle, minimum size=1.6mm},
  fmTri/.style={fmk=#1, regular polygon, regular polygon sides=3, minimum size=2.45mm, shape border rotate=180},
  pics/fmGua/.style={code={\fill[#1] (-0.95,-0.32) rectangle (0.95,0.32); \fill[#1] (-0.32,-0.95) rectangle (0.32,0.95);}},
  fchip/.style={fill=envT, rounded corners=1.5pt, minimum height=5.2mm, minimum width=23mm, inner xsep=0.8mm,
    font=\figfs,
    execute at begin node={\medmuskip=2mu plus 1mu\thickmuskip=3mu plus 1mu}},
  fout/.style={fbox=#1, fill=#1T, minimum height=3.4mm, minimum width=4.2mm, font=\figfs},
  fs1g/.style={text=sOne, font=\fontsize{9}{9}\selectfont},
  fs2g/.style={text=sTwo, font=\fontsize{9}{9}\selectfont},
}
\newcommand{\figmk}[4]{%
  \pgfmathsetmacro\figmkx{#1}\pgfmathsetmacro\figmky{#2}%
  \ifnum\pdfstrcmp{#3}{Gua}=0 \pic at (\figmkx,\figmky) {fmGua=#4};%
  \else\ifnum\pdfstrcmp{#3}{Cas}=0 \node[fmCas=#4] at ({\figmkx},{\figmky-0.3}) {};%
  \else\ifnum\pdfstrcmp{#3}{Tri}=0 \node[fmTri=#4] at ({\figmkx},{\figmky+0.3}) {};%
  \else \node[fm#3=#4] at (\figmkx,\figmky) {};\fi\fi\fi}
\newsavebox\figmkboxRep \newsavebox\figmkboxCas \newsavebox\figmkboxAss \newsavebox\figmkboxGua
\newsavebox\figmkboxTri \newsavebox\figtapboxobj \newsavebox\figtapboxmeta
\AtBeginDocument{%
  \sbox\figmkboxRep{\tikz[jevfig, baseline=-0.75mm]{\figmk{0}{0}{Rep}{ink2}}}%
  \sbox\figmkboxCas{\tikz[jevfig, baseline=-0.75mm]{\figmk{0}{0}{Cas}{ink2}}}%
  \sbox\figmkboxAss{\tikz[jevfig, baseline=-0.75mm]{\figmk{0}{0}{Ass}{ink2}}}%
  \sbox\figmkboxGua{\tikz[jevfig, baseline=-0.75mm]{\figmk{0}{0}{Gua}{ink2}}}%
  \sbox\figmkboxTri{\tikz[jevfig, baseline=-0.75mm]{\figmk{0}{0}{Tri}{ink2}}}%
  \sbox\figtapboxobj{\tikz[jevfig, baseline=-0.75mm]{\node[fobjtap, minimum height=2.6mm]{};}}%
  \sbox\figtapboxmeta{\tikz[jevfig, baseline=-0.75mm]{\node[fmetatap, minimum height=2.6mm]{};}}}
\newcommand{\figmki}[1]{\usebox{\csname figmkbox#1\endcsname}}
\newcommand{\figtapi}[1]{\usebox{\csname figtapbox#1\endcsname}}
\newcommand{\figtapmarks}[3]{%
  \foreach \m [count=\i] in #3 {\xdef\figtapn{\i}}%
  \foreach \m [count=\i] in #3 {\figmk{#1+(\i-1)*2.35-(\figtapn-1)*1.175}{#2}{\m}{ink2}}}
\InputIfFileExists{generated/results}{}{}     %
\InputIfFileExists{generated/primaries}{}{}   %
\InputIfFileExists{generated/stage2}{}{}      %
\InputIfFileExists{generated/data_stats}{}{}  %
\providecommand{\res}[1]{\ifcsname jevres@#1\endcsname\csname jevres@#1\endcsname\else\tbd\fi}
\providecommand{\defres}[2]{\expandafter\def\csname jevres@#1\endcsname{#2}}
\defres{swe_verified_100/action_guard/cost_usd}{0.554}
\defres{swe_verified_100/action_guard/faster}{no}
\defres{swe_verified_100/action_guard/n}{100}
\defres{swe_verified_100/action_guard/resolved}{76}
\defres{swe_verified_100/action_guard/screen}{pass}
\defres{swe_verified_100/action_guard/sens/success}{77.0}
\defres{swe_verified_100/action_guard/sens/verdict}{passes the exploratory success screen; not faster among jointly solved tasks}
\defres{swe_verified_100/action_guard/sens/vs_baseline/delta}{$+$2.0}
\defres{swe_verified_100/action_guard/steps}{58.5}
\defres{swe_verified_100/action_guard/success}{76.0}
\defres{swe_verified_100/action_guard/time_both_solved/d_median_s}{$+$9.1}
\defres{swe_verified_100/action_guard/time_both_solved/n}{69}
\defres{swe_verified_100/action_guard/time_both_solved/sign_p}{0.95}
\defres{swe_verified_100/action_guard/verdict}{passes the exploratory success screen; not faster among jointly solved tasks}
\defres{swe_verified_100/action_guard/vs_baseline/cost_savings}{$+$1.6}
\defres{swe_verified_100/action_guard/vs_baseline/delta}{$+$1.0}
\defres{swe_verified_100/action_guard/vs_baseline/delta_lb90}{$-$4.0}
\defres{swe_verified_100/action_guard/vs_baseline/losses}{6}
\defres{swe_verified_100/action_guard/vs_baseline/wins}{7}
\defres{swe_verified_100/action_guard/wall_p50_s}{132}
\defres{swe_verified_100/action_guard/wall_p90_s}{425}
\defres{swe_verified_100/baseline/cost_usd}{0.563}
\defres{swe_verified_100/baseline/n}{100}
\defres{swe_verified_100/baseline/resolved}{75}
\defres{swe_verified_100/baseline/steps}{60.6}
\defres{swe_verified_100/baseline/success}{75.0}
\defres{swe_verified_100/baseline/wall_p50_s}{111}
\defres{swe_verified_100/baseline/wall_p90_s}{435}
\defres{swe_verified_100/n_tasks}{100}
\defres{swe_verified_100/route/cost_usd}{0.464}
\defres{swe_verified_100/route/faster}{no}
\defres{swe_verified_100/route/n}{100}
\defres{swe_verified_100/route/resolved}{77}
\defres{swe_verified_100/route/screen}{pass}
\defres{swe_verified_100/route/sens/success}{78.0}
\defres{swe_verified_100/route/sens/verdict}{passes the exploratory success screen; not faster among jointly solved tasks}
\defres{swe_verified_100/route/sens/vs_baseline/delta}{$+$3.0}
\defres{swe_verified_100/route/steps}{54.1}
\defres{swe_verified_100/route/success}{77.0}
\defres{swe_verified_100/route/time_both_solved/d_median_s}{$-$0.1}
\defres{swe_verified_100/route/time_both_solved/n}{69}
\defres{swe_verified_100/route/time_both_solved/sign_p}{0.50}
\defres{swe_verified_100/route/verdict}{passes the exploratory success screen; not faster among jointly solved tasks}
\defres{swe_verified_100/route/vs_baseline/cost_savings}{$+$17.6}
\defres{swe_verified_100/route/vs_baseline/delta}{$+$2.0}
\defres{swe_verified_100/route/vs_baseline/delta_lb90}{$-$3.0}
\defres{swe_verified_100/route/vs_baseline/losses}{6}
\defres{swe_verified_100/route/vs_baseline/wins}{8}
\defres{swe_verified_100/route/wall_p50_s}{116}
\defres{swe_verified_100/route/wall_p90_s}{326}
\defres{swe_verified_100/stop_guard/cost_usd}{0.619}
\defres{swe_verified_100/stop_guard/faster}{no}
\defres{swe_verified_100/stop_guard/n}{100}
\defres{swe_verified_100/stop_guard/resolved}{74}
\defres{swe_verified_100/stop_guard/screen}{fail}
\defres{swe_verified_100/stop_guard/sens/success}{74.0}
\defres{swe_verified_100/stop_guard/sens/verdict}{does not pass the exploratory success screen; not faster among jointly solved tasks}
\defres{swe_verified_100/stop_guard/sens/vs_baseline/delta}{$-$1.0}
\defres{swe_verified_100/stop_guard/steps}{63.5}
\defres{swe_verified_100/stop_guard/success}{74.0}
\defres{swe_verified_100/stop_guard/time_both_solved/d_median_s}{$+$7.8}
\defres{swe_verified_100/stop_guard/time_both_solved/n}{68}
\defres{swe_verified_100/stop_guard/time_both_solved/sign_p}{0.94}
\defres{swe_verified_100/stop_guard/verdict}{does not pass the exploratory success screen; not faster among jointly solved tasks}
\defres{swe_verified_100/stop_guard/vs_baseline/cost_savings}{$-$10.0}
\defres{swe_verified_100/stop_guard/vs_baseline/delta}{$-$1.0}
\defres{swe_verified_100/stop_guard/vs_baseline/delta_lb90}{$-$6.0}
\defres{swe_verified_100/stop_guard/vs_baseline/losses}{7}
\defres{swe_verified_100/stop_guard/vs_baseline/wins}{6}
\defres{swe_verified_100/stop_guard/wall_p50_s}{125}
\defres{swe_verified_100/stop_guard/wall_p90_s}{556}
\defres{tbsci/baseline/cost_usd}{1.13}
\defres{tbsci/baseline/n}{45}
\defres{tbsci/baseline/s1_requests}{0.0}
\defres{tbsci/baseline/solved}{0}
\defres{tbsci/baseline/steps}{72.3}
\defres{tbsci/baseline/success}{0.0}
\defres{tbsci/baseline/tests_passed_share}{36.4}
\defres{tbsci/baseline/timeouts}{0}
\defres{tbsci/baseline/wall_p50_s}{350}
\defres{tbsci/baseline/wall_p90_s}{1287}
\defres{tbsci/jev_gate_stop/cost_usd}{1.31}
\defres{tbsci/jev_gate_stop/faster}{n.e.}
\defres{tbsci/jev_gate_stop/n}{45}
\defres{tbsci/jev_gate_stop/s1_requests}{65.1}
\defres{tbsci/jev_gate_stop/screen}{pass}
\defres{tbsci/jev_gate_stop/solved}{1}
\defres{tbsci/jev_gate_stop/steps}{82.0}
\defres{tbsci/jev_gate_stop/success}{2.2}
\defres{tbsci/jev_gate_stop/tests_passed_share}{36.9}
\defres{tbsci/jev_gate_stop/timeouts}{1}
\defres{tbsci/jev_gate_stop/verdict}{passes the exploratory success screen; faster not estimable (no jointly solved task)}
\defres{tbsci/jev_gate_stop/vs_baseline/cost_savings}{$-$16.2}
\defres{tbsci/jev_gate_stop/vs_baseline/d_run_median_all_s}{$+$13.3}
\defres{tbsci/jev_gate_stop/vs_baseline/delta}{$+$2.2}
\defres{tbsci/jev_gate_stop/vs_baseline/delta_lb90}{$+$0.0}
\defres{tbsci/jev_gate_stop/vs_baseline/losses}{0}
\defres{tbsci/jev_gate_stop/vs_baseline/wins}{1}
\defres{tbsci/jev_gate_stop/wall_p50_s}{352}
\defres{tbsci/jev_gate_stop/wall_p90_s}{995}
\defres{tbsci/jev_stop_reflect/cost_usd}{0.95}
\defres{tbsci/jev_stop_reflect/faster}{n.e.}
\defres{tbsci/jev_stop_reflect/n}{45}
\defres{tbsci/jev_stop_reflect/s1_requests}{5.1}
\defres{tbsci/jev_stop_reflect/screen}{pass}
\defres{tbsci/jev_stop_reflect/solved}{0}
\defres{tbsci/jev_stop_reflect/steps}{59.1}
\defres{tbsci/jev_stop_reflect/success}{0.0}
\defres{tbsci/jev_stop_reflect/tests_passed_share}{38.8}
\defres{tbsci/jev_stop_reflect/timeouts}{0}
\defres{tbsci/jev_stop_reflect/verdict}{passes the exploratory success screen; faster not estimable (no jointly solved task)}
\defres{tbsci/jev_stop_reflect/vs_baseline/cost_savings}{$+$15.9}
\defres{tbsci/jev_stop_reflect/vs_baseline/d_run_median_all_s}{$-$20.8}
\defres{tbsci/jev_stop_reflect/vs_baseline/delta}{$+$0.0}
\defres{tbsci/jev_stop_reflect/vs_baseline/delta_lb90}{$+$0.0}
\defres{tbsci/jev_stop_reflect/vs_baseline/losses}{0}
\defres{tbsci/jev_stop_reflect/vs_baseline/wins}{0}
\defres{tbsci/jev_stop_reflect/wall_p50_s}{311}
\defres{tbsci/jev_stop_reflect/wall_p90_s}{948}
\defres{tbsci/k}{3}
\defres{tbsci/n_target}{20}
\defres{tbsci/n_tasks}{15}
\defres{tbsci/n_trials}{45}
\defres{tbsci/pool}{37}
\InputIfFileExists{generated/secondary}{}{}   %
\providecommand{\res}[1]{\ifcsname jevres@#1\endcsname\csname jevres@#1\endcsname\else\tbd\fi}
\providecommand{\defres}[2]{\expandafter\def\csname jevres@#1\endcsname{#2}}
\defres{spend/amortisation/episodes}{5{,}759}
\defres{spend/amortisation/saving_per_episode}{0.0335}
\defres{spend/amortisation/training_usd}{192.93}
\defres{spend/api/development_pilots}{196.06}
\defres{spend/api/exploratory}{8.47}
\defres{spend/api/figures}{2.17}
\defres{spend/api/hard_tasks}{413.63}
\defres{spend/api/reviews}{16.75}
\defres{spend/api/round6_evaluation}{13.22}
\defres{spend/api/stage1_calibration}{16.59}
\defres{spend/api/stage1_main}{931.07}
\defres{spend/api/stage1_tau2}{235.20}
\defres{spend/api/stage2_calibration_checks}{24.94}
\defres{spend/api/stage2_evaluation}{320.21}
\defres{spend/api/stage2_sol}{14.89}
\defres{spend/api/total}{2{,}383.27}
\defres{spend/api/training_round5}{142.36}
\defres{spend/api/training_round6}{42.43}
\defres{spend/api/training_v1_data}{5.28}
\defres{spend/gpu/explore/hours}{1.6}
\defres{spend/gpu/explore/usd}{12.09}
\defres{spend/gpu/round5/hours}{15.1}
\defres{spend/gpu/round5/usd}{45.29}
\defres{spend/s24_host/hours_upper}{0.37}
\defres{spend/s24_host/per_episode_upper}{0.00164}
\defres{spend/s24_host/tariff_q}{0.00196}
\defres{spend/s24_host/usd_per_h}{2.24}
\defres{spend/s24_host/usd_upper}{0.82}
\defres{spend/unpriced_calls}{23}
\providecommand{\res}[1]{\ifcsname jevres@#1\endcsname\csname jevres@#1\endcsname\else\tbd\fi}
\providecommand{\defres}[2]{\expandafter\def\csname jevres@#1\endcsname{#2}}
\defres{rl/chain/base_model}{Qwen/Qwen3-4B}
\defres{rl/chain/frozen}{GRPO (iteration 15)}
\defres{rl/chain/grpo/reward_live}{0.6846}
\defres{rl/chain/grpo/success_live}{100.0}
\defres{rl/chain/grpo_i15/reward_live}{0.6822}
\defres{rl/chain/grpo_i15/success_live}{100.0}
\defres{rl/chain/kept}{SFT v1 + GRPO (iteration 15)}
\defres{rl/chain/opd/reward_live}{0.6730}
\defres{rl/chain/opd/success_live}{100.0}
\defres{rl/chain/opd_i12/reward_live}{0.6724}
\defres{rl/chain/opd_i12/success_live}{100.0}
\defres{rl/chain/sft_v1/reward_live}{0.6788}
\defres{rl/chain/sft_v1/success_live}{99.6}
\defres{rl/dev/base/gate_acc}{72.9}
\defres{rl/dev/base/v1_acc}{70.1}
\defres{rl/dev/base/v1_ece}{0.278}
\defres{rl/dev/base/v1_recovery}{66.3}
\defres{rl/dev/base/v1_stop}{80.4}
\defres{rl/dev/base/v1_tool_params}{81.6}
\defres{rl/dev/base/v1_tool_select}{51.6}
\defres{rl/dev/final/gate_acc}{100.0}
\defres{rl/dev/final/go}{GO}
\defres{rl/dev/final/s1_ms_p50}{314}
\defres{rl/dev/final/ts_replace_success}{100.0}
\defres{rl/dev/final/ts_replace_ts}{100.0}
\defres{rl/dev/final/v1_acc}{99.7}
\defres{rl/dev/final/v1_ece}{0.004}
\defres{rl/dev/final/v1_recovery}{99.8}
\defres{rl/dev/final/v1_stop}{99.4}
\defres{rl/dev/final/v1_tool_params}{100.0}
\defres{rl/dev/final/v1_tool_select}{99.7}
\defres{rl/dev/grpo/gate_acc}{100.0}
\defres{rl/dev/grpo/go}{GO}
\defres{rl/dev/grpo/s1_ms_p50}{390}
\defres{rl/dev/grpo/ts_replace_success}{100.0}
\defres{rl/dev/grpo/ts_replace_ts}{100.0}
\defres{rl/dev/grpo/v1_acc}{99.4}
\defres{rl/dev/grpo/v1_ece}{0.003}
\defres{rl/dev/grpo/v1_recovery}{99.5}
\defres{rl/dev/grpo/v1_stop}{99.1}
\defres{rl/dev/grpo/v1_tool_params}{100.0}
\defres{rl/dev/grpo/v1_tool_select}{99.1}
\defres{rl/dev/grpo_i15/gate_acc}{100.0}
\defres{rl/dev/grpo_i15/go}{GO}
\defres{rl/dev/grpo_i15/s1_ms_p50}{443}
\defres{rl/dev/grpo_i15/ts_replace_success}{100.0}
\defres{rl/dev/grpo_i15/ts_replace_ts}{100.0}
\defres{rl/dev/grpo_i15/v1_acc}{99.7}
\defres{rl/dev/grpo_i15/v1_ece}{0.004}
\defres{rl/dev/grpo_i15/v1_recovery}{99.8}
\defres{rl/dev/grpo_i15/v1_stop}{99.4}
\defres{rl/dev/grpo_i15/v1_tool_params}{100.0}
\defres{rl/dev/grpo_i15/v1_tool_select}{99.7}
\defres{rl/dev/intern/gate_acc}{86.2}
\defres{rl/dev/intern/s1_ms_p50}{139}
\defres{rl/dev/intern/ts_replace_success}{26.0}
\defres{rl/dev/intern/ts_replace_ts}{54.4}
\defres{rl/dev/intern/v1_acc}{84.3}
\defres{rl/dev/intern/v1_recovery}{82.0}
\defres{rl/dev/intern/v1_stop}{93.0}
\defres{rl/dev/intern/v1_tool_params}{87.5}
\defres{rl/dev/intern/v1_tool_select}{74.8}
\defres{rl/dev/jev/gate_acc}{97.1}
\defres{rl/dev/jev/s1_ms_p50}{157}
\defres{rl/dev/jev/ts_replace_success}{80.0}
\defres{rl/dev/jev/ts_replace_ts}{97.0}
\defres{rl/dev/jev/v1_acc}{95.8}
\defres{rl/dev/jev/v1_ece}{0.013}
\defres{rl/dev/jev/v1_recovery}{94.4}
\defres{rl/dev/jev/v1_stop}{98.7}
\defres{rl/dev/jev/v1_tool_params}{97.9}
\defres{rl/dev/jev/v1_tool_select}{91.6}
\defres{rl/dev/llm_only/reward_ccps}{0.6250}
\defres{rl/dev/llm_only/success}{100.0}
\defres{rl/dev/opd/gate_acc}{94.8}
\defres{rl/dev/opd/go}{GO}
\defres{rl/dev/opd/s1_ms_p50}{416}
\defres{rl/dev/opd/ts_replace_success}{100.0}
\defres{rl/dev/opd/ts_replace_ts}{100.0}
\defres{rl/dev/opd/v1_acc}{98.9}
\defres{rl/dev/opd/v1_ece}{0.021}
\defres{rl/dev/opd/v1_recovery}{98.0}
\defres{rl/dev/opd/v1_stop}{98.4}
\defres{rl/dev/opd/v1_tool_params}{99.8}
\defres{rl/dev/opd/v1_tool_select}{98.5}
\defres{rl/dev/opd_i12/gate_acc}{100.0}
\defres{rl/dev/opd_i12/go}{GO}
\defres{rl/dev/opd_i12/s1_ms_p50}{413}
\defres{rl/dev/opd_i12/ts_replace_success}{100.0}
\defres{rl/dev/opd_i12/ts_replace_ts}{100.0}
\defres{rl/dev/opd_i12/v1_acc}{98.9}
\defres{rl/dev/opd_i12/v1_ece}{0.023}
\defres{rl/dev/opd_i12/v1_recovery}{97.9}
\defres{rl/dev/opd_i12/v1_stop}{98.5}
\defres{rl/dev/opd_i12/v1_tool_params}{99.9}
\defres{rl/dev/opd_i12/v1_tool_select}{98.4}
\defres{rl/dev/sft_v0/gate_acc}{99.5}
\defres{rl/dev/sft_v0/go}{GO}
\defres{rl/dev/sft_v0/s1_ms_p50}{263}
\defres{rl/dev/sft_v0/ts_replace_success}{88.0}
\defres{rl/dev/sft_v0/ts_replace_ts}{90.3}
\defres{rl/dev/sft_v0/v1_acc}{95.2}
\defres{rl/dev/sft_v0/v1_ece}{0.022}
\defres{rl/dev/sft_v0/v1_recovery}{92.2}
\defres{rl/dev/sft_v0/v1_stop}{98.2}
\defres{rl/dev/sft_v0/v1_tool_params}{95.3}
\defres{rl/dev/sft_v0/v1_tool_select}{92.8}
\defres{rl/dev/sft_v1/gate_acc}{100.0}
\defres{rl/dev/sft_v1/go}{GO}
\defres{rl/dev/sft_v1/s1_ms_p50}{367}
\defres{rl/dev/sft_v1/ts_replace_success}{100.0}
\defres{rl/dev/sft_v1/ts_replace_ts}{100.0}
\defres{rl/dev/sft_v1/v1_acc}{99.7}
\defres{rl/dev/sft_v1/v1_ece}{0.029}
\defres{rl/dev/sft_v1/v1_recovery}{99.8}
\defres{rl/dev/sft_v1/v1_stop}{99.4}
\defres{rl/dev/sft_v1/v1_tool_params}{100.0}
\defres{rl/dev/sft_v1/v1_tool_select}{99.7}
\defres{rl/gpu}{L40S}
\defres{rl/latency/intern/p50_ms_c1}{414}
\defres{rl/latency/intern/p50_ms_c4}{1{,}648}
\defres{rl/latency/intern/p95_ms_c1}{416}
\defres{rl/latency/intern/req_per_s_c1}{2.4}
\defres{rl/latency/jev/p50_ms_c1}{157}
\defres{rl/latency/jev/p50_ms_c4}{161}
\defres{rl/latency/jev/p95_ms_c1}{203}
\defres{rl/latency/jev/req_per_s_c1}{6.2}
\defres{rl/latency/jevagentrl/p50_ms_c1}{307}
\defres{rl/latency/jevagentrl/p50_ms_c4}{1{,}213}
\defres{rl/latency/jevagentrl/p95_ms_c1}{310}
\defres{rl/latency/jevagentrl/req_per_s_c1}{3.3}
\defres{rl2/base}{StartLux-Decision-4B}
\defres{rl2/dev/grpo/gate_acc}{99.5}
\defres{rl2/dev/grpo/v1_acc}{99.6}
\defres{rl2/dev/grpo/v1_ece}{0.002}
\defres{rl2/dev/grpo/v1_recovery}{99.7}
\defres{rl2/dev/grpo/v1_tool_select}{99.5}
\defres{rl2/dev/grpo_i15/gate_acc}{99.5}
\defres{rl2/dev/grpo_i15/v1_acc}{99.6}
\defres{rl2/dev/grpo_i15/v1_ece}{0.006}
\defres{rl2/dev/grpo_i15/v1_recovery}{99.7}
\defres{rl2/dev/grpo_i15/v1_tool_select}{99.7}
\defres{rl2/dev/opd/gate_acc}{100.0}
\defres{rl2/dev/opd/v1_acc}{98.9}
\defres{rl2/dev/opd/v1_ece}{0.021}
\defres{rl2/dev/opd/v1_recovery}{97.7}
\defres{rl2/dev/opd/v1_tool_select}{98.1}
\defres{rl2/dev/opd_i12/gate_acc}{99.0}
\defres{rl2/dev/opd_i12/v1_acc}{98.7}
\defres{rl2/dev/opd_i12/v1_ece}{0.017}
\defres{rl2/dev/opd_i12/v1_recovery}{97.9}
\defres{rl2/dev/opd_i12/v1_tool_select}{97.3}
\defres{rl2/dev/sft_v2/gate_acc}{99.5}
\defres{rl2/dev/sft_v2/v1_acc}{99.6}
\defres{rl2/dev/sft_v2/v1_ece}{0.027}
\defres{rl2/dev/sft_v2/v1_recovery}{99.7}
\defres{rl2/dev/sft_v2/v1_tool_select}{99.5}
\defres{rl2/dev/zeroshot_startlux/gate_acc}{90.5}
\defres{rl2/dev/zeroshot_startlux/v1_acc}{94.2}
\defres{rl2/dev/zeroshot_startlux/v1_ece}{0.007}
\defres{rl2/dev/zeroshot_startlux/v1_tool_select}{88.8}
\defres{rl2/frozen}{SFT v2}
\defres{rl2/go}{GO}
\defres{rl2/kept}{SFT v2}
\defres{rl2/rl_claim}{no}
\defres{rl2/stress/grpo/A}{93.9}
\defres{rl2/stress/grpo/P}{90.9}
\defres{rl2/stress/grpo/R}{0.404}
\defres{rl2/stress/grpo/S}{60.0}
\defres{rl2/stress/grpo_i15/A}{94.9}
\defres{rl2/stress/grpo_i15/P}{92.3}
\defres{rl2/stress/grpo_i15/R}{0.410}
\defres{rl2/stress/grpo_i15/S}{62.5}
\defres{rl2/stress/intern/A}{61.1}
\defres{rl2/stress/intern/P}{56.8}
\defres{rl2/stress/intern/R}{-0.046}
\defres{rl2/stress/intern/S}{11.5}
\defres{rl2/stress/jev/A}{69.1}
\defres{rl2/stress/jev/P}{86.5}
\defres{rl2/stress/jev/R}{0.323}
\defres{rl2/stress/jev/S}{49.0}
\defres{rl2/stress/n}{200}
\defres{rl2/stress/opd/A}{93.9}
\defres{rl2/stress/opd/P}{96.2}
\defres{rl2/stress/opd/R}{0.453}
\defres{rl2/stress/opd/S}{65.0}
\defres{rl2/stress/opd_i12/A}{94.7}
\defres{rl2/stress/opd_i12/P}{92.8}
\defres{rl2/stress/opd_i12/R}{0.473}
\defres{rl2/stress/opd_i12/S}{66.5}
\defres{rl2/stress/sft_v2/A}{95.4}
\defres{rl2/stress/sft_v2/P}{93.3}
\defres{rl2/stress/sft_v2/R}{0.502}
\defres{rl2/stress/sft_v2/S}{71.5}
\defres{rl2/zeroshot/qwen/v1_acc}{70.6}
\defres{rl2/zeroshot/startlux/v1_acc}{94.2}
\InputIfFileExists{generated/fig1_episode}{}{}  %
\makeatletter
\providecommand{\res}[1]{\ifcsname jevres@#1\endcsname\csname jevres@#1\endcsname\else\tbd\fi}
\providecommand{\defres}[2]{\expandafter\def\csname jevres@#1\endcsname{#2}}
\makeatother
\defres{t2ref/n_questions}{22{,}704}
\defres{t2ref/airline_test/tool_params/labelled}{1709}
\defres{t2ref/airline_test/tool_params/occurrences}{3363}
\defres{t2ref/airline_test/tool_params/coverage}{50.8}
\defres{t2ref/airline_test/tool_params/tasks}{29}
\defres{t2ref/airline_test/tool_select/labelled}{11992}
\defres{t2ref/airline_test/tool_select/occurrences}{11992}
\defres{t2ref/airline_test/tool_select/coverage}{100.0}
\defres{t2ref/airline_test/tool_select/tasks}{35}
\defres{t2ref/retail/tool_params/labelled}{2314}
\defres{t2ref/retail/tool_params/occurrences}{3973}
\defres{t2ref/retail/tool_params/coverage}{58.2}
\defres{t2ref/retail/tool_params/tasks}{108}
\defres{t2ref/retail/tool_select/labelled}{7942}
\defres{t2ref/retail/tool_select/occurrences}{7942}
\defres{t2ref/retail/tool_select/coverage}{100.0}
\defres{t2ref/retail/tool_select/tasks}{114}
\defres{t2ref/airline_test/tool_select/n}{11992}
\defres{t2ref/airline_test/tool_select/tasks}{35}
\defres{t2ref/airline_test/tool_select/frozen/agree}{56.0}
\defres{t2ref/airline_test/tool_select/frozen/ci/lo}{50.1}
\defres{t2ref/airline_test/tool_select/frozen/ci/hi}{61.6}
\defres{t2ref/airline_test/tool_select/intern/agree}{46.2}
\defres{t2ref/airline_test/tool_select/intern/ci/lo}{41.3}
\defres{t2ref/airline_test/tool_select/intern/ci/hi}{50.8}
\defres{t2ref/airline_test/tool_select/base/agree}{54.6}
\defres{t2ref/airline_test/tool_select/base/ci/lo}{47.3}
\defres{t2ref/airline_test/tool_select/base/ci/hi}{61.6}
\defres{t2ref/airline_test/tool_select/frozen_vs_intern/delta}{+9.9}
\defres{t2ref/airline_test/tool_select/frozen_vs_intern/ci/lo}{+5.0}
\defres{t2ref/airline_test/tool_select/frozen_vs_intern/ci/hi}{+14.8}
\defres{t2ref/airline_test/tool_select/frozen_vs_base/delta}{+1.5}
\defres{t2ref/airline_test/tool_select/frozen_vs_base/ci/lo}{-3.1}
\defres{t2ref/airline_test/tool_select/frozen_vs_base/ci/hi}{+6.1}
\defres{t2ref/airline_test/tool_select/majority/agree}{65.4}
\defres{t2ref/airline_test/tool_select/majority/label}{respond\_to\_user}
\defres{t2ref/airline_test/tool_select/jevsub/n}{11992}
\defres{t2ref/airline_test/tool_select/jevsub/jev/agree}{61.4}
\defres{t2ref/airline_test/tool_select/jevsub/frozen/agree}{56.0}
\defres{t2ref/airline_test/tool_select/jevsub/frozen/delta}{-5.3}
\defres{t2ref/airline_test/tool_select/jevsub/frozen/ci/lo}{-11.1}
\defres{t2ref/airline_test/tool_select/jevsub/frozen/ci/hi}{+0.3}
\defres{t2ref/airline_test/tool_select/jevsub/intern/agree}{46.2}
\defres{t2ref/airline_test/tool_select/jevsub/intern/delta}{-15.2}
\defres{t2ref/airline_test/tool_select/jevsub/intern/ci/lo}{-19.2}
\defres{t2ref/airline_test/tool_select/jevsub/intern/ci/hi}{-11.1}
\defres{t2ref/airline_test/tool_select/jevsub/base/agree}{54.6}
\defres{t2ref/airline_test/tool_select/jevsub/base/delta}{-6.8}
\defres{t2ref/airline_test/tool_select/jevsub/base/ci/lo}{-14.3}
\defres{t2ref/airline_test/tool_select/jevsub/base/ci/hi}{+0.5}
\defres{t2ref/airline_test/tool_params/n}{1709}
\defres{t2ref/airline_test/tool_params/tasks}{29}
\defres{t2ref/airline_test/tool_params/frozen/agree}{85.7}
\defres{t2ref/airline_test/tool_params/frozen/ci/lo}{77.9}
\defres{t2ref/airline_test/tool_params/frozen/ci/hi}{92.6}
\defres{t2ref/airline_test/tool_params/intern/agree}{81.0}
\defres{t2ref/airline_test/tool_params/intern/ci/lo}{71.7}
\defres{t2ref/airline_test/tool_params/intern/ci/hi}{89.3}
\defres{t2ref/airline_test/tool_params/base/agree}{79.6}
\defres{t2ref/airline_test/tool_params/base/ci/lo}{69.6}
\defres{t2ref/airline_test/tool_params/base/ci/hi}{88.7}
\defres{t2ref/airline_test/tool_params/frozen_vs_intern/delta}{+4.7}
\defres{t2ref/airline_test/tool_params/frozen_vs_intern/ci/lo}{+1.5}
\defres{t2ref/airline_test/tool_params/frozen_vs_intern/ci/hi}{+7.9}
\defres{t2ref/airline_test/tool_params/frozen_vs_base/delta}{+6.1}
\defres{t2ref/airline_test/tool_params/frozen_vs_base/ci/lo}{+2.9}
\defres{t2ref/airline_test/tool_params/frozen_vs_base/ci/hi}{+9.8}
\defres{t2ref/airline_test/tool_params/jevsub/n}{1709}
\defres{t2ref/airline_test/tool_params/jevsub/jev/agree}{87.4}
\defres{t2ref/airline_test/tool_params/jevsub/frozen/agree}{85.7}
\defres{t2ref/airline_test/tool_params/jevsub/frozen/delta}{-1.8}
\defres{t2ref/airline_test/tool_params/jevsub/frozen/ci/lo}{-5.8}
\defres{t2ref/airline_test/tool_params/jevsub/frozen/ci/hi}{+2.1}
\defres{t2ref/airline_test/tool_params/jevsub/intern/agree}{81.0}
\defres{t2ref/airline_test/tool_params/jevsub/intern/delta}{-6.5}
\defres{t2ref/airline_test/tool_params/jevsub/intern/ci/lo}{-12.5}
\defres{t2ref/airline_test/tool_params/jevsub/intern/ci/hi}{-0.7}
\defres{t2ref/airline_test/tool_params/jevsub/base/agree}{79.6}
\defres{t2ref/airline_test/tool_params/jevsub/base/delta}{-7.8}
\defres{t2ref/airline_test/tool_params/jevsub/base/ci/lo}{-14.5}
\defres{t2ref/airline_test/tool_params/jevsub/base/ci/hi}{-1.8}
\defres{t2ref/retail/tool_select/n}{7942}
\defres{t2ref/retail/tool_select/tasks}{114}
\defres{t2ref/retail/tool_select/frozen/agree}{52.9}
\defres{t2ref/retail/tool_select/frozen/ci/lo}{50.2}
\defres{t2ref/retail/tool_select/frozen/ci/hi}{55.6}
\defres{t2ref/retail/tool_select/intern/agree}{40.9}
\defres{t2ref/retail/tool_select/intern/ci/lo}{37.9}
\defres{t2ref/retail/tool_select/intern/ci/hi}{43.9}
\defres{t2ref/retail/tool_select/base/agree}{37.0}
\defres{t2ref/retail/tool_select/base/ci/lo}{34.4}
\defres{t2ref/retail/tool_select/base/ci/hi}{39.5}
\defres{t2ref/retail/tool_select/frozen_vs_intern/delta}{+12.0}
\defres{t2ref/retail/tool_select/frozen_vs_intern/ci/lo}{+9.1}
\defres{t2ref/retail/tool_select/frozen_vs_intern/ci/hi}{+15.1}
\defres{t2ref/retail/tool_select/frozen_vs_base/delta}{+16.0}
\defres{t2ref/retail/tool_select/frozen_vs_base/ci/lo}{+13.2}
\defres{t2ref/retail/tool_select/frozen_vs_base/ci/hi}{+18.9}
\defres{t2ref/retail/tool_select/majority/agree}{50.7}
\defres{t2ref/retail/tool_select/majority/label}{respond\_to\_user}
\defres{t2ref/retail/tool_select/jevsub/n}{7942}
\defres{t2ref/retail/tool_select/jevsub/jev/agree}{59.4}
\defres{t2ref/retail/tool_select/jevsub/frozen/agree}{52.9}
\defres{t2ref/retail/tool_select/jevsub/frozen/delta}{-6.5}
\defres{t2ref/retail/tool_select/jevsub/frozen/ci/lo}{-8.9}
\defres{t2ref/retail/tool_select/jevsub/frozen/ci/hi}{-4.1}
\defres{t2ref/retail/tool_select/jevsub/intern/agree}{40.9}
\defres{t2ref/retail/tool_select/jevsub/intern/delta}{-18.5}
\defres{t2ref/retail/tool_select/jevsub/intern/ci/lo}{-21.5}
\defres{t2ref/retail/tool_select/jevsub/intern/ci/hi}{-15.7}
\defres{t2ref/retail/tool_select/jevsub/base/agree}{37.0}
\defres{t2ref/retail/tool_select/jevsub/base/delta}{-22.5}
\defres{t2ref/retail/tool_select/jevsub/base/ci/lo}{-25.6}
\defres{t2ref/retail/tool_select/jevsub/base/ci/hi}{-19.4}
\defres{t2ref/retail/tool_params/n}{2314}
\defres{t2ref/retail/tool_params/tasks}{108}
\defres{t2ref/retail/tool_params/frozen/agree}{86.0}
\defres{t2ref/retail/tool_params/frozen/ci/lo}{83.1}
\defres{t2ref/retail/tool_params/frozen/ci/hi}{88.7}
\defres{t2ref/retail/tool_params/intern/agree}{76.2}
\defres{t2ref/retail/tool_params/intern/ci/lo}{73.1}
\defres{t2ref/retail/tool_params/intern/ci/hi}{79.3}
\defres{t2ref/retail/tool_params/base/agree}{69.0}
\defres{t2ref/retail/tool_params/base/ci/lo}{65.1}
\defres{t2ref/retail/tool_params/base/ci/hi}{73.0}
\defres{t2ref/retail/tool_params/frozen_vs_intern/delta}{+9.8}
\defres{t2ref/retail/tool_params/frozen_vs_intern/ci/lo}{+7.1}
\defres{t2ref/retail/tool_params/frozen_vs_intern/ci/hi}{+12.5}
\defres{t2ref/retail/tool_params/frozen_vs_base/delta}{+17.0}
\defres{t2ref/retail/tool_params/frozen_vs_base/ci/lo}{+13.7}
\defres{t2ref/retail/tool_params/frozen_vs_base/ci/hi}{+20.4}
\defres{t2ref/retail/tool_params/jevsub/n}{2314}
\defres{t2ref/retail/tool_params/jevsub/jev/agree}{90.7}
\defres{t2ref/retail/tool_params/jevsub/frozen/agree}{86.0}
\defres{t2ref/retail/tool_params/jevsub/frozen/delta}{-4.7}
\defres{t2ref/retail/tool_params/jevsub/frozen/ci/lo}{-7.3}
\defres{t2ref/retail/tool_params/jevsub/frozen/ci/hi}{-1.8}
\defres{t2ref/retail/tool_params/jevsub/intern/agree}{76.2}
\defres{t2ref/retail/tool_params/jevsub/intern/delta}{-14.6}
\defres{t2ref/retail/tool_params/jevsub/intern/ci/lo}{-17.9}
\defres{t2ref/retail/tool_params/jevsub/intern/ci/hi}{-10.9}
\defres{t2ref/retail/tool_params/jevsub/base/agree}{69.0}
\defres{t2ref/retail/tool_params/jevsub/base/delta}{-21.7}
\defres{t2ref/retail/tool_params/jevsub/base/ci/lo}{-25.0}
\defres{t2ref/retail/tool_params/jevsub/base/ci/hi}{-18.4}
\defres{t2ref/all/tool_select/n}{19934}
\defres{t2ref/all/tool_select/tasks}{149}
\defres{t2ref/all/tool_select/frozen/agree}{53.7}
\defres{t2ref/all/tool_select/frozen/ci/lo}{51.2}
\defres{t2ref/all/tool_select/frozen/ci/hi}{56.2}
\defres{t2ref/all/tool_select/intern/agree}{42.1}
\defres{t2ref/all/tool_select/intern/ci/lo}{39.5}
\defres{t2ref/all/tool_select/intern/ci/hi}{44.8}
\defres{t2ref/all/tool_select/base/agree}{41.1}
\defres{t2ref/all/tool_select/base/ci/lo}{38.3}
\defres{t2ref/all/tool_select/base/ci/hi}{44.0}
\defres{t2ref/all/tool_select/frozen_vs_intern/delta}{+11.5}
\defres{t2ref/all/tool_select/frozen_vs_intern/ci/lo}{+9.0}
\defres{t2ref/all/tool_select/frozen_vs_intern/ci/hi}{+14.1}
\defres{t2ref/all/tool_select/frozen_vs_base/delta}{+12.6}
\defres{t2ref/all/tool_select/frozen_vs_base/ci/lo}{+10.0}
\defres{t2ref/all/tool_select/frozen_vs_base/ci/hi}{+15.2}
\defres{t2ref/all/tool_select/majority/agree}{54.1}
\defres{t2ref/all/tool_select/majority/label}{respond\_to\_user}
\defres{t2ref/all/tool_select/jevsub/n}{19934}
\defres{t2ref/all/tool_select/jevsub/jev/agree}{59.9}
\defres{t2ref/all/tool_select/jevsub/frozen/agree}{53.7}
\defres{t2ref/all/tool_select/jevsub/frozen/delta}{-6.2}
\defres{t2ref/all/tool_select/jevsub/frozen/ci/lo}{-8.5}
\defres{t2ref/all/tool_select/jevsub/frozen/ci/hi}{-3.9}
\defres{t2ref/all/tool_select/jevsub/intern/agree}{42.1}
\defres{t2ref/all/tool_select/jevsub/intern/delta}{-17.8}
\defres{t2ref/all/tool_select/jevsub/intern/ci/lo}{-20.1}
\defres{t2ref/all/tool_select/jevsub/intern/ci/hi}{-15.4}
\defres{t2ref/all/tool_select/jevsub/base/agree}{41.1}
\defres{t2ref/all/tool_select/jevsub/base/delta}{-18.8}
\defres{t2ref/all/tool_select/jevsub/base/ci/lo}{-21.9}
\defres{t2ref/all/tool_select/jevsub/base/ci/hi}{-15.6}
\defres{t2ref/all/tool_params/n}{4023}
\defres{t2ref/all/tool_params/tasks}{137}
\defres{t2ref/all/tool_params/frozen/agree}{85.9}
\defres{t2ref/all/tool_params/frozen/ci/lo}{83.2}
\defres{t2ref/all/tool_params/frozen/ci/hi}{88.6}
\defres{t2ref/all/tool_params/intern/agree}{77.2}
\defres{t2ref/all/tool_params/intern/ci/lo}{74.1}
\defres{t2ref/all/tool_params/intern/ci/hi}{80.2}
\defres{t2ref/all/tool_params/base/agree}{71.3}
\defres{t2ref/all/tool_params/base/ci/lo}{67.4}
\defres{t2ref/all/tool_params/base/ci/hi}{75.0}
\defres{t2ref/all/tool_params/frozen_vs_intern/delta}{+8.7}
\defres{t2ref/all/tool_params/frozen_vs_intern/ci/lo}{+6.5}
\defres{t2ref/all/tool_params/frozen_vs_intern/ci/hi}{+11.0}
\defres{t2ref/all/tool_params/frozen_vs_base/delta}{+14.7}
\defres{t2ref/all/tool_params/frozen_vs_base/ci/lo}{+11.9}
\defres{t2ref/all/tool_params/frozen_vs_base/ci/hi}{+17.6}
\defres{t2ref/all/tool_params/jevsub/n}{4023}
\defres{t2ref/all/tool_params/jevsub/jev/agree}{90.0}
\defres{t2ref/all/tool_params/jevsub/frozen/agree}{85.9}
\defres{t2ref/all/tool_params/jevsub/frozen/delta}{-4.1}
\defres{t2ref/all/tool_params/jevsub/frozen/ci/lo}{-6.3}
\defres{t2ref/all/tool_params/jevsub/frozen/ci/hi}{-1.6}
\defres{t2ref/all/tool_params/jevsub/intern/agree}{77.2}
\defres{t2ref/all/tool_params/jevsub/intern/delta}{-12.8}
\defres{t2ref/all/tool_params/jevsub/intern/ci/lo}{-15.8}
\defres{t2ref/all/tool_params/jevsub/intern/ci/hi}{-9.6}
\defres{t2ref/all/tool_params/jevsub/base/agree}{71.3}
\defres{t2ref/all/tool_params/jevsub/base/delta}{-18.8}
\defres{t2ref/all/tool_params/jevsub/base/ci/lo}{-21.9}
\defres{t2ref/all/tool_params/jevsub/base/ci/hi}{-15.8}
\providecommand{\res}[1]{\ifcsname jevres@#1\endcsname\csname jevres@#1\endcsname\else\tbd\fi}
\providecommand{\defres}[2]{\expandafter\def\csname jevres@#1\endcsname{#2}}
\defres{recipe/base/sft_gap}{-0.1}
\defres{recipe/base/zeroshot_gap}{+23.6}
\defres{recipe/calib/base/acc_minus_conf}{-27.8}
\defres{recipe/calib/base/mean_conf}{97.9}
\defres{recipe/calib/base/temp_global}{6.31}
\defres{recipe/calib/frozen/temp_global}{0.78}
\defres{recipe/calib/frozen/v1_acc_temp}{99.7}
\defres{recipe/calib/frozen/v1_ece_temp}{0.001}
\defres{recipe/calib/grpo_i15/acc_minus_conf}{+0.3}
\defres{recipe/calib/grpo_i15/mean_conf}{99.4}
\defres{recipe/calib/sft_v1/acc_minus_conf}{+2.8}
\defres{recipe/calib/sft_v1/mean_conf}{96.9}
\defres{recipe/calib/sft_v1/temp_global}{0.40}
\defres{recipe/calib/v2/temp_global}{0.51}
\defres{recipe/cost/cal_arms_usd}{8.03}
\defres{recipe/cost/dev_slices_share}{67.2}
\defres{recipe/cost/dev_slices_usd}{95.69}
\defres{recipe/cost/gpu_usd_per_h}{3.004}
\defres{recipe/cost/sft_gpu_usd}{6.71}
\defres{recipe/cost/sft_only_usd}{11.99}
\defres{recipe/data/v0/after_error_rows}{0}
\defres{recipe/data/v0/gold_rows}{1{,}225}
\defres{recipe/data/v0/rows}{1{,}225}
\defres{recipe/data/v0/state_chars_median}{563}
\defres{recipe/data/v0/tasks}{77}
\defres{recipe/data/v1/after_error_rows}{4{,}843}
\defres{recipe/data/v1/after_error_share}{30.9}
\defres{recipe/data/v1/build_usd}{5.28}
\defres{recipe/data/v1/dev_recovery_rows}{606}
\defres{recipe/data/v1/dev_rows}{2{,}377}
\defres{recipe/data/v1/episodes}{1{,}000}
\defres{recipe/data/v1/episodes_usd}{4.65}
\defres{recipe/data/v1/gold_fallback_rows}{864}
\defres{recipe/data/v1/gold_fallback_share}{5.5}
\defres{recipe/data/v1/guard_no_rows}{673}
\defres{recipe/data/v1/knockout_rows}{2{,}133}
\defres{recipe/data/v1/nearmiss_offplan_share}{72.0}
\defres{recipe/data/v1/nearmiss_rows}{8{,}083}
\defres{recipe/data/v1/nearmiss_share}{51.5}
\defres{recipe/data/v1/offplan_episodes}{570}
\defres{recipe/data/v1/rows}{15{,}694}
\defres{recipe/data/v1/rows_ratio}{12.8}
\defres{recipe/data/v1/state_chars_median}{1{,}808}
\defres{recipe/data/v1/target_argmax_gold_share}{100.0}
\defres{recipe/data/v1/target_gold_mass_mean}{96.9}
\defres{recipe/data/v1/tasks}{587}
\defres{recipe/data/v1/teacher_rows}{14{,}830}
\defres{recipe/data/v1/teacher_train_acc}{94.5}
\defres{recipe/data/v1/teacher_usd}{0.63}
\defres{recipe/deploy/latency/jev/ms_per_question_c1}{52}
\defres{recipe/deploy/latency/ratio_c1}{2.0}
\defres{recipe/deploy/latency/ratio_c4}{7.5}
\defres{recipe/deploy/latency/rl/ms_per_question_c1}{102}
\defres{recipe/deploy/tauhat/jev_flash_tp}{0.99}
\defres{recipe/deploy/tauhat/jev_flash_tp_n}{1{,}324}
\defres{recipe/deploy/tauhat/jev_flash_tp_wrong}{56}
\defres{recipe/deploy/tauhat/jev_flash_ts}{1.00}
\defres{recipe/deploy/tauhat/jev_flash_ts_n}{1{,}125}
\defres{recipe/deploy/tauhat/jev_flash_ts_wrong}{67}
\defres{recipe/deploy/tauhat/rl_flash_tp}{0.9999999}
\defres{recipe/deploy/tauhat/rl_flash_tp_accept}{80.5}
\defres{recipe/deploy/tauhat/rl_flash_tp_n}{1{,}328}
\defres{recipe/deploy/tauhat/rl_flash_tp_wrong}{13}
\defres{recipe/deploy/tauhat/rl_flash_ts}{0.50}
\defres{recipe/deploy/tauhat/rl_flash_ts_accept}{100.0}
\defres{recipe/deploy/tauhat/rl_flash_ts_n}{1{,}125}
\defres{recipe/deploy/tauhat/rl_flash_ts_wrong}{0}
\defres{recipe/dev/base/onplan_acc}{71.4}
\defres{recipe/dev/base/recovery_gap}{+5.0}
\defres{recipe/dev/base/recovery_tool_select}{49.6}
\defres{recipe/dev/grpo/onplan_acc}{99.4}
\defres{recipe/dev/grpo/recovery_gap}{-0.1}
\defres{recipe/dev/grpo/recovery_tool_select}{99.2}
\defres{recipe/dev/grpo_i15/onplan_acc}{99.7}
\defres{recipe/dev/grpo_i15/recovery_gap}{-0.2}
\defres{recipe/dev/grpo_i15/recovery_tool_select}{99.6}
\defres{recipe/dev/jev/onplan_acc}{96.3}
\defres{recipe/dev/jev/recovery_gap}{+1.9}
\defres{recipe/dev/opd/onplan_acc}{99.2}
\defres{recipe/dev/opd/recovery_gap}{+1.2}
\defres{recipe/dev/opd/recovery_tool_select}{97.1}
\defres{recipe/dev/opd_i12/onplan_acc}{99.3}
\defres{recipe/dev/opd_i12/recovery_gap}{+1.5}
\defres{recipe/dev/opd_i12/recovery_tool_select}{96.3}
\defres{recipe/dev/sft_v0/onplan_acc}{96.2}
\defres{recipe/dev/sft_v0/recovery_gap}{+3.9}
\defres{recipe/dev/sft_v0/recovery_tool_select}{88.4}
\defres{recipe/dev/sft_v1/onplan_acc}{99.7}
\defres{recipe/dev/sft_v1/recovery_gap}{-0.2}
\defres{recipe/dev/sft_v1/recovery_tool_select}{99.6}
\defres{recipe/labels/c/frozen_minus_jev}{+7.7}
\defres{recipe/labels/c/frozen_minus_jev_offplan}{+13.2}
\defres{recipe/labels/c/sft_v1_minus_jev}{+7.7}
\defres{recipe/labels/c/sft_v1_minus_jev_offplan}{+12.3}
\defres{recipe/labels/c/v2_minus_jev}{+7.9}
\defres{recipe/labels/c/v2_minus_jev_offplan}{+13.6}
\defres{recipe/labels/recovery/student_minus_teacher}{+5.4}
\defres{recipe/labels/tool_select/student_minus_teacher}{+8.0}
\defres{recipe/labels/v1/student_minus_teacher}{+3.9}
\defres{recipe/stage/grpo/api_usd}{26.53}
\defres{recipe/stage/grpo/gpu_h}{4.5}
\defres{recipe/stage/grpo/gpu_min}{270}
\defres{recipe/stage/grpo/gpu_usd}{13.52}
\defres{recipe/stage/opd/api_usd}{12.11}
\defres{recipe/stage/opd/gpu_h}{3.7}
\defres{recipe/stage/opd/gpu_min}{222}
\defres{recipe/stage/opd/gpu_usd}{11.11}
\defres{recipe/stage/opd/teacher_usd}{0.38}
\defres{recipe/stage/rl_extra/api_usd}{38.64}
\defres{recipe/stage/rl_extra/gpu_h}{8.2}
\defres{recipe/stage/rl_extra/gpu_usd}{24.63}
\defres{recipe/stage/sft_v1/gpu_h}{2.2}
\defres{recipe/stage/sft_v1/gpu_min}{134}
\defres{recipe/stage/sft_v1/gpu_usd}{6.71}
\defres{recipe/stages/grpo/delta_recovery}{-0.3}
\defres{recipe/stages/grpo/delta_reward_live}{+0.0058}
\defres{recipe/stages/grpo/delta_v1_acc}{-0.3}
\defres{recipe/stages/grpo_i15/delta_recovery}{+0.0}
\defres{recipe/stages/grpo_i15/delta_reward_live}{+0.0034}
\defres{recipe/stages/grpo_i15/delta_v1_acc}{+0.0}
\defres{recipe/stages/opd/delta_recovery}{-1.8}
\defres{recipe/stages/opd/delta_reward_live}{-0.0058}
\defres{recipe/stages/opd/delta_v1_acc}{-0.8}
\defres{recipe/stages/opd_i12/delta_recovery}{-2.0}
\defres{recipe/stages/opd_i12/delta_reward_live}{-0.0064}
\defres{recipe/stages/opd_i12/delta_v1_acc}{-0.8}
\defres{recipe/stages/v2/grpo/delta_stress_reward}{-0.098}
\defres{recipe/stages/v2/grpo/delta_stress_success}{-11.5}
\defres{recipe/stages/v2/grpo/delta_v1_acc}{+0.0}
\defres{recipe/stages/v2/grpo_i15/delta_stress_reward}{-0.092}
\defres{recipe/stages/v2/grpo_i15/delta_stress_success}{-9.0}
\defres{recipe/stages/v2/grpo_i15/delta_v1_acc}{+0.0}
\defres{recipe/stages/v2/opd/delta_stress_reward}{-0.049}
\defres{recipe/stages/v2/opd/delta_stress_success}{-6.5}
\defres{recipe/stages/v2/opd/delta_v1_acc}{-0.7}
\defres{recipe/stages/v2/opd_i12/delta_stress_reward}{-0.029}
\defres{recipe/stages/v2/opd_i12/delta_stress_success}{-5.0}
\defres{recipe/stages/v2/opd_i12/delta_v1_acc}{-1.0}
\defres{recipe/states/delta_recovery}{+7.6}
\defres{recipe/states/delta_recovery_tool_select}{+11.2}
\defres{recipe/states/delta_tool_select}{+6.9}
\defres{recipe/states/delta_ts_replace_own_acc}{+9.7}
\defres{recipe/states/delta_ts_replace_success}{+12.0}
\defres{recipe/states/delta_v1_acc}{+4.5}
\InputIfFileExists{generated/economics}{}{}   %
\InputIfFileExists{generated/failure_audit}{}{}  %
\InputIfFileExists{generated/explore}{}{}     %
\InputIfFileExists{generated/stage2_sol}{}{}  %
\InputIfFileExists{generated/leakage}{}{}     %
\InputIfFileExists{generated/difficulty}{}{}  %
\InputIfFileExists{generated/harness_final}{}{}
\InputIfFileExists{generated/e4_swe}{}{}      %
\InputIfFileExists{generated/e4b_swe}{}{}     %
\InputIfFileExists{generated/e6a}{}{}         %
\InputIfFileExists{generated/e6b}{}{}         %
\InputIfFileExists{generated/e6e}{}{}         %
\InputIfFileExists{generated/e6d}{}{}         %
\InputIfFileExists{generated/e6c}{}{}         %
\InputIfFileExists{generated/e6f}{}{}         %
\providecommand{\defres}[2]{\expandafter\def\csname jevres@#1\endcsname{#2}}
\defres{recipe_abl/D10/c_acc}{97.8}
\defres{recipe_abl/D10/c_acc_offplan}{84.6}
\defres{recipe_abl/D10/c_ece}{0.034}
\defres{recipe_abl/D10/rows}{1{,}627}
\defres{recipe_abl/D10/sft_min}{60}
\defres{recipe_abl/D10/tasks}{59}
\defres{recipe_abl/D10/v1_acc}{97.0}
\defres{recipe_abl/D10/v1_ece}{0.017}
\defres{recipe_abl/D10/v1_onplan}{97.2}
\defres{recipe_abl/D10/v1_recovery}{96.2}
\defres{recipe_abl/D10/v1_recovery_tool_select}{94.2}
\defres{recipe_abl/D10/v1_stop}{99.0}
\defres{recipe_abl/D10/v1_tool_params}{97.0}
\defres{recipe_abl/D10/v1_tool_select}{95.5}
\defres{recipe_abl/D25/c_acc}{98.5}
\defres{recipe_abl/D25/c_acc_offplan}{84.9}
\defres{recipe_abl/D25/c_ece}{0.016}
\defres{recipe_abl/D25/rows}{3{,}831}
\defres{recipe_abl/D25/sft_min}{136}
\defres{recipe_abl/D25/tasks}{147}
\defres{recipe_abl/D25/v1_acc}{99.2}
\defres{recipe_abl/D25/v1_ece}{0.030}
\defres{recipe_abl/D25/v1_onplan}{99.4}
\defres{recipe_abl/D25/v1_recovery}{98.5}
\defres{recipe_abl/D25/v1_recovery_tool_select}{97.5}
\defres{recipe_abl/D25/v1_stop}{99.0}
\defres{recipe_abl/D25/v1_tool_params}{99.5}
\defres{recipe_abl/D25/v1_tool_select}{99.1}
\defres{recipe_abl/D50/c_acc}{99.1}
\defres{recipe_abl/D50/c_acc_offplan}{92.6}
\defres{recipe_abl/D50/c_ece}{0.012}
\defres{recipe_abl/D50/rows}{7{,}586}
\defres{recipe_abl/D50/sft_min}{266}
\defres{recipe_abl/D50/tasks}{293}
\defres{recipe_abl/D50/v1_acc}{99.6}
\defres{recipe_abl/D50/v1_ece}{0.028}
\defres{recipe_abl/D50/v1_onplan}{99.7}
\defres{recipe_abl/D50/v1_recovery}{99.5}
\defres{recipe_abl/D50/v1_recovery_tool_select}{99.6}
\defres{recipe_abl/D50/v1_stop}{99.3}
\defres{recipe_abl/D50/v1_tool_params}{99.8}
\defres{recipe_abl/D50/v1_tool_select}{99.8}
\defres{recipe_abl/L_gold_25/c_acc}{98.9}
\defres{recipe_abl/L_gold_25/c_acc_offplan}{89.3}
\defres{recipe_abl/L_gold_25/c_ece}{0.014}
\defres{recipe_abl/L_gold_25/rows}{3{,}831}
\defres{recipe_abl/L_gold_25/sft_min}{136}
\defres{recipe_abl/L_gold_25/tasks}{147}
\defres{recipe_abl/L_gold_25/v1_acc}{99.3}
\defres{recipe_abl/L_gold_25/v1_ece}{0.002}
\defres{recipe_abl/L_gold_25/v1_onplan}{99.5}
\defres{recipe_abl/L_gold_25/v1_recovery}{98.7}
\defres{recipe_abl/L_gold_25/v1_recovery_tool_select}{98.3}
\defres{recipe_abl/L_gold_25/v1_stop}{98.8}
\defres{recipe_abl/L_gold_25/v1_tool_params}{99.6}
\defres{recipe_abl/L_gold_25/v1_tool_select}{99.3}
\defres{recipe_abl/L_teach_25/c_acc}{93.2}
\defres{recipe_abl/L_teach_25/c_acc_offplan}{76.2}
\defres{recipe_abl/L_teach_25/c_ece}{0.059}
\defres{recipe_abl/L_teach_25/rows}{3{,}831}
\defres{recipe_abl/L_teach_25/sft_min}{136}
\defres{recipe_abl/L_teach_25/tasks}{147}
\defres{recipe_abl/L_teach_25/v1_acc}{95.5}
\defres{recipe_abl/L_teach_25/v1_ece}{0.014}
\defres{recipe_abl/L_teach_25/v1_onplan}{95.9}
\defres{recipe_abl/L_teach_25/v1_recovery}{94.6}
\defres{recipe_abl/L_teach_25/v1_recovery_tool_select}{90.5}
\defres{recipe_abl/L_teach_25/v1_stop}{98.5}
\defres{recipe_abl/L_teach_25/v1_tool_params}{96.8}
\defres{recipe_abl/L_teach_25/v1_tool_select}{92.1}
\defres{recipe_abl/S_canon_25/c_acc}{95.5}
\defres{recipe_abl/S_canon_25/c_acc_offplan}{83.1}
\defres{recipe_abl/S_canon_25/c_ece}{0.033}
\defres{recipe_abl/S_canon_25/rows}{2{,}438}
\defres{recipe_abl/S_canon_25/sft_min}{85}
\defres{recipe_abl/S_canon_25/tasks}{147}
\defres{recipe_abl/S_canon_25/v1_acc}{98.1}
\defres{recipe_abl/S_canon_25/v1_ece}{0.031}
\defres{recipe_abl/S_canon_25/v1_onplan}{98.9}
\defres{recipe_abl/S_canon_25/v1_recovery}{96.0}
\defres{recipe_abl/S_canon_25/v1_recovery_tool_select}{93.4}
\defres{recipe_abl/S_canon_25/v1_stop}{98.5}
\defres{recipe_abl/S_canon_25/v1_tool_params}{99.4}
\defres{recipe_abl/S_canon_25/v1_tool_select}{96.7}
\defres{recipe_abl/S_match_25/c_acc}{97.4}
\defres{recipe_abl/S_match_25/c_acc_offplan}{79.0}
\defres{recipe_abl/S_match_25/c_ece}{0.006}
\defres{recipe_abl/S_match_25/rows}{2{,}438}
\defres{recipe_abl/S_match_25/sft_min}{88}
\defres{recipe_abl/S_match_25/tasks}{146}
\defres{recipe_abl/S_match_25/v1_acc}{98.7}
\defres{recipe_abl/S_match_25/v1_ece}{0.032}
\defres{recipe_abl/S_match_25/v1_onplan}{99.1}
\defres{recipe_abl/S_match_25/v1_recovery}{97.4}
\defres{recipe_abl/S_match_25/v1_recovery_tool_select}{96.7}
\defres{recipe_abl/S_match_25/v1_stop}{98.7}
\defres{recipe_abl/S_match_25/v1_tool_params}{99.0}
\defres{recipe_abl/S_match_25/v1_tool_select}{98.3}
\defres{recipe_abl/contrast/anchor_vs_published/c_acc_delta}{-0.0}
\defres{recipe_abl/contrast/anchor_vs_published/c_acc_hi}{+0.0}
\defres{recipe_abl/contrast/anchor_vs_published/c_acc_lo}{-0.0}
\defres{recipe_abl/contrast/canon_vs_match_25/c_acc_delta}{-1.9}
\defres{recipe_abl/contrast/canon_vs_match_25/c_acc_hi}{-1.0}
\defres{recipe_abl/contrast/canon_vs_match_25/c_acc_lo}{-2.8}
\defres{recipe_abl/contrast/canon_vs_match_25/v1_acc_diff}{-0.5}
\defres{recipe_abl/contrast/canon_vs_match_25/v1_recovery_diff}{-1.3}
\defres{recipe_abl/contrast/canon_vs_match_25/v1_recovery_tool_select_diff}{-3.3}
\defres{recipe_abl/contrast/gold_vs_mixed_25/c_acc_delta}{+0.4}
\defres{recipe_abl/contrast/gold_vs_mixed_25/c_acc_hi}{+0.7}
\defres{recipe_abl/contrast/gold_vs_mixed_25/c_acc_lo}{+0.0}
\defres{recipe_abl/contrast/gold_vs_mixed_25/v1_acc_diff}{+0.1}
\defres{recipe_abl/contrast/gold_vs_mixed_25/v1_recovery_diff}{+0.2}
\defres{recipe_abl/contrast/gold_vs_mixed_25/v1_recovery_tool_select_diff}{+0.8}
\defres{recipe_abl/contrast/size_10_vs_50/c_acc_delta}{-1.3}
\defres{recipe_abl/contrast/size_10_vs_50/c_acc_hi}{-0.3}
\defres{recipe_abl/contrast/size_10_vs_50/c_acc_lo}{-2.1}
\defres{recipe_abl/contrast/size_10_vs_50/v1_acc_diff}{-2.7}
\defres{recipe_abl/contrast/size_10_vs_50/v1_recovery_diff}{-3.3}
\defres{recipe_abl/contrast/size_10_vs_50/v1_recovery_tool_select_diff}{-5.4}
\defres{recipe_abl/contrast/size_10_vs_anchor/c_acc_delta}{-1.3}
\defres{recipe_abl/contrast/size_10_vs_anchor/c_acc_hi}{-0.5}
\defres{recipe_abl/contrast/size_10_vs_anchor/c_acc_lo}{-2.1}
\defres{recipe_abl/contrast/size_10_vs_anchor/v1_acc_diff}{-2.7}
\defres{recipe_abl/contrast/size_10_vs_anchor/v1_recovery_diff}{-3.6}
\defres{recipe_abl/contrast/size_10_vs_anchor/v1_recovery_tool_select_diff}{-5.4}
\defres{recipe_abl/contrast/size_25_vs_50/c_acc_delta}{-0.6}
\defres{recipe_abl/contrast/size_25_vs_50/c_acc_hi}{-0.2}
\defres{recipe_abl/contrast/size_25_vs_50/c_acc_lo}{-1.0}
\defres{recipe_abl/contrast/size_25_vs_50/v1_acc_diff}{-0.4}
\defres{recipe_abl/contrast/size_25_vs_50/v1_recovery_diff}{-1.0}
\defres{recipe_abl/contrast/size_25_vs_50/v1_recovery_tool_select_diff}{-2.1}
\defres{recipe_abl/contrast/size_25_vs_anchor/c_acc_delta}{-0.6}
\defres{recipe_abl/contrast/size_25_vs_anchor/c_acc_hi}{-0.3}
\defres{recipe_abl/contrast/size_25_vs_anchor/c_acc_lo}{-1.0}
\defres{recipe_abl/contrast/size_25_vs_anchor/v1_acc_diff}{-0.5}
\defres{recipe_abl/contrast/size_25_vs_anchor/v1_recovery_diff}{-1.3}
\defres{recipe_abl/contrast/size_25_vs_anchor/v1_recovery_tool_select_diff}{-2.1}
\defres{recipe_abl/contrast/size_50_vs_anchor/c_acc_delta}{-0.1}
\defres{recipe_abl/contrast/size_50_vs_anchor/c_acc_hi}{+0.2}
\defres{recipe_abl/contrast/size_50_vs_anchor/c_acc_lo}{-0.3}
\defres{recipe_abl/contrast/size_50_vs_anchor/v1_acc_diff}{-0.1}
\defres{recipe_abl/contrast/size_50_vs_anchor/v1_recovery_diff}{-0.3}
\defres{recipe_abl/contrast/size_50_vs_anchor/v1_recovery_tool_select_diff}{+0.0}
\defres{recipe_abl/contrast/teacher_vs_mixed_25/c_acc_delta}{-5.4}
\defres{recipe_abl/contrast/teacher_vs_mixed_25/c_acc_hi}{-4.0}
\defres{recipe_abl/contrast/teacher_vs_mixed_25/c_acc_lo}{-6.6}
\defres{recipe_abl/contrast/teacher_vs_mixed_25/v1_acc_diff}{-3.7}
\defres{recipe_abl/contrast/teacher_vs_mixed_25/v1_recovery_diff}{-4.0}
\defres{recipe_abl/contrast/teacher_vs_mixed_25/v1_recovery_tool_select_diff}{-7.0}
\defres{recipe_abl/pub_base/c_acc}{50.1}
\defres{recipe_abl/pub_base/c_acc_offplan}{26.7}
\defres{recipe_abl/pub_base/c_ece}{0.160}
\defres{recipe_abl/pub_frozen/c_acc}{99.2}
\defres{recipe_abl/pub_frozen/c_acc_offplan}{94.7}
\defres{recipe_abl/pub_frozen/c_ece}{0.010}
\defres{recipe_abl/pub_jev/c_acc}{91.5}
\defres{recipe_abl/pub_jev/c_acc_offplan}{81.5}
\defres{recipe_abl/pub_jev/c_ece}{0.043}
\defres{recipe_abl/pub_sft_v1/c_acc}{99.2}
\defres{recipe_abl/pub_sft_v1/c_acc_offplan}{93.8}
\defres{recipe_abl/pub_sft_v1/c_ece}{0.011}
\defres{recipe_abl/sft_v1/c_acc}{99.2}
\defres{recipe_abl/sft_v1/c_acc_offplan}{93.7}
\defres{recipe_abl/sft_v1/c_ece}{0.011}
\defres{recipe_abl/sft_v1/v1_acc}{99.7}
\defres{recipe_abl/sft_v1/v1_ece}{0.030}
\defres{recipe_abl/sft_v1/v1_onplan}{99.7}
\defres{recipe_abl/sft_v1/v1_recovery}{99.8}
\defres{recipe_abl/sft_v1/v1_recovery_tool_select}{99.6}
\defres{recipe_abl/sft_v1/v1_stop}{99.4}
\defres{recipe_abl/sft_v1/v1_tool_params}{100.0}
\defres{recipe_abl/sft_v1/v1_tool_select}{99.7}
\providecommand{\defres}[2]{\expandafter\def\csname jevres@#1\endcsname{#2}}
\defres{e5/4b/lat_c1_p50}{317}
\defres{e5/4b/lat_c1_p90}{439}
\defres{e5/4b/lat_c1_s1_per_step_p50}{375}
\defres{e5/4b/lat_c4_p50}{876}
\defres{e5/4b/lat_c4_p90}{1{,}298}
\defres{e5/4b/lat_c4_s1_per_step_p50}{1{,}725}
\defres{e5/4b/serve_peak_gib}{21.3}
\defres{e5/4b_end/lat_c1_p50}{317}
\defres{e5/4b_end/lat_c1_p90}{440}
\defres{e5/4b_end/lat_c1_s1_per_step_p50}{375}
\defres{e5/contrast/mc1_vs_rl/c_all_delta}{-0.6}
\defres{e5/contrast/mc1_vs_rl/c_all_hi}{-0.3}
\defres{e5/contrast/mc1_vs_rl/c_all_lo}{-0.9}
\defres{e5/contrast/mc1_vs_rl/c_ask_delta}{+0.0}
\defres{e5/contrast/mc1_vs_rl/c_ask_hi}{+0.0}
\defres{e5/contrast/mc1_vs_rl/c_ask_lo}{+0.0}
\defres{e5/contrast/mc1_vs_rl/c_offplan_delta}{-2.4}
\defres{e5/contrast/mc1_vs_rl/c_offplan_hi}{-0.5}
\defres{e5/contrast/mc1_vs_rl/c_offplan_lo}{-5.0}
\defres{e5/contrast/mc1_vs_sftv1/c_all_delta}{-0.6}
\defres{e5/contrast/mc1_vs_sftv1/c_all_hi}{-0.4}
\defres{e5/contrast/mc1_vs_sftv1/c_all_lo}{-0.9}
\defres{e5/contrast/mc1_vs_sftv1/c_ask_delta}{+0.0}
\defres{e5/contrast/mc1_vs_sftv1/c_ask_hi}{+0.0}
\defres{e5/contrast/mc1_vs_sftv1/c_ask_lo}{+0.0}
\defres{e5/contrast/mc1_vs_sftv1/c_offplan_delta}{-1.5}
\defres{e5/contrast/mc1_vs_sftv1/c_offplan_hi}{+0.9}
\defres{e5/contrast/mc1_vs_sftv1/c_offplan_lo}{-4.3}
\defres{e5/contrast/q06_vs_rl/c_all_delta}{-0.1}
\defres{e5/contrast/q06_vs_rl/c_all_hi}{+0.3}
\defres{e5/contrast/q06_vs_rl/c_all_lo}{-0.4}
\defres{e5/contrast/q06_vs_rl/c_ask_delta}{+0.0}
\defres{e5/contrast/q06_vs_rl/c_ask_hi}{+0.0}
\defres{e5/contrast/q06_vs_rl/c_ask_lo}{+0.0}
\defres{e5/contrast/q06_vs_rl/c_offplan_delta}{-6.0}
\defres{e5/contrast/q06_vs_rl/c_offplan_hi}{-1.5}
\defres{e5/contrast/q06_vs_rl/c_offplan_lo}{-11.0}
\defres{e5/contrast/q06_vs_sftv1/c_all_delta}{-0.1}
\defres{e5/contrast/q06_vs_sftv1/c_all_hi}{+0.3}
\defres{e5/contrast/q06_vs_sftv1/c_all_lo}{-0.3}
\defres{e5/contrast/q06_vs_sftv1/c_ask_delta}{+0.0}
\defres{e5/contrast/q06_vs_sftv1/c_ask_hi}{+0.0}
\defres{e5/contrast/q06_vs_sftv1/c_ask_lo}{+0.0}
\defres{e5/contrast/q06_vs_sftv1/c_offplan_delta}{-5.1}
\defres{e5/contrast/q06_vs_sftv1/c_offplan_hi}{-0.3}
\defres{e5/contrast/q06_vs_sftv1/c_offplan_lo}{-10.4}
\defres{e5/contrast/q17_vs_rl/c_all_delta}{-0.3}
\defres{e5/contrast/q17_vs_rl/c_all_hi}{+0.0}
\defres{e5/contrast/q17_vs_rl/c_all_lo}{-0.7}
\defres{e5/contrast/q17_vs_rl/c_ask_delta}{+0.0}
\defres{e5/contrast/q17_vs_rl/c_ask_hi}{+0.0}
\defres{e5/contrast/q17_vs_rl/c_ask_lo}{+0.0}
\defres{e5/contrast/q17_vs_rl/c_offplan_delta}{-4.5}
\defres{e5/contrast/q17_vs_rl/c_offplan_hi}{-2.1}
\defres{e5/contrast/q17_vs_rl/c_offplan_lo}{-7.5}
\defres{e5/contrast/q17_vs_sftv1/c_all_delta}{-0.3}
\defres{e5/contrast/q17_vs_sftv1/c_all_hi}{-0.1}
\defres{e5/contrast/q17_vs_sftv1/c_all_lo}{-0.6}
\defres{e5/contrast/q17_vs_sftv1/c_ask_delta}{+0.0}
\defres{e5/contrast/q17_vs_sftv1/c_ask_hi}{+0.0}
\defres{e5/contrast/q17_vs_sftv1/c_ask_lo}{+0.0}
\defres{e5/contrast/q17_vs_sftv1/c_offplan_delta}{-3.6}
\defres{e5/contrast/q17_vs_sftv1/c_offplan_hi}{-0.7}
\defres{e5/contrast/q17_vs_sftv1/c_offplan_lo}{-6.7}
\defres{e5/live/e5_final/api_cost}{0.0127}
\defres{e5/live/e5_final/arg_cov}{80.4}
\defres{e5/live/e5_final/cost}{0.0223}
\defres{e5/live/e5_final/s1_s}{1.09}
\defres{e5/live/e5_final/s2_s}{13.00}
\defres{e5/live/e5_final/success}{100.0}
\defres{e5/live/e5_final/tool_cov}{39.0}
\defres{e5/live/e5_final/wall_s}{14.12}
\defres{e5/live/e5_final_vs_harness_final/cost_saving}{-57.8}
\defres{e5/live/e5_final_vs_harness_final/cost_saving_no_q}{-391.2}
\defres{e5/live/e5_final_vs_harness_final/success_delta}{+0.0}
\defres{e5/live/e5_final_vs_harness_final/wall_delta_s}{-2.84}
\defres{e5/live/e5_final_vs_harness_final/wall_saving}{16.8}
\defres{e5/live/e5_final_vs_llm_only/cost_saving}{-23.3}
\defres{e5/live/e5_final_vs_llm_only/cost_saving_no_q}{29.8}
\defres{e5/live/e5_final_vs_llm_only/success_delta}{+0.0}
\defres{e5/live/e5_final_vs_llm_only/wall_delta_s}{+5.66}
\defres{e5/live/e5_final_vs_llm_only/wall_saving}{-67.0}
\defres{e5/live/harness_final/api_cost}{0.0026}
\defres{e5/live/harness_final/arg_cov}{86.0}
\defres{e5/live/harness_final/cost}{0.0141}
\defres{e5/live/harness_final/s1_s}{3.90}
\defres{e5/live/harness_final/s2_s}{13.03}
\defres{e5/live/harness_final/success}{100.0}
\defres{e5/live/harness_final/tool_cov}{100.0}
\defres{e5/live/harness_final/wall_s}{16.96}
\defres{e5/live/harness_final_vs_llm_only/cost_saving}{21.9}
\defres{e5/live/harness_final_vs_llm_only/cost_saving_no_q}{85.7}
\defres{e5/live/harness_final_vs_llm_only/success_delta}{+0.0}
\defres{e5/live/harness_final_vs_llm_only/wall_delta_s}{+8.50}
\defres{e5/live/harness_final_vs_llm_only/wall_saving}{-100.6}
\defres{e5/live/lat/e5_final/all/mean}{2.07}
\defres{e5/live/lat/e5_final/all/n}{565}
\defres{e5/live/lat/e5_final/all/n_slow5}{62}
\defres{e5/live/lat/e5_final/all/p50}{1.01}
\defres{e5/live/lat/e5_final/all/p90}{5.82}
\defres{e5/live/lat/e5_final/all/slow_share}{50.7}
\defres{e5/live/lat/e5_final/tool_params/mean}{4.64}
\defres{e5/live/lat/e5_final/tool_params/n}{152}
\defres{e5/live/lat/e5_final/tool_params/n_slow5}{59}
\defres{e5/live/lat/e5_final/tool_params/p50}{1.72}
\defres{e5/live/lat/e5_final/tool_params/p90}{11.76}
\defres{e5/live/lat/e5_final/tool_params/slow_share}{80.3}
\defres{e5/live/lat/e5_final/tool_select/mean}{1.12}
\defres{e5/live/lat/e5_final/tool_select/n}{413}
\defres{e5/live/lat/e5_final/tool_select/n_slow5}{3}
\defres{e5/live/lat/e5_final/tool_select/p50}{0.97}
\defres{e5/live/lat/e5_final/tool_select/p90}{1.33}
\defres{e5/live/lat/e5_final/tool_select/slow_share}{5.7}
\defres{e5/live/lat/harness_final/all/mean}{4.36}
\defres{e5/live/lat/harness_final/all/n}{269}
\defres{e5/live/lat/harness_final/all/n_slow5}{93}
\defres{e5/live/lat/harness_final/all/p50}{1.36}
\defres{e5/live/lat/harness_final/all/p90}{11.54}
\defres{e5/live/lat/harness_final/all/slow_share}{78.5}
\defres{e5/live/lat/harness_final/tool_params/mean}{4.36}
\defres{e5/live/lat/harness_final/tool_params/n}{269}
\defres{e5/live/lat/harness_final/tool_params/n_slow5}{93}
\defres{e5/live/lat/harness_final/tool_params/p50}{1.36}
\defres{e5/live/lat/harness_final/tool_params/p90}{11.54}
\defres{e5/live/lat/harness_final/tool_params/slow_share}{78.5}
\defres{e5/live/lat/llm_only/all/mean}{1.13}
\defres{e5/live/lat/llm_only/all/n}{675}
\defres{e5/live/lat/llm_only/all/n_slow5}{5}
\defres{e5/live/lat/llm_only/all/p50}{0.98}
\defres{e5/live/lat/llm_only/all/p90}{1.37}
\defres{e5/live/lat/llm_only/all/slow_share}{6.6}
\defres{e5/live/lat/llm_only/tool_select/mean}{1.13}
\defres{e5/live/lat/llm_only/tool_select/n}{675}
\defres{e5/live/lat/llm_only/tool_select/n_slow5}{5}
\defres{e5/live/lat/llm_only/tool_select/p50}{0.98}
\defres{e5/live/lat/llm_only/tool_select/p90}{1.37}
\defres{e5/live/lat/llm_only/tool_select/slow_share}{6.6}
\defres{e5/live/llm_only/api_cost}{0.0181}
\defres{e5/live/llm_only/cost}{0.0181}
\defres{e5/live/llm_only/s1_s}{0.00}
\defres{e5/live/llm_only/s2_s}{8.44}
\defres{e5/live/llm_only/success}{100.0}
\defres{e5/live/llm_only/tool_cov}{0.0}
\defres{e5/live/llm_only/wall_s}{8.46}
\defres{e5/live/n}{90}
\defres{e5/live/q_usd_per_h}{2.448}
\defres{e5/mc1/c_acc}{98.6}
\defres{e5/mc1/c_acc_ask}{100.0}
\defres{e5/mc1/c_acc_offplan}{92.3}
\defres{e5/mc1/c_ece}{0.017}
\defres{e5/mc1/lat_c1_p50}{107}
\defres{e5/mc1/lat_c1_p90}{127}
\defres{e5/mc1/lat_c1_s1_per_step_p50}{141}
\defres{e5/mc1/lat_c4_p50}{174}
\defres{e5/mc1/lat_c4_p90}{319}
\defres{e5/mc1/lat_c4_s1_per_step_p50}{327}
\defres{e5/mc1/serve_peak_gib}{4.4}
\defres{e5/mc1/sft_min}{101}
\defres{e5/mc1/speedup_c1_p50}{2.95}
\defres{e5/mc1/speedup_c1_p90}{3.45}
\defres{e5/mc1/speedup_c4_p50}{5.03}
\defres{e5/mc1/speedup_c4_p90}{4.06}
\defres{e5/mc1/train_peak_gib}{9.5}
\defres{e5/mc1/v1_acc}{99.28}
\defres{e5/mc1/v1_ece}{0.034}
\defres{e5/mc1/v1_knockout}{100.00}
\defres{e5/mc1/v1_recovery}{99.01}
\defres{e5/mc1/v1_stop}{98.95}
\defres{e5/mc1/v1_tool_params}{99.88}
\defres{e5/mc1/v1_tool_select}{98.98}
\defres{e5/q06/c_acc}{99.1}
\defres{e5/q06/c_acc_ask}{100.0}
\defres{e5/q06/c_acc_offplan}{88.7}
\defres{e5/q06/c_ece}{0.011}
\defres{e5/q06/lat_c1_p50}{127}
\defres{e5/q06/lat_c1_p90}{159}
\defres{e5/q06/lat_c1_s1_per_step_p50}{171}
\defres{e5/q06/lat_c4_p50}{245}
\defres{e5/q06/lat_c4_p90}{448}
\defres{e5/q06/lat_c4_s1_per_step_p50}{465}
\defres{e5/q06/serve_peak_gib}{11.7}
\defres{e5/q06/sft_min}{119}
\defres{e5/q06/speedup_c1_p50}{2.50}
\defres{e5/q06/speedup_c1_p90}{2.77}
\defres{e5/q06/speedup_c4_p50}{3.57}
\defres{e5/q06/speedup_c4_p90}{2.90}
\defres{e5/q06/train_peak_gib}{8.5}
\defres{e5/q06/v1_acc}{99.79}
\defres{e5/q06/v1_ece}{0.032}
\defres{e5/q06/v1_knockout}{100.00}
\defres{e5/q06/v1_recovery}{99.67}
\defres{e5/q06/v1_stop}{99.70}
\defres{e5/q06/v1_tool_params}{100.00}
\defres{e5/q06/v1_tool_select}{99.66}
\defres{e5/q17/c_acc}{98.9}
\defres{e5/q17/c_acc_ask}{100.0}
\defres{e5/q17/c_acc_offplan}{90.2}
\defres{e5/q17/c_ece}{0.016}
\defres{e5/q17/lat_c1_p50}{172}
\defres{e5/q17/lat_c1_p90}{215}
\defres{e5/q17/lat_c1_s1_per_step_p50}{221}
\defres{e5/q17/lat_c4_p50}{360}
\defres{e5/q17/lat_c4_p90}{603}
\defres{e5/q17/lat_c4_s1_per_step_p50}{697}
\defres{e5/q17/serve_peak_gib}{14.1}
\defres{e5/q17/sft_min}{190}
\defres{e5/q17/speedup_c1_p50}{1.84}
\defres{e5/q17/speedup_c1_p90}{2.04}
\defres{e5/q17/speedup_c4_p50}{2.44}
\defres{e5/q17/speedup_c4_p90}{2.15}
\defres{e5/q17/train_peak_gib}{14.4}
\defres{e5/q17/v1_acc}{99.58}
\defres{e5/q17/v1_ece}{0.029}
\defres{e5/q17/v1_knockout}{100.00}
\defres{e5/q17/v1_recovery}{99.67}
\defres{e5/q17/v1_stop}{98.95}
\defres{e5/q17/v1_tool_params}{100.00}
\defres{e5/q17/v1_tool_select}{99.66}
\defres{e5/rl/c_acc}{99.2}
\defres{e5/rl/c_acc_ask}{100.0}
\defres{e5/rl/c_acc_offplan}{94.7}
\defres{e5/rl/c_ece}{0.010}
\defres{e5/rl/v1_acc}{99.71}
\defres{e5/rl/v1_ece}{0.004}
\defres{e5/rl/v1_knockout}{99.54}
\defres{e5/rl/v1_recovery}{99.83}
\defres{e5/rl/v1_stop}{99.40}
\defres{e5/rl/v1_tool_params}{100.00}
\defres{e5/rl/v1_tool_select}{99.66}
\defres{e5/sftv1/c_acc}{99.2}
\defres{e5/sftv1/c_acc_ask}{100.0}
\defres{e5/sftv1/c_acc_offplan}{93.8}
\defres{e5/sftv1/c_ece}{0.011}
\defres{e5/sftv1/v1_acc}{99.71}
\defres{e5/sftv1/v1_ece}{0.029}
\defres{e5/sftv1/v1_knockout}{99.85}
\defres{e5/sftv1/v1_recovery}{99.83}
\defres{e5/sftv1/v1_stop}{99.40}
\defres{e5/sftv1/v1_tool_params}{100.00}
\defres{e5/sftv1/v1_tool_select}{99.66}
\InputIfFileExists{generated/e7_hard}{}{}
\InputIfFileExists{generated/e7b_hard}{}{}    %
\InputIfFileExists{generated/e7c_cal}{}{}     %
\InputIfFileExists{generated/e7c_hard}{}{}    %
\InputIfFileExists{generated/e7e_hard}{}{}    %
\InputIfFileExists{generated/e7d_tier}{}{}    %
\InputIfFileExists{generated/e7f_hard}{}{}    %
\InputIfFileExists{generated/e11_alfworld}{}{}%
\InputIfFileExists{generated/e23_alfworld_unseen}{}{}%
\InputIfFileExists{generated/e21_shift}{}{}  %
\InputIfFileExists{generated/e21b_shift_cascade}{}{}%
\InputIfFileExists{generated/e24_public_l2}{}{}%
\InputIfFileExists{generated/e25_harnesses}{}{}%
\InputIfFileExists{generated/e27_small_shift}{}{}%
\InputIfFileExists{generated/e28_sl08b}{}{}%
\InputIfFileExists{generated/e29_port}{}{}%
\InputIfFileExists{generated/e29b_route}{}{}%
\InputIfFileExists{generated/e30_cost}{}{}%
\InputIfFileExists{generated/e31_rl_only}{}{}%
\InputIfFileExists{generated/module_map}{}{}%
\InputIfFileExists{generated/e22_tau2}{}{}%
\InputIfFileExists{generated/hard_ni}{}{}    %
\InputIfFileExists{generated/e9_frontier}{}{}%
\InputIfFileExists{generated/e8_gates}{}{}    %
\providecommand{\defres}[2]{\expandafter\def\csname jevres@#1\endcsname{#2}}
\defres{checker/eval/n_success}{17{,}120}
\defres{checker/eval/embedded_only}{0}
\defres{checker/eval/flagged_totals}{2}
\defres{checker/eval/second_total}{0}
\defres{checker/eval/out_of_order}{11}
\defres{checker/eval/out_of_order_pct}{0.06}
\defres{checker/hard/writes}{9{,}636}
\defres{checker/hard/unbound_writes}{0}
\defres{checker/hard/label_states}{9{,}650}
\defres{checker/hard/label_unready}{0}
\InputIfFileExists{generated/coverage}{}{}    %
\InputIfFileExists{generated/e4r_swe}{}{}     %
\InputIfFileExists{generated/e12_train}{}{}   %
\InputIfFileExists{generated/e13_llms}{}{}    %
\InputIfFileExists{generated/e14_small}{}{}   %
\InputIfFileExists{generated/e15_baselines}{}{}%
\InputIfFileExists{generated/drift}{}{}        %
\InputIfFileExists{generated/template_baseline}{}{}%
\InputIfFileExists{generated/e16_best_open}{}{}%
\InputIfFileExists{generated/e19_hard_random}{}{}%
\InputIfFileExists{generated/e17_e18}{}{}     %
\InputIfFileExists{generated/e20_ensembles}{}{}%

\ifdefined\jevinterim\else\def\jevinterim{0}\fi
\ifnum\jevinterim=1
\usepackage{eso-pic}
\AddToShipoutPictureFG{\AtPageUpperLeft{\put(\LenToUnit{0.5\paperwidth},\LenToUnit{-1.25cm}){\makebox[0pt]{%
  \colorbox{red!8}{\color{red!65!black}\sffamily\footnotesize\textbf{Interim draft (4 Oct 2026):} $\tau^2$ reference-agreement replay
  pending.}}}}}
\fi

\ifdefined\hypersetup
\ifarxiv
  \hypersetup{pdfauthor={Kunlun Zhu, Tao Feng, Xu Han, Heng Wang, Ling Yang, Xiangru Tang, Jiaxuan You, Heng Ji},
    pdfsubject={}, pdfkeywords={},
    pdftitle={JevAgentHarness: Benchmarking, Deploying and Training Decision Models That Share Control with LLMs}}
\else
  \hypersetup{pdfauthor={}, pdfsubject={}, pdfkeywords={},
    pdftitle={JevAgentHarness: Benchmarking, Deploying and Training Decision Models That Share Control with LLMs}}
\fi
\fi
\ifdefined\pdfsuppressptexinfo \pdfsuppressptexinfo=-1 \fi

\title{JevAgentHarness: Benchmarking, Deploying and Training \\
Decision Models That Share Control with LLMs}

\ifarxiv
\author{%
  \textbf{Kunlun Zhu\textsuperscript{1}},
  \textbf{Tao Feng\textsuperscript{1}},
  \textbf{Xu Han\textsuperscript{1}},
  \textbf{Heng Wang\textsuperscript{1}},
  \\
  \textbf{Ling Yang\textsuperscript{2}},
  \textbf{Xiangru Tang\textsuperscript{3}},
  \textbf{Jiaxuan You\textsuperscript{1}},
  \textbf{Heng Ji\textsuperscript{1}}
  \\
  \normalsize
  \textsuperscript{1}University of Illinois Urbana-Champaign\quad
  \textsuperscript{2}Princeton University\quad
  \textsuperscript{3}Google
}
\else
\author{Anonymous ACL submission}
\fi

\begin{document}
\maketitle

\begin{abstract}
Large language model (LLM) agents now fix software, operate computers and serve customers, but their harnesses, the
programs that run them, call a large LLM at every step, which makes long tasks slow and costly. Most of these calls end
in a multiple-choice decision, such as which tool to call next. Decision models such as \jev{} answer these questions
directly, at a fraction of an LLM call's time and price. We ask how a harness can delegate these decisions to a
decision model with little loss of success. To evaluate decision models in agentic tasks, we build \benchname{}, which
tests them from isolated questions up to control of a running agent, on public benchmarks and our \gym{} environment.
The decision model most accurate on isolated questions solved only \keyB{\res{e21/sl9b/l3/success}\%} of agent tasks,
and without a fitted confidence threshold, delegating tool choices lowered success by
\resabs{mm/cell/flash/tool_select/replace/delta} points. Building on these findings, we then design \harnessname{},
which fits this threshold for each choice on the agent's own trajectories and lets the LLM decide below it. Because
existing decision models never see the states an agent reaches after its own mistakes, we train \ours{}, an open 4B
model, on such states with \jev{}'s answers as labels. On tasks unlike its training data, the agent solved
\keyM{\res{e21/e12/l3/success}\%} of them with \ours{} and \res{e21/jev/l3/success}\% with \jev{}. On tasks like their
training data, 4B models trained this way made agents \keyH{\res{stage2/S2-4/cost_savings}--\res{e11/cascade/saving}\%}
cheaper and up to \keyA{\fpeval{round(\csname jevres@e21b/e12_in/ref_wall\endcsname/\csname jevres@e21b/e12_in/wall\endcsname,1)} times} faster, without losing 3 or more points of task success.
\ifarxiv The models, data, and harness will be released at
\url{https://github.com/JevAgentHarness/JevAgentHarness}.
\else We will release the benchmark, harness, and the models.\fi
\end{abstract}

\definecolor{rkNavy}{HTML}{2B3A55}   \definecolor{rkGrey}{HTML}{5C6476}   \definecolor{rkRule}{HTML}{DCE1E9}
\definecolor{rkBlue}{HTML}{EDF3FC}   \definecolor{rkBlueD}{HTML}{D9E6F8}
\definecolor{rkGreen}{HTML}{EEF7F0}  \definecolor{rkGreenD}{HTML}{D8EDDF}
\definecolor{rkYellow}{HTML}{FFF7E2} \definecolor{rkYellowD}{HTML}{FBEAC2}
\definecolor{rkPink}{HTML}{FDEDED}   \definecolor{rkPinkD}{HTML}{F7DADA}
\definecolor{rkPeach}{HTML}{FEF0E4}  \definecolor{rkPeachD}{HTML}{F9DCC4}
\definecolor{rkBeige}{HTML}{F7F1E6}  \definecolor{rkBeigeD}{HTML}{E9DCC4}
\definecolor{rkRed}{HTML}{B5483B}    \definecolor{rkOk}{HTML}{2F8152}
\definecolor{rkOchre}{HTML}{C28E2E}  \definecolor{rkSage}{HTML}{6C9878}
\providecommand{\rkT}{\fontsize{8.5}{10}\selectfont\rmfamily\bfseries\itshape}   %
\providecommand{\rkB}{\fontsize{7}{8.3}\selectfont\rmfamily}                     %
\providecommand{\rkI}{\fontsize{7}{8.3}\selectfont\rmfamily\itshape}             %
\providecommand{\rkN}{\fontsize{13}{14}\selectfont\rmfamily\bfseries}            %
\providecommand{\rkIcon}[2]{\includegraphics[height=#1mm]{figures/art/r14_#2}}
\providecommand{\rkIconK}[3]{\includegraphics[width=#1mm,height=#2mm,keepaspectratio]{figures/art/r14_#3}}
\providecommand{\rkWord}[2]{%
  \ifcsname jevres@#1\endcsname
    \ifnum\pdfmatch{^not (tested|run)}{\csname jevres@#1\endcsname}=1 \textcolor{red}{\textbf{?}}%
    \else\ifnum\pdfmatch{^(NI|superior|rejects)([^a-z]|$)}{\csname jevres@#1\endcsname}=1
      \ifnum\pdfmatch{(^|[^a-z])not([^a-z]|$)}{#2}=1 \textcolor{red}{\textbf{?}}\else #2\fi
    \else\ifnum\pdfmatch{(^|[^a-z])not([^a-z]|$)}{\csname jevres@#1\endcsname}=1
      \ifnum\pdfmatch{(^|[^a-z])not([^a-z]|$)}{#2}=1 #2\else\textcolor{red}{\textbf{?}}\fi
    \else \textcolor{red}{\textbf{?}}\fi\fi\fi
  \else \textcolor{red}{\textbf{?}}\fi}
\providecommand{\rkSign}[1]{%
  \ifcsname jevres@#1\endcsname
    \ifnum\pdfmatch{^not (tested|run)}{\csname jevres@#1\endcsname}=1 2%
    \else\ifnum\pdfmatch{^(NI|superior|rejects)([^a-z]|$)}{\csname jevres@#1\endcsname}=1 1%
    \else\ifnum\pdfmatch{(^|[^a-z])not([^a-z]|$)}{\csname jevres@#1\endcsname}=1 0%
    \else 2\fi\fi\fi
  \else 2\fi}
\tikzset{
  rkframe/.style={draw=rkRule, line width=0.5pt, fill=white, rounded corners=3mm},
  rkchip/.style={draw=rkRule, line width=0.45pt, fill=white, rounded corners=1.3mm},
  rkarr/.style={-{Stealth[length=1.25mm,width=1.0mm]}, line width=0.5pt, draw=rkNavy!70},
  rkdash/.style={rkarr, dash pattern=on 1.2pt off 0.9pt},
  rkshadow/.style={blur shadow={shadow blur steps=6, shadow xshift=0.15mm, shadow yshift=-0.25mm,
    shadow opacity=14, shadow blur radius=0.6mm}},
}
\providecommand{\rkTile}[6]{%
  \shade[rounded corners=1.8mm, inner color=#4, outer color=#4D]
    ({#1-#3/2},{#2-#3/2}) rectangle ({#1+#3/2},{#2+#3/2});
  \node at (#1,#2) {\rkIconK{#6}{#6}{#5}};}
\providecommand{\rkOkC}[3]{%
  \fill[white] (#1,#2) circle ({#3+0.35});  \fill[rkOk] (#1,#2) circle (#3);
  \draw[white, line width=0.75pt, line cap=round, line join=round]
    ({#1-0.48*#3},{#2+0.02*#3}) -- ({#1-0.12*#3},{#2-0.34*#3}) -- ({#1+0.48*#3},{#2+0.36*#3});}
\providecommand{\rkNoC}[3]{%
  \fill[white] (#1,#2) circle ({#3+0.35});  \fill[rkRed] (#1,#2) circle (#3);
  \draw[white, line width=0.75pt, line cap=round]
    ({#1-0.36*#3},{#2-0.36*#3}) -- ({#1+0.36*#3},{#2+0.36*#3}) ({#1-0.36*#3},{#2+0.36*#3}) -- ({#1+0.36*#3},{#2-0.36*#3});}
\providecommand{\rkQC}[3]{%
  \fill[white] (#1,#2) circle ({#3+0.35});  \fill[red] (#1,#2) circle (#3);
  \node[text=white, font=\rkB\bfseries] at (#1,#2) {?};}
\providecommand{\rkMark}[4]{%
  \ifcase\rkSign{#4}\relax \rkNoC{#1}{#2}{#3}\or \rkOkC{#1}{#2}{#3}\else \rkQC{#1}{#2}{#3}\fi}
\providecommand{\rkStep}[3]{%
  \fill[white] (4.3,#1) circle (2.75);  \fill[#2] (4.3,#1) circle (2.3);
  \node[text=white, font=\fontsize{8.5}{9}\selectfont\rmfamily\bfseries\itshape] at (4.3,#1) {#3};}

\providecommand{\fiImg}[4]{\node at (#1,#2) {\includegraphics[height=#3mm]{figures/art/#4}};}
\providecommand{\fiTile}[6]{%
  \shade[rounded corners=1.6mm, inner color=#4, outer color=#4D] ({#1-#3/2},{#2-#3/2}) rectangle ({#1+#3/2},{#2+#3/2});
  \fiImg{#1}{#2}{#6}{#5}}
\definecolor{fiBar}{HTML}{6F9BD9}   %

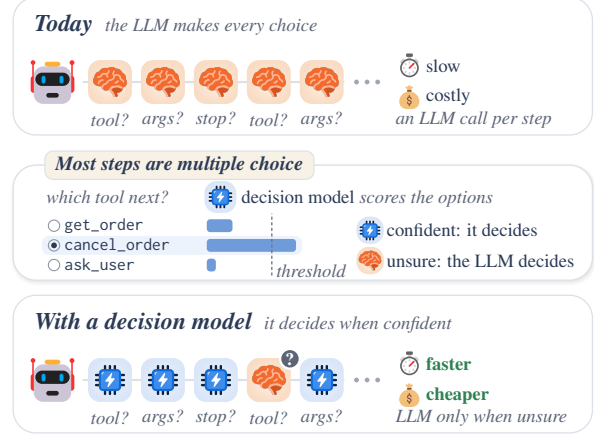
\begin{figure}[t]
\centering
\begin{tikzpicture}[jevfig]
\tikzset{
  fiHead/.style={font=\fontsize{9}{10}\selectfont\rmfamily\bfseries\itshape, text=rkNavy, anchor=base west},
  fiSub/.style={font=\rkI, text=rkGrey, anchor=base west},
  fiQ/.style={font=\rkI, text=rkGrey},
  fiOpt/.style={font=\fontsize{6.6}{7}\selectfont\ttfamily, text=rkNavy, anchor=west},
  fiTxt/.style={font=\rkB, text=rkNavy, anchor=west},
}
\newcommand{\fiLane}[1]{%
  \fiImg{5.4}{#1}{6.6}{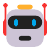}
  \draw[rkRule, line width=0.8pt] (9.4,#1) -- (44.0,#1);
  \foreach \x in {45.6,47.0,48.4} \fill[rkNavy!45] (\x,#1) circle (0.32);
  \foreach \l/\x in {tool?/13.0, args?/20.0, stop?/27.0, tool?/34.0, args?/41.0}
    \node[fiQ] at (\x,#1-5.0) {\l};}
\newcommand{\fiStepL}[2]{\fiTile{#1}{#2}{5.8}{rkPeach}{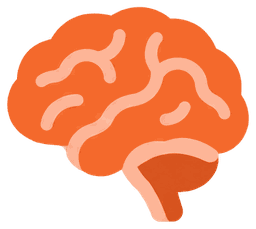}{4.1}}
\newcommand{\fiStepD}[2]{\fiTile{#1}{#2}{5.8}{rkBlue}{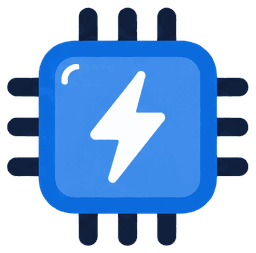}{4.1}}
\newcommand{\fiEnd}[4]{%
  \fiImg{52.6}{#1+2.2}{3.6}{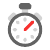}  \node[fiTxt] at (54.8,#1+2.2) {#2};
  \fiImg{52.6}{#1-2.0}{3.6}{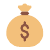}  \node[fiTxt] at (54.8,#1-2.0) {#3};
  \node[fiQ, anchor=west] at (50.8,#1-5.0) {#4};}

\draw[rkframe] (0.2,39.6) rectangle (76.8,57.8);
\node[fiHead] (h1) at (3.0,53.4) {Today};
\node[fiSub] at ([xshift=1.6mm]h1.base east) {the LLM makes every choice};
\fiLane{46.6}
\foreach \x in {13.0,20.0,27.0,34.0,41.0} {\fiStepL{\x}{46.6}}
\fiEnd{46.6}{slow}{costly}{an LLM call per step}

\draw[rkframe, rkshadow] (0.2,20.4) rectangle (76.8,35.6);
\node[draw=rkRule, line width=0.45pt, fill=rkBeige, rounded corners=1.1mm, inner xsep=1.4mm, inner ysep=0.7mm,
  font=\fontsize{7.5}{8.6}\selectfont\rmfamily\bfseries\itshape, text=rkNavy, anchor=west] at (4.4,35.6)
  {Most steps are multiple choice};
\fill[rkBlue, rounded corners=0.9mm] (4.0,23.75) rectangle (38.4,26.25);
\node[fiQ, anchor=base west] at (4.6,30.6) {which tool next?};
\foreach \o/\y in {get\_order/27.6, cancel\_order/25.0, ask\_user/22.4} {
  \draw[rkNavy!55, line width=0.4pt] (5.6,\y) circle (0.75);
  \node[fiOpt] at (7.0,\y) {\o};}
\fill[rkNavy] (5.6,25.0) circle (0.38);
\fiTile{27.6}{31.4}{4.2}{rkBlue}{r12_dmodel.png}{3.1}
\node[fiTxt, anchor=base west] at (30.4,30.6) {decision model \textcolor{rkGrey}{\textit{scores the options}}};
\foreach \y/\w in {27.6/3.4, 25.0/11.8, 22.4/1.2}
  \fill[fiBar, rounded corners=0.45mm] (25.8,\y-0.85) rectangle (25.8+\w,\y+0.85);
\draw[rkNavy!70, line width=0.5pt, dash pattern=on 1.2pt off 0.9pt] (34.4,21.0) -- (34.4,29.2);
\node[fiQ, anchor=west] at (35.0,21.7) {threshold};
\fiTile{47.4}{27.0}{3.6}{rkBlue}{r12_dmodel.png}{2.7}
\node[fiTxt] at (49.6,27.0) {confident: it decides};
\fiTile{47.4}{23.0}{3.6}{rkPeach}{r12_llm.png}{2.7}
\node[fiTxt] at (49.6,23.0) {unsure: the LLM decides};

\draw[rkframe] (0.2,0.2) rectangle (76.8,18.4);
\node[fiHead] (h2) at (3.0,14.0) {With a decision model};
\node[fiSub] at ([xshift=1.6mm]h2.base east) {it decides when confident};
\fiLane{7.2}
\foreach \x in {13.0,20.0,27.0,41.0} {\fiStepD{\x}{7.2}}
\fiStepL{34.0}{7.2}
\fill[white] (36.8,10.0) circle (1.45); \fill[rkGrey] (36.8,10.0) circle (1.15);
\node[text=white, font=\fontsize{6.6}{6.6}\selectfont\rmfamily\bfseries] at (36.8,10.0) {?};
\fiEnd{7.2}{\textcolor{rkOk}{\textbf{faster}}}{\textcolor{rkOk}{\textbf{cheaper}}}{LLM only when unsure}
\end{tikzpicture}
\caption{\textbf{Intuition of \harnessname{}.} Most of an agent's LLM calls end in a multiple-choice decision, and a small
decision model can make the routine ones in a fraction of the time and price. The LLM makes the choices the decision model is unsure of.}
\label{fig:teaser}
\end{figure}
\section{Introduction}
\label{sec:intro}

Large language models have become capable agents. Recent frontier models such as GPT-6 Astra
\citep{openai2026gpt6astra} and Claude Opus 5.5 \citep{anthropic2026opus55} report their largest gains on agentic
work: they resolve issues in real code repositories \citep{jimenez2024swebench}, operate desktop and web applications
\citep{xie2024osworld}, complete tasks in a terminal
\citep{merrill2026terminalbench} and serve customers under domain policies \citep{barres2025tau2}. The length of the
software tasks that frontier models can complete has doubled about every seven months since 2019
\citep{kwa2025timehorizon}. These capabilities are put to use through agent \emph{harnesses} such as Claude Code, Codex
and OpenHands \citep{anthropic2026claudecode,codexcli2026,wang2025openhands}, programs that give the model tools,
manage its context and memory, and decide when to stop \citep{yao2023react,yang2024sweagent,guo2026harnesssurvey}.

\input{figures/fig2_pipeline}%

Yet agents remain slow and costly on long tasks, because most harnesses call the large model at every step
\citep{kapoor2025agentsthatmatter,wang2025efficientagents}. Evaluating agents for one public leaderboard cost about
\$40{,}000 \citep{kapoor2026hal}. Most of these calls, however, end in a multiple-choice decision. In
Figure~\ref{fig:teaser}, a customer-service agent must either look up an order, cancel it or ask the customer.

\emph{Decision models} such as \jev{} \citep{typesafe2026jev} and its open counterparts
\citep{interndecision2026,startlux2026decision,openjev2026} make such choices directly. Given the task, the history and
one multiple-choice question, a decision model returns a probability for each listed option instead of generating
text, much like the fast, intuitive System~1 rather than the slow, deliberate System~2 \citep{kahneman2011thinking}.
In one recorded episode, a call to \jev{} took
\resratio{fig1/latency/s1}{fig1/latency/s2}{0}\% of an LLM call's time and \resratio{fig1/cost/s1}{fig1/cost/s2}{1}\% of
its price. Existing systems use decision models for game actions, memory, escalation and
judging \citep{ma2026jevstar,jiang2026jevmem,wu2026reflex,li2026jevjudge}. Because a harness lists the actions at each step, it can \emph{delegate} a choice to a decision model when that model is confident and otherwise \emph{defer} to the LLM (Figure~\ref{fig:teaser}). We call this \emph{decision modeling} for agents \citep{chen2025decisionflow}.

Decision models, however, are built and tested on fixed questions and do not adapt to the agent they serve.
Whether one works inside an agent has not been tested. Routers and cascades \citep{chen2024frugalgpt,ong2025routellm},
small function-calling models \citep{zhang2025xlam,lin2024hammer} and fast--slow agents
\citep{lin2023swiftsage,christakopoulou2024talkerreasoner} also move work away from the large model, and they too have
been tested on fixed questions or on a single kind of choice, and agent benchmarks score only whole tasks. Inside an
agent, a decision model's mistakes carry over to later steps \citep{zhu2025agentdebug}, and one choice changes the next. It also faces states that follow the agent's own errors, which fixed question sets rarely contain, and every query to it adds latency. We therefore ask:

\noindent\begin{tikzpicture}
\node[draw=sOne, line width=0.6pt, fill=sOneT, rounded corners=2pt, inner xsep=2.0mm, inner ysep=1.6mm,
  text width=\columnwidth-4.0mm-1.2pt, align=left] {\textit{How can an agent harness delegate routine choices to a
  fast, cheap decision model when the model is reliable, so that agentic tasks become faster and cheaper with little loss of
  success?}};
\end{tikzpicture}

\input{figures/fig_where_how}%
\noindent Our hypothesis is that a decision model helps an agent when it is chosen by tests inside the agent, used only when it is
confident, and trained on the states that the agent itself reaches. We build one part for each of these three
conditions (Figure~\ref{fig:pipeline}). \textbf{\benchname{}} tests decision models at three levels, each closer to
real use: on isolated questions with no agent around, at each step of recorded agent runs, where answers are scored but
not carried out (step level), and in control of a running agent, where each choice changes what comes next (task
level). We use public benchmarks and \gym{},
an environment we built that knows the acceptable next steps at every state, so that each choice can be scored.
\textbf{\harnessname{}} lets an existing harness put each routine choice to a decision model as a typed multiple-choice question and uses the answer only when the model's
confidence clears a threshold fitted on the agent's own trajectories. \textbf{\ours{}} is an open 4B decision model that
learns from \jev{}'s answers on the states the agent visits. We will release all three.
\textbf{\ours{} chooses better than \jev{} inside the agent:} on tasks unlike its training data,
the agent solved \keyM{\res{e21/e12/l3/success}\%} of them when \ours{} made its tool choices and
\res{e21/jev/l3/success}\% when \jev{} made them.
\textbf{\harnessname{} makes agents cheaper and faster:} with 4B models trained on each agent's states, customer-service and
household agents
were \keyH{\res{stage2/S2-4/cost_savings}--\res{e11/cascade/saving}\%} cheaper and up to
\keyA{\fpeval{round(\csname jevres@e21b/e12_in/ref_wall\endcsname/\csname jevres@e21b/e12_in/wall\endcsname,1)} times}
faster, without losing 3 or more points of success.

\section{Where and How to Delegate}
\label{sec:design}
Whether a choice is worth handing to the decision model depends on which choice it is and on how the decision model's
answer is combined with the LLM's. We define the task (Figure~\ref{fig:wherehow}) and derive a rule for what to delegate
(\S\ref{sec:design:theory}). The setup applies to any harness that can list its options at a step.

\InputIfFileExists{generated/module_map}{}{}
\paragraph{Decision points.}\label{sec:design:defs}\label{sec:design:where}\label{sec:design:how}
An episode is one agent run on a task, and its state $s$ holds what the agent knows at a step. A \emph{decision
point} $d$ is a kind of choice that recurs across steps and whose options $O_d(s)$ the harness can list. We study eight
decision points (Figure~\ref{fig:wherehow:a}). Three choose what the agent does: the \emph{tool}, its \emph{arguments} and, without tools, the next
\emph{action}. Five control the loop: whether to \emph{write memory}, \emph{reflect} after a failure, \emph{stop}, or
\emph{route} the step to a cheaper LLM, and which stored notes to \emph{read back}.

\paragraph{Decision models.}\label{sec:background}
A decision model reads $s$ as text with one question and its options (App.~\ref{app:prompts}) and returns a
probability $p(o \mid s)$ for each option. Its answer is the likeliest option, and its \emph{confidence} $c(s)$ is
that option's probability (\jev{} \citep{typesafe2026jev} returns this probability directly). Open models like Intern-Decision and \ours{} score all options in one forward pass.

\paragraph{Who decides.}
An \emph{integration mode} fixes who proposes an answer and who has the last word (Figure~\ref{fig:wherehow:b}). We
mostly use the \emph{cascade} (\figmki{Cas}\,\modeT{\modeCas}): the decision model answers first, and the harness uses it when its confidence clears a threshold, $c(s) \ge \tau_d$. Otherwise the LLM decides as usual. In the other modes, the
decision model always decides (\figmki{Rep}\,\modeT{\modeRep}), shortlists options for the LLM (\figmki{Ass}\,\modeT{\modeAss}), approves or rejects the LLM's choice (\figmki{Gua}\,\modeT{\modeGua}), or decides whether an optional step such as reflection runs (\figmki{Tri}\,\modeT{\modeTri}). We implement all \res{mm/count/flash/pairs} point--mode combinations (App.~\ref{app:theory}). A decision point with its mode and threshold is a \emph{delegation}, such as the tool-choice cascade. We seek the cheapest set of delegations whose success stays within 3 points of the LLM alone.

\paragraph{A decision-making example.}
Consider a customer-service agent asked to cancel an order (App.~Figure~\ref{fig:decision}). At each step,
the harness asks which of \res{fig1/n_options} tools to call. At step~\res{fig1/defer/step}, \jev{} split its vote between asking the customer
(\res{fig1/defer/opt/1/p}) and cancelling (\res{fig1/defer/opt/2/p}), so the LLM decided and asked. At
step~\res{fig1/accept/step}, \jev{} was confident about cancelling the order. \jev{} then chose the order from options copied from the chat and the reason from a fixed list. Both answers were confident, so the call ran
without the LLM. An unsure answer or a free-text argument would cost one LLM call.

\subsection{Deciding what to delegate}
\label{sec:design:theory}\label{sec:design:bp}
\paragraph{A threshold rule.}
Handing a choice to the decision model can save an LLM call but may lead to a wrong choice. If its confidence equals its chance of being right (App.~\ref{app:theory}), handing over is worth it exactly when $c(s) \ge q_L - \Delta C / V$, where
$q_L$ is how often the LLM gets this choice right, $\Delta C$ the price of the saved call and $V$ the cost of a wrong
choice. This value is
the best threshold $\tau_d$. The decision model may thus be a little less accurate than the LLM as long as the
saving covers the extra risk.

\paragraph{Fitting thresholds.}
We first run the agent on separate tasks (150 per agent LLM) while the decision model answers every question silently (\emph{shadow
mode}). From these trajectories we estimate, per decision point, how often the LLM was right ($q_L$), a call's price ($\Delta C$) and $V$, the drop in task success after one wrong choice, valued at \$1 per solved task. The trajectories also give
the decision model's accuracy at each confidence, so the threshold is the lowest confidence above which accepting is worth it,
and the confidence need not match its exact chance of being right (App.~\ref{app:theory}). If none qualifies, the LLM keeps deciding. With \jev{} and Flash, the fitted thresholds were \res{calibration/c1/jev/flash/tool_select/tauhat} for tools and \res{calibration/c1/jev/flash/tool_params/tauhat} for arguments, so only near-certain answers are used.

\section{\benchname{}}
\label{sec:bench}
A decision model can score well on a test and still fail inside an agent. Existing tests ask single questions with
fixed answers, so a wrong answer never affects what happens next. Agent benchmarks run whole tasks but report only
whether a task succeeded, without showing which choice went wrong. \benchname{} tests decision models at three levels
on public data and on \gym{}, with every call priced and timed (App.~\ref{app:cost}, Table~\ref{tab:suites}).

\paragraph{L1: isolated questions.}
The decision model answers single questions with no agent around it. We turn the \res{diff/e1/bfcl_tool_select/n} tool-choice and \res{diff/e1/bfcl_arguments/n} argument items of BFCL
v4 \citep{patil2025bfcl} and the \res{diff/e1/when2call/n} When2Call items \citep{ross2025when2call} into multiple-choice questions.

\paragraph{L2: step level, on recorded agent runs.}
The decision model answers at each step of recorded agent runs, but its answer is only scored, so the run continues as recorded and one mistake cannot cause
the next, as when driving policies are tested on recorded drives. Each answer is checked against an
\emph{answer key} of acceptable next steps. All models answer the same \res{cov/ask/all/occurrences} states of \res{stage2/S2-3/corpus/tasks} new \gym{} tasks (the \emph{common states}).

\paragraph{L3: task level, in control of the agent.}
The decision model now makes every tool choice of a running agent (every action in ALFWorld) with no threshold, while
the LLM writes the arguments, and each choice is carried out. A wrong choice changes
what the agent sees next, so mistakes can pile up. This level measures task success. It is the hardest level and the closest to deployment. Our agent loop runs $\tau^2$-bench \citep{barres2025tau2}, ALFWorld \citep{shridhar2021alfworld} and \gym{} tasks,
and hooks let us test existing harnesses without changing their code: mini-SWE-agent and SWE-agent on SWE-bench Verified
\citep{jimenez2024swebench}, and the pi terminal agent on Terminal-Bench 2.0 and Terminal-Bench-Science \citep{merrill2026terminalbench,tbscience2026}.

Levels 2 and 3 need an environment that knows which next steps are acceptable at every state an agent can reach,
including after its own mistakes. Public agent benchmarks check only the final state or the progress of a run
\citep{lu2025toolsandbox,ma2024agentboard}, have a fixed set of tasks and are slow and costly to rerun. We therefore built
\gym{}, a simulated customer service desk for a small shop, with typed tools over a small database as in $\tau$-bench
\citep{yao2025taubench} and requests from templates as in TextWorld \citep{cote2018textworld}. Every task offers the same 68 tools plus \emph{finish} in shuffled order: seven core tools, one that asks the customer, 31 near-duplicates of core tools and 29 unrelated ones. Because each task comes
from a scripted plan, \gym{} derives an answer key at every reachable state from which the task can still be solved, even after detours. Training, threshold
fitting and testing use disjoint random seeds. Near-duplicate distractor tools and \emph{must-ask states}, where the right move is to ask the customer, make the tasks hard. The answer key is hidden from all models, and in \emph{strict} mode any change to the shop's data that is not in the plan ends the episode (App.~\ref{app:jevgym}). Two further task sets, added after the main study, test the models beyond their training data. \emph{Shifted tasks} (\gym{}-Shift, App.~\ref{app:e21}) rename the tools and change the
wording and kinds of requests. \emph{Hard tasks} (\gym{}-Hard, App.~\ref{app:e7}) have \res{e7/setting/multi_customers} or \res{e7b/setting/multi_customers} customers each and a new rule for cancelling or shipping.

\section{\harnessname{}}
\label{sec:harness}
\harnessname{} moves a harness's routine choices to a decision model. It poses each listed choice as a multiple-choice question and uses the answer only when the model is confident enough.
Otherwise the LLM decides as usual. The harness has three parts (App.~Figure~\ref{fig:harness}).

\paragraph{A typed decision interface.} Every choice uses \jev{}'s input format, whose options are the tools,
fixed-list or yes/no argument values, or text spans to copy from the context. A tool call is split into a tool choice
and one choice per such argument, so the decision model can pick both while the LLM writes only free text. Any model that accepts this format can be used (App.~\ref{app:harness}).

\paragraph{A confidence threshold fitted on trajectories.} Each decision point gets the threshold of
\S\ref{sec:design:theory}, fitted on shadow-mode trajectories, so the decision model takes only choices it makes reliably. Below the threshold the LLM decides, so an unsure answer wastes one cheap decision-model call but does not risk the task. Thresholds are refitted when the LLM, decision model or task rules change.

\paragraph{Adapters, serving and accounting.} Adapters ask the same questions inside existing harnesses through their hooks: strong-or-fast LLM routing in mini-SWE-agent, guards and triggers in pi, and command
hooks for Codex CLI and Claude Code. If an adapter or the decision model fails, the harness runs as before
(App.~Table~\ref{tab:affordances}).

\section{\ours{}}
\label{sec:rl}\label{sec:training}
We train our own decision model for two reasons. An API model such as \jev{} is billed per call and cannot learn a harness's own
rules, and off-the-shelf models learn from fixed questions and successful demonstrations. Such data rarely contain the
states an agent reaches after its own mistakes, where a decision model must also answer (App.~\ref{app:devdiag}).
We therefore post-train a small open model, Qwen3-4B with LoRA
\citep{yang2025qwen3,hu2022lora}, on states the harness visits: first by supervised fine-tuning (SFT),
then by RL in the agent loop.

\paragraph{Supervised fine-tuning.}
We run the harness on \res{recipe/data/v1/tasks} training tasks and record every decision state $s$,
\res{recipe/data/v1/rows} in all. To include states after errors, a scripted policy repeats mistakes seen during development \citep{ross2011dagger,laskey2017dart}, and \res{recipe/data/v1/after_error_share}\% of the states follow a mistake (App.~\ref{app:rl}). Where the
teacher, \jev{}, ranks the correct answer $a^\star$ first, the target mixes the one-hot label of $a^\star$ with the teacher's probabilities,
$y_a = \alpha\,\mathbb{1}[a = a^\star] + (1-\alpha)\,p_{\jev{}}(a \mid s)$ with $\alpha = 0.3$. Otherwise the target is the one-hot label alone. SFT fits this target with cross-entropy plus a squared-error (Brier) term that keeps the probabilities calibrated:
\begin{equation}
\mathcal{L}_{\text{SFT}} = \mathbb{E}_{s}\textstyle\sum_{a}\bigl[-y_a \log p_a + \mu\,(p_a - y_a)^{2}\bigr],
\label{eq:sft}
\end{equation}
with $p_a = p_\theta(a \mid s)$ and $\mu = 0.5$. We use LoRA ($r{=}32$), 2 epochs and learning rate $10^{-4}$.

\paragraph{GRPO inside the agent loop.}
The SFT model then decides on training tasks while a fixed agent LLM does the rest, so it learns from the
consequences of its own choices. For each task we run a group of $G = 8$ episodes. Episode $i$ earns a reward for success, $S_i \in \{0,1\}$, reduced for the $N_i$ LLM calls it used, minus a penalty when it made an irreversible wrong call ($E_i = 1$):
\begin{equation}
R_i = S_i\bigl(1 - \lambda \min(1, N_i/N_0)\bigr) - \eta E_i,
\label{eq:reward}
\end{equation}
with $\lambda = 0.5$, $N_0 = 10$ and $\eta = 0.2$, so a solved task with ten or more LLM calls earns half. Each episode's advantage is its reward relative to the group, $\hat A_i = (R_i - \bar
R)/\sigma_R$, and every decision $a_t$ in the episode shares it in the GRPO objective \citep{shao2024deepseekmath}, plus a step-level term at states that recur within the group \citep{feng2025gigpo}, with a KL penalty to the SFT model (App.~\ref{app:rl}). Flash-Lite is the agent LLM, and iteration 15 is kept.

\paragraph{Evaluation.}
Stage~2 swaps \jev{} for the frozen \ours{} and refits the thresholds. We later trained it further on public tool-use data.

\section{Experimental Protocol}
\label{sec:setup}
\emph{Stage~1} tests each delegation alone and keeps those that are cheaper and \emph{non-inferior} to the LLM alone. Non-inferior means that a one-sided test at the 5\% level, on the same tasks run both ways, shows a loss of less than 3 points in task success (5 on $\tau^2$-bench; App.~\ref{app:metrics}). \emph{Stage~2} tests the passing delegations together on new tasks, because delegations can interact. We \emph{pre-registered} the models, tasks and tests, and registered the Stage-2 plan before reading any Stage-1 result, so that choosing the combination could not bias its test.

\paragraph{Stage 1: each delegation alone.}
\label{sec:setup:stage1}
The decision model is \jev{} (and Intern-Decision-4B for three delegations). The agent LLM is
Gemini~3.5 Flash (\emph{Flash}) with native tool calls. The cheaper, weaker Gemini~3.5 Flash-Lite (\emph{Flash-Lite})
receives routed steps, simulates the $\tau^2$-bench user and is a second agent LLM on \gym{}
(App.~\ref{app:cost}). The LLM alone and simple controls without a decision model run on the same tasks as each delegation (App.~\ref{app:metagold}). With eleven registered tests, a few could pass by chance, so Holm's method \citep{holm1979simple} makes each test stricter and keeps the chance of any wrong pass at 5\%. Other results are descriptive.

\paragraph{Stage 2: the delegations together.}
\label{sec:setup:stage2}
A script chose the combination by a rule fixed in advance. At each decision point it took the delegation that passed Stage~1 with the largest saving, or kept the LLM if none passed (App.~\ref{app:stage2}). We ran the
combination on \res{stage2/S2-1/n_pairs} new \gym{} tasks and on $\tau^2$-bench telecom tasks. Four
Holm-corrected tests check two things. The first two check that the combination still works. It should be
non-inferior to the LLM alone at lower cost, and cheaper than its best single delegation without losing success, or combining would not be worth it. The other two check that a model we train can take \jev{}'s place. On the common states,
\ours{} should beat Intern-Decision-4B, a model of its size that never saw the agent's
states, and be non-inferior to its teacher \jev{}. The combination should also stay non-inferior to the LLM alone
when \ours{} replaces \jev{}.

\section{Results}
\label{sec:results}
\csname @secpenalty\endcsname=0

\definecolor{rkNavy}{HTML}{2B3A55}  \definecolor{rkGrey}{HTML}{5C6476}  \definecolor{rkOk}{HTML}{2F8152}
\definecolor{rkRed}{HTML}{B5483B}
\definecolor{krHoldF}{HTML}{DCEFE1} \definecolor{krNotF}{HTML}{FBE1DF} \definecolor{krDescF}{HTML}{F3EDE2}
\definecolor{krRefF}{HTML}{E5EEFA}  \definecolor{krBand}{HTML}{F3F6FB}
\newcommand{\krPill}[3]{\tikz[baseline=(krp.base)]\node[fill=#1, text=#2, rounded corners=1.7pt, inner xsep=2.4pt,
  inner ysep=0.75pt, text height=1.55ex, text depth=0.4ex, font=\footnotesize](krp){#3};}
\makeatletter
\newcommand{\krSign}[1]{%
  \ifcsname jevres@#1\endcsname
    \ifnum\pdfmatch{^not (tested|run)}{\csname jevres@#1\endcsname}=1 2%
    \else\ifnum\pdfmatch{^(NI|superior|rejects)([^a-z]|$)}{\csname jevres@#1\endcsname}=1 1%
    \else\ifnum\pdfmatch{(^|[^a-z])not([^a-z]|$)}{\csname jevres@#1\endcsname}=1 0%
    \else 2\fi\fi\fi
  \else 2\fi}
\makeatother
\newcommand{\krOut}[1]{\ifcase\krSign{#1}\relax \krPill{krNotF}{rkRed}{\resword{#1}}%
  \or \krPill{krHoldF}{rkOk}{\resword{#1}}\else \resword{#1}\fi}
\newcommand{\krDesc}{\krPill{krDescF}{rkGrey}{descriptive}}
\newcommand{\krExp}{\krPill{krDescF}{rkGrey}{exploratory}}
\newcommand{\krRef}{\krPill{krRefF}{rkNavy}{reference}}
\newcommand{\krSave}[1]{\textcolor{rkOk}{#1}}%
\newcommand{\krID}[1]{{\color{rkGrey}\footnotesize #1}}%
\newcommand{\krBandRow}[1]{\rowcolor{krBand}\multicolumn{9}{@{\hspace{1.2mm}}l}{\rule{0pt}{2.4ex}\textit{#1}}\\}
\newsavebox\krMkLo \newsavebox\krMkCas \newsavebox\krMkPr \newsavebox\krMkRl
\sbox\krMkLo{\tikz[jevfig, baseline=-0.75mm]{\node[fmk=sTwo, diamond, minimum size=2.3mm] {};}}
\sbox\krMkCas{\tikz[jevfig, baseline=-0.75mm]{\node[fmCas=sOne] at (0,-0.3) {};}}
\sbox\krMkPr{\tikz[jevfig, baseline=-0.75mm]{\node[fmk=sOne, circle, minimum size=2.0mm] {};}}
\sbox\krMkRl{\tikz[jevfig, baseline=-0.75mm]{\node[circle, draw=sOne, line width=0.8pt, fill=white,
  minimum size=2.2mm] {}; \node[fmk=sOne, circle, minimum size=0.85mm] {};}}
\newcommand{\krMk}[1]{\makebox[3.2mm][l]{\usebox{#1}}}
\newcommand{\krMkNone}{\hspace*{3.2mm}}

\makeatletter
\newcommand{\mmref}{\@ifundefined{r@fig:modulemap}{\@ifundefined{r@fig:module-map}{\@ifundefined{r@fig:module_map}%
  {App.~Table~\ref{tab:tests}}{Figure~\ref{fig:module_map}}}{Figure~\ref{fig:module-map}}}{Figure~\ref{fig:modulemap}}}
\makeatother
\newcommand{\resans}[1]{\noindent\textbf{#1}}

\definecolor{kbEdge}{HTML}{2F8152}  \definecolor{kbFill}{HTML}{EEF7F0}%
\expandafter\ifx\csname ifmmDetail\endcsname\relax\expandafter\newif\csname ifmmDetail\endcsname\fi
\ifdefined\mmMapLoaded\else\def\mmMapLoaded{1}%
\InputIfFileExists{generated/module_map}{}{}%
\definecolor{mmNavy}{HTML}{2B3A55}   \definecolor{mmGrey}{HTML}{5C6476}   \definecolor{mmRule}{HTML}{DCE1E9}
\definecolor{mmBlue}{HTML}{E8F0FB}   \definecolor{mmBlueD}{HTML}{C8D9F1}
\definecolor{mmGreen}{HTML}{E8F4EA}  \definecolor{mmGreenD}{HTML}{BCDCC4}  \definecolor{mmGreenS}{HTML}{5FA874}
\definecolor{mmYellow}{HTML}{FFF3D6} \definecolor{mmYellowD}{HTML}{EBD39A} \definecolor{mmYellowS}{HTML}{D9A23A}
\definecolor{mmPink}{HTML}{FBE5E5}   \definecolor{mmPinkD}{HTML}{E8BDBD}   \definecolor{mmPinkS}{HTML}{CC6B6B}
\definecolor{mmNR}{HTML}{ECEEF2}     \definecolor{mmNRD}{HTML}{D2D7DF}     \definecolor{mmNA}{HTML}{FAFBFC}
\definecolor{mmOk}{HTML}{2F8152}     \definecolor{mmRed}{HTML}{B5483B}
\newcommand{\mmB}{\fontsize{7}{8.2}\selectfont\rmfamily}                  %
\newcommand{\mmI}{\fontsize{7}{8.2}\selectfont\rmfamily\itshape}          %
\newcommand{\mmH}{\fontsize{7.5}{8.6}\selectfont\rmfamily}                %
\newcommand{\mmCode}[1]{\ifcsname jevres@#1\endcsname\csname jevres@#1\endcsname\else 8\fi}
\newcommand{\mmFillOf}[1]{\ifcase#1 mmNR\or mmGreen\or mmYellow\or mmYellow\or mmPink\or mmNR\or mmNR\or mmNR\or
  mmNR\or mmNA\else mmNR\fi}
\newcommand{\mmDrawOf}[1]{\ifcase#1 mmNRD\or mmGreenD\or mmYellowD\or mmYellowD\or mmPinkD\or mmNRD\or mmNRD\or
  mmNRD\or mmNRD\or mmRule\else mmNRD\fi}
\newcommand{\mmStrongOf}[1]{\ifcase#1 mmNRD\or mmGreenS\or mmYellowS\or mmYellowS\or mmPinkS\else mmNRD\fi}
\newcommand{\mmNum}[1]{\ifcsname jevres@#1\endcsname
  \expandafter\expandafter\expandafter\mmNumA\csname jevres@#1\endcsname\relax\else\tbd\fi}
\def\mmNumA#1#2\relax{\if-#1\textminus#2\else#1#2\fi}
\newcommand{\mmSign}[1]{%
  \ifcsname jevres@#1\endcsname
    \ifnum\pdfmatch{^not (tested|run)}{\csname jevres@#1\endcsname}=1 2%
    \else\ifnum\pdfmatch{^(NI|superior|rejects)([^a-z]|$)}{\csname jevres@#1\endcsname}=1 1%
    \else\ifnum\pdfmatch{(^|[^a-z])not([^a-z]|$)}{\csname jevres@#1\endcsname}=1 0%
    \else 2\fi\fi\fi
  \else 2\fi}
\newcommand{\mmBadgeCode}[4]{%
  \fill[white] (#1,#2) circle ({#3+0.25});
  \ifcase#4\relax
    \fill[mmRed] (#1,#2) circle (#3);
    \draw[white, line width=0.55pt, line cap=round]
      ({#1-0.38*#3},{#2-0.38*#3}) -- ({#1+0.38*#3},{#2+0.38*#3}) ({#1-0.38*#3},{#2+0.38*#3}) -- ({#1+0.38*#3},{#2-0.38*#3});
  \or
    \fill[mmOk] (#1,#2) circle (#3);
    \draw[white, line width=0.55pt, line cap=round, line join=round]
      ({#1-0.48*#3},{#2+0.02*#3}) -- ({#1-0.12*#3},{#2-0.36*#3}) -- ({#1+0.48*#3},{#2+0.38*#3});
  \else
    \fill[mmGrey] (#1,#2) circle (#3);
    \node[text=white, font=\fontsize{5}{5}\selectfont\bfseries] at (#1,#2) {?};
  \fi}
\newcommand{\mmBadge}[4]{\edef\mmSg{\mmSign{#4}}\mmBadgeCode{#1}{#2}{#3}{\mmSg}}
\newcommand{\mmTile}[7]{%
  \edef\mmC{\mmCode{mm/cell/#1/#2/#3/vcode}}%
  \ifnum\mmC=0 \ifcsname jevres@mm/xcell/#1/#2/#3/vcode\endcsname \def\mmC{-1}\fi\fi
  \ifnum\mmC=9 \ifcsname jevres@mm/xcell/#1/#2/#3/vcode\endcsname \def\mmC{-1}\fi\fi
  \ifnum\mmC=-1 \ifmmDetail\mmXTile{#1}{#2}{#3}{#4}{#5}{#6}{#7}\else\mmXPlain{#1}{#2}{#3}{#4}{#5}{#6}{#7}\fi
  \else\mmTileStage{#1}{#2}{#3}{#4}{#5}{#6}{#7}\fi}
\newcommand{\mmXPlain}[7]{%
  \edef\mmXC{\csname jevres@mm/xcell/#1/#2/#3/vcode\endcsname}%
  \edef\mmXF{\mmFillOf{\mmXC}}\edef\mmXD{\mmDrawOf{\mmXC}}%
  \draw[draw=\mmXD, fill=\mmXF, line width=0.45pt, rounded corners=0.9mm] (#4,#5) rectangle ({#4+#6},{#5+#7});
  \node[font=\mmB, text=mmNavy, anchor=base] at ({#4+#6/2},{#5+#7/2-0.95})
    {\mmNum{mm/xcell/#1/#2/#3/fig_delta}\hspace{0.45mm}{\color{mmGrey!70}\rule[-0.25mm]{0.3pt}{2.3mm}}\hspace{0.45mm}%
     \mmNum{mm/xcell/#1/#2/#3/fig_saving}};}
\newcommand{\mmXTile}[7]{%
  \edef\mmXC{\csname jevres@mm/xcell/#1/#2/#3/vcode\endcsname}%
  \edef\mmXF{\mmFillOf{\mmXC}}\edef\mmXS{\mmStrongOf{\mmXC}}%
  \fill[\mmXF, rounded corners=0.9mm] (#4,#5) rectangle ({#4+#6},{#5+#7});
  \begin{scope}
    \clip[rounded corners=0.9mm] (#4,#5) rectangle ({#4+#6},{#5+#7});
    \foreach \mmh in {0,1.3,...,18.2} {\draw[white, line width=0.55pt] ({#4+\mmh-#7},#5) -- ({#4+\mmh},{#5+#7});}
  \end{scope}
  \draw[draw=\mmXS, line width=0.45pt, dash pattern=on 1.1pt off 0.7pt, rounded corners=0.9mm]
    (#4,#5) rectangle ({#4+#6},{#5+#7});
  \ifcsname jevres@mm/xcell/#1/#2/#3/equiv\endcsname
    \node[font=\mmB, text=mmNavy, anchor=base, fill=\mmXF, rounded corners=0.4mm, inner xsep=0.2mm, inner ysep=0.25mm]
      at ({#4+#6/2},{#5+0.55})
      {\mmNum{mm/xcell/#1/#2/#3/fig_delta}\hspace{0.45mm}{\color{mmGrey!70}\rule[-0.25mm]{0.3pt}{2.3mm}}\hspace{0.45mm}%
       \mmNum{mm/xcell/#1/#2/#3/fig_saving}};
    \node[font=\mmI, text=mmGrey, anchor=base west, fill=\mmXF, rounded corners=0.4mm, inner xsep=0.25mm,
      inner ysep=0.1mm] at ({#4+0.5},{#5+2.8})
      {\edef\mmEq{\csname jevres@mm/xcell/#1/#2/#3/equiv\endcsname}=\,\mmEqShort{\mmEq}};
  \else
    \node[font=\mmB, text=mmNavy, anchor=base, fill=\mmXF, rounded corners=0.4mm, inner xsep=0.2mm, inner ysep=0.25mm]
      at ({#4+#6/2},{#5+#7/2-0.95})
      {\mmNum{mm/xcell/#1/#2/#3/fig_delta}\hspace{0.45mm}{\color{mmGrey!70}\rule[-0.25mm]{0.3pt}{2.3mm}}\hspace{0.45mm}%
       \mmNum{mm/xcell/#1/#2/#3/fig_saving}};
  \fi
  \mmXGlyph{#4+#6-1.15}{#5+#7-0.4}}
\newcommand{\mmEqShort}[1]{\ifnum\pdfstrcmp{#1}{trigger}=0 trig.\else\ifnum\pdfstrcmp{#1}{cascade}=0 casc.\else#1\fi\fi}
\newcommand{\mmXGlyph}[2]{%
  \draw[draw=white, fill=white, line width=0.5pt] ({#1},{#2+0.95}) -- ({#1+0.95},{#2}) -- ({#1},{#2-0.95}) -- ({#1-0.95},{#2}) -- cycle;
  \draw[draw=mmNavy, fill=white, line width=0.4pt] ({#1},{#2+0.7}) -- ({#1+0.7},{#2}) -- ({#1},{#2-0.7}) -- ({#1-0.7},{#2}) -- cycle;}
\newcommand{\mmUndefMark}[5]{%
  \ifmmDetail\else\node[font=\mmB, text=mmGrey!80] at ({#4},{#5}) {--};\fi
  \ifmmDetail\ifcsname jevres@mm/cell/#1/#2/#3/equiv\endcsname
    \node[font=\mmI, text=mmGrey] at ({#4},{#5}) {=\,\csname jevres@mm/cell/#1/#2/#3/equiv\endcsname};
  \else\ifcsname jevres@mm/cell/#1/#2/#3/undef\endcsname
    \node[font=\mmB, text=mmGrey!80] at ({#4},{#5}) {--};
  \fi\fi\fi}
\newcommand{\mmTileStage}[7]{%
  \edef\mmC{\mmCode{mm/cell/#1/#2/#3/vcode}}%
  \edef\mmF{\mmFillOf{\mmC}}\edef\mmD{\mmDrawOf{\mmC}}%
  \ifnum\mmC=9
    \draw[draw=mmRule, fill=mmNA, line width=0.35pt, dash pattern=on 0.9pt off 0.7pt, rounded corners=0.9mm]
      (#4,#5) rectangle ({#4+#6},{#5+#7});
    \mmUndefMark{#1}{#2}{#3}{#4+#6/2}{#5+#7/2}%
  \else
    \draw[draw=\mmD, fill=\mmF, line width=0.45pt, rounded corners=0.9mm] (#4,#5) rectangle ({#4+#6},{#5+#7});
  \fi
  \ifnum\mmC=0
    \node[font=\mmI, text=mmGrey] at ({#4+#6/2},{#5+#7/2}) {not run};
  \fi
  \ifnum\mmC=8
    \node[font=\mmB] at ({#4+#6/2},{#5+#7/2}) {\tbd};
  \fi
  \ifnum\mmC>0 \ifnum\mmC<5
    \ifmmDetail\ifcsname jevres@mm/cell/#1/#2/#3/alt_vcode\endcsname
      \edef\mmS{\mmStrongOf{\csname jevres@mm/cell/#1/#2/#3/alt_vcode\endcsname}}%
      \begin{scope}
        \clip[rounded corners=0.9mm] (#4,#5) rectangle ({#4+#6},{#5+#7});
        \fill[\mmS] ({#4+#6},{#5}) -- ({#4+#6-2.1},{#5}) -- ({#4+#6},{#5+2.1}) -- cycle;
      \end{scope}
    \fi\fi
    \node[font=\mmB, text=mmNavy, anchor=base] at ({#4+#6/2},{#5+#7/2-0.95})
      {\mmNum{mm/cell/#1/#2/#3/fig_delta}\hspace{0.45mm}{\color{mmGrey!70}\rule[-0.25mm]{0.3pt}{2.3mm}}\hspace{0.45mm}%
       \mmNum{mm/cell/#1/#2/#3/fig_saving}};
    \ifmmDetail\ifcsname jevres@mm/cell/#1/#2/#3/reg\endcsname
      \edef\mmR{\csname jevres@mm/cell/#1/#2/#3/reg\endcsname}%
      \mmBadge{#4+#6-1.15}{#5+#7-0.4}{0.75}{primaries/\mmR/decision}
    \fi\fi
  \fi\fi}
\newcommand{\mmRowLabel}[1]{\ifcase#1 Tool choice\or Tool args.\ifmmDetail$^{a}$\fi\or Memory write\or
  Memory read\ifmmDetail$^{a}$\fi\or Reflection\or Stop\or Routing\ifmmDetail$^{b}$\fi\or Action\fi}
\newcount\mmUndefCount
\newcommand{\mmCountUndef}[1]{\global\mmUndefCount=0
  \foreach \mmdp in {tool_select, tool_params, memory_write, memory_read, reflection, stop, routing, action} {
    \foreach \m in {replace, cascade, assist, guard, trigger} {
      \edef\mmCc{\mmCode{mm/cell/#1/\mmdp/\m/vcode}}%
      \ifnum\mmCc=9 \ifcsname jevres@mm/xcell/#1/\mmdp/\m/vcode\endcsname\else
        \ifcsname jevres@mm/cell/#1/\mmdp/\m/equiv\endcsname\else\global\advance\mmUndefCount by 1 \fi\fi\fi}}}
\newcommand{\mmMap}[1]{%
  \mmCountUndef{#1}%
  \begin{tikzpicture}[jevfig]
    \ifmmDetail\useasboundingbox (0,-2.6) rectangle (77,53.3);\else\useasboundingbox (0,5.2) rectangle (77,53.3);\fi
    \foreach \m/\h [count=\j from 0] in {replace/\modeRep, cascade/\modeCas, assist/\modeAss, guard/\modeGua,
        trigger/\modeTri} {
      \fill[mmBlue, rounded corners=0.9mm] ({15.3+\j*12.32},50.3) rectangle ({15.3+\j*12.32+11.8},53.3);
      \node[font=\mmH, text=mmNavy, anchor=base] at ({15.3+\j*12.32+5.9},51.1) {\h};}
    \node[font=\mmI, text=mmGrey, anchor=base west] at (0,51.1) {$\Delta$\,pt\,{\color{mmGrey!70}|}\,saved\,\%};
    \foreach \mmdp [count=\i from 0] in {tool_select, tool_params, memory_write, memory_read, reflection, stop,
        routing, action} {
      \pgfmathsetmacro\yb{45.25-\i*5.15}
      \node[font=\mmB, text=mmNavy, anchor=east] at (14.85,{\yb+2.3}) {\mmRowLabel{\i}};
      \foreach \m [count=\j from 0] in {replace, cascade, assist, guard, trigger} {
        \mmTile{#1}{\mmdp}{\m}{15.3+\j*12.32}{\yb}{11.8}{4.6}}}
    \begin{scope}[font=\mmB, text=mmNavy]
      \foreach \mmcol/\mmlab/\x in {mmGreen/{kept success, cheaper}/0.2, mmYellow/{small loss or no saving}/27.6,
          mmPink/{lost 3+ points}/57.0} {
        \draw[draw=\mmcol D, fill=\mmcol, line width=0.4pt, rounded corners=0.5mm] (\x,5.7) rectangle ({\x+3.0},7.7);
        \node[anchor=base west] at ({\x+3.8},6.0) {\mmlab};}
      \def\mmArgLlm{#1}\def\mmFlashLlm{flash}%
      \ifx\mmArgLlm\mmFlashLlm\else
      \draw[draw=mmNRD, fill=mmNR, line width=0.4pt, rounded corners=0.5mm] (65.0,5.7) rectangle (68.0,7.7);
      \node[anchor=base west] at (68.8,6.0) {not run};
      \fi
      \ifmmDetail
      \mmBadgeCode{0.95}{3.9}{0.75}{1}
      \mmBadgeCode{3.25}{3.9}{0.75}{0}
      \node[anchor=base west] at (4.6,3.2) {planned test: passed / failed};
      \fill[mmYellowS] (40.0,2.9) -- (42.2,2.9) -- (42.2,5.1) -- cycle;
      \node[anchor=base west] at (43.0,3.2) {a repeat run disagreed};
      \ifnum\mmUndefCount>0
      \draw[draw=mmRule, fill=mmNA, line width=0.35pt, dash pattern=on 0.9pt off 0.7pt, rounded corners=0.5mm]
        (0.2,-2.6) rectangle (3.2,-0.6);
      \node[text=mmGrey!80] at (1.7,-1.6) {--};
      \node[anchor=base west] at (4.0,-2.3) {undefined};
      \fi
      \node[font=\mmI, text=mmGrey, anchor=base west] at (18.2,-2.3) {=\,mode};
      \node[anchor=base west] at (26.0,-2.3) {same as that mode};
      \mmXGlyph{52.0}{-1.6}
      \node[anchor=base west] at (53.6,-2.3) {added later$^{\dagger}$};
      \node[anchor=base west] at (0.2,0.4) {$^{a}$With tool choice or memory write};
      \node[anchor=base west] at (40.6,0.4) {$^{b}$vs.\ Flash-Lite, routing to Flash};
      \fi
    \end{scope}
  \end{tikzpicture}}
\fi
\input{figures/fig_frontier}%
\newif\ifefLive
\makeatletter
\gdef\efArm{}
\@for\ef@a:=retrained,rl_tier,rl_refit,tier,jevagent_best_rl\do{%
  \ifefLive\else\ifcsname jevres@e7f/\ef@a/success\endcsname\global\efLivetrue\xdef\efArm{\ef@a}\fi\fi}
\makeatother
\newif\ifebLive
\makeatletter
\gdef\ebArm{}
\@for\eb@a:=rl,trained,retrained,ours,rl_alf,sft\do{%
  \ifebLive\else\ifcsname jevres@e11/\eb@a/success\endcsname\global\ebLivetrue\xdef\ebArm{\eb@a}\fi\fi}
\makeatother
\definecolor{hdOurs}{HTML}{7D58B0}%
\newsavebox\hsMkJ \newsavebox\hsMkO \newsavebox\hsMkN
\sbox\hsMkN{\tikz[jevfig, baseline=-0.75mm]{\node[circle, draw=envC!70, line width=0.5pt, fill=white, minimum size=2.0mm] {};}}
\sbox\hsMkJ{\tikz[jevfig, baseline=-0.75mm]{\node[fmk=sOne, circle, minimum size=2.0mm] {};}}
\sbox\hsMkO{\tikz[jevfig, baseline=-0.75mm]{\node[circle, draw=hdOurs, line width=0.8pt, fill=white,
  minimum size=2.2mm] {}; \node[fmk=hdOurs, circle, minimum size=0.85mm] {};}}
\newsavebox\fittobox
\providecommand{\fitto}[2]{\sbox\fittobox{#2}\ifdim\wd\fittobox>#1\resizebox{#1}{!}{\usebox\fittobox}\else\usebox\fittobox\fi}
\newcommand{\hsBand}[1]{\rowcolor{krBand}\multicolumn{8}{@{\hspace{1.2mm}}l}{\rule{0pt}{2.4ex}\textit{#1}}\\}
\newcommand{\hsOut}[1]{\ifcsname jevres@hardni/#1/ni\endcsname
  \ifnum\pdfstrcmp{\csname jevres@hardni/#1/ni\endcsname}{yes}=0 \krPill{krHoldF}{rkOk}{non-inferior\dg}%
  \else\krPill{krDescF}{rkGrey}{not shown\dg}\fi\else\krPill{krDescF}{rkGrey}{\dg}\fi}
\newcommand{\hsOutKey}[1]{\ifcsname jevres@#1\endcsname
  \ifnum\pdfstrcmp{\csname jevres@#1\endcsname}{yes}=0 \krPill{krHoldF}{rkOk}{non-inferior\dg}%
  \else\krPill{krDescF}{rkGrey}{not shown\dg}\fi\else\krPill{krDescF}{rkGrey}{\dg}\fi}
\newcommand{\krPillS}[3]{\tikz[baseline=(krp.base)]\node[fill=#1, text=#2, rounded corners=1.5pt, inner xsep=1.4pt,
  inner ysep=0.5pt, text height=1.35ex, text depth=0.35ex, font=\scriptsize](krp){#3};}
\newcommand{\krOutS}[1]{\ifcase\krSign{#1}\relax \krPillS{krNotF}{rkRed}{\resword{#1}}%
  \or \krPillS{krHoldF}{rkOk}{\resword{#1}}\else \resword{#1}\fi}
\newcommand{\krDescS}{\krPillS{krDescF}{rkGrey}{descriptive}}
\newcommand{\hsOutS}[1]{\ifcsname jevres@hardni/#1/ni\endcsname
  \ifnum\pdfstrcmp{\csname jevres@hardni/#1/ni\endcsname}{yes}=0 \krPillS{krHoldF}{rkOk}{non-inferior\dg}%
  \else\krPillS{krDescF}{rkGrey}{not shown\dg}\fi\else\krPillS{krDescF}{rkGrey}{\dg}\fi}
\newcommand{\krNumOut}[2]{\ifcase\krSign{#1}\relax \krPillS{krNotF}{rkRed}{#2}\or \krPillS{krHoldF}{rkOk}{#2}\else
  #2\fi}
\newcommand{\krNumNI}[2]{\ifcsname jevres@#1\endcsname\ifnum\pdfstrcmp{\csname jevres@#1\endcsname}{yes}=0
  \krPillS{krHoldF}{rkOk}{#2}\else\krPillS{krDescF}{rkGrey}{#2}\fi\else #2\fi}
\makeatletter
\newcommand{\tbNum}[1]{\ifcsname jevres@#1\endcsname\mmNum{#1}\else--\fi}
\newcommand{\tbSave}[1]{\ifcsname jevres@#1\endcsname
  \ifnum\pdfmatch{^[$]?-}{\csname jevres@#1\endcsname}=1 \mmNum{#1}\else\krSave{\mmNum{#1}}\fi\else--\fi}
\makeatother
\newcommand{\tbBand}[2]{\rowcolor{krBand}\multicolumn{#1}{@{\hspace{1.0mm}}l}{\rule{0pt}{2.2ex}\textit{#2}}\\}
\ExplSyntaxOn
\cs_new_protected:Npn \__jevhs_setnum:Nn #1#2
  { \tl_gset:Nx #1 { \use:c { jevres@#2 } } \tl_gremove_all:Nn #1 { $ } \tl_gremove_all:Nn #1 { + } }
\NewDocumentCommand{\hsSetNum}{mm}{ \__jevhs_setnum:Nn #1 {#2} }
\ExplSyntaxOff
\makeatletter
\newcommand{\hsPair}[3]{\gdef\hs@d{0}\ifcsname jevres@#1\endcsname\ifcsname jevres@#2\endcsname
  \hsSetNum\hs@a{#1}\hsSetNum\hs@b{#2}\xdef\hs@d{\fpeval{(#3)*((\hs@b)-(\hs@a))}}\fi\fi
  \ifdim\hs@d pt>0pt \resna{#1}\else\textbf{\resna{#1}}\fi & \ifdim\hs@d pt<0pt \resna{#2}\else\textbf{\resna{#2}}\fi}
\newcommand{\hsLoss}[2]{\ifcsname jevres@#1\endcsname\hsSetNum\hs@x{#1}\ifdim\hs@x pt>-3pt #2\else
  \textcolor{rkRed}{#2}\fi\else#2\fi}
\newcommand{\hsSave}[2]{\ifcsname jevres@#1\endcsname\hsSetNum\hs@s{#1}%
  \ifcsname jevres@#2\endcsname\hsSetNum\hs@x{#2}\else\gdef\hs@x{0}\fi
  \ifdim\hs@s pt<0pt \textcolor{rkRed}{\mmNum{#1}}\else\ifdim\hs@x pt>-3pt \krSave{\textbf{\mmNum{#1}}}\else
  \krSave{\mmNum{#1}}\fi\fi\else--\fi}
\makeatother
\newif\ifeRLive
\ifcsname jevres@e4r/jevagent/success\endcsname\eRLivetrue\fi
\def\eRk{e4\ifeRLive r\fi}%
\makeatletter
\def\eTh@ref#1{\ifcsname jevres@e13/#1/llm_only/success\endcsname#1\else luna\fi}
\newcommand{\eThSave}[1]{\ifcsname jevres@e13/#1/jev_refit/saving\endcsname
  \ifdim\csname jevres@e13/#1/jev_refit/saving\endcsname pt<-100pt
    \ifcsname jevres@e13/#1/jev_refit/cost_ref_k\endcsname\textcolor{rkRed}{$\times$\fpeval{round(\csname jevres@e13/#1/jev_refit/cost_k\endcsname
      /\csname jevres@e13/#1/jev_refit/cost_ref_k\endcsname,1)}}
    \else\hsSave{e13/#1/jev_refit/saving}{e13/#1/jev_refit/delta}\fi
  \else\hsSave{e13/#1/jev_refit/saving}{e13/#1/jev_refit/delta}\fi\else--\fi}
\newcommand{\eThirteenRow}[1]{\krMk\hsMkJ \ifcsname jevres@e13/#1/name\endcsname\res{e13/#1/name}\else
  \res{e13/luna/name}, cache kept\fi{} as agent LLM: tool cascade refitted, \jev{}
  & \resna{e13/#1/jev_refit/n} & \hsPair{e13/\eTh@ref{#1}/llm_only/success}{e13/#1/jev_refit/success}{1}
  & \hsLoss{e13/#1/jev_refit/delta}{\resmna{e13/#1/jev_refit/delta}} & \eThSave{#1}
  & \hsPair{e13/\eTh@ref{#1}/llm_only/wall_mean}{e13/#1/jev_refit/wall_mean}{-1}\\}
\makeatother
\gdef\eThirteenRows{}
\makeatletter
\@for\eTh@k:=luna,luna_cache,deepseek,claude\do{%
  \ifcsname jevres@e13/\eTh@k/jev_refit/success\endcsname
    \xdef\eThirteenRows{\unexpanded\expandafter{\eThirteenRows}\noexpand\eThirteenRow{\eTh@k}}\fi}
\makeatother
\begin{table*}[t]
\centering\small
\setlength{\tabcolsep}{4pt}
\renewcommand{\arraystretch}{1.06}
\fitto{\textwidth}{%
\begin{tabular}{@{}l r r r r r r r@{}}
\toprule
& & \multicolumn{2}{c}{Success (\%)} & & Cost & \multicolumn{2}{c}{Time (s)}\\
\cmidrule(lr){3-4}\cmidrule(lr){7-8}
Delegation & $n$ & ref. & with & $\Delta$ pt & saved \% & ref. & with\\
\midrule
\hsBand{Hard \gym{} tasks: new tasks, \res{e7b/setting/multi_customers} customers each, a rule no decision model was
  trained on}
\krMk\hsMkJ Chosen combination (tool and argument cascades + reflection trigger), \jev{} & \res{e7b/jevagent_best/n} & \hsPair{e7b/llm_only/success}{e7b/jevagent_best/success}{1}
  & \hsLoss{e7b/jevagent_best/delta}{\resm{e7b/jevagent_best/delta}} & \hsSave{e7b/jevagent_best/saving}{e7b/jevagent_best/delta}
  & \hsPair{e7b/llm_only/wall_mean}{e7b/jevagent_best/wall_mean}{-1}\\
\krMk\hsMkJ Tool cascade refitted on the rule, \jev{} & \res{e7e/jev_refit/n} & \hsPair{e7e/llm_only/success}{e7e/jev_refit/success}{1}
  & \hsLoss{e7e/jev_refit/delta}{\resm{e7e/jev_refit/delta}} & \hsSave{e7e/jev_refit/saving}{e7e/jev_refit/delta}
  & \hsPair{e7e/llm_only/wall_mean}{e7e/jev_refit/wall_mean}{-1}\\
\eThirteenRows
\hsBand{Coding agents: \jev{} sends each step to a strong or a fast LLM (ref.: always strong)}
\ifeRLive
\krMk\hsMkJ mini-SWE-agent, SWE-bench Verified: DeepSeek-V4-Pro or -Flash & \res{e4r/n_eval} & \hsPair{e4r/always_deep/success}{e4r/jevagent/success}{1}
  & \hsLoss{e4r/jevagent/vs_deep/delta}{\res{e4r/jevagent/vs_deep/delta}} & \hsSave{e4r/jevagent/vs_deep/saving_pct}{e4r/jevagent/vs_deep/delta}
  & \hsPair{e4r/always_deep/wall_mean_s}{e4r/jevagent/wall_mean_s}{-1}\\
\else
\krMk\hsMkJ DeepSeek-V4-Pro (strong) or DeepSeek-V4-Flash (fast) & \res{e4/n_eval} & \hsPair{e4/always_deep/success}{e4/jevagent/success}{1}
  & \hsLoss{e4/jevagent/vs_deep/delta}{\res{e4/jevagent/vs_deep/delta}} & \hsSave{e4/jevagent/vs_deep/saving_pct}{e4/jevagent/vs_deep/delta}
  & \hsPair{e4/always_deep/wall_mean_s}{e4/jevagent/wall_mean_s}{-1}\\
\fi
\krMk\hsMkJ mini-SWE-agent: Gemini 3.8 Flash, high or low thinking & \res{e4b/jevagent/n} & \hsPair{e4b/always_deep/success}{e4b/jevagent/success}{1}
  & \hsLoss{e4b/jevagent/vs_deep/delta}{\res{e4b/jevagent/vs_deep/delta}} & \hsSave{e4b/jevagent/vs_deep/saving_pct}{e4b/jevagent/vs_deep/delta}
  & \hsPair{e4b/always_deep/wall_mean_s}{e4b/jevagent/wall_mean_s}{-1}\\
\ifcsname jevres@e25/tb/jev_router/success\endcsname
\krMk\hsMkJ SWE-agent, SWE-bench Verified: Gemini 3.5 Flash or Flash-Lite & \res{e25/swe/jev_router/n} & \hsPair{e25/swe/baseline/success}{e25/swe/jev_router/success}{1}
  & \hsLoss{e25/swe/jev_router/delta}{\res{e25/swe/jev_router/delta}} & \hsSave{e25/swe/jev_router/saving}{e25/swe/jev_router/delta}
  & \hsPair{e25/swe/baseline/wall_mean_s}{e25/swe/jev_router/wall_mean_s}{-1}\\
\krMk\hsMkJ pi, Terminal-Bench 2.0: Gemini 3.5 Flash or Flash-Lite & \res{e25/tb/jev_router/n} & \hsPair{e25/tb/baseline/success}{e25/tb/jev_router/success}{1}
  & \hsLoss{e25/tb/jev_router/delta}{\res{e25/tb/jev_router/delta}} & \hsSave{e25/tb/jev_router/saving}{e25/tb/jev_router/delta}
  & \hsPair{e25/tb/baseline/wall_mean_s}{e25/tb/jev_router/wall_mean_s}{-1}\\
\fi
\hsBand{ALFWorld text games}
\krMk\hsMkJ ALFWorld: action cascade, \jev{} & \res{e6a/cascade/n} & \hsPair{e6a/llm_only/success}{e6a/cascade/success}{1}
  & \hsLoss{e6a/cascade/delta}{\resm{e6a/cascade/delta}} & \hsSave{e6a/cascade/saving}{e6a/cascade/delta}
  & \hsPair{e6a/llm_only/wall_mean}{e6a/cascade/wall_mean}{-1}\\
\ifebLive
\krMk\hsMkO ALFWorld: action cascade, 4B trained on its states & \res{e11/cascade/n} & \hsPair{e11/llm_only/success}{e11/cascade/success}{1}
  & \hsLoss{e11/cascade/delta}{\resm{e11/cascade/delta}} & \hsSave{e11/cascade/saving}{e11/cascade/delta}
  & \hsPair{e11/llm_only/wall_mean}{e11/cascade/wall_mean}{-1}\\
\ifcsname jevres@e23/cascade/success\endcsname
\krMk\hsMkO ALFWorld, unseen rooms: same model and threshold & \res{e23/cascade/n} & \hsPair{e23/llm_only/success}{e23/cascade/success}{1}
  & \hsLoss{e23/cascade/delta}{\resm{e23/cascade/delta}} & \hsSave{e23/cascade/saving}{e23/cascade/delta}
  & \hsPair{e23/llm_only/wall_mean}{e23/cascade/wall_mean}{-1}\\
\fi
\fi
\bottomrule
\end{tabular}}
\caption{\textbf{Delegating choices to a decision model in harder settings.} Task success (\%) of
the LLM alone (ref.) and with the delegation on the same $n$ tasks, the change $\Delta$ (points), the share of cost
saved ($\times$: cost as a multiple of the reference's) and the mean seconds per task. Bold marks the better result and savings
that kept success, red a loss of 3 points or more or a higher cost. All rows come from experiments added after Stage~1. Row notes: App.~\ref{app:tablenotes}.}
\label{tab:mainresults}
\end{table*}

\definecolor{fdEdge1}{HTML}{2A78D6}\definecolor{fdFill1}{HTML}{E8F1FB}
\definecolor{fdEdge2}{HTML}{1E8A5A}\definecolor{fdFill2}{HTML}{E4F4EB}
\definecolor{fdEdge3}{HTML}{B8730A}\definecolor{fdFill3}{HTML}{FCF1DC}
\definecolor{fdEdge4}{HTML}{7D58B0}\definecolor{fdFill4}{HTML}{F0EAF7}
\newcommand{\finding}[2]{\par\vspace{2pt}\noindent\begin{tikzpicture}
\node[draw=fdEdge#1, line width=0.7pt, fill=fdFill#1, rounded corners=2pt, inner xsep=1.8mm, inner ysep=0.9mm,
  text width=\columnwidth-3.6mm-1.4pt, align=left, font=\small] {{\color{fdEdge#1}\textbf{Finding #1.}}~#2};
\end{tikzpicture}\par\vspace{-3pt}}

The LLM alone solved \res{stage2/S2-1/ni/success_ref}\% of the \gym{} tasks, so
Table~\ref{tab:mainresults} reports harder settings. Table~\ref{tab:mainresults}, Figure~\ref{fig:frontier} and the
shifted tasks come from experiments added after Stage~1.

\subsection{RQ1: Do isolated scores predict success?}
\label{sec:results:static}\label{sec:results:loop}
On isolated questions (Figure~\ref{fig:frontier:a}), \jev{} was \res{e9/jev/vs/flash/gapabs} points less accurate than
Flash at 1/\res{e9/jev/vs/flash/ratio} of its cost (App.~\ref{app:e9}).\ifcsname jevres@e30/e12/usdk\endcsname{}
On one L4 GPU, \ours{} and \oursS{} cost about
\fpeval{round(\csname jevres@e9/jev/usdk\endcsname/\csname jevres@e30/e12/usdk\endcsname,0)} and
\fpeval{round(\csname jevres@e9/jev/usdk\endcsname/\csname jevres@e30/e28/usdk\endcsname,0)} times less per decision
than \jev{} (App.~\ref{app:e30}).\fi{}
At the task level, the ranking differed from the isolated one (Figure~\ref{fig:frontier:b}). StartLux-9B was the most
accurate model on BFCL tool choice (\res{explore/startlux_live/sl9b/e1_acc}\%), yet it solved only \res{e21/sl9b/l3/success}\% of tasks. \ours{} was less accurate on BFCL (\res{e12/e12/e1/acc/toolchoice}\%) but solved \res{e21/e12/l3/success}\% of tasks, and \jev{} solved \res{e21/jev/l3/success}\%. Step-level accuracy ranked the models nearly
as task success did (Kendall correlation \res{e21/kendall/l2_vs_l3}). The gap lies at must-ask states, where \ours{} answered \res{e21/e12/l2/mustask}\% correctly and \jev{} answered \res{e21/jev/l2/mustask}\%.

\finding{1}{\textbf{Isolated scores misrank decision models.} The best decision model on isolated questions solved only \res{e21/sl9b/l3/success}\% of agent tasks, while \ours{} solved \res{e21/e12/l3/success}\%.}

\begin{figure}[t]
\centering
\mmMap{flash}
\caption[Module map]{\textbf{Delegating tool and argument choices in a cascade saved cost without
losing success.} Each cell: change in task success against the LLM alone (points) $|$ cost saved (\%, negative when costlier), with \jev{}
and Gemini~3.5 Flash. Full map: App.~\ref{app:modulemap}.}
\label{fig:modulemap}
\end{figure}

\subsection{RQ2: Which decisions can be delegated?}
\label{sec:results:stage1}\label{sec:results:where}\label{sec:results:how}
\label{sec:results:harness}\label{sec:results:hard}
Stage~1 tested each delegation alone (Figure~\ref{fig:modulemap}). The cascade, which uses an answer only above its threshold, saved cost at tool and argument choices. With Flash as the agent LLM, \jev{} made \res{primaries/C1F/jev/tool_coverage}\% of the tool choices
and \res{primaries/C1F/jev/arg_coverage}\% of the argument choices, which cut cost by
\res{primaries/C1F/jev/readouts/cost_savings}\% without changing the outcome of any of the \res{primaries/C1F/jev/n_pairs}
tasks, and it passed its registered test. With Flash-Lite, the fitted threshold left every tool choice to the LLM, so the cascade
saved nothing. Modes without a threshold lost success. Letting \jev{} make every tool choice lost
\resabs{mm/cell/flash/tool_select/replace/delta} points, and a three-tool shortlist from \jev{} lost
\resabs{primaries/C2F/jev/delta} points while cutting cost by \res{primaries/C2F/jev/readouts/cost_savings}\%, though
it beat a keyword shortlist by far (\res{primaries/C3/jev/success_arm}\% of tasks solved against
\res{primaries/C3/jev/success_ref}\%). A guard on stopping kept
success but cost \resabs{primaries/C6F/jev/readouts/cost_savings}\% more, and letting \jev{} send Flash-Lite's steps to
Flash cost \resabs{primaries/C4/jev/readouts/cost_savings}\% more than Flash-Lite alone without a clear gain in success.
Refitted on $\tau^2$-bench, the cheapest delegation that passed lost \resabs{primaries/C7/jev/delta} points, so we could not show it transfers.

\finding{2}{\textbf{A fitted threshold protects success.} It cut cost by \res{primaries/C1F/jev/readouts/cost_savings}\% without losing a single task. A fixed threshold lost \resabs{mm/tool_select/cascade/jev_ts_cascade_tau0.5/delta} points.}

\makeatletter
\newcommand{\tpTime}[2]{\ifcsname jevres@#1\endcsname\ifcsname jevres@#2\endcsname
  \edef\tp@v{\fpeval{round(100*(\csname jevres@#2\endcsname-\csname jevres@#1\endcsname)/\csname jevres@#2\endcsname,1)}}%
  \expandafter\tp@signed\tp@v\relax\else--\fi\else--\fi}
\def\tp@signed#1#2\relax{\if-#1\textcolor{rkRed}{\textminus#2}\else\krSave{\textbf{#1#2}}\fi}
\makeatother
\begin{table}[t]
\centering\footnotesize
\setlength{\tabcolsep}{2.6pt}
\renewcommand{\arraystretch}{1.0}
\newcommand{\tpGap}{\addlinespace[1.4pt]}
\def\tpSrvC{stage2/S2-4/cost_savings,hf_ded/cost/saving,e21b/e12_in/saving,e18/final/saving_vs_llm}
\def\tpSrvT{econ/jevagent_best_rl/wall/mean=econ/jevagent_best_rl/wall/ref_mean,e21b/e12_in/wall=e21b/e12_in/ref_wall,e18/final/wall=e18/final/llm/wall}
\fitto{\columnwidth}{%
\begin{tabular}{@{}l l r r r@{}}
\toprule
Part & Without, then with & $\Delta$ pt & \multicolumn{2}{c}{Saved \%: cost, time}\\
\midrule
Threshold & fixed at 0.5 & \hsLoss{mm/tool_select/cascade/jev_ts_cascade_tau0.5/delta}{\tbNum{mm/tool_select/cascade/jev_ts_cascade_tau0.5/delta}}
  & \hsSave{mm/tool_select/cascade/jev_ts_cascade_tau0.5/saving}{mm/tool_select/cascade/jev_ts_cascade_tau0.5/delta} & --\\
& fitted in shadow mode & \tbNum{primaries/C1F/jev/delta} & \hsSave{primaries/C1F/jev/readouts/cost_savings}{primaries/C1F/jev/delta} & --\\
\tpGap
Check\dg & own confidence & -- & \hsSave{e8/toolparams/g1tau/save}{} & --\\
& + re-ask if unsure & -- & \hsSave{e8/toolparams/g1g3/save}{} & --\\
& LLM check instead & -- & \hsSave{e8/toolparams/g4/save}{} & --\\
\tpGap
Combined & tool cascade alone & \tbNum{mm/tool_select/cascade/jev_ts_cascade_tauhat/fresh/delta}
  & \hsSave{mm/tool_select/cascade/jev_ts_cascade_tauhat/fresh/saving}{mm/tool_select/cascade/jev_ts_cascade_tauhat/fresh/delta}
  & \tpTime{econ/best_single/wall/mean}{econ/best_single/wall/ref_mean}\\
& + reflection trigger & \tbNum{stage2/S2-1/ni/est} & \hsSave{stage2/S2-1/cost_savings}{stage2/S2-1/ni/est}
  & \tpTime{econ/jevagent_best/wall/mean}{econ/jevagent_best/wall/ref_mean}\\
\tpGap
Serving & 4B, shared GPU & \tbNum{stage2/S2-4/delta} & \hsSave{stage2/S2-4/cost_savings}{stage2/S2-4/delta}
  & \tpTime{econ/jevagent_best_rl/wall/mean}{econ/jevagent_best_rl/wall/ref_mean}\\
& 4B, batched, own GPU\dg & \tbNum{hf_ded/success/delta} & \hsSave{hf_ded/cost/saving}{hf_ded/success/delta} & \hsSave{hf_ded/wall_s/saving}{hf_ded/success/delta}\\
\ifcsname jevres@e21b/e12_in/saving\endcsname
& 4B + public data, own GPU\dg & \tbNum{e21b/e12_in/delta} & \hsSave{e21b/e12_in/saving}{e21b/e12_in/delta}
  & \tpTime{e21b/e12_in/wall}{e21b/e12_in/ref_wall}\\
\fi
\ifcsname jevres@e18/final/saving_vs_llm\endcsname
& 0.8B, learns to defer\dg & $+$\res{e18/final/vsllm/success} & \hsSave{e18/final/saving_vs_llm}{e18/final/vsllm/success}
  & \tpTime{e18/final/wall}{e18/final/llm/wall}\\
\fi
\bottomrule
\end{tabular}}
\caption{\textbf{What each part of \harnessname{} contributes.} Each row compares the agent without and with one part:
the change in success (points) and the cost and time saved (\%) vs.\ the LLM alone.
Bold green marks savings that kept success, red a loss or an added cost. \dg{}Added after Stage~1. Row notes: App.~\ref{app:tablenotes}.}
\label{tab:parts}
\end{table}

\def\trLone{e9/base/acc,e9/sftv1/acc,e9/rl/acc,e9/e12/acc,e9/jev/acc,e9/intern/acc,e9/sl4b/acc,e9/e12c/acc,e9/sl0.8b/acc,e9/e28/acc}
\def\trLall{e21/base/l2/all,e21/sft_v1/l2/all,e21/rl/l2/all,e21/e12/l2/all,e21/jev/l2/all,e21/intern/l2/all,e21/sl4b/l2/all,e21/e12c/l2/all,e28/base/l2/all,e28/l2/all}
\def\trLask{e21/base/l2/mustask,e21/sft_v1/l2/mustask,e21/rl/l2/mustask,e21/e12/l2/mustask,e21/jev/l2/mustask,e21/intern/l2/mustask,e21/sl4b/l2/mustask,e21/e12c/l2/mustask,e28/base/l2/mustask,e28/l2/mustask}
\def\trLthree{e21/base/l3/success,e21/sft_v1/l3/success,e21/rl/l3/success,e21/e12/l3/success,e21/jev/l3/success,e21/intern/l3/success,e21/sl4b/l3/success,e21/e12c/l3/success,e28/base/l3/success,e28/l3/success}
\providecommand{\trS}[4]{#1 & #2 & \resbest{e21/#3/l2/all}{\trLall} & \resbest{e21/#3/l2/mustask}{\trLask} & \resbest{e21/#3/l3/success}{\trLthree} & #4\\}
\begin{table}[t]
\centering\scriptsize
\setlength{\tabcolsep}{1.35pt}
\renewcommand{\arraystretch}{1.0}
\fitto{\columnwidth}{%
\begin{tabular}{@{}l r@{\hspace{6pt}}r r@{\hspace{6pt}}r@{\hspace{4pt}}r@{}}
\toprule
& L1 & \multicolumn{2}{c}{L2 step level} & L3 task &\\
\cmidrule(lr){3-4}
Model & isolated & all & must-ask & success & Saved\\
\midrule
\tbBand{6}{Shifted tasks\dg: \ours{}, step by step from Qwen3-4B}
\trS{Untrained}{\resbest{e9/base/acc}{\trLone}}{base}{--}
\trS{+ SFT}{\resbest{e9/sftv1/acc}{\trLone}}{sft_v1}{--}
\trS{+ GRPO (Stage~2)}{\resbest{e9/rl/acc}{\trLone}}{rl}{\krNumOut{stage2/S2-4/decision}{\res{stage2/S2-4/cost_savings}}}
\trS{\quad + public data}{\resbest{e9/e12/acc}{\trLone}}{e12}{\ifcsname jevres@e21b/e12_in/saving\endcsname\ifnum\pdfstrcmp{\csname jevres@e21b/e12_in/ni/verdict\endcsname}{non-inferior}=0 \krPillS{krHoldF}{rkOk}{\res{e21b/e12_in/saving}}\else\res{e21b/e12_in/saving}\fi\else--\fi}
\tbBand{6}{Shifted tasks\dg: off-the-shelf decision models}
\trS{\jev{} (teacher)}{\resbest{e9/jev/acc}{\trLone}}{jev}{\krNumOut{stage2/S2-1/decision}{\res{stage2/S2-1/cost_savings}}}
\trS{Intern-Decision-4B}{\resbest{e9/intern/acc}{\trLone}}{intern}{--}
\trS{StartLux-4B}{\resbest{e9/sl4b/acc}{\trLone}}{sl4b}{--}
\trS{\quad + our recipe}{\resbest{e9/e12c/acc}{\trLone}}{e12c}{--}
\ifcsname jevres@e28/l3/success\endcsname
StartLux-0.8B & \resbest{e9/sl0.8b/acc}{\trLone} & \resbest{e28/base/l2/all}{\trLall} & \resbest{e28/base/l2/mustask}{\trLask} & \resbest{e28/base/l3/success}{\trLthree} & --\\
\quad + our recipe = \oursS{} & \resbest{e9/e28/acc}{\trLone} & \resbest{e28/l2/all}{\trLall} & \resbest{e28/l2/mustask}{\trLask} & \resbest{e28/l3/success}{\trLthree} & --\\
\fi
\tbBand{6}{ALFWorld\dg: SFT on states of its training games}
+ SFT & -- & \ifcsname jevres@e24/e11/pooled/acceptable\endcsname\res{e24/e11/pooled/acceptable}\else\res{e11/offline/trained/dev/acc}\fi
  & -- & \res{e11/replace/success} & \krNumNI{hardni/e11/ni}{\res{e11/cascade/saving}}\\
\bottomrule
\end{tabular}}
\caption{\textbf{Training on the agent's own states made a 4B model accurate, even on shifted tasks.} Accuracy on
isolated questions (L1), step-level accuracy (all states and must-ask states) and task-level success on shifted tasks (L2, L3), and
the cost saved when the model is deployed. \dg{}Added after Stage~1. Column notes: App.~\ref{app:tablenotes}.}
\label{tab:train}
\end{table}

\subsection{RQ3: What changes in deployment?}
\label{sec:results:stage2}\label{sec:results:bp}\label{sec:results:deploy}
\harnessname{} cut cost and time on \gym{}, ALFWorld and mini-SWE-agent. With \ours{} on its own GPU, the
\gym{} agent was \res{e21b/e12_in/saving}\% cheaper and
\fpeval{round(\csname jevres@e21b/e12_in/ref_wall\endcsname/\csname jevres@e21b/e12_in/wall\endcsname,1)} times faster, without losing 3 or more points of success, and on ALFWorld a 4B model trained there made the agent \res{e11/cascade/saving}\% cheaper and about twice as fast, with
higher success (Table~\ref{tab:mainresults}). Coding agents gained less. When \jev{} sent each step to a strong or fast LLM, cost fell by
\res{e4b/jevagent/vs_deep/saving_pct}--\res{e4r/jevagent/vs_deep/saving_pct}\% and success varied by harness (App.~\ref{app:e25}). On hard \gym{} tasks, delegation kept or raised success but saved only \res{e7b/jevagent_best/saving}--\res{e7e/jev_refit/saving}\% and ran slower. The agent was also slower on a GPU shared by about four runs.

\finding{3}{\textbf{Delegation cut cost and time without losing success.} With \ours{} the agent ran \res{e21b/e12_in/saving}\% cheaper and \fpeval{round(\csname jevres@e21b/e12_in/ref_wall\endcsname/\csname jevres@e21b/e12_in/wall\endcsname,1)}$\times$ faster. A 4B model trained on ALFWorld made the ALFWorld agent \res{e11/cascade/saving}\% cheaper.}

\ifcsname jevres@stage2/intervention/jevagent_best_rl/tool_coverage\endcsname{} On \res{stage2/intervention/jevagent_best_rl/n} new tasks, \ours{} made
\res{stage2/intervention/jevagent_best_rl/tool_coverage}\% of the tool choices and
\res{stage2/intervention/jevagent_best_rl/arg_coverage}\% of the argument choices, saving
\fpeval{round(\csname jevres@stage2/intervention/jevagent_best_rl/llm_calls_avoided\endcsname,1)} LLM calls per task.\fi\ifcsname jevres@e30/e12/lat_p50\endcsname{} On one L4 GPU, it took a median \res{e30/e12/lat_p50}\,ms per answer, while a Flash call took \res{fig1/latency/s2}\,ms.\fi

\subsection{RQ4: Can a 4B model learn the decisions?}
\label{sec:results:rl}\label{sec:results:training}
\ifcsname jevres@e21/sft_v1/l2/all\endcsname SFT on the agent's own states raised step-level accuracy on shifted states from
\res{e21/base/l2/all}\% to \res{e21/sft_v1/l2/all}\% (Table~\ref{tab:train})\else SFT on the agent's own states raised accuracy on the common states from about half to nearly all
(Table~\ref{tab:train})\fi{}, while isolated accuracy rose by only \resgap{e9/sftv1/acc}{e9/base/acc}{1} points. Most of the gain comes from SFT. It took
\res{recipe/stage/sft_v1/gpu_h} GPU-hours. GRPO added \res{recipe/stage/grpo/gpu_h} GPU-hours and
\$\res{recipe/stage/grpo/api_usd} of LLM calls without a measurable gain\ifcsname jevres@e31/rlonly/l3/success\endcsname{}. Run without SFT, GRPO did not improve the untrained model, which solved \res{e31/rlonly/l3/success}\% of shifted tasks after GRPO and \res{e21/base/l3/success}\% before (App.~\ref{app:e31}, Table~\ref{tab:paradigms})\fi{}. With \jev{}, the chosen combination lost \resabs{stage2/S2-1/ni/est} points and failed the non-inferiority test. With \ours{} in place of \jev{}, it passed at \res{stage2/S2-4/cost_savings}\% lower cost on \res{stage2/S2-4/n_pairs} new tasks. On the common states, \ours{} beat Intern-Decision-4B and was non-inferior to \jev{}.\ifcsname jevres@e12/e12/c/acc\endcsname{} Further training on public tool-use data raised its isolated
accuracy from \res{e9/rl/acc}\% to \res{e9/e12/acc}\%, while task-level success on shifted tasks fell from
\res{e21/rl/l3/success}\% to \res{e21/e12/l3/success}\% (App.~\ref{app:e12}).\fi\ifcsname jevres@e28/l3/success\endcsname{} %
Our recipe turned StartLux-0.8B into \oursS{} (Table~\ref{tab:train}).\fi{}
\ifebLive{}\ifcsname jevres@e24/e11/pooled/acceptable\endcsname{}On ALFWorld, a 4B model trained there answered \res{e24/e11/pooled/acceptable}\% of recorded states correctly, and the best off-the-shelf model answered \res{e24/sl9b/pooled/acceptable}\% (App.~\ref{app:e24}).\fi\fi%

\finding{4}{\textbf{Training on agent states raised success.} Qwen3-4B's success on shifted tasks rose from \res{e21/base/l3/success}\% to \res{e21/sft_v1/l3/success}\%, and StartLux-4B's from \res{e21/sl4b/l3/success}\% to \res{e21/e12c/l3/success}\%.}

\subsection{Where delegation fails}
\label{sec:results:fail}
Three failure modes recur. A confident wrong choice at an irreversible step ends the task. \ifdim\csname jevres@drift/jev/first/irreversible_of_failed\endcsname pt=100pt
All \res{drift/jev/first/failed} of \jev{}'s failed tasks began\else Of \jev{}'s \res{drift/jev/first/failed}
failed tasks, \res{drift/jev/first/irreversible_of_failed}\% began\fi{} with an irreversible wrong tool choice, usually cancelling where the agent should have asked. Models trained on agent states made no such error
(App.~\ref{app:drift}). The decision model can also face a new rule. On hard tasks with \res{e7/setting/multi_customers} customers,
\ours{}, trained without this rule, made most cancel-or-ship decisions but got only \res{e7/jevagent_best_rl/op_acc}\% of them right,
and the agent solved \res{e7/jevagent_best_rl/success}\% of tasks against \res{e7/llm_only/success}\% for the LLM
alone. Refitting the threshold restored success with \jev{} (Table~\ref{tab:mainresults}), and retraining \ours{} on the rule
raised its held-out accuracy from \res{e7d/frozen/heldout/op/acc}\% to
\res{e7d/tier/heldout/op/acc}\% (App.~\ref{app:e7}). Finally, modes that still call the LLM, such as guards, save little.

\relax

\section{Related Work}
\label{sec:related}

\paragraph{Decision models.}
\jev{} and open models like Intern-Decision, StartLux-Decision and OpenJev
\citep{typesafe2026jev,interndecision2026,startlux2026decision,openjev2026} score listed options, while DecisionFlow \citep{chen2025decisionflow} structures LLM decisions. REFLEX hands an action to the LLM when its confidence falls below one global threshold \citep{wu2026reflex}, and we fit one per decision point.

\paragraph{Decision evaluation.}
Decision and tool-use benchmarks score fixed questions
\citep{jevbench2026,ling2026jevinthewild,patil2025bfcl,ross2025when2call}, and agent environments score whole episodes
\citep{yao2025taubench,trivedi2024appworld,lu2025toolsandbox}. Models with equal scores can still decide differently \citep{zhang2026samescores}. \benchname{} adds step- and task-level tests.

\paragraph{Routing, deferral and training.}
Routers and cascades choose the model for a call
\citep{chen2024frugalgpt,ong2025routellm,aggarwal2024automix}. Learned routers, compared in LLMRouter \citep{feng2026llmrouter}, use query--LLM graphs, multi-LLM logs or multi-round RL \citep{feng2025graphrouter,feng2026fusionfactory,zhang2025routerr1}. A decision model instead scores listed options and defers if unsure, like learning to defer
\citep{geifman2017selective,mozannar2020consistent}. Agentic RL
trains the LLM \citep{jin2025searchr1,wang2025ragen}, while we train the decision model (App.~\ref{app:related}).

\section{Conclusion}
\label{sec:conclusion}
Our results support the hypothesis that decision models help agents when tested, used and trained inside the agent. Only tests inside the agent (\benchname{}) showed which decision models work there. Trained on the agent's states, \ours{} beat \jev{} inside the agent and, with thresholds fitted on agent trajectories (\harnessname{}), made most routine choices. Models trained this way made the agent \res{stage2/S2-4/cost_savings}--\res{e11/cascade/saving}\% cheaper and up to \fpeval{round(\csname jevres@e21b/e12_in/ref_wall\endcsname/\csname jevres@e21b/e12_in/wall\endcsname,1)} times faster, without losing 3 or more points of success.

\clearpage
\section*{Limitations}
All tasks and prompts are in English, and most runs use Gemini agent LLMs, with DeepSeek and OpenAI models in a
few. With other languages or agent LLMs, other kinds of choices may be worth delegating. Our main tests are non-inferiority
tests. They show that task success drops by less than a fixed margin (3 points, 5 on $\tau^2$-bench), and showing equal success would need an equivalence test. The thresholds and the trained decision model are fitted to an agent's
own runs, so a new agent, new tools or a new task family call for refitting the thresholds and, for the best accuracy,
training on that agent's states, as we did for ALFWorld. In coding agents, whose actions are free text, the decision
model can only choose which LLM writes a step. There the savings were small (\res{e4b/jevagent/vs_deep/saving_pct}--\res{e4r/jevagent/vs_deep/saving_pct}\%)\ifcsname jevres@e29b/go/decision\endcsname, and sending the steps that only read files to a fast LLM gave no reliable saving, because the decision model could hardly tell those steps from the others (App.~\ref{app:e29b})\fi. Prices and latencies are those we recorded, and they change with
providers, hardware and how the decision model is served (App.~\ref{app:cost}). We did not compare with learned routers
such as RouteLLM on the same tasks, and GRPO added no measurable gain over SFT in our runs. In deployment, the share of
answers above the threshold could be tracked to notice when the tasks drift and the thresholds need refitting.

\section*{Ethical Considerations}
We will release the benchmark, harness, recipes, trained models and data
(App.~\ref{app:licences}). Training on \jev{} outputs and redistributing derived weights depend on the provider's terms,
and StartLux-Decision weights are licensed CC BY-NC 4.0. All benchmarks are public, and we collect no personal data: \gym{}
customers, orders and e-mail addresses are generated from templates. Lower cost can make a fast model's failures on
safety-relevant decisions easy to overlook. With the cheaper Flash-Lite as agent LLM, the stop guard accepted
\res{e2b_jevgym_meta_control/meta_shadow_s2lite/decisions/gold/probes/stop/far_nat/far}\% of the LLM's own wrong
attempts to end the task (App.~Table~\ref{tab:guardfar}), so a decision-model guard should never be the only control on an irreversible
action. If an adapter fails, the step runs without it (\S\ref{sec:harness}). Adapters therefore control only cost and speed and should not be used as security boundaries. We did not test adversarial
observations or policy injection, which can flip a decision model's typed answers
\citep{xu2026jevout,hu2026jevadvbench,wu2026decisionhijacking}, so our results hold for benign states only. Cheaper
agent control also lowers the cost of misuse at scale.

\makeatletter
\ifacl@anonymize\else\ifarxiv\else
\section*{Acknowledgments}
\todo{Funding, compute credits and people (final version only).}
AI assistance: \todo{authors to confirm scope (ARR AI writing/coding policy): AI coding assistants (Claude Code)
wrote and tested most of the code and drafted and edited parts of the text under the authors' direction; Gemini
and GPT models gave simulated reviews of drafts; every output was checked by the authors, who take full
responsibility for the paper.}
\fi\fi
\makeatother

\bibliography{references}

\clearpage %
\appendix
\setcounter{topnumber}{4}\setcounter{bottomnumber}{2}\setcounter{totalnumber}{6}\setcounter{dbltopnumber}{3}
\renewcommand{\topfraction}{0.92}\renewcommand{\bottomfraction}{0.6}\renewcommand{\textfraction}{0.08}
\renewcommand{\floatpagefraction}{0.72}\renewcommand{\dbltopfraction}{0.88}\renewcommand{\dblfloatpagefraction}{0.72}
\makeatletter %
\setlength{\@fptop}{0pt}\setlength{\@fpsep}{12pt plus 4pt}\setlength{\@fpbot}{0pt plus 1fil}
\setlength{\@dblfptop}{0pt}\setlength{\@dblfpsep}{12pt plus 4pt}\setlength{\@dblfpbot}{0pt plus 1fil}
\def\@fb@topbarrier{}%
\makeatother
\raggedbottom %

\makeatletter
\newcommand{\seclabel}[1]{{\protected@edef\@currentlabel{\thesection}\label{#1}}}
\makeatother

\makeatletter
\newcommand{\appTocLeaders}{\nobreak\leaders\hbox to 0.55em{\hss.\hss}\hfill\nobreak}
\etocsetstyle{section}{\parindent0pt\parskip0pt}
  {\addvspace{2.6pt}\leftskip1.5em\rightskip1.8em plus 1fil\parfillskip-\rightskip\relax
   \leavevmode\normalsize\bfseries}
  {\llap{\makebox[1.5em][l]{\etocnumber}}\etocname\appTocLeaders\makebox[1.8em][r]{\etocpage}\par}
  {}
\etocsetstyle{subsection}{}
  {\parindent0pt\leftskip3.7em\rightskip1.8em plus 1fil\parfillskip-\rightskip\relax\leavevmode\small\mdseries}
  {\llap{\makebox[2.2em][l]{\etocnumber}}\etocname\appTocLeaders\makebox[1.8em][r]{\etocpage}\par}
  {}
\etocsettocstyle{\section*{Appendix Contents}\vspace{-0.6ex}%
  {\small The first, unnumbered part gives compute, data and AI use. Sections A--E follow \S\S2--6 of the paper, F gives models, costs and licences, G the result tables of
  \S7, H further related work and I three recorded trajectories as case studies. \oursV{} is \ours{} as frozen after SFT and
  GRPO for Stage~2, before its training on public tool-use data.\par}\vspace{0.5ex}}{\par\vspace{1ex}}
\makeatother
\etocsetnexttocdepth{subsection}
\addcontentsline{toc}{part}{Appendix}%
{\ifdefined\hypersetup\hypersetup{linkcolor=black}\fi\localtableofcontents}%

\section*{Compute, Data and AI Use}
\addcontentsline{toc}{section}{Compute, Data and AI Use}
\paragraph{Compute, software and runs.}
The open decision models have 0.8B to 27B parameters, as their names state, and the API models (\jev{}, Gemini, GPT,
DeepSeek) are used through their providers. We trained and served the open models on cloud hosts with one NVIDIA L40S,
A10G or L4 GPU. Table~\ref{tab:recipe} gives the GPU hours of each training stage. Training used PyTorch, Hugging Face
Transformers, TRL and PEFT (the StartLux track used Miles), and serving used vLLM or llama.cpp. Unless a table states
attempts, each arm ran each task once at temperature~0, and every number comes from that single run.

\paragraph{Data and annotation.}
\gym{} tasks, customers, orders and e-mail addresses are generated from templates, so the tasks and agent states we
release contain no personal information and no offensive text. Public benchmark items are used as distributed. Labels
come from \gym{}'s answer key and the benchmarks' own answers and checkers, and the $\tau^2$-bench user is an LLM
simulator, so no human annotators or study participants took part.

\paragraph{Use of AI assistants.}
LLMs play three roles in this work. They are the objects of study, called through their APIs in every experiment
(agent LLMs and decision models). \jev{} and Gemini also answered agent states to build the training data of \ours{}
(\S\ref{sec:rl}). Finally, AI assistants helped write and test code, check references and revise the writing, under
the authors' direction. The paper was not generated automatically. The authors designed the study, checked every
result and every sentence, and take full responsibility for the content.

\section{Delegation Theory}
\label{app:theory}
\paragraph{Objective.}
A combination of delegations $\phi$ gives each decision point that the decision model handles a mode and a threshold,
and the LLM keeps the rest. With expected cost $C(\phi)$ per episode and task success $S(\phi)$, we seek
$\min_{\phi} C(\phi)$ subject to $S(\phi) \ge S(\phi_{\varnothing}) - \delta$, where $\phi_{\varnothing}$ is LLM-only
and $\delta$ = 3 points on \gym{} (\S\ref{sec:setup}).

\paragraph{Notation.}
At step $t$ the harness holds the history $h_t$. Decision point $d$ has options $\mathcal{A}_d$, possibly with an
extra option $\bot$ (none of them, or end the task). By default the LLM decides, $a \sim \pi_L(\cdot \mid h_t)$. A
decision model $f$ reads a text rendering $s_t$ of $h_t$ and returns $p_f(\cdot \mid s_t)$. Its answer is
$\hat a = \arg\max_a p_f(a \mid s_t)$ with confidence $c = p_f(\hat a \mid s_t)$, and it is \emph{calibrated} at $d$ if
$g_d(c) := P(\hat a \text{ correct} \mid c) = c$. With threshold $\tau$, width $k$ and LLM proposal $a_L$:
\begin{align*}
\text{\modeRep:}\;& a = \hat a \\
\text{\modeCas:}\;& a = \hat a \text{ if } c \ge \tau, \text{ else } a \sim \pi_L(\cdot \mid h_t)\\
\text{\modeAss:}\;& a \sim \pi_L\big(\cdot \mid h_t, \mathrm{top}\text{-}k(p_f) \cup \{\bot\}\big)\\
\text{\modeGua:}\;& \text{accept } a_L \text{ iff } p_f(\text{yes} \mid s_t, a_L) \ge \tau\\
\text{\modeTri:}\;& \text{invoke module } M \text{ iff } p_f(\text{yes} \mid s_t) \ge \tau
\end{align*}

\begin{table}[!htbp]
\centering
\small
\setlength{\tabcolsep}{3pt}
\resizebox{0.86\columnwidth}{!}{%
\begin{tabular}{llccccc}
\toprule
Decision point & Q & Rep & Cas & Ass & Gua & Tri\\
\midrule
Tool selection & \Choice & \checkmark & \checkmark & \checkmark & \checkmark & $\circ$\\
Tool arguments & \Choice & \checkmark & \checkmark & $\circ$ & $\circ$ & $\circ$\\
Memory write & \Noul & $\circ$ & $\circ$ & $\circ$ & $\circ$ & \checkmark\\
Memory read & \Choice & $\circ$ & $\circ$ & \checkmark & $\circ$ & $\circ$\\
Reflection & \Noul & $\circ$ & $\circ$ & $\circ$ & $\circ$ & \checkmark\\
Stopping & \Noul & $\circ$ & $\circ$ & $\circ$ & \checkmark & \checkmark\\
Routing & \Noul & $\circ$ & $\circ$ & $\circ$ & $\circ$ & \checkmark\\
Action-only & \Choice & \checkmark & \checkmark & \checkmark & $\circ$ & $\circ$\\
\bottomrule
\end{tabular}}
\caption{Implemented pairs of decision point and mode. \checkmark: a Stage-1 configuration. $\circ$: added later as
an exploratory row (App.~\ref{app:modulemap-x}). \modeCas{} with $\tau = 0$ is \modeRep{}, and LLM-only is \modeAss{}
with $k = |\mathcal{A}_d|$ or \modeGua{} with $\tau = 0$.}
\label{tab:design}
\end{table}

\paragraph{Proposition~1.}
Under \modeCas{}, each query to the decision model costs $\kappa$ and returns $\hat a$ with confidence $c$. A rule $\psi$ accepts $\hat a$ with probability $\psi(c)$ and otherwise defers to the LLM at extra cost $\Delta C$. Let
$q_L(c) = P(\text{LLM correct} \mid c)$, and let an error cost $V$ whatever the wrong option, with losses adding across
decisions. Among rules that see the state only through $c$, expected loss is minimised by
\begin{equation}
\text{accept } \hat a \;\iff\; g_d(c) - q_L(c) \;\ge\; -\Delta C / V ,
\label{eq:tau-general}
\end{equation}
and, if the decision model is calibrated ($g_d(c) = c$) and the LLM's accuracy $q_L$ does not depend on $c$, by the
threshold rule
\begin{equation}
c \;\ge\; \tau^\star \;=\; \max\bigl(0,\; q_L - \Delta C / V\bigr).
\label{eq:tau}
\end{equation}
\emph{Proof sketch.} Let $A(c) = g_d(c) - q_L(c) + \Delta C/V$. Accepting costs $(1-g_d(c))V$ in expected errors, deferring costs $(1-q_L(c))V + \Delta C$, and $\kappa$ is paid either way, so the expected loss is
$\kappa + \mathbb{E}_c[(1-q_L(c))V + \Delta C] - V\,\mathbb{E}_c[\psi(c)A(c)]$. Since $\psi(c)A(c) \le A(c)^{+}$, with
equality iff $\psi$ accepts where $A(c) > 0$ and defers where $A(c) < 0$, Eq.~\ref{eq:tau-general} is optimal. Under
calibration and a constant $q_L$, $A(c) = c - q_L + \Delta C/V$ increases in $c$, which gives Eq.~\ref{eq:tau}.
$\square$ An over-confident model ($g_d(c) = c - \xi$) needs a threshold higher by $\xi$, so we fit $g_d$ on held-out
decisions.

\paragraph{Triggers and guards.}
A trigger that invokes module $M$ at calibrated need probability $p$ pays iff $p \ge (C_M + H_M)/(B_M + H_M)$, for
benefit $B_M$ when needed, harm $H_M$ when not and cost $C_M$. A guard follows the same rule with $B_M = \eta V$,
$H_M = \delta V$ and $C_M = \Delta C$, where $\eta$ and $\delta$ are the chances that a rejection fixes a wrong
proposal or breaks a right one.

\paragraph{Proposition~2.}
Take a gate with need prevalence $\pi$, resolution $J = \text{TPR} - \text{FPR}$, $n$ opportunities, query price
$\kappa$, and benefit $B_M$ or harm $H_M$ of firing when needed or not. Under additive effects, common opportunities
and a need-independent firing cost, its expected gain over a random gate that fires at the same rate is
\begin{equation}
\Delta U \;=\; n\,\big[(B_M + H_M)\,\pi(1-\pi)\,J - \kappa\big],
\label{eq:voi}
\end{equation}
which grows with the resolution $J$ whatever the accuracy.
\emph{Proof sketch.} The gate fires at rate $\nu = \pi\,\text{TPR} + (1-\pi)\,\text{FPR}$ and the random gate at $\nu$
independently of need. The difference per opportunity is
$\pi B_M(\text{TPR} - \nu) - (1-\pi)H_M(\text{FPR} - \nu) - \kappa$, with
$\text{TPR} - \nu = (1-\pi)J$ and $\text{FPR} - \nu = -\pi J$. $\square$
An always-\emph{no} gate is $1-\pi$ accurate but gains nothing ($J = 0$). On \gym{}, $B_M = W e$ and $H_M = W h$
(App.~\ref{app:metagold}).

\paragraph{Deployed thresholds.}
The registered cascade tests plug estimates of $g_d$, $q_L$, $\Delta C$ and $V$ from shadow-mode trajectories into
Eq.~\ref{eq:tau-general}. We sort paired decisions into equal-mass confidence bins of at least 20 each. $\hat\tau^\star$ is the lowest confidence from which every higher bin has a non-negative estimated advantage, and the harness always defers if there is none. We set $\lambda = 0$ and estimate $\hat V$ as the success difference between calibration episodes with every decision at $d$ right and those with one wrong. For tool selection and arguments the fits are \res{calibration/c1/jev/flash/tool_select/tauhat} and
\res{calibration/c1/jev/flash/tool_params/tauhat} for \jev{} with Flash (other fits in \texttt{results/\allowbreak
calibration/}).

\paragraph{Registered hypotheses.}
H1 says that the plug-in threshold has lower held-out regret than a fixed default and that recalibration returns
over-confident models to the cost--success front. H2 says that at a matched firing rate a gate's gain over a random
gate is proportional to $J$ (Eq.~\ref{eq:voi}). H3 says that among configurations non-inferior to LLM-only, those
delegating bounded, recoverable, meta-level decisions dominate the front, and those delegating copy-dependent
arguments or irreversible actions do not. H4 says that Monte-Carlo targets beat teacher distillation. H1 and H3 were \emph{falsified}, and H2 could \emph{not be estimated} (App.~\ref{app:metagold}). H4 was \emph{\res{secondary/h4/verdict}}, because Stage~1 ended without a Monte-Carlo checkpoint.
Fitted thresholds beat the default $\tau = 0.8$ in
\res{secondary/h1/wins} of \res{secondary/h1/n_estimable} cells. After temperature scaling, \res{secondary/h1/clause2/n_return} of the \res{secondary/h1/clause2/n_over_confident} over-confident cells returned to the cost--success front. In
the tool-cascade test with Flash, \jev{}'s cascade delegates \res{secondary/h3/counterexample/credited} decisions yet was non-inferior
($\Delta = \resm{secondary/h3/counterexample/delta}$ points) at \res{secondary/h3/counterexample/cost_savings}\%
savings, a registered counterexample to H3.

\section{\benchname{}}
\label{app:bench}

\subsection{Suites and the \gym{} Environment}
\seclabel{app:jevgym}
Table~\ref{tab:suites} lists every suite of \benchname{}.

\begin{table}[!htbp]
\centering\footnotesize
\setlength{\tabcolsep}{2.5pt}
\resizebox{\columnwidth}{!}{%
\begin{tabular}{@{}lrl@{}}
\toprule
Suite & Size & Decision points\\
\midrule
\multicolumn{3}{@{}l}{\emph{L1: isolated}}\\
BFCL v4 (6 categories) & 1{,}114 & tool, args, none\\
When2Call (test) & 928 & call/ask/refuse/answer\\
\multicolumn{3}{@{}l}{\emph{L2: step-level, recorded states with an answer key}}\\
\gym{} common states & \res{stage2/S2-3/corpus/tasks} & tool\\
\multicolumn{3}{@{}l}{\emph{L3: task-level, our own agent loop}}\\
\gym{} & 500 & seven (no action-only)\\
$\tau^2$ airline + retail & 164$\times$3$^\ddagger$ & tool, args, guard, route\\
\multicolumn{3}{@{}l}{\emph{L3: hard tasks, real harnesses$^\S$}}\\
SWE-bench Verified & 100$\times$1 & guard, stop, route, reflect\\
Terminal-Bench-Science & \res{tbsci/n_tasks}$\times$\res{tbsci/k} & guard, stop, reflect\\
\multicolumn{3}{@{}l}{\emph{Stage 2: fresh suites}}\\
\gym{}, new seeds & 500 & chosen combination\\
$\tau^2$ telecom (sample) & \res{stage2/S2-L3a/n_pairs}${\times}1$ & chosen combination\\
\bottomrule
\end{tabular}}
\caption{Suites (size $\times$ attempts, L1 without development items). Sizes are registered except for the hard
tasks. $^\ddagger$149 tested (15 airline tasks refit $\hat\tau^\star$). $^\S$Exploratory: \res{tbsci/n_tasks} of 37
Terminal-Bench-Science tasks qualified under a registered rule. ALFWorld and Terminal-Bench~2.0 (budget) were
registered but not run. ALFWorld later ran as exploratory rows (App.~\ref{app:modulemap-x}).}
\label{tab:suites}
\end{table}

\paragraph{World, tasks and success.}
Each \gym{} task samples a world with 12 users, 2--4 orders per user and stock for 10 items. At the evaluation
settings every task offers the same 68 tools in shuffled order, plus \emph{finish}. Seven are core tools (find a
user, list orders, cancel, refund, check stock, ship, send a message), one asks the user, 31 are near-duplicates of core tools that read or write, and 29 are unrelated tools such as a weather lookup. The evaluation set uses the two hardest of seven task
families. In \emph{clarify} a needed detail is missing, so the agent must ask a scripted customer before any
state-changing call. In \emph{multi}, 2--12 customers each make a cancel or ship request, and the task asks for the
total amount of the processed orders. Canonical plans have a median of
\res{data_stats/jevgym_eval/by_family/clarify/plan_steps_median} steps on \emph{clarify} and
\res{data_stats/jevgym_eval/by_family/multi/plan_steps_median} on \emph{multi}. In \emph{strict} mode, used in
evaluation, a state-changing call that deviates from every gold call or precedes a required question fails the
episode. Success is checked on the final world state and, where the task asks for a number, on the final answer.

\paragraph{From the design principles to the generator.}
\emph{(i) An answer key at every state.} A task is generated with its plan, and from a copy of any solvable state,
also after the agent's own detours, the answer key finishes the task and counts the steps it needs
(App.~\ref{app:metagold}). \emph{(ii) Disjoint seeds.} A task is a deterministic function of its seed, and training,
threshold fitting, both test stages and the later experiments use separate seed ranges (App.~\ref{app:hygiene}).
\emph{(iii) Difficulty.} Switches set the distractor tools, a missing detail the agent must ask for, the number of
customers,
\ifcsname jevres@e21/rl/l3/success\endcsname renamed tools with new task kinds (App.~\ref{app:e21}) \fi and a new
business rule (\gym{}-Hard, App.~\ref{app:e7}). \emph{(iv) Cost.} A scripted customer answers the agent, so an episode costs only the model calls (App.~\ref{app:cost}).
\paragraph{A worked episode.}
\InputIfFileExists{generated/fig1_episode}{}{}
\definecolor{foneRed}{HTML}{B5483B}%
\definecolor{foneBarA}{HTML}{5C6476}%
\definecolor{foneBarB}{HTML}{C9CDD6}%
\providecommand{\foneDec}[4]{%
  \fill[#3] (1.7,#1+0.8) circle (1.55);
  \node[font=\figfs\sffamily\bfseries, text=white] at (1.7,#1+0.8) {\res{fig1/#2/step}};
  \pgfmathsetmacro\fpa{\fbL*\res{fig1/#2/opt/1/pnum}}
  \pgfmathsetmacro\fpb{\fbL*\res{fig1/#2/opt/2/pnum}}
  \fill[foneBarA] (\fbX,#1-0.9) rectangle (\fbX+\fpa,#1+2.5);
  \ifdim\fpb pt>3pt \fill[foneBarB] (\fbX+\fpa,#1-0.9) rectangle (\fbX+\fpa+\fpb,#1+2.5);
    \node[anchor=base, font=\ttfamily\figfs, text=ink] at (\fbX+\fpa+0.5*\fpb,#1) {\res{fig1/#2/opt/2/name}};\fi
  \node[anchor=base, font=\ttfamily\figfs, text=white] at (\fbX+0.5*\fpa,#1) {\res{fig1/#2/opt/1/name}};
  \node[anchor=base west, font=\figfs\sffamily, text=#3] at (\fbX+\fbL+1.6,#1) {#4};}

\begin{figure}[!htbp]
\centering
\begin{tikzpicture}[jevfig]
\def\fW{77}
\node[fnote, anchor=east] (fonelg) at (\fW,-1.75) {decision model};
\fill[sOne] ([xshift=-1.5mm]fonelg.west) circle (0.95);
\node[fnote, anchor=east] (fonelg2) at ([xshift=-4.2mm]fonelg.west) {LLM};
\fill[sTwo] ([xshift=-1.5mm]fonelg2.west) circle (0.95);
\pgfmathtruncatemacro\fn{0+\res{fig1/n_steps}}
\pgfmathsetmacro\fdx{(\fW-14)/(\fn-1)}
\draw[envC!45, line width=0.6pt] (7,-7.2) -- (\fW-7,-7.2);
\foreach \i in {1,...,\fn} {
  \pgfmathsetmacro\fx{7+(\i-1)*\fdx}
  \ifnum\pdfstrcmp{\res{fig1/step/\i/by}}{s1}=0 \def\fcol{sOne}\else\def\fcol{sTwo}\fi
  \node[font=\figfs\sffamily, text=ink2] at (\fx,-4.9) {\i};
  \fill[\fcol] (\fx,-7.2) circle (0.95);
  \node[anchor=base, font=\ttfamily\figfs, text=\fcol] at (\fx,-11.1) {\res{fig1/step/\i/tool}};}
\def\fbX{4.6}\def\fbL{44.6}
\node[fnote, anchor=base west] at (\fbX,-15.6) {probabilities of the top two tools};
\node[fnote, anchor=base west] at (\fbX+\fbL+1.6,-15.6) {confidence};
\foneDec{-19.6}{defer}{sTwo}{\res{fig1/defer/conf} $<\tau$: LLM decides}
\foneDec{-24.4}{accept}{sOne}{\res{fig1/accept/conf} $\ge\tau$: no LLM call}
\draw[ink, line width=0.45pt, dash pattern=on 1.1pt off 0.8pt] ({\fbX+\fbL*\res{fig1/tauhat}},-25.8) --
  ({\fbX+\fbL*\res{fig1/tauhat}},-17.0);
\node[anchor=base east, font=\figfs] at ({\fbX+\fbL*\res{fig1/tauhat}-0.5},-15.6) {$\tau$};
\node[fnote, anchor=base west] at (0,-29.6) {one call: LLM \$\res{fig1/cost/s2}, \res{fig1/latency/s2}\,ms;
  decision model \$\res{fig1/cost/s1}, \res{fig1/latency/s1}\,ms};
\end{tikzpicture}
\caption{\textbf{In one recorded trajectory on a new \gym{} task, the decision model answered only when it was
confident.} At
step~\res{fig1/defer/step}, \jev{}'s confidence was below the fitted threshold $\tau$, so the LLM chose the tool; at
step~\res{fig1/accept/step} it reached $\tau$, and the tool ran without an LLM call. Bars: \jev{}'s probabilities for
the top two of the \res{fig1/n_options} offered tools (\jev{} reports its confidence separately).}
\label{fig:decision}
\end{figure}
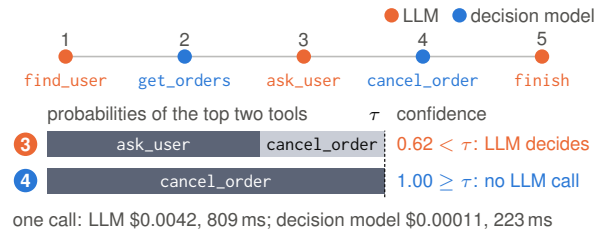
Figure~\ref{fig:decision} follows one recorded trajectory of the chosen combination on a new task, picked by a fixed rule
(threshold $\hat\tau^\star = \res{fig1/tauhat}$). A deferred step pays for both the decision-model call and the LLM call.

\subsection{Example Tasks}
\label{app:jevgym-examples}
Table~\ref{tab:gymexamples} shows two tasks generated at development seeds, which no evaluation uses, with the
evaluation settings. The answer key accepts every open plan step that makes progress and uses only values the agent
has seen. A near-duplicate tool that only reads returns a plausible result that does not advance the task.

\begin{table*}[!t]
\centering\footnotesize
\setlength{\tabcolsep}{4pt}
\begin{tabular}{@{}p{1.8cm}>{\raggedright\arraybackslash}p{13.8cm}@{}}
\toprule
\multicolumn{2}{@{}l}{\textbf{(a) Ask before acting} (\emph{clarify}, which order is missing; development seed 600)}\\
\midrule
Request & ``The customer Lena Wang asked us to cancel their pending order. Please do it.''\\
Hidden state & Lena Wang (U2883, basic tier) has two pending orders, ORD-3872 (monitor, \$817.57) and ORD-4002 (mouse,
\$278.07). Asked which one, the scripted customer replies ``please cancel order ORD-4002 (the mouse).''\\
Plan & \texttt{find\_user(Lena Wang)} $\rightarrow$ \texttt{get\_orders(U2883, pending)} $\rightarrow$
\texttt{ask\_user(order\_id)} $\rightarrow$ \texttt{cancel\_order(ORD-4002, customer\_request)} $\rightarrow$
\texttt{finish(done)}\\
Answer key & At the start: \texttt{find\_user} or \texttt{ask\_user}. After the two lookups: only
\texttt{ask\_user} for the order (three steps to go). There, \texttt{get\_order(ORD-4002)} and a question about the
refund method are detours, and \texttt{finish} is premature.\\
Fails the task & \texttt{cancel\_order(ORD-4002, customer\_request)} before asking, although it names the right order;
cancelling ORD-3872; \texttt{cancel\_order\_item(ORD-4002, mouse, customer\_request)}. Each ends the episode.\\
\midrule
\multicolumn{2}{@{}l}{\textbf{(b) A rule instead of named operations} (\gym{}-Hard, tier rule, ten customers;
development seed 7003)}\\
\midrule
Request & ``Handle the pending order of each of these customers, in order: (1) Dana Ali; (2) Rosa Silva; \ldots; (10)
Priya Garcia. For each customer, look up their pending order and cancel it if the customer is basic tier and the
amount is above \$300.00, or silver tier and the amount is above \$600.00 (never cancel a gold-tier customer's order);
otherwise ship it. \ldots''\\
Hidden state & Close calls among the ten: Rosa Silva (gold, \$583.10) ship; Jia Ali (basic, \$310.70) cancel; Priya
Garcia (silver, \$601.53) cancel; Hana Garcia (silver, \$120.61) ship. The correct total is 3552.19.\\
Plan & The same three steps per customer, the operation set by the rule, then \texttt{finish}: 31 steps within a
budget of \res{e7/setting/max_steps}.\\
Answer key & After each customer's two lookups, the operation the rule gives, or looking up the next customer; the
other operation ends the episode.\\
\bottomrule
\end{tabular}
\caption{Two tasks generated at development seeds with the evaluation settings. ``Answer key'' lists the acceptable
next steps at one state. Detours are allowed but cost a step.}
\label{tab:gymexamples}
\end{table*}
\subsection{Typed Questions and Decider Prompts}
\label{app:prompts}
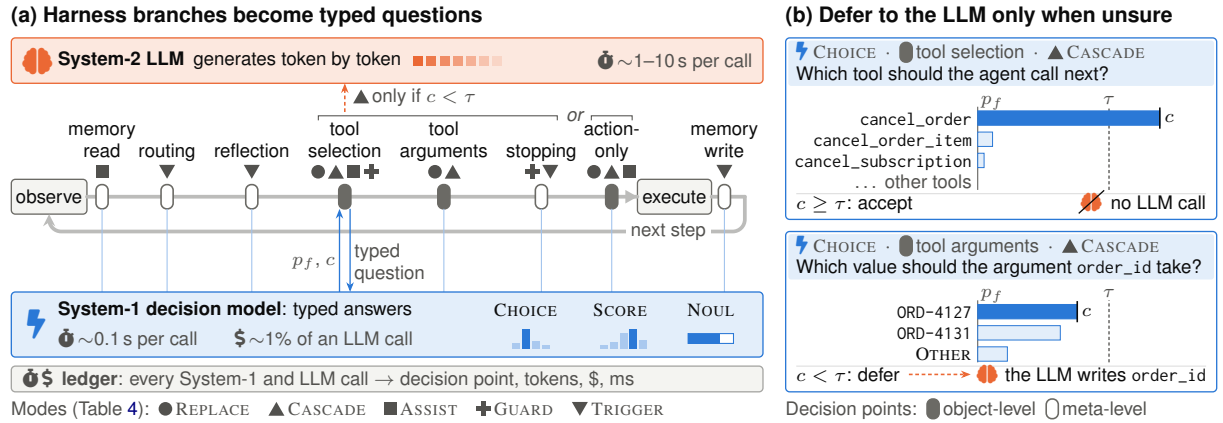
\begin{figure*}[!ht]
\centering
\begin{tikzpicture}[jevfig]
\def\Wa{99.4}   %
\node[fttl] at (0,54.5) {(a) Harness branches become typed questions};

\draw[fbox=sTwo, fill=sTwoT] (0,44.4) rectangle (\Wa,50.4);
\node[anchor=west, text=sTwo, font=\fontsize{10}{10}\selectfont] at (1.5,47.4) {\faBrain};
\node[flbl, anchor=west] (s2t) at (6.2,47.4) {\textbf{\systwo{} LLM}\enspace generates token by token};
\foreach \i in {0,...,6} {\pgfmathtruncatemacro\pct{100-12*\i}
  \fill[sTwo!\pct] ([xshift=1.6mm+\i*1.75mm, yshift=-0.65mm]s2t.east) rectangle ++(1.25,1.3);}
\node[fnote, anchor=east] at (\Wa-1.4,47.4) {\faStopwatch\,${\sim}$1--10\,s per call};

\def\yt{29.4}           %
\def\ymk{32.7}          %
\def\ylb{34.7}          %
\def\xx{\Wa-2.4}        %
\node[fbox=envC, fill=envT, minimum width=10.0mm, minimum height=4.6mm, flbl] (obs) at (5.0,\yt) {observe};
\node[fbox=envC, fill=envT, minimum width=9.8mm, minimum height=4.6mm, flbl] (exe) at (87.7,\yt) {execute};
\draw[frail] (obs.east) -- (exe.west);
\draw[frail] (exe.east) -- (\xx,\yt);
\draw[frail, -{Stealth[length=1.6mm,width=1.6mm]}, rounded corners=1.6mm]
  (\xx,\yt) -- (\xx,24.8) -- (5.0,24.8) -- (obs.south);
\fill[envC!45] ($(exe.west)+(-1.6,-0.9)$) -- (exe.west) -- ($(exe.west)+(-1.6,0.9)$) -- cycle;
\node[fnote, fill=white, inner xsep=0.6mm, inner ysep=0.3mm] at (86.9,24.8) {next step};

\foreach \n/\x/\o/\la/\lb/\ms in {
    mr/12.0/0/memory/read/{Ass},
    rt/20.6/0//routing/{Tri},
    rf/31.8/0//reflection/{Tri},
    ts/44.1/1/tool/selection/{Rep,Cas,Ass,Gua},
    ta/57.2/1/tool/arguments/{Rep,Cas},
    st/70.1/0//stopping/{Gua,Tri},
    ao/79.4/1/action-/only/{Rep,Cas,Ass},
    mw/94.3/0/memory/write/{Tri}} {
  \ifnum\o=1 \node[fobjtap] (\n) at (\x,\yt) {}; \else \node[fmetatap] (\n) at (\x,\yt) {}; \fi
  \figtapmarks{\x}{\ymk}{\ms}
  \node[flbl, anchor=base] at (\x,\ylb) {\lb};
  \node[flbl, anchor=base] at (\x,\ylb+2.9) {\la};
  \ifnum\pdfstrcmp{\n}{ts}=0 \else \draw[sOne!45, line width=0.4pt] (\n.south) -- (\x,16.8); \fi
}
\def\yb{40.1}
\draw[ink2, line width=0.4pt] (40.6,\yb-0.6) |- (72.2,\yb) -- ++(0,-0.6);
\draw[ink2, line width=0.4pt] (76.7,\yb-0.6) |- (82.1,\yb) -- ++(0,-0.6);
\node[fnote, font=\figfs\sffamily\itshape, anchor=base] at (74.45,\yb-0.75) {or};

\draw[farr, draw=sOne] ($(ts.south)+(0.7,0)$) -- (44.8,16.8);
\draw[farr, draw=sOne] (43.4,16.8) -- ($(ts.south)+(-0.7,0)$);
\node[fnote, anchor=west, align=left] at (45.4,20.6) {typed\\question};
\node[fnote, anchor=east, align=right] at (42.8,20.6) {$p_f$, $c$};
\draw[fdefer] (44.1,40.6) -- (44.1,44.4);
\node[fnote, anchor=west] at (45.2,42.55) {\figmki{Cas}\,only if $c<\tau$};

\draw[fbox=sOne, fill=sOneT] (0,8.2) rectangle (\Wa,16.8);
\node[text=sOne, font=\fontsize{11}{11}\selectfont] at (3.3,12.5) {\faBolt};
\node[flbl, anchor=west] at (6.2,14.4) {\textbf{\sysone{} decision model}: typed answers};
\node[fnote, anchor=west] at (6.2,10.6) {\faStopwatch\,${\sim}$0.1\,s per call\qquad\faDollarSign\,${\sim}$1\% of an LLM call};
\node[flbl] at (68.0,14.4) {\figrm{\Choice}};
\foreach \h [count=\i] in {0.7,2.6,1.0,0.5}
  {\ifnum\i=2 \def\cc{sOne}\else\def\cc{sOne!40}\fi \fill[\cc] (65.0+\i*1.3,9.3) rectangle ++(0.95,\h);}
\node[flbl] at (80.5,14.4) {\figrm{\Score}};
\foreach \h [count=\i] in {0.4,0.9,2.0,2.6,1.1}
  {\ifnum\i=4 \def\cc{sOne}\else\def\cc{sOne!40}\fi \fill[\cc] (76.6+\i*1.3,9.3) rectangle ++(0.95,\h);}
\node[flbl] at (92.5,14.4) {\figrm{\Noul}};
\draw[sOne, line width=0.4pt, fill=white] (89.5,10.0) rectangle (95.5,11.3);
\fill[sOne] (89.5,10.0) rectangle (93.7,11.3);

\draw[fbox=envC!60, fill=envT] (0,3.6) rectangle (\Wa,7.0);
\node[fnote, anchor=west] at (1.5,5.3) {\faStopwatch\,\faDollarSign\enspace\textbf{ledger}: every \sysone{} and LLM call $\to$ decision point, tokens, \$, ms};

\node[fnote, anchor=base west] at (0,0.6) {%
  Modes (Table~\ref{tab:design}):\enspace
  \figmki{Rep}\,\figrm{\modeRep}\quad \figmki{Cas}\,\figrm{\modeCas}\quad \figmki{Ass}\,\figrm{\modeAss}\quad
  \figmki{Gua}\,\figrm{\modeGua}\quad \figmki{Tri}\,\figrm{\modeTri}};

\begin{scope}[shift={(102.4,0)}]
\def\W{57.0}    %
\def\xo{25.4}   %
\def\bw{28.0}   %
\def\tpos{0.62} %
\pgfmathsetmacro\xt{\xo+\tpos*\bw}
\node[fttl] at (0,54.5) {(b) Defer to the LLM only when unsure};
\newcommand{\qcard}[4]{%
  \draw[fbox=sOne, fill=white] (0,#2) rectangle (\W,#1);
  \fill[sOneT, rounded corners=1.1pt] (0.35,#1-6.35) rectangle (\W-0.35,#1-0.35);
  \node[fnote, anchor=base west] at (1.4,#1-2.65) {#3};
  \node[flbl, anchor=base west] at (1.4,#1-5.5) {#4};
  \node[fnote, anchor=base west] at (\xo+0.4,#1-8.7) {$p_f$};
  \node[fnote, anchor=base] at (\xt,#1-8.7) {$\tau$};
  \draw[ink2, line width=0.3pt] (\xo,#2+3.9) -- (\xo,#1-9.1);
  \draw[ink2, line width=0.5pt, dash pattern=on 1.2pt off 0.9pt] (\xt,#2+3.9) -- (\xt,#1-9.1);
  \draw[black!14, line width=0.3pt] (1.4,#2+3.4) -- (\W-1.4,#2+3.4);}
\newcommand{\qbar}[5]{%
  \pgfmathsetmacro\yy{#1-10.6-(#2-1)*2.8}\pgfmathsetmacro\xe{\xo+#4*\bw}
  \node[anchor=base east, font=\ttfamily\figfs] at (\xo-0.8,\yy-0.85) {#3};
  \ifnum#5=1 \draw[fill=sOne, draw=sOne, line width=0.3pt] (\xo,\yy-0.95) rectangle (\xe,\yy+0.95);
    \draw[ink, line width=0.6pt] (\xe,\yy-1.4) -- (\xe,\yy+1.4);
    \node[fnote, text=ink, anchor=west] at (\xe+0.6,\yy) {$c$};
  \else \draw[fill=sOneT, draw=sOne, line width=0.3pt] (\xo,\yy-0.95) rectangle (\xe,\yy+0.95); \fi}
\def\T{50.4}\def\B{26.4}
\qcard{\T}{\B}{\textcolor{sOne}{\faBolt}\;\figrm{\Choice} \,$\cdot$\, \figtapi{obj}\,tool selection
  \,$\cdot$\, \figmki{Cas}\,\figrm{\modeCas}}{Which tool should the agent call next?}
\qbar{\T}{1}{cancel\_order}{0.86}{1}
\qbar{\T}{2}{cancel\_order\_item}{0.07}{0}
\qbar{\T}{3}{cancel\_subscription}{0.03}{0}
\node[fnote, anchor=base east] at (\xo-0.8,\T-19.0-0.85) {\ldots\ other tools};
\node[flbl, anchor=base west] at (1.4,\B+1.4) {$c \ge \tau$: accept};
\node[flbl, anchor=base east] (nsv) at (\W-1.4,\B+1.4) {no LLM call};
\node[anchor=base east, text=sTwo, font=\figfs] (br1) at ([xshift=-1.2mm]nsv.base west) {\faBrain};
\draw[ink, line width=0.5pt] ([shift={(-0.3mm,-0.3mm)}]br1.south west) -- ([shift={(0.3mm,0.3mm)}]br1.north east);
\def\T{24.8}\def\B{3.6}
\qcard{\T}{\B}{\textcolor{sOne}{\faBolt}\;\figrm{\Choice} \,$\cdot$\, \figtapi{obj}\,tool arguments
  \,$\cdot$\, \figmki{Cas}\,\figrm{\modeCas}}{Which value should the argument \texttt{order\_id} take?}
\qbar{\T}{1}{ORD-4127}{0.47}{1}
\qbar{\T}{2}{ORD-4131}{0.39}{0}
\qbar{\T}{3}{\textrm{\textsc{Other}}}{0.14}{0}
\node[flbl, anchor=base west] (dv) at (1.4,\B+1.4) {$c < \tau$: defer};
\node[flbl, anchor=base east] (wr) at (\W-1.4,\B+1.4) {the LLM writes \texttt{order\_id}};
\node[anchor=base east, text=sTwo, font=\figfs] (br2) at ([xshift=-1.0mm]wr.base west) {\faBrain};
\draw[fdefer] ([xshift=1.2mm, yshift=0.9mm]dv.base east) -- ([xshift=-0.8mm, yshift=0.9mm]br2.base west);
\node[fnote, anchor=base west] at (0,0.6) {%
  Decision points:\enspace\figtapi{obj}\,object-level\enspace \figtapi{meta}\,meta-level};
\end{scope}
\end{tikzpicture}
\caption{\textbf{Each harness branch over a computable option set becomes a typed question to a fast decision
model, whose answer decides when and how the LLM is called.} (a)~The eight decision points in harness order
(WHERE) and the modes implemented at each (HOW). (b)~Two \modeCas{} decisions (illustrative distributions; Eq.~\ref{eq:tau} sets $\tau$): \sysone{} is
sure of the tool but not of the copied \texttt{order\_id}; in this factorised path the accepted tool skips tool
generation and only the deferred argument needs an LLM call. ${\sim}$:~indicative (\jev{}: vendor-reported;
measured ratios vary with state length, backend and reads).}
\label{fig:loop}
\end{figure*}
Figure~\ref{fig:loop} places the eight decision points in the harness loop. App.~\ref{app:prompts-full} gives the
question asked at each point, the state text and every prompt. An LLM decider read through log-probabilities sums
the probabilities of tokens equal to an option label (after stripping whitespace and trailing punctuation), gives
absent labels $10^{-6}$ and renormalises. Without log-probabilities the generated token takes all the mass, and a
verbalised confidence goes to the answer with the rest spread uniformly. Trained models use the full-logit readout of
App.~\ref{app:training}.

\subsection{Prompts}
\label{app:prompts-full}
\providecommand{\promptline}[1]{\noindent\colorbox{gray!7}{\makebox[\columnwidth][l]{\textcolor{gray!55}{\vrule width 1pt}\hspace{3pt}\strut#1}}\par}
\providecommand{\promptgap}{{\setlength{\baselineskip}{0.5\baselineskip}\noindent\colorbox{gray!7}{\makebox[\columnwidth][l]{\textcolor{gray!55}{\vrule width 1pt height \dimexpr\baselineskip-\dp\strutbox\relax depth \dp\strutbox}}}\par}}
\ifnum\maxdeadcycles<1000 \global\maxdeadcycles=1000 \fi
This subsection lists every prompt our code sends to a model in the experiments, generated from the code by \texttt{scripts/paper\_prompts.py}. Examples come from development seed 600, which no evaluation uses. \textit{$\langle$name$\rangle$} marks a value filled in at run time, an italic [note] marks our shortening, and a hook marks a line we wrapped. No prompt contains the label of the decision it asks about. The answer key and every label stay in the decision log. Training data and teacher labels add no prompt of their own, since teachers answer the same requests and training rows use the trained-model format.

\paragraph{Decision-model inputs.}
Every decision model reads the same state text and one typed question, a choice among options or a yes/no question. The reference loop builds the state below. The questions come next, then the request that carries them.

\par\smallskip{\footnotesize\raggedright\interlinepenalty=10000\noindent\textbf{P1. Decision-model state (reference loop).} Source: \texttt{src/\allowbreak{}jevharness/\allowbreak{}harness/\allowbreak{}agent.py: JevAgent.\_s1\_state}, \texttt{src/\allowbreak{}jevharness/\allowbreak{}harness/\allowbreak{}config.py}. MEMORY and LATEST REFLECTION appear only when the arm has them. The history keeps the last 8 steps and cuts each result to 400 characters. The current observation is cut to 2,000 characters and the whole state to 12,000. A cut keeps the head and the tail around \texttt{...[truncated]...}.\par}\nobreak\vspace{1pt}\nobreak
\begingroup\ttfamily\scriptsize\setlength{\fboxsep}{0pt}\setlength{\parskip}{0pt}%
\promptline{TASK:\ \textrm{\textit{$\langle$task instruction$\rangle$}}}\nobreak
\promptgap
\promptline{MEMORY:}
\promptline{\textrm{\textit{$\langle$recalled memories, one per line$\rangle$}}}
\promptgap
\promptline{LATEST\ REFLECTION:}
\promptline{\textrm{\textit{$\langle$the agent LLM's latest reflection$\rangle$}}}
\promptgap
\promptline{HISTORY:}
\promptline{[\textrm{\textit{$\langle$step$\rangle$}}]\ \textrm{\textit{$\langle$tool(arguments)$\rangle$}}\ -{}>{}\ \textrm{\textit{$\langle$result$\rangle$}}}
\promptgap
\promptline{CURRENT\ OBSERVATION:}\nobreak
\promptline{\textrm{\textit{$\langle$latest tool result$\rangle$}}}\nobreak
\endgroup\smallskip

\par\smallskip{\footnotesize\raggedright\interlinepenalty=10000\noindent\textbf{P2. State at development seed 600, step 2.} Source: \texttt{src/\allowbreak{}jevharness/\allowbreak{}harness/\allowbreak{}agent.py}, \texttt{src/\allowbreak{}jevharness/\allowbreak{}bench/\allowbreak{}suites/\allowbreak{}jevgym.py}. A \gym{} clarify task in the evaluation settings (60 look-alike tools, 20 log lines per lookup), after the agent looked the customer up and listed the pending orders. Our shortening of the log lines is marked in italics. The model sees all of them.\par}\nobreak\vspace{1pt}\nobreak
\begingroup\ttfamily\scriptsize\setlength{\fboxsep}{0pt}\setlength{\parskip}{0pt}%
\promptline{TASK:\ The\ customer\ Lena\ Wang\ asked\ us\ to\ cancel\ their}\nobreak
\promptline{\rlap{\textcolor{gray}{$\hookrightarrow$}}\phantom{xx}pending\ order.\ Please\ do\ it.}
\promptgap
\promptline{HISTORY:}
\promptline{[0]\ find\_user(\{\textquotedbl{}name\textquotedbl{}:\ \textquotedbl{}Lena\ Wang\textquotedbl{}\})\ -{}>{}\ \{\textquotedbl{}user\_id\textquotedbl{}:}
\promptline{\rlap{\textcolor{gray}{$\hookrightarrow$}}\phantom{xx}\textquotedbl{}U2883\textquotedbl{},\ \textquotedbl{}name\textquotedbl{}:\ \textquotedbl{}Lena\ Wang\textquotedbl{},\ \textquotedbl{}tier\textquotedbl{}:\ \textquotedbl{}basic\textquotedbl{}\}}
\promptline{log:\ heartbeat\ ok\ \#911760}
\promptline{\textrm{\textit{[4 more log lines]}}}
\promptline{log:\ heartb}
\promptline{...[truncated]...}
\promptline{artbeat\ ok\ \#777025}
\promptline{\textrm{\textit{[6 more log lines]}}}
\promptline{log:\ heartbeat\ ok\ \#417838}
\promptline{[1]\ get\_orders(\{\textquotedbl{}user\_id\textquotedbl{}:\ \textquotedbl{}U2883\textquotedbl{},\ \textquotedbl{}status\textquotedbl{}:\ \textquotedbl{}pending\textquotedbl{}\})}
\promptline{\rlap{\textcolor{gray}{$\hookrightarrow$}}\phantom{xx}-{}>{}\ [\{\textquotedbl{}order\_id\textquotedbl{}:\ \textquotedbl{}ORD-{}3872\textquotedbl{},\ \textquotedbl{}user\_id\textquotedbl{}:\ \textquotedbl{}U2883\textquotedbl{},\ \textquotedbl{}item\textquotedbl{}:}
\promptline{\rlap{\textcolor{gray}{$\hookrightarrow$}}\phantom{xx}\textquotedbl{}monitor\textquotedbl{},\ \textquotedbl{}status\textquotedbl{}:\ \textquotedbl{}pending\textquotedbl{},\ \textquotedbl{}amount\textquotedbl{}:\ 817.57\},}
\promptline{\rlap{\textcolor{gray}{$\hookrightarrow$}}\phantom{xx}\{\textquotedbl{}order\_id\textquotedbl{}:\ \textquotedbl{}ORD-{}4002\textquotedbl{},\ \textquotedbl{}user\_id\textquotedbl{}:\ \textquotedbl{}U2883\textquotedbl{},\ \textquotedbl{}item\textquotedbl{}:}
\promptline{\rlap{\textcolor{gray}{$\hookrightarrow$}}\phantom{xx}\textquotedbl{}mouse\textquotedbl{},\ \textquotedbl{}status\textquotedbl{}:\ \textquotedbl{}pending\textquotedbl{},\ \textquotedbl{}amount\textquotedbl{}:\ 27}
\promptline{...[truncated]...}
\promptline{\phantom{x}heartbeat\ ok\ \#885494}
\promptline{log:\ heartbeat\ ok\ \#605733}
\promptline{\textrm{\textit{[5 more log lines]}}}
\promptline{log:\ heartbeat\ ok\ \#947812}
\promptgap
\promptline{CURRENT\ OBSERVATION:}
\promptline{[\{\textquotedbl{}order\_id\textquotedbl{}:\ \textquotedbl{}ORD-{}3872\textquotedbl{},\ \textquotedbl{}user\_id\textquotedbl{}:\ \textquotedbl{}U2883\textquotedbl{},\ \textquotedbl{}item\textquotedbl{}:}
\promptline{\rlap{\textcolor{gray}{$\hookrightarrow$}}\phantom{xx}\textquotedbl{}monitor\textquotedbl{},\ \textquotedbl{}status\textquotedbl{}:\ \textquotedbl{}pending\textquotedbl{},\ \textquotedbl{}amount\textquotedbl{}:\ 817.57\},}
\promptline{\rlap{\textcolor{gray}{$\hookrightarrow$}}\phantom{xx}\{\textquotedbl{}order\_id\textquotedbl{}:\ \textquotedbl{}ORD-{}4002\textquotedbl{},\ \textquotedbl{}user\_id\textquotedbl{}:\ \textquotedbl{}U2883\textquotedbl{},\ \textquotedbl{}item\textquotedbl{}:}
\promptline{\rlap{\textcolor{gray}{$\hookrightarrow$}}\phantom{xx}\textquotedbl{}mouse\textquotedbl{},\ \textquotedbl{}status\textquotedbl{}:\ \textquotedbl{}pending\textquotedbl{},\ \textquotedbl{}amount\textquotedbl{}:\ 278.07\}]}
\promptline{log:\ heartbeat\ ok\ \#149716}
\promptline{\textrm{\textit{[18 more log lines]}}}\nobreak
\promptline{log:\ heartbeat\ ok\ \#947812}\nobreak
\endgroup\smallskip

\par\smallskip{\footnotesize\raggedright\interlinepenalty=10000\noindent\textbf{P3. Questions at tool calls.} Source: \texttt{src/\allowbreak{}jevharness/\allowbreak{}harness/\allowbreak{}prompts.py}, \texttt{src/\allowbreak{}jevharness/\allowbreak{}harness/\allowbreak{}agent.py}, \texttt{src/\allowbreak{}jevharness/\allowbreak{}harness/\allowbreak{}tools.py}. Options carry the first line of each tool description. The decision model fills an argument only when its values are bounded, as with an enum or a boolean, or can be copied from the state. The copy options are the spans the code extracts. Free text always goes to the LLM.\par}\nobreak\vspace{1pt}\nobreak
\begingroup\ttfamily\scriptsize\setlength{\fboxsep}{0pt}\setlength{\parskip}{0pt}%
\promptline{[Tool\ choice:\ one\ of\ 69\ options]}\nobreak
\promptline{Which\ tool\ should\ the\ agent\ call\ next?}
\promptline{get\_orders\_archive\ -{}-{}\ List\ a\ customer\textquotesingle{}s\ orders\ from\ the}
\promptline{\rlap{\textcolor{gray}{$\hookrightarrow$}}\phantom{xx}archive\ of\ a\ past\ calendar\ year\ (orders\ older\ than\ 12}
\promptline{\rlap{\textcolor{gray}{$\hookrightarrow$}}\phantom{xx}months\ only).}
\promptline{set\_reminder\ -{}-{}\ Set\ a\ reminder.}
\promptline{ship\_replacement\ -{}-{}\ Ship\ a\ free\ replacement\ for\ a\ damaged,}
\promptline{\rlap{\textcolor{gray}{$\hookrightarrow$}}\phantom{xx}already\ delivered\ order.}
\promptline{\textrm{\textit{[65 more tools]}}}
\promptline{finish\ -{}-{}\ Call\ when\ the\ task\ is\ complete.\ Provide\ the\ final}
\promptline{\rlap{\textcolor{gray}{$\hookrightarrow$}}\phantom{xx}answer\ (or\ a\ short\ summary).}
\promptgap
\promptline{[Tool\ arguments:\ one\ question\ per\ argument]}
\promptline{Which\ value\ should\ the\ argument\ \textasciigrave{}\textrm{\textit{$\langle$argument$\rangle$}}\textasciigrave{}\ of\ tool}
\promptline{\rlap{\textcolor{gray}{$\hookrightarrow$}}\phantom{xx}\textasciigrave{}\textrm{\textit{$\langle$tool$\rangle$}}\textasciigrave{}\ take?\ (\textrm{\textit{$\langle$description in the schema$\rangle$}})}
\promptline{Example:\ Which\ value\ should\ the\ argument\ \textasciigrave{}name\textasciigrave{}\ of\ tool}
\promptline{\rlap{\textcolor{gray}{$\hookrightarrow$}}\phantom{xx}\textasciigrave{}find\_user\textasciigrave{}\ take?\ (customer\ full\ name)}
\promptline{Options:\ the\ 3\ values\ copied\ from\ the\ state\ (Lena\ Wang,}
\promptline{\rlap{\textcolor{gray}{$\hookrightarrow$}}\phantom{xx}Lena,\ Wang),\ then}
\promptline{<{}OTHER:\ none\ of\ these\ /\ needs\ generation>{}}
\promptline{Example:\ Which\ value\ should\ the\ argument\ \textasciigrave{}status\textasciigrave{}\ of\ tool}
\promptline{\rlap{\textcolor{gray}{$\hookrightarrow$}}\phantom{xx}\textasciigrave{}get\_orders\textasciigrave{}\ take?\ ()}
\promptline{Options:\ the\ enum\ all,\ pending,\ shipped,\ delivered,}
\promptline{\rlap{\textcolor{gray}{$\hookrightarrow$}}\phantom{xx}cancelled}
\promptgap
\promptline{[Tool\ guard:\ yes/no]}
\promptline{The\ agent\ proposes\ to\ call\ \textasciigrave{}\textrm{\textit{$\langle$tool$\rangle$}}\textasciigrave{}\ with\ arguments}
\promptline{\rlap{\textcolor{gray}{$\hookrightarrow$}}\phantom{xx}\textrm{\textit{$\langle$arguments as JSON$\rangle$}}.\ Is\ this\ the\ right\ next\ step\ toward}\nobreak
\promptline{\rlap{\textcolor{gray}{$\hookrightarrow$}}\phantom{xx}completing\ the\ task?}\nobreak
\endgroup\smallskip

\par\smallskip{\footnotesize\raggedright\interlinepenalty=10000\noindent\textbf{P4. Questions at the other decision points.} Source: \texttt{src/\allowbreak{}jevharness/\allowbreak{}harness/\allowbreak{}prompts.py}, \texttt{src/\allowbreak{}jevharness/\allowbreak{}harness/\allowbreak{}memory.py}, \texttt{src/\allowbreak{}jevharness/\allowbreak{}harness/\allowbreak{}agent.py}. The stop guard sees the whole proposed answer when it has at most 800 characters, and otherwise its first 200 and last 600. A memory option is labelled with the first 60 characters of the memory, and the full text is its description.\par}\nobreak\vspace{1pt}\nobreak
\begingroup\ttfamily\scriptsize\setlength{\fboxsep}{0pt}\setlength{\parskip}{0pt}%
\promptline{[Memory\ write:\ the\ state\ ends\ with\ the\ candidate]}\nobreak
\promptline{CANDIDATE\ MEMORY:}
\promptline{\textrm{\textit{$\langle$candidate$\rangle$}}}
\promptline{\phantom{xx}candidate:\ \textrm{\textit{$\langle$tool$\rangle$}}(\textrm{\textit{$\langle$arguments$\rangle$}})\ -{}>{}\ \textrm{\textit{$\langle$result$\rangle$}}}
\promptline{Will\ this\ information\ likely\ be\ needed\ later\ to\ complete}
\promptline{\rlap{\textcolor{gray}{$\hookrightarrow$}}\phantom{xx}the\ task\ (facts,\ IDs,\ constraints,\ user\ preferences,}
\promptline{\rlap{\textcolor{gray}{$\hookrightarrow$}}\phantom{xx}results)?}
\promptgap
\promptline{[Memory\ read:\ one\ option\ per\ stored\ memory]}
\promptline{Which\ stored\ memory\ is\ most\ relevant\ to\ the\ current\ step?}
\promptline{[m\textrm{\textit{$\langle$id$\rangle$}}]\ \textrm{\textit{$\langle$its first 60 characters$\rangle$}}\ -{}-{}\ \textrm{\textit{$\langle$memory text$\rangle$}}}
\promptline{(none\ of\ these\ memories\ is\ relevant)}
\promptgap
\promptline{[Memory\ read\ trigger:\ the\ state\ ends\ with]}
\promptline{STORED\ MEMORIES:\ \textrm{\textit{$\langle$n$\rangle$}}\ earlier\ observations\ (not\ shown)}
\promptline{Does\ the\ agent\ need\ information\ it\ stored\ earlier\ (facts,}
\promptline{\rlap{\textcolor{gray}{$\hookrightarrow$}}\phantom{xx}IDs,\ results\ from\ steps\ no\ longer\ in\ the\ recent\ history)}
\promptline{\rlap{\textcolor{gray}{$\hookrightarrow$}}\phantom{xx}to\ choose\ its\ next\ step\ correctly?}
\promptgap
\promptline{[Memory\ read\ guard:\ the\ state\ ends\ with\ the\ selection]}
\promptline{SELECTED\ MEMORIES:}
\promptline{-{}\ [m\textrm{\textit{$\langle$id$\rangle$}}]\ \textrm{\textit{$\langle$memory text$\rangle$}}}
\promptline{The\ agent\ proposes\ to\ recall\ these\ stored\ memories\ for\ its}
\promptline{\rlap{\textcolor{gray}{$\hookrightarrow$}}\phantom{xx}next\ step:\ \textrm{\textit{$\langle$label$\rangle$}},\ \textrm{\textit{$\langle$label$\rangle$}}.\ Are\ these\ the\ memories\ it}
\promptline{\rlap{\textcolor{gray}{$\hookrightarrow$}}\phantom{xx}needs,\ with\ no\ needed\ memory\ missing?}
\promptgap
\promptline{[Reflection]}
\promptline{Should\ the\ agent\ pause\ and\ re-{}check\ before\ its\ next\ action:}
\promptline{\rlap{\textcolor{gray}{$\hookrightarrow$}}\phantom{xx}is\ it\ stuck\ or\ off\ track,\ or\ likely\ to\ make\ a\ wrong}
\promptline{\rlap{\textcolor{gray}{$\hookrightarrow$}}\phantom{xx}action\ or\ give\ a\ wrong\ final\ answer\ next?}
\promptline{Earlier\ wording\ (measurement\ arms):}
\promptline{Is\ the\ agent\ stuck,\ repeating\ itself,\ or\ did\ the\ last}
\promptline{\rlap{\textcolor{gray}{$\hookrightarrow$}}\phantom{xx}action\ fail\ or\ move\ away\ from\ the\ goal?}
\promptgap
\promptline{[Stop\ guard,\ when\ the\ LLM\ calls\ finish]}
\promptline{The\ agent\ wants\ to\ stop\ and\ submit\ the\ final\ answer}
\promptline{\rlap{\textcolor{gray}{$\hookrightarrow$}}\phantom{xx}\textrm{\textit{$\langle$answer, JSON-quoted$\rangle$}}.\ Is\ the\ task\ actually\ complete,\ and}
\promptline{\rlap{\textcolor{gray}{$\hookrightarrow$}}\phantom{xx}is\ this\ answer\ correct?}
\promptline{[Stop\ trigger,\ at\ every\ other\ step]}
\promptline{Has\ the\ task\ already\ been\ fully\ completed,\ so\ the\ agent}
\promptline{\rlap{\textcolor{gray}{$\hookrightarrow$}}\phantom{xx}should\ stop\ now?}
\promptgap
\promptline{[Routing]}
\promptline{Does\ the\ next\ step\ require\ careful\ multi-{}step\ reasoning,}
\promptline{\rlap{\textcolor{gray}{$\hookrightarrow$}}\phantom{xx}planning,\ or\ recovery\ from\ an\ error\ (i.e.,\ a\ strong}
\promptline{\rlap{\textcolor{gray}{$\hookrightarrow$}}\phantom{xx}model)\ rather\ than\ a\ routine\ action?}
\promptgap
\promptline{[Action-{}only\ environments:\ choice\ and\ guard]}
\promptline{Which\ action\ should\ the\ agent\ take\ next?}
\promptline{Options:\ the\ admissible\ actions}
\promptline{The\ agent\ proposes\ to\ take\ the\ action\ \textasciigrave{}\textrm{\textit{$\langle$action$\rangle$}}\textasciigrave{}.\ Is\ this}
\promptline{\rlap{\textcolor{gray}{$\hookrightarrow$}}\phantom{xx}the\ right\ next\ step\ toward\ completing\ the\ task?}
\promptgap
\promptline{[Verification\ triggers:\ does\ the\ LLM\textquotesingle{}s\ proposal\ need\ a}
\promptline{\rlap{\textcolor{gray}{$\hookrightarrow$}}\phantom{xx}second\ look?]}
\promptline{The\ agent\ proposes\ to\ call\ \textasciigrave{}\textrm{\textit{$\langle$tool$\rangle$}}\textasciigrave{}\ with\ arguments}
\promptline{\rlap{\textcolor{gray}{$\hookrightarrow$}}\phantom{xx}\textrm{\textit{$\langle$arguments$\rangle$}}.\ Should\ the\ agent\ re-{}check\ this\ choice\ of}
\promptline{\rlap{\textcolor{gray}{$\hookrightarrow$}}\phantom{xx}tool\ before\ the\ call\ runs:\ could\ another\ tool\ be\ the}
\promptline{\rlap{\textcolor{gray}{$\hookrightarrow$}}\phantom{xx}right\ next\ step?}
\promptline{The\ agent\ proposes\ to\ call\ \textasciigrave{}\textrm{\textit{$\langle$tool$\rangle$}}\textasciigrave{}\ with\ arguments}
\promptline{\rlap{\textcolor{gray}{$\hookrightarrow$}}\phantom{xx}\textrm{\textit{$\langle$arguments$\rangle$}}.\ Should\ the\ agent\ re-{}check\ these\ argument}
\promptline{\rlap{\textcolor{gray}{$\hookrightarrow$}}\phantom{xx}values\ before\ the\ call\ runs:\ could\ a\ value\ be\ wrong\ or}
\promptline{\rlap{\textcolor{gray}{$\hookrightarrow$}}\phantom{xx}missing?}
\promptline{The\ agent\ proposes\ to\ take\ the\ action\ \textasciigrave{}\textrm{\textit{$\langle$action$\rangle$}}\textasciigrave{}.\ Should}
\promptline{\rlap{\textcolor{gray}{$\hookrightarrow$}}\phantom{xx}the\ agent\ re-{}check\ this\ action\ before\ it\ runs:\ could}\nobreak
\promptline{\rlap{\textcolor{gray}{$\hookrightarrow$}}\phantom{xx}another\ action\ be\ the\ right\ next\ step?}\nobreak
\endgroup\smallskip

\par\smallskip{\footnotesize\raggedright\interlinepenalty=10000\noindent\textbf{P5. \jev{} request (wire format).} Source: \texttt{src/\allowbreak{}jevharness/\allowbreak{}decision/\allowbreak{}jev.py: question\_to\_wire}. The example keeps 2 of the 69 tool options and finish. In the reference loop each decision point is one request. A Noul (yes/no) question carries \texttt{criteria} only when the harness gives them, as mini-SWE-agent and pi do. \jev{}, OpenJev, StartLux-Decision and our trained models all receive this request.\par}\nobreak\vspace{1pt}\nobreak
\begingroup\ttfamily\scriptsize\setlength{\fboxsep}{0pt}\setlength{\parskip}{0pt}%
\promptline{\{}\nobreak
\promptline{\phantom{x}\textquotedbl{}state\textquotedbl{}:\ \textquotedbl{}\textrm{\textit{[the state shown above]}}\textquotedbl{},}
\promptline{\phantom{x}\textquotedbl{}model\textquotedbl{}:\ \textquotedbl{}typesafe/jev-{}1.13\textquotedbl{},}
\promptline{\phantom{x}\textquotedbl{}questions\textquotedbl{}:\ \{}
\promptline{\phantom{xx}\textquotedbl{}q0\textquotedbl{}:\ \{}
\promptline{\phantom{xxx}\textquotedbl{}type\textquotedbl{}:\ \textquotedbl{}choice\textquotedbl{},}
\promptline{\phantom{xxx}\textquotedbl{}instructions\textquotedbl{}:\ \textquotedbl{}Which\ tool\ should\ the\ agent\ call\ next?\textquotedbl{},}
\promptline{\phantom{xxx}\textquotedbl{}criteria\textquotedbl{}:\ \{}
\promptline{\phantom{xxxx}\textquotedbl{}get\_orders\_archive\textquotedbl{}:\ \textquotedbl{}List\ a\ customer\textquotesingle{}s\ orders\ from}
\promptline{\rlap{\textcolor{gray}{$\hookrightarrow$}}\phantom{xx}the\ archive\ of\ a\ past\ calendar\ year\ (orders\ older\ than\ 12}
\promptline{\rlap{\textcolor{gray}{$\hookrightarrow$}}\phantom{xx}months\ only).\textquotedbl{},}
\promptline{\phantom{xxxx}\textquotedbl{}set\_reminder\textquotedbl{}:\ \textquotedbl{}Set\ a\ reminder.\textquotedbl{},}
\promptline{\phantom{xxxx}\textquotedbl{}finish\textquotedbl{}:\ \textquotedbl{}Call\ when\ the\ task\ is\ complete.\ Provide\ the}
\promptline{\rlap{\textcolor{gray}{$\hookrightarrow$}}\phantom{xx}final\ answer\ (or\ a\ short\ summary).\textquotedbl{}}
\promptline{\phantom{xxx}\}}
\promptline{\phantom{xx}\},}
\promptline{\phantom{xx}\textquotedbl{}q1\textquotedbl{}:\ \{}
\promptline{\phantom{xxx}\textquotedbl{}type\textquotedbl{}:\ \textquotedbl{}noul\textquotedbl{},}
\promptline{\phantom{xxx}\textquotedbl{}instructions\textquotedbl{}:\ \textquotedbl{}Has\ the\ task\ already\ been\ fully}
\promptline{\rlap{\textcolor{gray}{$\hookrightarrow$}}\phantom{xx}completed,\ so\ the\ agent\ should\ stop\ now?\textquotedbl{}}
\promptline{\phantom{xx}\}}
\promptline{\phantom{x}\}}\nobreak
\promptline{\}}\nobreak
\endgroup\smallskip

\paragraph{How each decision model reads a question.}
\jev{}, OpenJev and StartLux-Decision receive the request above and build their prompts inside their own servers \citep{typesafe2026jev,openjev2026,startlux2026decision}. Every other decision model gets one of the prompts below, where \textit{$\langle$state$\rangle$} is the state text shown above.

\par\smallskip{\footnotesize\raggedright\interlinepenalty=10000\noindent\textbf{P6. Intern-Decision (one pass for all questions).} Source: \texttt{src/\allowbreak{}jevharness/\allowbreak{}decision/\allowbreak{}intern\_decision.py: compile\_row}. The vendor's prompt compiler, copied with its licence. The state is JSON-encoded. Each field's answer is read at its \texttt{<{}decision>{}} placeholder, so k questions cost one forward pass.\par}\nobreak\vspace{1pt}\nobreak
\begingroup\ttfamily\scriptsize\setlength{\fboxsep}{0pt}\setlength{\parskip}{0pt}%
\promptline{[system]}\nobreak
\promptline{You\ are\ a\ careful\ decision\ assistant.\ Use\ the\ state\ and}
\promptline{\rlap{\textcolor{gray}{$\hookrightarrow$}}\phantom{xx}decision\ schema\ in\ the\ user\ message\ to\ make\ the\ requested}
\promptline{\rlap{\textcolor{gray}{$\hookrightarrow$}}\phantom{xx}decisions.\ For\ every\ field,\ choose\ exactly\ one\ answer}
\promptline{\rlap{\textcolor{gray}{$\hookrightarrow$}}\phantom{xx}symbol\ (e.g.\ A,\ B,\ C,\ ...)\ from\ its\ listed\ options\ and}
\promptline{\rlap{\textcolor{gray}{$\hookrightarrow$}}\phantom{xx}return\ one\ valid\ JSON\ object\ mapping\ each\ field\ name\ to}
\promptline{\rlap{\textcolor{gray}{$\hookrightarrow$}}\phantom{xx}its\ chosen\ symbol.\ Use\ the\ field\ names\ and\ symbols}
\promptline{\rlap{\textcolor{gray}{$\hookrightarrow$}}\phantom{xx}exactly\ as\ given.\ Do\ not\ include\ explanations,\ Markdown,}
\promptline{\rlap{\textcolor{gray}{$\hookrightarrow$}}\phantom{xx}or\ extra\ text.}
\promptline{[user]}
\promptline{Return\ one\ answer\ for\ every\ field\ using\ the\ supplied\ answer}
\promptline{\rlap{\textcolor{gray}{$\hookrightarrow$}}\phantom{xx}symbols.}
\promptgap
\promptline{\#\#\ State}
\promptline{\textquotedbl{}\textrm{\textit{$\langle$state$\rangle$}}\textquotedbl{}}
\promptline{\#\#\ Decision\ schema}
\promptline{tool\_select:\ Which\ tool\ should\ the\ agent\ call\ next?}
\promptline{\phantom{xxxx}A\ =\ get\_orders\_archive:\ List\ a\ customer\textquotesingle{}s\ orders\ from}
\promptline{\rlap{\textcolor{gray}{$\hookrightarrow$}}\phantom{xx}the\ archive\ of\ a\ past\ calendar\ year\ (orders\ older\ than\ 12}
\promptline{\rlap{\textcolor{gray}{$\hookrightarrow$}}\phantom{xx}months\ only).}
\promptline{\phantom{xxxx}B\ =\ set\_reminder:\ Set\ a\ reminder.}
\promptline{\phantom{xxxx}C\ =\ finish:\ Call\ when\ the\ task\ is\ complete.\ Provide\ the}
\promptline{\rlap{\textcolor{gray}{$\hookrightarrow$}}\phantom{xx}final\ answer\ (or\ a\ short\ summary).}
\promptline{stop:\ Has\ the\ task\ already\ been\ fully\ completed,\ so\ the}
\promptline{\rlap{\textcolor{gray}{$\hookrightarrow$}}\phantom{xx}agent\ should\ stop\ now?}
\promptline{\phantom{xxxx}A\ =\ no:\ The\ answer\ is\ no\ (negative,\ or\ disagree\ with}
\promptline{\rlap{\textcolor{gray}{$\hookrightarrow$}}\phantom{xx}the\ claim).}
\promptline{\phantom{xxxx}B\ =\ yes:\ The\ answer\ is\ yes\ (affirmative,\ or\ align\ with}
\promptline{\rlap{\textcolor{gray}{$\hookrightarrow$}}\phantom{xx}the\ claim).}
\promptline{[assistant]}
\promptline{\{}
\promptline{\phantom{xxxx}\textquotedbl{}tool\_select\textquotedbl{}:\ \textquotedbl{}<{}decision>{}\textquotedbl{},}
\promptline{\phantom{xxxx}\textquotedbl{}stop\textquotedbl{}:\ \textquotedbl{}<{}decision>{}\textquotedbl{}}\nobreak
\promptline{\}}\nobreak
\endgroup\smallskip

\par\smallskip{\footnotesize\raggedright\interlinepenalty=10000\noindent\textbf{P7. \ours{} and the other trained models.} Source: \texttt{src/\allowbreak{}jevharness/\allowbreak{}training/\allowbreak{}data.py: SYSTEM, render}, \texttt{src/\allowbreak{}jevharness/\allowbreak{}training/\allowbreak{}serve\_jev.py}. Each question gets its own prompt in the checkpoint's chat template, with thinking off and the assistant turn opened. The answer is read from the next-token logits of the option codes. Training rows use this exact format.\par}\nobreak\vspace{1pt}\nobreak
\begingroup\ttfamily\scriptsize\setlength{\fboxsep}{0pt}\setlength{\parskip}{0pt}%
\promptline{[system]}\nobreak
\promptline{You\ are\ a\ fast\ decision\ module\ inside\ an\ AI\ agent.\ Read\ the}
\promptline{\rlap{\textcolor{gray}{$\hookrightarrow$}}\phantom{xx}state\ and\ answer\ the\ question\ with\ exactly\ one\ token:\ the}
\promptline{\rlap{\textcolor{gray}{$\hookrightarrow$}}\phantom{xx}label\ of\ your\ answer.}
\promptline{[user]}
\promptline{STATE:}
\promptline{\textrm{\textit{$\langle$state$\rangle$}}}
\promptgap
\promptline{QUESTION:\ Which\ tool\ should\ the\ agent\ call\ next?}
\promptline{OPTIONS:}
\promptline{A.\ get\_orders\_archive\ -{}-{}\ List\ a\ customer\textquotesingle{}s\ orders\ from\ the}
\promptline{\rlap{\textcolor{gray}{$\hookrightarrow$}}\phantom{xx}archive\ of\ a\ past\ calendar\ year\ (orders\ older\ than\ 12}
\promptline{\rlap{\textcolor{gray}{$\hookrightarrow$}}\phantom{xx}months\ only).}
\promptline{B.\ set\_reminder\ -{}-{}\ Set\ a\ reminder.}
\promptline{C.\ finish\ -{}-{}\ Call\ when\ the\ task\ is\ complete.\ Provide\ the}
\promptline{\rlap{\textcolor{gray}{$\hookrightarrow$}}\phantom{xx}final\ answer\ (or\ a\ short\ summary).}
\promptgap
\promptline{Answer\ with\ the\ option\ label\ only.}
\promptgap
\promptline{[user,\ a\ yes/no\ question]}
\promptline{STATE:}
\promptline{\textrm{\textit{$\langle$state$\rangle$}}}
\promptgap
\promptline{QUESTION:\ Has\ the\ task\ already\ been\ fully\ completed,\ so\ the}
\promptline{\rlap{\textcolor{gray}{$\hookrightarrow$}}\phantom{xx}agent\ should\ stop\ now?}
\promptline{OPTIONS:}
\promptline{A.\ yes}
\promptline{B.\ no}
\promptgap\nobreak
\promptline{Answer\ with\ the\ option\ label\ only.}\nobreak
\endgroup\smallskip

\par\smallskip{\footnotesize\raggedright\interlinepenalty=10000\noindent\textbf{P8. LLM decider read through log-probabilities.} Source: \texttt{src/\allowbreak{}jevharness/\allowbreak{}decision/\allowbreak{}llm\_logprob.py}. The model generates one token at temperature 0 and returns the top 20 log-probabilities. The probabilities of tokens that equal an option label are summed per label.\par}\nobreak\vspace{1pt}\nobreak
\begingroup\ttfamily\scriptsize\setlength{\fboxsep}{0pt}\setlength{\parskip}{0pt}%
\promptline{[system]}\nobreak
\promptline{You\ are\ a\ fast\ decision\ module\ inside\ an\ AI\ agent.\ Read\ the}
\promptline{\rlap{\textcolor{gray}{$\hookrightarrow$}}\phantom{xx}state\ and\ answer\ the\ question\ with\ exactly\ one\ token:\ the}
\promptline{\rlap{\textcolor{gray}{$\hookrightarrow$}}\phantom{xx}label\ of\ your\ answer.\ Do\ not\ explain.}
\promptline{[user]}
\promptline{STATE:}
\promptline{\textrm{\textit{$\langle$state$\rangle$}}}
\promptgap
\promptline{QUESTION:\ Which\ tool\ should\ the\ agent\ call\ next?}
\promptline{OPTIONS:}
\promptline{A.\ get\_orders\_archive\ -{}-{}\ List\ a\ customer\textquotesingle{}s\ orders\ from\ the}
\promptline{\rlap{\textcolor{gray}{$\hookrightarrow$}}\phantom{xx}archive\ of\ a\ past\ calendar\ year\ (orders\ older\ than\ 12}
\promptline{\rlap{\textcolor{gray}{$\hookrightarrow$}}\phantom{xx}months\ only).}
\promptline{B.\ set\_reminder\ -{}-{}\ Set\ a\ reminder.}
\promptline{C.\ finish\ -{}-{}\ Call\ when\ the\ task\ is\ complete.\ Provide\ the}
\promptline{\rlap{\textcolor{gray}{$\hookrightarrow$}}\phantom{xx}final\ answer\ (or\ a\ short\ summary).}
\promptgap
\promptline{Answer\ with\ the\ option\ label\ only.}
\promptgap
\promptline{[user,\ a\ yes/no\ question]}
\promptline{STATE:}
\promptline{\textrm{\textit{$\langle$state$\rangle$}}}
\promptgap
\promptline{QUESTION:\ Has\ the\ task\ already\ been\ fully\ completed,\ so\ the}
\promptline{\rlap{\textcolor{gray}{$\hookrightarrow$}}\phantom{xx}agent\ should\ stop\ now?}
\promptgap\nobreak
\promptline{Answer\ Yes\ or\ No.}\nobreak
\endgroup\smallskip

\par\smallskip{\footnotesize\raggedright\interlinepenalty=10000\noindent\textbf{P9. LLM decider with verbalised probabilities.} Source: \texttt{src/\allowbreak{}jevharness/\allowbreak{}decision/\allowbreak{}llm\_generative.py}. The Gemini stand-in and the control in which the agent LLM answers its own questions use this prompt. All questions about one state go in one call.\par}\nobreak\vspace{1pt}\nobreak
\begingroup\ttfamily\scriptsize\setlength{\fboxsep}{0pt}\setlength{\parskip}{0pt}%
\promptline{[system]}\nobreak
\promptline{You\ are\ a\ fast\ decision\ module\ inside\ an\ AI\ agent.\ Read\ the}
\promptline{\rlap{\textcolor{gray}{$\hookrightarrow$}}\phantom{xx}state\ and\ answer\ each\ typed\ question\ independently.\ Be}
\promptline{\rlap{\textcolor{gray}{$\hookrightarrow$}}\phantom{xx}calibrated:\ probabilities\ must\ reflect\ how\ likely\ each}
\promptline{\rlap{\textcolor{gray}{$\hookrightarrow$}}\phantom{xx}option\ is\ correct.\ Reply\ with\ ONLY\ one\ JSON\ object,\ no}
\promptline{\rlap{\textcolor{gray}{$\hookrightarrow$}}\phantom{xx}prose,\ no\ code\ fences.}
\promptline{[user]}
\promptline{STATE:}
\promptline{\textrm{\textit{$\langle$state$\rangle$}}}
\promptgap
\promptline{QUESTIONS:}
\promptgap
\promptline{Q1\ (choose\ one\ option):\ Which\ tool\ should\ the\ agent\ call}
\promptline{\rlap{\textcolor{gray}{$\hookrightarrow$}}\phantom{xx}next?}
\promptline{OPTIONS:}
\promptline{A.\ get\_orders\_archive\ -{}-{}\ List\ a\ customer\textquotesingle{}s\ orders\ from\ the}
\promptline{\rlap{\textcolor{gray}{$\hookrightarrow$}}\phantom{xx}archive\ of\ a\ past\ calendar\ year\ (orders\ older\ than\ 12}
\promptline{\rlap{\textcolor{gray}{$\hookrightarrow$}}\phantom{xx}months\ only).}
\promptline{B.\ set\_reminder\ -{}-{}\ Set\ a\ reminder.}
\promptline{C.\ finish\ -{}-{}\ Call\ when\ the\ task\ is\ complete.\ Provide\ the}
\promptline{\rlap{\textcolor{gray}{$\hookrightarrow$}}\phantom{xx}final\ answer\ (or\ a\ short\ summary).}
\promptgap
\promptline{Q2\ (yes/no):\ Has\ the\ task\ already\ been\ fully\ completed,\ so}
\promptline{\rlap{\textcolor{gray}{$\hookrightarrow$}}\phantom{xx}the\ agent\ should\ stop\ now?}
\promptgap
\promptline{For\ EACH\ question\ give\ a\ probability\ for\ EVERY\ option\ key}
\promptline{\rlap{\textcolor{gray}{$\hookrightarrow$}}\phantom{xx}(they\ must\ sum\ to\ 1),\ reflecting\ how\ likely\ each\ option}
\promptline{\rlap{\textcolor{gray}{$\hookrightarrow$}}\phantom{xx}is\ the\ correct\ answer.}
\promptline{Reply\ with\ JSON\ exactly\ in\ this\ shape:}
\promptline{\{\textquotedbl{}Q1\textquotedbl{}:\ \{\textquotedbl{}A\textquotedbl{}:\ \textquotedbl{}<{}prob>{}\textquotedbl{},\ \textquotedbl{}B\textquotedbl{}:\ \textquotedbl{}<{}prob>{}\textquotedbl{},\ \textquotedbl{}C\textquotedbl{}:\ \textquotedbl{}<{}prob>{}\textquotedbl{}\},\ \textquotedbl{}Q2\textquotedbl{}:}
\promptline{\rlap{\textcolor{gray}{$\hookrightarrow$}}\phantom{xx}\{\textquotedbl{}yes\textquotedbl{}:\ \textquotedbl{}<{}prob>{}\textquotedbl{},\ \textquotedbl{}no\textquotedbl{}:\ \textquotedbl{}<{}prob>{}\textquotedbl{}\}\}}
\promptgap
\promptline{[last\ block\ in\ the\ label\ variant]}
\promptline{For\ EACH\ question\ give\ your\ answer\ key\ and\ the\ probability}
\promptline{\rlap{\textcolor{gray}{$\hookrightarrow$}}\phantom{xx}that\ it\ is\ correct.}
\promptline{Reply\ with\ JSON\ exactly\ in\ this\ shape:}
\promptline{\{\textquotedbl{}Q1\textquotedbl{}:\ \{\textquotedbl{}answer\textquotedbl{}:\ \textquotedbl{}<{}one\ of\ [\textquotesingle{}A\textquotesingle{},\ \textquotesingle{}B\textquotesingle{},\ \textquotesingle{}C\textquotesingle{}]>{}\textquotedbl{},\ \textquotedbl{}confidence\textquotedbl{}:}\nobreak
\promptline{\rlap{\textcolor{gray}{$\hookrightarrow$}}\phantom{xx}\textquotedbl{}<{}prob\ that\ the\ answer\ is\ correct>{}\textquotedbl{}\}\}}\nobreak
\endgroup\smallskip

\par\smallskip{\footnotesize\raggedright\interlinepenalty=10000\noindent\textbf{P10. Embedding baseline (L1 only).} Source: \texttt{src/\allowbreak{}jevharness/\allowbreak{}decision/\allowbreak{}embedding.py}. The score of an option is the cosine similarity of the two embeddings.\par}\nobreak\vspace{1pt}\nobreak
\begingroup\ttfamily\scriptsize\setlength{\fboxsep}{0pt}\setlength{\parskip}{0pt}%
\promptline{[query]}\nobreak
\promptline{\textrm{\textit{$\langle$state$\rangle$}}}
\promptline{\textrm{\textit{$\langle$question$\rangle$}}}
\promptline{[one\ text\ per\ option]}\nobreak
\promptline{\textrm{\textit{$\langle$option$\rangle$}}:\ \textrm{\textit{$\langle$description$\rangle$}}}\nobreak
\endgroup\smallskip

\paragraph{The agent LLM on \gym{}.}
The agent LLM sees a short system prompt, the task and the tool results, plus the notes below when a decision point acts on its turn.

\par\smallskip{\footnotesize\raggedright\interlinepenalty=10000\noindent\textbf{P11. Agent LLM: system prompt and first message on \gym{}.} Source: \texttt{src/\allowbreak{}jevharness/\allowbreak{}harness/\allowbreak{}agent.py}, \texttt{src/\allowbreak{}jevharness/\allowbreak{}bench/\allowbreak{}suites/\allowbreak{}jevgym.py}, \texttt{src/\allowbreak{}jevharness/\allowbreak{}harness/\allowbreak{}tools.py}. The task's tools (68 at these settings, shuffled per task) are passed as function declarations, and the LLM sees every tool result as a tool message.\par}\nobreak\vspace{1pt}\nobreak
\begingroup\ttfamily\scriptsize\setlength{\fboxsep}{0pt}\setlength{\parskip}{0pt}%
\promptline{[system]}\nobreak
\promptline{You\ are\ a\ capable\ autonomous\ agent.\ Solve\ the\ user\textquotesingle{}s\ task}
\promptline{\rlap{\textcolor{gray}{$\hookrightarrow$}}\phantom{xx}by\ calling\ the\ available\ tools.\ Call\ one\ tool\ per\ turn.}
\promptline{\rlap{\textcolor{gray}{$\hookrightarrow$}}\phantom{xx}When\ the\ task\ is\ complete,\ call\ \textasciigrave{}finish\textasciigrave{}\ with\ the\ final}
\promptline{\rlap{\textcolor{gray}{$\hookrightarrow$}}\phantom{xx}answer.}
\promptline{[user]}
\promptline{\textrm{\textit{$\langle$task instruction$\rangle$}}}
\promptgap
\promptline{\textrm{\textit{$\langle$first observation$\rangle$}}}
\promptgap
\promptline{[first\ observation]}
\promptline{You\ have\ access\ to\ the\ customer-{}operations\ tools.\ Policy:}
\promptline{\rlap{\textcolor{gray}{$\hookrightarrow$}}\phantom{xx}never\ guess\ -{}-{}\ if\ a\ detail\ you\ need\ to\ act\ (who\ the}
\promptline{\rlap{\textcolor{gray}{$\hookrightarrow$}}\phantom{xx}customer\ is,\ which\ order,\ the\ refund\ method,\ the\ shipping}
\promptline{\rlap{\textcolor{gray}{$\hookrightarrow$}}\phantom{xx}speed)\ is\ not\ given\ or\ is\ ambiguous,\ ask\ the\ customer}
\promptline{\rlap{\textcolor{gray}{$\hookrightarrow$}}\phantom{xx}with\ ask\_user\ before\ taking\ any\ action.}
\promptgap
\promptline{[tool\ \textasciigrave{}finish\textasciigrave{},\ offered\ with\ the\ task\textquotesingle{}s\ tools]}
\promptline{Call\ when\ the\ task\ is\ complete.\ Provide\ the\ final\ answer}\nobreak
\promptline{\rlap{\textcolor{gray}{$\hookrightarrow$}}\phantom{xx}(or\ a\ short\ summary).}\nobreak
\endgroup\smallskip

\par\smallskip{\footnotesize\raggedright\interlinepenalty=10000\noindent\textbf{P12. \gym{} task instructions.} Source: \texttt{src/\allowbreak{}jevharness/\allowbreak{}bench/\allowbreak{}suites/\allowbreak{}jevgym.py: JevGymEnv.\_build*, \_ask}. The first five replies give a clarify task's missing detail. A question about another detail gets the sixth reply in a clarify task and the seventh in every other task.\par}\nobreak\vspace{1pt}\nobreak
\begingroup\ttfamily\scriptsize\setlength{\fboxsep}{0pt}\setlength{\parskip}{0pt}%
\promptline{[cancel]\ The\ customer\ \textrm{\textit{$\langle$customer$\rangle$}}\ asked\ us\ to\ cancel\ their}\nobreak
\promptline{\rlap{\textcolor{gray}{$\hookrightarrow$}}\phantom{xx}pending\ order.\ Please\ do\ it.}
\promptline{[refund]\ Refund\ the\ full\ amount\ of\ \textrm{\textit{$\langle$customer$\rangle$}}\textquotesingle{}s\ most\ recent}
\promptline{\rlap{\textcolor{gray}{$\hookrightarrow$}}\phantom{xx}delivered\ order\ (highest\ order\ id)\ as\ store\ credit.}
\promptline{[stock]\ How\ many\ units\ of\ \textquotesingle{}\textrm{\textit{$\langle$item$\rangle$}}\textquotesingle{}\ do\ we\ have\ in\ stock?}
\promptline{\rlap{\textcolor{gray}{$\hookrightarrow$}}\phantom{xx}Answer\ with\ the\ number.}
\promptline{[ship]\ Ship\ \textrm{\textit{$\langle$customer$\rangle$}}\textquotesingle{}s\ pending\ order.\ Use\ express}
\promptline{\rlap{\textcolor{gray}{$\hookrightarrow$}}\phantom{xx}shipping\ if\ the\ customer\ is\ gold\ tier,\ otherwise\ standard.}
\promptline{[total]\ What\ is\ the\ total\ amount\ of\ all\ of\ \textrm{\textit{$\langle$customer$\rangle$}}\textquotesingle{}s}
\promptline{\rlap{\textcolor{gray}{$\hookrightarrow$}}\phantom{xx}delivered\ orders?\ Answer\ with\ the\ number\ rounded\ to\ 2}
\promptline{\rlap{\textcolor{gray}{$\hookrightarrow$}}\phantom{xx}decimals.}
\promptline{[clarify:\ order\ missing]\ The\ customer\ \textrm{\textit{$\langle$customer$\rangle$}}\ asked\ us}
\promptline{\rlap{\textcolor{gray}{$\hookrightarrow$}}\phantom{xx}to\ cancel\ their\ pending\ order.\ Please\ do\ it.}
\promptline{[clarify:\ method\ missing]\ Refund\ the\ full\ amount\ of}
\promptline{\rlap{\textcolor{gray}{$\hookrightarrow$}}\phantom{xx}\textrm{\textit{$\langle$customer$\rangle$}}\textquotesingle{}s\ most\ recent\ delivered\ order\ (highest\ order}
\promptline{\rlap{\textcolor{gray}{$\hookrightarrow$}}\phantom{xx}id).}
\promptline{[clarify:\ speed\ missing]\ Ship\ \textrm{\textit{$\langle$customer$\rangle$}}\textquotesingle{}s\ pending\ order.}
\promptline{[clarify:\ name\ missing]\ A\ customer\ wrote\ in\ asking\ us\ to}
\promptline{\rlap{\textcolor{gray}{$\hookrightarrow$}}\phantom{xx}ship\ their\ pending\ order\ (express\ if\ they\ are\ gold\ tier,}
\promptline{\rlap{\textcolor{gray}{$\hookrightarrow$}}\phantom{xx}otherwise\ standard),\ but\ did\ not\ say\ who\ they\ are.}
\promptline{[multi]\ Process\ these\ requests\ in\ order:}
\promptline{\rlap{\textcolor{gray}{$\hookrightarrow$}}\phantom{xx}\textrm{\textit{$\langle$numbered requests$\rangle$}}.\textrm{\textit{$\langle$shipping rule$\rangle$}}\ When\ all\ are\ done,}
\promptline{\rlap{\textcolor{gray}{$\hookrightarrow$}}\phantom{xx}answer\ with\ the\ total\ amount\ of\ the\ orders\ you\ processed,}
\promptline{\rlap{\textcolor{gray}{$\hookrightarrow$}}\phantom{xx}rounded\ to\ 2\ decimals.}
\promptline{\phantom{xx}numbered\ requests,\ joined\ by\ \textquotedbl{};\ \textquotedbl{}:\ (\textrm{\textit{$\langle$i$\rangle$}})\ \textrm{\textit{$\langle$request$\rangle$}}}
\promptline{\phantom{xx}request:\ cancel\ \textrm{\textit{$\langle$customer$\rangle$}}\textquotesingle{}s\ pending\ order\ (the\ customer}
\promptline{\rlap{\textcolor{gray}{$\hookrightarrow$}}\phantom{xx}requested\ it)}
\promptline{\phantom{xxxxxxx}or:\ ship\ \textrm{\textit{$\langle$customer$\rangle$}}\textquotesingle{}s\ pending\ order}
\promptline{\phantom{xx}shipping\ rule\ (if\ any\ request\ ships):\ Use\ express}
\promptline{\rlap{\textcolor{gray}{$\hookrightarrow$}}\phantom{xx}shipping\ for\ gold-{}tier\ customers\ and\ standard\ shipping}
\promptline{\rlap{\textcolor{gray}{$\hookrightarrow$}}\phantom{xx}otherwise.}
\promptgap
\promptline{[customer\ replies\ to\ ask\_user]}
\promptline{Customer:\ Sorry\ for\ the\ confusion\ -{}-{}\ please\ cancel\ order}
\promptline{\rlap{\textcolor{gray}{$\hookrightarrow$}}\phantom{xx}\textrm{\textit{$\langle$order id$\rangle$}}\ (the\ \textrm{\textit{$\langle$item$\rangle$}}).}
\promptline{Customer:\ Please\ refund\ it\ as\ store\ credit.}
\promptline{Customer:\ Please\ refund\ it\ to\ my\ original\ payment\ method.}
\promptline{Customer:\ \textrm{\textit{$\langle$Speed$\rangle$}}\ shipping,\ please.}
\promptline{Customer:\ Oh,\ sorry\ -{}-{}\ my\ name\ is\ \textrm{\textit{$\langle$customer$\rangle$}}.}
\promptline{Customer:\ I\ think\ I\ already\ gave\ you\ everything\ you\ need\ on}
\promptline{\rlap{\textcolor{gray}{$\hookrightarrow$}}\phantom{xx}that\ point.}
\promptline{Customer:\ I\ have\ nothing\ to\ add\ -{}-{}\ please\ go\ ahead\ with\ the}\nobreak
\promptline{\rlap{\textcolor{gray}{$\hookrightarrow$}}\phantom{xx}request\ as\ described.}\nobreak
\endgroup\smallskip

\par\smallskip{\footnotesize\raggedright\interlinepenalty=10000\noindent\textbf{P13. \gym{}-Hard instruction.} Source: \texttt{src/\allowbreak{}jevharness/\allowbreak{}bench/\allowbreak{}suites/\allowbreak{}jevgym\_hard.py: \_build\_multi, policy\_text, answer\_text}. The runs used ten or twelve customers per task. Each customer's tier and amount appear only in the lookup results. The code also has a rule on the amount alone, which no run used.\par}\nobreak\vspace{1pt}\nobreak
\begingroup\ttfamily\scriptsize\setlength{\fboxsep}{0pt}\setlength{\parskip}{0pt}%
\promptline{Handle\ the\ pending\ order\ of\ each\ of\ these\ customers,\ in}\nobreak
\promptline{\rlap{\textcolor{gray}{$\hookrightarrow$}}\phantom{xx}order:\ (1)\ \textrm{\textit{$\langle$customer 1$\rangle$}};\ (2)\ \textrm{\textit{$\langle$customer 2$\rangle$}};}
\promptline{\rlap{\textcolor{gray}{$\hookrightarrow$}}\phantom{xx}\textrm{\textit{[one per customer]}}.\ For\ each\ customer,\ look\ up\ their}
\promptline{\rlap{\textcolor{gray}{$\hookrightarrow$}}\phantom{xx}pending\ order\ and\ \textrm{\textit{$\langle$rule$\rangle$}}.\ Cancellations\ use\ the\ reason}
\promptline{\rlap{\textcolor{gray}{$\hookrightarrow$}}\phantom{xx}customer\_request;\ use\ express\ shipping\ for\ gold-{}tier}
\promptline{\rlap{\textcolor{gray}{$\hookrightarrow$}}\phantom{xx}customers\ and\ standard\ shipping\ otherwise.}
\promptline{\rlap{\textcolor{gray}{$\hookrightarrow$}}\phantom{xx}\textrm{\textit{$\langle$answer sentence$\rangle$}}}
\promptgap
\promptline{[rule]}
\promptline{cancel\ it\ if\ the\ customer\ is\ basic\ tier\ and\ the\ amount\ is}
\promptline{\rlap{\textcolor{gray}{$\hookrightarrow$}}\phantom{xx}above\ \$300.00,\ or\ silver\ tier\ and\ the\ amount\ is\ above}
\promptline{\rlap{\textcolor{gray}{$\hookrightarrow$}}\phantom{xx}\$600.00\ (never\ cancel\ a\ gold-{}tier\ customer\textquotesingle{}s\ order);}
\promptline{\rlap{\textcolor{gray}{$\hookrightarrow$}}\phantom{xx}otherwise\ ship\ it}
\promptgap
\promptline{[answer\ sentence:\ total,\ first\ runs]}
\promptline{When\ all\ are\ done,\ answer\ with\ the\ total\ amount\ of\ the}
\promptline{\rlap{\textcolor{gray}{$\hookrightarrow$}}\phantom{xx}orders\ you\ processed,\ rounded\ to\ 2\ decimals.}
\promptline{[answer\ sentence:\ itemized,\ from\ E7e]}
\promptline{When\ all\ are\ done,\ answer\ with\ one\ line\ per\ customer\ giving}
\promptline{\rlap{\textcolor{gray}{$\hookrightarrow$}}\phantom{xx}the\ action\ you\ took\ and\ the\ order\ amount,\ then\ the\ total}\nobreak
\promptline{\rlap{\textcolor{gray}{$\hookrightarrow$}}\phantom{xx}amount\ of\ the\ orders\ you\ processed,\ rounded\ to\ 2\ decimals.}\nobreak
\endgroup\smallskip

\par\smallskip{\footnotesize\raggedright\interlinepenalty=10000\noindent\textbf{P14. Messages the harness adds to the LLM's context.} Source: \texttt{src/\allowbreak{}jevharness/\allowbreak{}harness/\allowbreak{}prompts.py}, \texttt{src/\allowbreak{}jevharness/\allowbreak{}harness/\allowbreak{}agent.py}. The reflection reply and the stop-guard rejection stay in the context. They are appended to the last tool message when one precedes them, because Gemini 3 returns an empty reply after a user turn there. The other notes go with one call only. Tool assist adds no text and offers the LLM only the model's top tools and finish. Routing adds no text either. It sends the step to the cheaper LLM with the same messages.\par}\nobreak\vspace{1pt}\nobreak
\begingroup\ttfamily\scriptsize\setlength{\fboxsep}{0pt}\setlength{\parskip}{0pt}%
\promptline{[Reflection:\ asked\ as\ a\ user\ turn,\ the\ reply\ kept]}\nobreak
\promptline{Pause.\ In\ <{}=4\ sentences:\ assess\ progress\ toward\ the\ task,}
\promptline{\rlap{\textcolor{gray}{$\hookrightarrow$}}\phantom{xx}identify\ any\ mistake\ so\ far,\ and\ state\ the\ best\ next}
\promptline{\rlap{\textcolor{gray}{$\hookrightarrow$}}\phantom{xx}step.\ Do\ not\ call\ tools.}
\promptline{[Reflection]\ \textrm{\textit{$\langle$the LLM's reply$\rangle$}}}
\promptgap
\promptline{[Recalled\ memories,\ before\ the\ LLM\textquotesingle{}s\ turn]}
\promptline{[Memory]}
\promptline{-{}\ [m\textrm{\textit{$\langle$id$\rangle$}}]\ \textrm{\textit{$\langle$memory text$\rangle$}}}
\promptline{-{}\ [m\textrm{\textit{$\langle$id$\rangle$}}]\ \textrm{\textit{$\langle$memory text$\rangle$}}}
\promptgap
\promptline{[Stop\ guard\ rejected\ a\ finish]}
\promptline{[Verifier]\ The\ task\ may\ not\ be\ complete\ yet,\ or\ the\ final}
\promptline{\rlap{\textcolor{gray}{$\hookrightarrow$}}\phantom{xx}answer\ may\ be\ wrong.\ Re-{}check\ the\ requests\ and\ recompute}
\promptline{\rlap{\textcolor{gray}{$\hookrightarrow$}}\phantom{xx}the\ answer\ before\ calling\ finish.}
\promptline{[Stop\ trigger\ fired:\ a\ forced\ finish\ call]}
\promptline{The\ task\ appears\ complete.\ Call\ finish\ with\ the\ final}
\promptline{\rlap{\textcolor{gray}{$\hookrightarrow$}}\phantom{xx}answer.}
\promptgap
\promptline{[Tool\ guard\ rejected\ the\ call]}
\promptline{[Verifier]\ Your\ proposed\ call\ \textasciigrave{}\textrm{\textit{$\langle$tool$\rangle$}}(\textrm{\textit{$\langle$arguments$\rangle$}})\textasciigrave{}\ was}
\promptline{\rlap{\textcolor{gray}{$\hookrightarrow$}}\phantom{xx}judged\ likely\ wrong.\ Reconsider\ and\ choose\ the\ correct}
\promptline{\rlap{\textcolor{gray}{$\hookrightarrow$}}\phantom{xx}next\ tool\ call.}
\promptline{[Argument\ guard\ rejected\ one\ argument]}
\promptline{[Verifier]\ In\ your\ proposed\ call\ \textasciigrave{}\textrm{\textit{$\langle$tool$\rangle$}}(\textrm{\textit{$\langle$arguments$\rangle$}})\textasciigrave{},\ the}
\promptline{\rlap{\textcolor{gray}{$\hookrightarrow$}}\phantom{xx}value\ of\ the\ argument\ \textasciigrave{}\textrm{\textit{$\langle$argument$\rangle$}}\textasciigrave{}\ was\ judged\ likely}
\promptline{\rlap{\textcolor{gray}{$\hookrightarrow$}}\phantom{xx}wrong.\ Reconsider\ and\ choose\ the\ correct\ next\ tool\ call.}
\promptgap
\promptline{[Verification\ pass\ at\ tool\ choice:\ every\ tool\ offered]}
\promptline{[Check]\ Before\ your\ proposed\ call\ \textasciigrave{}\textrm{\textit{$\langle$tool$\rangle$}}(\textrm{\textit{$\langle$arguments$\rangle$}})\textasciigrave{}}
\promptline{\rlap{\textcolor{gray}{$\hookrightarrow$}}\phantom{xx}runs,\ re-{}read\ the\ task\ and\ the\ observations\ so\ far\ and}
\promptline{\rlap{\textcolor{gray}{$\hookrightarrow$}}\phantom{xx}check\ that\ this\ is\ the\ right\ tool\ to\ call\ now.\ If\ it\ is,}
\promptline{\rlap{\textcolor{gray}{$\hookrightarrow$}}\phantom{xx}make\ the\ same\ call\ again;\ if\ not,\ make\ the\ right\ call}
\promptline{\rlap{\textcolor{gray}{$\hookrightarrow$}}\phantom{xx}instead.}
\promptline{[Verification\ pass\ at\ tool\ arguments:\ the\ call\ forced\ to}
\promptline{\rlap{\textcolor{gray}{$\hookrightarrow$}}\phantom{xx}the\ proposed\ tool]}
\promptline{[Check]\ Before\ your\ proposed\ call\ \textasciigrave{}\textrm{\textit{$\langle$tool$\rangle$}}(\textrm{\textit{$\langle$arguments$\rangle$}})\textasciigrave{}}
\promptline{\rlap{\textcolor{gray}{$\hookrightarrow$}}\phantom{xx}runs,\ re-{}read\ the\ task\ and\ the\ observations\ so\ far\ and}
\promptline{\rlap{\textcolor{gray}{$\hookrightarrow$}}\phantom{xx}check\ every\ argument\ value.\ Call\ \textasciigrave{}\textrm{\textit{$\langle$tool$\rangle$}}\textasciigrave{}\ again,\ with\ the}
\promptline{\rlap{\textcolor{gray}{$\hookrightarrow$}}\phantom{xx}same\ arguments\ if\ they\ are\ right\ or\ with\ corrected}
\promptline{\rlap{\textcolor{gray}{$\hookrightarrow$}}\phantom{xx}arguments\ if\ not.}
\promptgap
\promptline{[Decision\ model\ chose\ the\ tool\ and\ the\ LLM\ fills\ the\ rest]}
\promptline{You\ must\ now\ call\ \textasciigrave{}\textrm{\textit{$\langle$tool$\rangle$}}\textasciigrave{}.\ Already-{}determined\ arguments:}
\promptline{\rlap{\textcolor{gray}{$\hookrightarrow$}}\phantom{xx}\textrm{\textit{$\langle$the model's arguments$\rangle$}}.\ Provide\ the\ remaining\ arguments.}
\promptline{[Argument\ assist:\ the\ model\textquotesingle{}s\ shortlists]}
\promptline{You\ must\ now\ call\ \textasciigrave{}\textrm{\textit{$\langle$tool$\rangle$}}\textasciigrave{}.\ Choose\ these\ arguments\ from\ the}
\promptline{\rlap{\textcolor{gray}{$\hookrightarrow$}}\phantom{xx}shortlists:}
\promptline{-{}\ \textasciigrave{}\textrm{\textit{$\langle$argument$\rangle$}}\textasciigrave{}:\ one\ of\ [\textrm{\textit{$\langle$value$\rangle$}},\ \textrm{\textit{$\langle$value$\rangle$}}]}
\promptline{-{}\ \textasciigrave{}\textrm{\textit{$\langle$argument$\rangle$}}\textasciigrave{}:\ one\ of\ [\textrm{\textit{$\langle$value$\rangle$}}]\ (or\ another\ value)}
\promptline{Already-{}determined\ arguments:\ \{\textquotedbl{}\textrm{\textit{$\langle$argument$\rangle$}}\textquotedbl{}:\ \textrm{\textit{$\langle$value$\rangle$}}\}.}\nobreak
\promptline{Provide\ all\ arguments.}\nobreak
\endgroup\smallskip

\par\smallskip{\footnotesize\raggedright\interlinepenalty=10000\noindent\textbf{P15. Yes/no questions the agent LLM answers itself.} Source: \texttt{src/\allowbreak{}jevharness/\allowbreak{}harness/\allowbreak{}memory.py}, \texttt{src/\allowbreak{}jevharness/\allowbreak{}harness/\allowbreak{}prompts.py}, \texttt{src/\allowbreak{}jevharness/\allowbreak{}harness/\allowbreak{}agent.py}. Each gate is one call without tools, read as yes when the reply starts with y. With advice, the memory gate puts the advice line between the state and its question.\par}\nobreak\vspace{1pt}\nobreak
\begingroup\ttfamily\scriptsize\setlength{\fboxsep}{0pt}\setlength{\parskip}{0pt}%
\promptline{[Memory-{}write\ gate\ (LLM\ decides,\ or\ the\ model\ defers)]}\nobreak
\promptline{\textrm{\textit{$\langle$state$\rangle$}}}
\promptgap
\promptline{Should\ this\ be\ stored\ in\ long-{}term\ memory\ because\ it\ will}
\promptline{\rlap{\textcolor{gray}{$\hookrightarrow$}}\phantom{xx}likely\ be\ needed\ later?\ Answer\ yes\ or\ no.}
\promptgap
\promptline{[Reflection\ and\ routing\ gates\ (the\ model\ defers)]}
\promptline{\textrm{\textit{$\langle$state$\rangle$}}}
\promptgap
\promptline{\textrm{\textit{$\langle$question$\rangle$}}\ Answer\ yes\ or\ no.}
\promptgap
\promptline{[Advise\ variants:\ the\ advice\ comes\ before\ the\ question]}
\promptline{\textrm{\textit{$\langle$state$\rangle$}}}
\promptgap
\promptline{\textrm{\textit{$\langle$advice$\rangle$}}}
\promptgap
\promptline{\textrm{\textit{$\langle$question$\rangle$}}\ Answer\ yes\ or\ no.}
\promptline{[Advisor]\ A\ fast\ decision\ model\ answered\ the\ question}
\promptline{\rlap{\textcolor{gray}{$\hookrightarrow$}}\phantom{xx}below:\ \textrm{\textit{$\langle$yes|no$\rangle$}}\ (confidence\ \textrm{\textit{$\langle$p$\rangle$}}).\ It\ can\ be\ wrong;\ decide}
\promptline{\rlap{\textcolor{gray}{$\hookrightarrow$}}\phantom{xx}yourself.}
\promptline{[Stop\ advise,\ added\ to\ the\ LLM\textquotesingle{}s\ turn]}
\promptline{[Advisor]\ A\ fast\ decision\ model\textquotesingle{}s\ answer\ to\ \textquotedbl{}Has\ the\ task}
\promptline{\rlap{\textcolor{gray}{$\hookrightarrow$}}\phantom{xx}already\ been\ fully\ completed,\ so\ the\ agent\ should\ stop}
\promptline{\rlap{\textcolor{gray}{$\hookrightarrow$}}\phantom{xx}now?\textquotedbl{}:\ \textrm{\textit{$\langle$yes|no$\rangle$}}\ (confidence\ \textrm{\textit{$\langle$p$\rangle$}}).\ It\ can\ be\ wrong;\ decide}
\promptline{\rlap{\textcolor{gray}{$\hookrightarrow$}}\phantom{xx}yourself\ whether\ to\ call\ finish\ now.}
\promptgap
\promptline{[Memory-{}read\ guard:\ the\ LLM\ selects]}
\promptline{\textrm{\textit{$\langle$state$\rangle$}}}
\promptgap
\promptline{STORED\ MEMORIES:}
\promptline{-{}\ [m\textrm{\textit{$\langle$id$\rangle$}}]\ \textrm{\textit{$\langle$text$\rangle$}}}
\promptgap
\promptline{\textrm{\textit{$\langle$rejection note, after a veto$\rangle$}}}
\promptgap
\promptline{Which\ of\ the\ stored\ memories\ above\ does\ the\ agent\textquotesingle{}s\ next}
\promptline{\rlap{\textcolor{gray}{$\hookrightarrow$}}\phantom{xx}step\ need?\ Answer\ with\ their\ ids\ separated\ by\ commas\ (for}
\promptline{\rlap{\textcolor{gray}{$\hookrightarrow$}}\phantom{xx}example:\ m3,\ m7),\ or\ with\ none.}
\promptline{[rejection\ note]}
\promptline{[Verifier]\ Your\ memory\ selection\ was\ judged\ likely\ wrong\ (a}
\promptline{\rlap{\textcolor{gray}{$\hookrightarrow$}}\phantom{xx}needed\ memory\ missing,\ or\ the\ wrong\ ones).\ Reconsider\ and}\nobreak
\promptline{\rlap{\textcolor{gray}{$\hookrightarrow$}}\phantom{xx}select\ again.}\nobreak
\endgroup\smallskip

\par\smallskip{\footnotesize\raggedright\interlinepenalty=10000\noindent\textbf{P16. Action-only environments (ALFWorld).} Source: \texttt{src/\allowbreak{}jevharness/\allowbreak{}harness/\allowbreak{}agent.py}, \texttt{src/\allowbreak{}jevharness/\allowbreak{}bench/\allowbreak{}suites/\allowbreak{}alfworld.py}. The task sentence and the observations are ALFWorld's own \citep{shridhar2021alfworld}. The reply's last \texttt{Action:} line is matched to the closest admissible action.\par}\nobreak\vspace{1pt}\nobreak
\begingroup\ttfamily\scriptsize\setlength{\fboxsep}{0pt}\setlength{\parskip}{0pt}%
\promptline{[system]}\nobreak
\promptline{You\ are\ an\ agent\ acting\ in\ a\ text\ environment.\ At\ each\ turn}
\promptline{\rlap{\textcolor{gray}{$\hookrightarrow$}}\phantom{xx}choose\ exactly\ one\ action\ from\ the\ list\ of\ admissible}
\promptline{\rlap{\textcolor{gray}{$\hookrightarrow$}}\phantom{xx}actions.\ Think\ briefly\ if\ needed,\ then\ end\ with\ a\ line}
\promptline{\rlap{\textcolor{gray}{$\hookrightarrow$}}\phantom{xx}\textasciigrave{}Action:\ <{}action>{}\textasciigrave{}\ copied\ verbatim\ from\ the\ list.}
\promptline{[user]}
\promptline{\textrm{\textit{$\langle$task instruction$\rangle$}}}
\promptgap
\promptline{\textrm{\textit{$\langle$first observation$\rangle$}}}
\promptline{[each\ turn,\ appended\ to\ the\ history]}
\promptline{Admissible\ actions:}
\promptline{-{}\ \textrm{\textit{$\langle$action$\rangle$}}}
\promptline{-{}\ \textrm{\textit{$\langle$action$\rangle$}}}
\promptline{[after\ each\ action]}
\promptline{Observation:\ \textrm{\textit{$\langle$observation$\rangle$}}}
\promptline{[action\ guard\ rejected\ the\ proposal]}
\promptline{[Verifier]\ Your\ proposed\ action\ \textasciigrave{}\textrm{\textit{$\langle$action$\rangle$}}\textasciigrave{}\ was\ judged}
\promptline{\rlap{\textcolor{gray}{$\hookrightarrow$}}\phantom{xx}likely\ wrong.\ Reconsider\ and\ choose\ the\ correct\ next}
\promptline{\rlap{\textcolor{gray}{$\hookrightarrow$}}\phantom{xx}action.}
\promptline{[verification\ pass,\ after\ the\ admissible\ list]}
\promptline{[Check]\ Before\ your\ proposed\ action\ \textasciigrave{}\textrm{\textit{$\langle$action$\rangle$}}\textasciigrave{}\ runs,}
\promptline{\rlap{\textcolor{gray}{$\hookrightarrow$}}\phantom{xx}re-{}read\ the\ task\ and\ the\ observations\ so\ far\ and\ check}
\promptline{\rlap{\textcolor{gray}{$\hookrightarrow$}}\phantom{xx}that\ it\ is\ the\ right\ next\ action.\ If\ it\ is,\ repeat\ it;\ if}
\promptline{\rlap{\textcolor{gray}{$\hookrightarrow$}}\phantom{xx}not,\ choose\ the\ right\ action\ instead.\ End\ with\ a\ line}\nobreak
\promptline{\rlap{\textcolor{gray}{$\hookrightarrow$}}\phantom{xx}\textasciigrave{}Action:\ <{}action>{}\textasciigrave{}\ copied\ verbatim\ from\ the\ list.}\nobreak
\endgroup\smallskip

\paragraph{Other harnesses and the static suites.}
Each integration renders its own state and asks its own questions. We list only the parts we wrote.

\par\smallskip{\footnotesize\raggedright\interlinepenalty=10000\noindent\textbf{P17. $\tau^2$-bench: state and questions.} Source: \texttt{src/\allowbreak{}jevharness/\allowbreak{}bench/\allowbreak{}suites/\allowbreak{}tau2.py}. The policy question drops the confirmation clause when the domain policy does not require explicit confirmation. The agent's system prompt and the user simulator are $\tau^2$-bench's own \citep{barres2025tau2}.\par}\nobreak\vspace{1pt}\nobreak
\begingroup\ttfamily\scriptsize\setlength{\fboxsep}{0pt}\setlength{\parskip}{0pt}%
\promptline{[state]}\nobreak
\promptline{DOMAIN\ POLICY\ (excerpt):}
\promptline{\textrm{\textit{$\langle$policy excerpt$\rangle$}}}
\promptgap
\promptline{CONVERSATION\ SO\ FAR\ (oldest\ first):}
\promptline{\textrm{\textit{$\langle$earlier messages$\rangle$}}}
\promptgap
\promptline{LATEST\ MESSAGE\ TO\ THE\ AGENT:}
\promptline{\textrm{\textit{$\langle$latest message$\rangle$}}}
\promptline{[each\ message\ of\ the\ conversation,\ one\ line]}
\promptline{[\textrm{\textit{$\langle$user|agent$\rangle$}}]\ \textrm{\textit{$\langle$text$\rangle$}}}
\promptline{[agent\ -{}>{}\ tool]\ \textrm{\textit{$\langle$tool(arguments)$\rangle$}};\ ...}
\promptline{[\textrm{\textit{$\langle$tool|tool ERROR$\rangle$}}]\ \textrm{\textit{$\langle$result$\rangle$}}}
\promptgap
\promptline{[Tool\ choice]}
\promptline{What\ should\ the\ customer-{}service\ agent\ do\ next?\ Choose\ the}
\promptline{\rlap{\textcolor{gray}{$\hookrightarrow$}}\phantom{xx}tool\ to\ call,\ or\ \textasciigrave{}respond\_to\_user\textasciigrave{}\ to\ send\ a\ message\ to}
\promptline{\rlap{\textcolor{gray}{$\hookrightarrow$}}\phantom{xx}the\ user.}
\promptline{Options:\ the\ domain\textquotesingle{}s\ tools,\ then}
\promptline{respond\_to\_user\ -{}-{}\ Send\ a\ text\ message\ to\ the\ user\ instead}
\promptline{\rlap{\textcolor{gray}{$\hookrightarrow$}}\phantom{xx}of\ calling\ a\ tool\ (ask\ for\ missing\ details\ or\ explicit}
\promptline{\rlap{\textcolor{gray}{$\hookrightarrow$}}\phantom{xx}confirmation,\ report\ a\ result,\ or\ refuse).}
\promptline{[Tool\ arguments:\ the\ reference\ loop\textquotesingle{}s\ question]}
\promptgap
\promptline{[Tool\ guard]}
\promptline{The\ agent\ proposes\ to\ \textrm{\textit{$\langle$proposal$\rangle$}}.\ Is\ this\ the\ right\ next}
\promptline{\rlap{\textcolor{gray}{$\hookrightarrow$}}\phantom{xx}step,\ consistent\ with\ the\ domain\ policy?}
\promptline{\phantom{xx}proposal:\ call\ \textasciigrave{}\textrm{\textit{$\langle$tool$\rangle$}}\textasciigrave{}\ with\ arguments\ \textrm{\textit{$\langle$arguments as JSON$\rangle$}}}
\promptline{\phantom{xxxxxxxx}or:\ reply\ to\ the\ user:\ \textquotedbl{}\textrm{\textit{$\langle$message$\rangle$}}\textquotedbl{}}
\promptline{[Policy\ guard,\ at\ state-{}changing\ calls]}
\promptline{The\ agent\ is\ about\ to\ call\ the\ state-{}changing\ tool\ \textasciigrave{}\textrm{\textit{$\langle$tool$\rangle$}}\textasciigrave{}}
\promptline{\rlap{\textcolor{gray}{$\hookrightarrow$}}\phantom{xx}with\ arguments\ \textrm{\textit{$\langle$arguments as JSON$\rangle$}}.\ Does\ this\ action}
\promptline{\rlap{\textcolor{gray}{$\hookrightarrow$}}\phantom{xx}comply\ with\ the\ domain\ policy,\ and\ has\ the\ user}
\promptline{\rlap{\textcolor{gray}{$\hookrightarrow$}}\phantom{xx}explicitly\ confirmed\ exactly\ this\ action\ (answered\ yes)}
\promptline{\rlap{\textcolor{gray}{$\hookrightarrow$}}\phantom{xx}in\ the\ conversation?}
\promptline{[Transfer\ guard]}
\promptline{The\ agent\ wants\ to\ transfer\ the\ user\ to\ a\ human\ agent}
\promptline{\rlap{\textcolor{gray}{$\hookrightarrow$}}\phantom{xx}(\textasciigrave{}transfer\_to\_human\_agents\textasciigrave{}).\ Per\ the\ policy,\ transfer}
\promptline{\rlap{\textcolor{gray}{$\hookrightarrow$}}\phantom{xx}only\ if\ the\ request\ cannot\ be\ handled\ within\ the\ scope\ of}
\promptline{\rlap{\textcolor{gray}{$\hookrightarrow$}}\phantom{xx}the\ agent\textquotesingle{}s\ tools\ and\ policy\ (or\ the\ user\ asks\ for\ a}
\promptline{\rlap{\textcolor{gray}{$\hookrightarrow$}}\phantom{xx}human\ where\ that\ is\ allowed).\ Is\ transferring\ now}
\promptline{\rlap{\textcolor{gray}{$\hookrightarrow$}}\phantom{xx}justified?}
\promptline{[Routing]}
\promptline{Does\ the\ next\ step\ require\ careful\ multi-{}step\ reasoning,}
\promptline{\rlap{\textcolor{gray}{$\hookrightarrow$}}\phantom{xx}policy\ interpretation\ or\ recovery\ from\ an\ error\ (i.e.,\ a}\nobreak
\promptline{\rlap{\textcolor{gray}{$\hookrightarrow$}}\phantom{xx}strong\ model)\ rather\ than\ a\ routine\ reply\ or\ lookup?}\nobreak
\endgroup\smallskip

\par\smallskip{\footnotesize\raggedright\interlinepenalty=10000\noindent\textbf{P18. $\tau^2$-bench: messages to the agent LLM.} Source: \texttt{src/\allowbreak{}jevharness/\allowbreak{}bench/\allowbreak{}suites/\allowbreak{}tau2.py}. A rejected proposal is not executed. The LLM answers again, with the note as a final user turn.\par}\nobreak\vspace{1pt}\nobreak
\begingroup\ttfamily\scriptsize\setlength{\fboxsep}{0pt}\setlength{\parskip}{0pt}%
\promptline{[Decision\ model\ chose\ the\ tool\ and\ the\ LLM\ fills\ the\ rest]}\nobreak
\promptline{You\ must\ now\ call\ \textasciigrave{}\textrm{\textit{$\langle$tool$\rangle$}}\textasciigrave{}.\ Already-{}determined\ arguments:}
\promptline{\rlap{\textcolor{gray}{$\hookrightarrow$}}\phantom{xx}\textrm{\textit{$\langle$the model's arguments$\rangle$}}.\ Provide\ the\ remaining\ arguments.}
\promptline{[Transfer\ guard\ rejected]}
\promptline{[Verifier]\ Transferring\ to\ a\ human\ agent\ does\ not\ look}
\promptline{\rlap{\textcolor{gray}{$\hookrightarrow$}}\phantom{xx}justified\ yet.\ Try\ to\ resolve\ the\ request\ with\ the}
\promptline{\rlap{\textcolor{gray}{$\hookrightarrow$}}\phantom{xx}available\ tools\ and\ the\ policy,\ or\ explain\ to\ the\ user}
\promptline{\rlap{\textcolor{gray}{$\hookrightarrow$}}\phantom{xx}what\ you\ can\ and\ cannot\ do.}
\promptline{[Policy\ guard\ rejected]}
\promptline{[Policy\ check]\ Your\ proposed\ action\ \textasciigrave{}\textrm{\textit{$\langle$tool$\rangle$}}(\textrm{\textit{$\langle$arguments$\rangle$}})\textasciigrave{}}
\promptline{\rlap{\textcolor{gray}{$\hookrightarrow$}}\phantom{xx}was\ judged\ likely\ to\ violate\ the\ domain\ policy\ or\ to\ lack}
\promptline{\rlap{\textcolor{gray}{$\hookrightarrow$}}\phantom{xx}the\ user\textquotesingle{}s\ explicit\ confirmation.\ Re-{}read\ the\ policy.\ If}
\promptline{\rlap{\textcolor{gray}{$\hookrightarrow$}}\phantom{xx}confirmation\ is\ required,\ list\ the\ action\ details\ and\ ask}
\promptline{\rlap{\textcolor{gray}{$\hookrightarrow$}}\phantom{xx}the\ user\ to\ confirm\ (yes)\ before\ acting.\ Decide\ again:}
\promptline{\rlap{\textcolor{gray}{$\hookrightarrow$}}\phantom{xx}respond\ to\ the\ user\ or\ call\ a\ tool.}
\promptline{[Tool\ guard\ rejected]}
\promptline{[Verifier]\ Your\ proposed\ next\ step\ (\textrm{\textit{$\langle$proposal$\rangle$}})\ was\ judged}
\promptline{\rlap{\textcolor{gray}{$\hookrightarrow$}}\phantom{xx}likely\ wrong.\ Reconsider\ and\ choose\ the\ correct\ next}\nobreak
\promptline{\rlap{\textcolor{gray}{$\hookrightarrow$}}\phantom{xx}action\ (a\ tool\ call\ or\ a\ message\ to\ the\ user).}\nobreak
\endgroup\smallskip

\par\smallskip{\footnotesize\raggedright\interlinepenalty=10000\noindent\textbf{P19. mini-SWE-agent: decision-model state.} Source: \texttt{src/\allowbreak{}jevharness/\allowbreak{}adapters/\allowbreak{}mini\_swe.py: \_s1\_state}, \texttt{adapters/\allowbreak{}mini\_swe/\allowbreak{}swebench\_jev.yaml}. The history keeps the last 8 commands and cuts each output to 400 characters. Only a blocked command carries the \texttt{[NOT EXECUTED]} tag. The questions of one step share this state and go out in one request.\par}\nobreak\vspace{1pt}\nobreak
\begingroup\ttfamily\scriptsize\setlength{\fboxsep}{0pt}\setlength{\parskip}{0pt}%
\promptline{TASK\ (issue\ to\ resolve):}\nobreak
\promptline{\textrm{\textit{$\langle$issue text$\rangle$}}}
\promptgap
\promptline{HISTORY\ (most\ recent\ last):}
\promptline{[\textrm{\textit{$\langle$step$\rangle$}}]\ \$\ \textrm{\textit{$\langle$command$\rangle$}}\ [NOT\ EXECUTED]\ -{}>{}\ rc=\textrm{\textit{$\langle$exit code$\rangle$}}:}
\promptline{\rlap{\textcolor{gray}{$\hookrightarrow$}}\phantom{xx}\textrm{\textit{$\langle$output$\rangle$}}}
\promptgap
\promptline{CURRENT\ OBSERVATION:}
\promptline{\textrm{\textit{$\langle$last output$\rangle$}}}
\promptgap
\promptline{STEP:\ \textrm{\textit{$\langle$step$\rangle$}}}
\promptgap
\promptline{[guard\ only]\ AGENT\textquotesingle{}S\ REASONING\ FOR\ THIS\ STEP:}
\promptline{\textrm{\textit{$\langle$the agent's reasoning$\rangle$}}}
\promptgap
\promptline{[stop\ guard\ only]\ The\ agent\ is\ now\ issuing\ the\ submit}\nobreak
\promptline{\rlap{\textcolor{gray}{$\hookrightarrow$}}\phantom{xx}command.}\nobreak
\endgroup\smallskip

\par\smallskip{\footnotesize\raggedright\interlinepenalty=10000\noindent\textbf{P20. mini-SWE-agent: questions and messages.} Source: \texttt{src/\allowbreak{}jevharness/\allowbreak{}adapters/\allowbreak{}mini\_swe.py}. The agent's prompts are mini-SWE-agent's SWE-bench configuration, unchanged \citep{yang2024sweagent}. The reflection, action-type and stop-trigger hooks are implemented but were not run.\par}\nobreak\vspace{1pt}\nobreak
\begingroup\ttfamily\scriptsize\setlength{\fboxsep}{0pt}\setlength{\parskip}{0pt}%
\promptline{[Routing:\ deep\ or\ fast\ path\ (E4,\ E4b\ and\ the\ H5\ route)]}\nobreak
\promptline{Does\ the\ next\ step\ require\ careful\ multi-{}step\ reasoning,}
\promptline{\rlap{\textcolor{gray}{$\hookrightarrow$}}\phantom{xx}planning,\ debugging,\ or\ recovery\ from\ an\ error\ (i.e.,\ a}
\promptline{\rlap{\textcolor{gray}{$\hookrightarrow$}}\phantom{xx}strong\ model)\ rather\ than\ a\ routine\ action\ such\ as}
\promptline{\rlap{\textcolor{gray}{$\hookrightarrow$}}\phantom{xx}reading\ a\ file,\ searching,\ or\ re-{}running\ a\ command?}
\promptgap
\promptline{[Command\ guard\ (H5)]}
\promptline{The\ coding\ agent\ proposes\ to\ run\ the\ bash\ command\ below\ in}
\promptline{\rlap{\textcolor{gray}{$\hookrightarrow$}}\phantom{xx}the\ repository.\ Is\ this\ command\ reasonable,\ on-{}track}
\promptline{\rlap{\textcolor{gray}{$\hookrightarrow$}}\phantom{xx}toward\ resolving\ the\ issue,\ and\ non-{}destructive\ (it\ does}
\promptline{\rlap{\textcolor{gray}{$\hookrightarrow$}}\phantom{xx}not\ delete\ or\ reset\ the\ repository,\ its\ tests,\ git}
\promptline{\rlap{\textcolor{gray}{$\hookrightarrow$}}\phantom{xx}history\ or\ the\ environment)?}
\promptline{COMMAND:}
\promptline{\textrm{\textit{$\langle$command$\rangle$}}}
\promptline{criteria:\ yes\ =\ The\ command\ is\ safe\ and\ a\ sensible\ next}
\promptline{\rlap{\textcolor{gray}{$\hookrightarrow$}}\phantom{xx}step\ for\ fixing\ the\ issue.}
\promptline{no\ =\ The\ command\ is\ destructive,\ violates\ the\ task}
\promptline{\rlap{\textcolor{gray}{$\hookrightarrow$}}\phantom{xx}boundaries\ (e.g.\ edits\ or\ deletes\ tests),\ or\ is}
\promptline{\rlap{\textcolor{gray}{$\hookrightarrow$}}\phantom{xx}unproductive\ /\ off-{}track.}
\promptgap
\promptline{[Submit\ guard\ (H5)]}
\promptline{The\ coding\ agent\ now\ wants\ to\ submit\ its\ patch\ and\ end\ the}
\promptline{\rlap{\textcolor{gray}{$\hookrightarrow$}}\phantom{xx}task.\ Has\ the\ fix\ been\ verified:\ was\ the\ issue}
\promptline{\rlap{\textcolor{gray}{$\hookrightarrow$}}\phantom{xx}reproduced,\ and\ were\ the\ modified\ code\ and\ the\ relevant}
\promptline{\rlap{\textcolor{gray}{$\hookrightarrow$}}\phantom{xx}tests\ actually\ run\ after\ the\ last\ source\ edit,\ with}
\promptline{\rlap{\textcolor{gray}{$\hookrightarrow$}}\phantom{xx}results\ showing\ that\ the\ issue\ is\ fixed?}
\promptline{criteria:\ yes\ =\ Verified:\ a\ reproduction\ and/or\ relevant}
\promptline{\rlap{\textcolor{gray}{$\hookrightarrow$}}\phantom{xx}tests\ were\ run\ after\ the\ final\ edit\ and\ pass.}
\promptline{no\ =\ Unverified:\ nothing\ was\ run\ after\ the\ final\ change,\ or}
\promptline{\rlap{\textcolor{gray}{$\hookrightarrow$}}\phantom{xx}the\ runs\ still\ show\ the\ bug/errors.}
\promptgap
\promptline{[blocked\ command,\ as\ its\ observation]}
\promptline{[Guard]\ This\ command\ was\ NOT\ executed:\ an\ automatic}
\promptline{\rlap{\textcolor{gray}{$\hookrightarrow$}}\phantom{xx}safety/progress\ check\ judged\ it\ likely\ destructive,}
\promptline{\rlap{\textcolor{gray}{$\hookrightarrow$}}\phantom{xx}off-{}track\ or\ unproductive\ (p\_ok=\textrm{\textit{$\langle$p$\rangle$}}\ <{}\ \textrm{\textit{$\langle$threshold$\rangle$}}).}
\promptline{\rlap{\textcolor{gray}{$\hookrightarrow$}}\phantom{xx}Reconsider\ your\ approach\ and\ issue\ a\ different\ command.}
\promptline{\rlap{\textcolor{gray}{$\hookrightarrow$}}\phantom{xx}If\ you\ are\ certain\ this\ exact\ command\ is\ necessary,}
\promptline{\rlap{\textcolor{gray}{$\hookrightarrow$}}\phantom{xx}re-{}issue\ it\ unchanged\ and\ it\ will\ run.}
\promptline{[deferred\ submission]}
\promptline{[Verifier]\ Submission\ deferred:\ the\ fix\ does\ not\ appear\ to}
\promptline{\rlap{\textcolor{gray}{$\hookrightarrow$}}\phantom{xx}be\ verified\ yet.\ Before\ submitting,\ (1)\ make\ sure\ you}
\promptline{\rlap{\textcolor{gray}{$\hookrightarrow$}}\phantom{xx}have\ reproduced\ the\ issue,\ (2)\ re-{}run\ your\ reproduction}
\promptline{\rlap{\textcolor{gray}{$\hookrightarrow$}}\phantom{xx}script\ and\ the\ most\ relevant\ existing\ tests\ against\ your}
\promptline{\rlap{\textcolor{gray}{$\hookrightarrow$}}\phantom{xx}modified\ code,\ and\ (3)\ fix\ any\ failures.\ Then\ create\ the}\nobreak
\promptline{\rlap{\textcolor{gray}{$\hookrightarrow$}}\phantom{xx}patch\ and\ submit\ again\ with\ the\ exact\ submit\ command.}\nobreak
\endgroup\smallskip

\par\smallskip{\footnotesize\raggedright\interlinepenalty=10000\noindent\textbf{P21. pi on Terminal-Bench-Science (H5).} Source: \texttt{adapters/\allowbreak{}pi/\allowbreak{}jev-{}tool-{}gate.ts}, \texttt{adapters/\allowbreak{}pi/\allowbreak{}jev-{}stop-{}verify.ts}, \texttt{adapters/\allowbreak{}pi/\allowbreak{}jev-{}reflect.ts}, \texttt{adapters/\allowbreak{}pi/\allowbreak{}jev-{}common.ts}. The agent's prompts are pi's own \citep{pi2026}, and the tasks are Terminal-Bench-Science's \citep{tbscience2026}. The router, memory, tool-pruning and compaction extensions were not run.\par}\nobreak\vspace{1pt}\nobreak
\begingroup\ttfamily\scriptsize\setlength{\fboxsep}{0pt}\setlength{\parskip}{0pt}%
\promptline{[state:\ a\ JSON\ object]}\nobreak
\promptline{\{\textquotedbl{}task\textquotedbl{}:\ \textrm{\textit{$\langle$latest user message$\rangle$}},\ \textquotedbl{}summary\textquotedbl{}:}
\promptline{\rlap{\textcolor{gray}{$\hookrightarrow$}}\phantom{xx}\textrm{\textit{$\langle$compaction summary, if any$\rangle$}},}
\promptline{\phantom{x}\textquotedbl{}history\textquotedbl{}:\ [\textquotedbl{}[\textrm{\textit{$\langle$i$\rangle$}}]\ \textrm{\textit{$\langle$tool(arguments)$\rangle$}}\ -{}>{}\ \textrm{\textit{$\langle$result$\rangle$}}\textquotedbl{},\ ...],}
\promptline{\phantom{x}\textquotedbl{}current\_observation\textquotedbl{}:\ \textrm{\textit{$\langle$latest tool result$\rangle$}},}
\promptline{\phantom{x}\textquotedbl{}proposed\_call\textquotedbl{}\ (guard)\ |\ \textquotedbl{}final\_response\textquotedbl{}\ (stop):\ \textrm{\textit{$\langle$value$\rangle$}}\}}
\promptgap
\promptline{[Tool\ guard]}
\promptline{The\ agent\ proposes\ to\ call\ \textasciigrave{}\textrm{\textit{$\langle$tool$\rangle$}}\textasciigrave{}\ with\ arguments}
\promptline{\rlap{\textcolor{gray}{$\hookrightarrow$}}\phantom{xx}\textrm{\textit{$\langle$arguments as JSON$\rangle$}}.\ Is\ this\ the\ right\ next\ step\ toward}
\promptline{\rlap{\textcolor{gray}{$\hookrightarrow$}}\phantom{xx}completing\ the\ task?}
\promptline{criteria:\ yes\ =\ The\ call\ is\ a\ correct\ and\ useful\ next\ step}
\promptline{no\ =\ The\ call\ is\ wrong,\ premature,\ unsafe,\ or\ off-{}task}
\promptline{[rejected:\ the\ tool\ result]}
\promptline{[Verifier]\ Your\ proposed\ call\ \textasciigrave{}\textrm{\textit{$\langle$tool(arguments)$\rangle$}}\textasciigrave{}\ was}
\promptline{\rlap{\textcolor{gray}{$\hookrightarrow$}}\phantom{xx}judged\ likely\ wrong.\ Reconsider\ and\ choose\ the\ correct}
\promptline{\rlap{\textcolor{gray}{$\hookrightarrow$}}\phantom{xx}next\ tool\ call.}
\promptgap
\promptline{[Stop\ guard,\ when\ the\ agent\ settles]}
\promptline{The\ agent\ wants\ to\ stop\ and\ submit.\ Is\ the\ task\ actually}
\promptline{\rlap{\textcolor{gray}{$\hookrightarrow$}}\phantom{xx}complete?}
\promptline{criteria:\ yes\ =\ Every\ requirement\ of\ the\ task\ is\ met,\ and}
\promptline{\rlap{\textcolor{gray}{$\hookrightarrow$}}\phantom{xx}the\ result\ was\ checked}
\promptline{no\ =\ Work\ remains,\ or\ the\ result\ is\ wrong\ or\ unverified}
\promptline{[rejected:\ a\ message,\ then\ one\ more\ turn]}
\promptline{[Verifier]\ The\ task\ does\ not\ look\ complete\ yet.\ Re-{}read\ the}
\promptline{\rlap{\textcolor{gray}{$\hookrightarrow$}}\phantom{xx}task,\ check\ every\ requirement\ against\ the\ current\ state}
\promptline{\rlap{\textcolor{gray}{$\hookrightarrow$}}\phantom{xx}(run\ the\ tests\ or\ inspect\ the\ outputs),\ and\ continue}
\promptline{\rlap{\textcolor{gray}{$\hookrightarrow$}}\phantom{xx}working\ before\ you\ finish.}
\promptgap
\promptline{[Reflection,\ after\ a\ failed\ or\ repeated\ call]}
\promptline{Is\ the\ agent\ stuck,\ repeating\ itself,\ or\ did\ the\ last}
\promptline{\rlap{\textcolor{gray}{$\hookrightarrow$}}\phantom{xx}action\ fail\ or\ move\ away\ from\ the\ goal?}
\promptline{criteria:\ yes\ =\ Stuck,\ looping,\ failing,\ or\ drifting\ from}
\promptline{\rlap{\textcolor{gray}{$\hookrightarrow$}}\phantom{xx}the\ goal}
\promptline{no\ =\ Making\ steady\ progress\ toward\ the\ goal}
\promptline{[fired:\ a\ message]}
\promptline{[Reflection]\ Pause\ before\ your\ next\ tool\ call.\ In\ at\ most\ 4}
\promptline{\rlap{\textcolor{gray}{$\hookrightarrow$}}\phantom{xx}sentences:\ assess\ progress\ toward\ the\ task,\ identify\ any}
\promptline{\rlap{\textcolor{gray}{$\hookrightarrow$}}\phantom{xx}mistake\ so\ far,\ and\ state\ the\ best\ next\ step.\ Then}\nobreak
\promptline{\rlap{\textcolor{gray}{$\hookrightarrow$}}\phantom{xx}continue\ working.}\nobreak
\endgroup\smallskip

\par\smallskip{\footnotesize\raggedright\interlinepenalty=10000\noindent\textbf{P22. L1 static decisions (BFCL, When2Call).} Source: \texttt{src/\allowbreak{}jevharness/\allowbreak{}bench/\allowbreak{}suites/\allowbreak{}static.py}, \texttt{src/\allowbreak{}jevharness/\allowbreak{}bench/\allowbreak{}decision\_eval.py}. The argument question names the reference function on purpose, so that it measures argument filling given the right function. The native function-calling baseline sees each benchmark's own messages and tools.\par}\nobreak\vspace{1pt}\nobreak
\begingroup\ttfamily\scriptsize\setlength{\fboxsep}{0pt}\setlength{\parskip}{0pt}%
\promptline{[BFCL\ state,\ one\ line\ per\ message]}\nobreak
\promptline{\textrm{\textit{$\langle$ROLE$\rangle$}}:\ \textrm{\textit{$\langle$message$\rangle$}}}
\promptline{[BFCL\ tool\ choice]}
\promptline{Which\ function\ should\ be\ called\ to\ handle\ the\ user\textquotesingle{}s}
\promptline{\rlap{\textcolor{gray}{$\hookrightarrow$}}\phantom{xx}request?}
\promptline{Options:\ the\ functions,\ then}
\promptline{none\ (no\ available\ function\ fits\ the\ request)\ -{}-{}\ Use\ when}
\promptline{\rlap{\textcolor{gray}{$\hookrightarrow$}}\phantom{xx}no\ function\ applies.}
\promptline{[BFCL\ argument,\ given\ the\ reference\ function]}
\promptline{The\ function\ \textasciigrave{}\textrm{\textit{$\langle$reference function$\rangle$}}\textasciigrave{}\ will\ be\ called.\ Which}
\promptline{\rlap{\textcolor{gray}{$\hookrightarrow$}}\phantom{xx}value\ should\ its\ argument\ \textasciigrave{}\textrm{\textit{$\langle$argument$\rangle$}}\textasciigrave{}\ take?}
\promptline{\rlap{\textcolor{gray}{$\hookrightarrow$}}\phantom{xx}(\textrm{\textit{$\langle$argument description$\rangle$}})}
\promptline{Options:\ enum\ or\ boolean\ values,\ or\ spans\ copied\ from\ the}
\promptline{\rlap{\textcolor{gray}{$\hookrightarrow$}}\phantom{xx}request,\ then\ <{}OTHER:\ none\ of\ these\ /\ needs\ generation>{}}
\promptgap
\promptline{[When2Call\ state]}
\promptline{AVAILABLE\ TOOLS:}
\promptline{-{}\ \textrm{\textit{$\langle$name$\rangle$}}:\ \textrm{\textit{$\langle$description$\rangle$}}}
\promptline{\phantom{xxxx}*\ \textrm{\textit{$\langle$argument$\rangle$}}\ (\textrm{\textit{$\langle$type$\rangle$}},\ \textrm{\textit{$\langle$required|optional$\rangle$}}):}
\promptline{\rlap{\textcolor{gray}{$\hookrightarrow$}}\phantom{xx}\textrm{\textit{$\langle$description$\rangle$}}}
\promptline{\textrm{\textit{[one block per tool]}}}
\promptgap
\promptline{USER:\ \textrm{\textit{$\langle$question$\rangle$}}}
\promptline{[When2Call]}
\promptline{How\ should\ the\ assistant\ respond\ to\ the\ user\textquotesingle{}s\ message?}
\promptline{-{}\ call\ one\ of\ the\ available\ tools}
\promptline{-{}\ ask\ the\ user\ for\ missing\ required\ information}
\promptline{-{}\ say\ it\ cannot\ help:\ no\ available\ tool\ fits\ and\ the}
\promptline{\rlap{\textcolor{gray}{$\hookrightarrow$}}\phantom{xx}question\ needs\ one}
\promptline{-{}\ answer\ directly\ without\ any\ tool}
\promptgap
\promptline{[Judge\ of\ the\ native\ function-{}calling\ baseline\textquotesingle{}s\ text}
\promptline{\rlap{\textcolor{gray}{$\hookrightarrow$}}\phantom{xx}replies]}
\promptline{You\ classify\ an\ AI\ assistant\textquotesingle{}s\ reply\ to\ a\ user\textquotesingle{}s\ request.}
\promptline{\rlap{\textcolor{gray}{$\hookrightarrow$}}\phantom{xx}The\ assistant\ had\ these\ tools:}
\promptline{\textrm{\textit{$\langle$tools$\rangle$}}}
\promptgap
\promptline{USER:\ \textrm{\textit{$\langle$request$\rangle$}}}
\promptgap
\promptline{ASSISTANT\ REPLY\ (it\ did\ not\ call\ a\ tool):\ \textrm{\textit{$\langle$reply$\rangle$}}}
\promptgap
\promptline{Which\ kind\ of\ reply\ is\ it?}
\promptline{-{}\ request\_for\_info:\ it\ asks\ the\ user\ for\ missing}
\promptline{\rlap{\textcolor{gray}{$\hookrightarrow$}}\phantom{xx}information\ or\ clarification\ it\ needs\ to\ proceed}
\promptline{-{}\ cannot\_answer:\ it\ says\ it\ cannot\ help\ /\ does\ not\ have\ a}
\promptline{\rlap{\textcolor{gray}{$\hookrightarrow$}}\phantom{xx}suitable\ tool\ or\ ability}
\promptline{-{}\ direct:\ it\ answers\ or\ acts\ on\ the\ request\ directly\ (gives}
\promptline{\rlap{\textcolor{gray}{$\hookrightarrow$}}\phantom{xx}an\ answer,\ or\ claims\ to\ have\ done\ it)}
\promptline{Reply\ with\ ONLY\ a\ JSON\ object:\ \{\textquotedbl{}label\textquotedbl{}:}\nobreak
\promptline{\rlap{\textcolor{gray}{$\hookrightarrow$}}\phantom{xx}\textquotedbl{}<{}request\_for\_info|cannot\_answer|direct>{}\textquotedbl{}\}}\nobreak
\endgroup\smallskip

\subsection{Gold Labels, Probes and Measurement Arms}
\label{app:metagold}
These definitions were fixed before the main runs. Labels are computed on a copy of the environment after the
decision model's answer is recorded and go only to the decision records. A canary test plants a gold value and checks that no request contains it.

\begin{table*}[!t]
\centering\footnotesize
\setlength{\tabcolsep}{5pt}
\begin{tabular}{@{}llll@{}}
\toprule
Guard & Rate (\%) [95\% CI] (proposals) & \systwo{} Flash & \systwo{} Flash-Lite\\
\midrule
Stop & FAR, model-generated wrong finishes & \res{e2b_jevgym_meta_control/meta_shadow/decisions/gold/probes/stop/far_model/far} [\respct{e2b_jevgym_meta_control/meta_shadow/decisions/gold/probes/stop/far_model/far_ci/lo}, \respct{e2b_jevgym_meta_control/meta_shadow/decisions/gold/probes/stop/far_model/far_ci/hi}] (\res{e2b_jevgym_meta_control/meta_shadow/decisions/gold/probes/stop/far_model/n}) & \res{e2b_jevgym_meta_control/meta_shadow_s2lite/decisions/gold/probes/stop/far_model/far} [\respct{e2b_jevgym_meta_control/meta_shadow_s2lite/decisions/gold/probes/stop/far_model/far_ci/lo}, \respct{e2b_jevgym_meta_control/meta_shadow_s2lite/decisions/gold/probes/stop/far_model/far_ci/hi}] (\res{e2b_jevgym_meta_control/meta_shadow_s2lite/decisions/gold/probes/stop/far_model/n})\\
 & FAR, the LLM's own wrong finishes & -- (0) & \res{e2b_jevgym_meta_control/meta_shadow_s2lite/decisions/gold/probes/stop/far_nat/far} [\respct{e2b_jevgym_meta_control/meta_shadow_s2lite/decisions/gold/probes/stop/far_nat/far_ci/lo}, \respct{e2b_jevgym_meta_control/meta_shadow_s2lite/decisions/gold/probes/stop/far_nat/far_ci/hi}] (\res{e2b_jevgym_meta_control/meta_shadow_s2lite/decisions/gold/probes/stop/far_nat/n})\\
 & FRR, the LLM's own right finishes & \res{e2b_jevgym_meta_control/meta_shadow/decisions/gold/probes/stop/frr/frr} [\respct{e2b_jevgym_meta_control/meta_shadow/decisions/gold/probes/stop/frr/frr_ci/lo}, \respct{e2b_jevgym_meta_control/meta_shadow/decisions/gold/probes/stop/frr/frr_ci/hi}] (\res{e2b_jevgym_meta_control/meta_shadow/decisions/gold/probes/stop/frr/n}) & \res{e2b_jevgym_meta_control/meta_shadow_s2lite/decisions/gold/probes/stop/frr/frr} [\respct{e2b_jevgym_meta_control/meta_shadow_s2lite/decisions/gold/probes/stop/frr/frr_ci/lo}, \respct{e2b_jevgym_meta_control/meta_shadow_s2lite/decisions/gold/probes/stop/frr/frr_ci/hi}] (\res{e2b_jevgym_meta_control/meta_shadow_s2lite/decisions/gold/probes/stop/frr/n})\\
Tool & FAR, fatal proposals (edits, transplants) & \res{e2b_jevgym_meta_control/meta_shadow/decisions/gold/probes/tool_select/far_fatal/far} [\respct{e2b_jevgym_meta_control/meta_shadow/decisions/gold/probes/tool_select/far_fatal/far_ci/lo}, \respct{e2b_jevgym_meta_control/meta_shadow/decisions/gold/probes/tool_select/far_fatal/far_ci/hi}] (\res{e2b_jevgym_meta_control/meta_shadow/decisions/gold/probes/tool_select/far_fatal/n}) & \res{e2b_jevgym_meta_control/meta_shadow_s2lite/decisions/gold/probes/tool_select/far_fatal/far} [\respct{e2b_jevgym_meta_control/meta_shadow_s2lite/decisions/gold/probes/tool_select/far_fatal/far_ci/lo}, \respct{e2b_jevgym_meta_control/meta_shadow_s2lite/decisions/gold/probes/tool_select/far_fatal/far_ci/hi}] (\res{e2b_jevgym_meta_control/meta_shadow_s2lite/decisions/gold/probes/tool_select/far_fatal/n})\\
 & FAR against progress (own, transplanted) & \res{e2b_jevgym_meta_control/meta_shadow/decisions/gold/probes/tool_select/headline/far} (\res{e2b_jevgym_meta_control/meta_shadow/decisions/gold/probes/tool_select/headline/n_wrong}) & \res{e2b_jevgym_meta_control/meta_shadow_s2lite/decisions/gold/probes/tool_select/headline/far} [\respct{e2b_jevgym_meta_control/meta_shadow_s2lite/decisions/gold/probes/tool_select/headline/far_ci/lo}, \respct{e2b_jevgym_meta_control/meta_shadow_s2lite/decisions/gold/probes/tool_select/headline/far_ci/hi}] (\res{e2b_jevgym_meta_control/meta_shadow_s2lite/decisions/gold/probes/tool_select/headline/n_wrong})\\
 & FRR against progress (own, transplanted) & \res{e2b_jevgym_meta_control/meta_shadow/decisions/gold/probes/tool_select/headline/frr} [\respct{e2b_jevgym_meta_control/meta_shadow/decisions/gold/probes/tool_select/headline/frr_ci/lo}, \respct{e2b_jevgym_meta_control/meta_shadow/decisions/gold/probes/tool_select/headline/frr_ci/hi}] (\res{e2b_jevgym_meta_control/meta_shadow/decisions/gold/probes/tool_select/headline/n_right}) & \res{e2b_jevgym_meta_control/meta_shadow_s2lite/decisions/gold/probes/tool_select/headline/frr} [\respct{e2b_jevgym_meta_control/meta_shadow_s2lite/decisions/gold/probes/tool_select/headline/frr_ci/lo}, \respct{e2b_jevgym_meta_control/meta_shadow_s2lite/decisions/gold/probes/tool_select/headline/frr_ci/hi}] (\res{e2b_jevgym_meta_control/meta_shadow_s2lite/decisions/gold/probes/tool_select/headline/n_right})\\
\bottomrule
\end{tabular}
\caption{Error rates of the \jev{} guards on \gym{} probes labelled by the answer key (Stage-1 measurement arms at
$\tau = 0.5$, task-clustered bootstrap intervals, descriptive). FAR: share of wrong proposals accepted. FRR: share of
right proposals rejected. Model-generated wrong finishes pool the LLM's own, forced-premature and transplanted ones.
Flash made no wrong finish of its own (--).}
\label{tab:guardfar}
\end{table*}

\paragraph{Verdicts and gold labels.}
Let $D(s)$ be the number of steps the answer key needs from a copy of $s$ to finish successfully ($\infty$ if it
cannot within the step budget, and such \emph{doomed} states are excluded). We label an attempt to finish \emph{progress} if it succeeds, a \emph{wrong answer} if everything but the answer is done, and \emph{premature} otherwise. Any other call
is \emph{fatal} if it makes $D$ infinite, \emph{progress} if it lowers $D$, and a \emph{detour} otherwise. Fatal,
premature and wrong-answer actions are \emph{failing}. A tool choice is right if it is an undone plan step that is
progress and whose inputs are visible. An argument is right if it matches the canonical value (up to case, whitespace
and 0.01 in amounts), and a guard verdict if it matches whether the proposal is progress. A reflection or routing
trigger is right if the next action without it is failing, a stop trigger if only the answer remains, and a memory
write if the text holds a value a remaining plan step consumes.
From its JSON schema alone, each argument is \emph{bounded} (enum, boolean, integer range of at most 20 values),
\emph{copyable} (numbers, and strings with a format or pattern or named like an identifier or a quantity) or
\emph{free}. The copy-dependent arguments of H3 (App.~\ref{app:theory}) are the copyable ones.

\paragraph{Measurement arms and probes.}
Two measurement arms, Flash on 200 tasks and Flash-Lite on 500, act like LLM-only while every gate runs in shadow, so
its label is observed at every state. On Flash-Lite, one-step branches replay states with the intervention and with a
\emph{placebo} that only resamples the default call. Relative to the placebo, $e$ is the gain in the chance of progress when the trigger fires before a failing action, and $h$ is the gain in the chance of failure when it fires before a progress action. A
\emph{probe} is a proposal the guard scores but that is never executed (Table~\ref{tab:guardfar}), and probes enter
no reported success, cost, test or training data. A gate cell is \emph{estimable} for H2 when its arm has at least 30
positive and 30 negative labels and a live and a matched-rate random arm exist. The reflection-trigger test has \emph{no headroom} when, at $W = \$1$, the 95\% CI of its one-step perfect-information ceiling (Eq.~\ref{eq:voi} with $J = 1$ and the measured $e + h$) lies within $\pm$3.3 points, the registered minimum detectable effect.

\subsection{Benchmark Difficulty: Hard States}
\label{app:difficulty}
\newcommand{\hardrow}[2]{#1 & \resna{diff/c/all/#2/acc} & \resna{diff/c/hard/#2/acc} & \resna{diff/c/after_off_plan/#2/acc}
  & \resna{diff/c/ask_gold/#2/acc} & \resna{diff/c/llm_states/#2/acc}\\}
\newcommand{\hardhead}[1]{\res{diff/c/all/#1} & \res{diff/c/hard/#1} & \res{diff/c/after_off_plan/#1}
  & \res{diff/c/ask_gold/#1} & \res{diff/c/llm_states/#1}\\}
\begin{table}[!htbp]
\centering\scriptsize
\setlength{\tabcolsep}{3pt}
\begin{tabular}{@{}lccccc@{}}
\toprule
& All & Hard & After & \texttt{ask\_user} & Flash\\
& states & (union) & off-plan & next & states\\
\midrule
States $n$ & \hardhead{n}
Tasks & \hardhead{tasks}
\midrule
\hardrow{Agent LLM (Flash)$^a$}{llm}
\midrule
\hardrow{\jev{}}{jev}
\hardrow{Intern-Decision-4B}{intern}
\hardrow{StartLux-27B}{sl27b}
\hardrow{StartLux-9B$^\ddagger$}{sl9b}
\hardrow{StartLux-4B}{sl4b}
\hardrow{StartLux-2B$^\ddagger$}{sl2b}
\hardrow{StartLux-0.8B$^\ddagger$}{sl0.8b}
\hardrow{Qwen3-4B (untrained)}{base}
\midrule
\hardrow{SFT v1$^\ast$}{sft_v1}
\hardrow{\oursV{}$^\ast$}{rl}
\hardrow{StartLux-4B SFT (track 2)$^{\ast\ddagger}$}{v2}
\midrule
Spread (points)$^b$ & \res{diff/c/all/spread} & \res{diff/c/hard/spread} & \res{diff/c/after_off_plan/spread}
  & \res{diff/c/ask_gold/spread} & \res{diff/c/llm_states/spread}\\
\bottomrule
\end{tabular}
\caption{Tool-choice accuracy (task-weighted \%) on the common states and on subsets defined by a rule after the main study (post hoc) but before any accuracy on them was computed. \emph{Hard}: after an off-plan step, or the next tool is \texttt{ask\_user}.
\emph{Flash states}: the LLM-only trajectories. $^a$Its own choice on its own trajectories. $^b$Over the eight models
of Figure~\ref{fig:levels}. $^\ast$Trained on \gym{}'s generator. $^\ddagger$Exploratory.}
\label{tab:hardstates}
\end{table}

Using stored answers only, Table~\ref{tab:hardstates} shows which models find \gym{} hard. Acting alone, Flash solved
\res{diff/ceiling/stage2_flash/success}\% of the 500 fresh Stage-2 tasks, so a decision model can lower cost but has
little success to add. The seven off-the-shelf decision models chose an acceptable tool at
\res{diff/c/all/offshelf/min}--\res{diff/c/all/offshelf/max}\% of all common states and at
\res{diff/c/hard/offshelf/min}--\res{diff/c/hard/offshelf/max}\% of the hard ones. The generator-trained rows are near
ceiling on every subset (App.~\ref{app:hygiene}).

\section{\harnessname{}}
\label{app:harness}
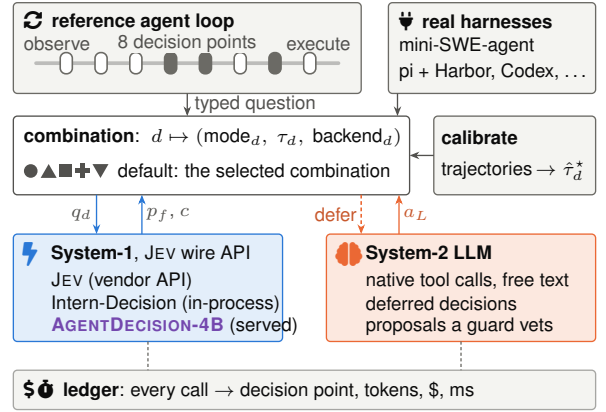
\begin{figure}[t]
\centering
\begin{tikzpicture}[jevfig]
\def\W{77.0}
\draw[fbox=envC, fill=envT] (0,48.4) rectangle (46.0,60.4);
\node[flbl, anchor=west] at (1.4,58.0) {\faIcon{sync-alt}\enspace\textbf{reference agent loop}};
\def\yk{52.4}
\node[fnote, anchor=west] at (1.4,55.0) {observe};
\node[fnote, anchor=east] at (44.6,55.0) {execute};
\draw[frail] (3.0,\yk) -- (43.0,\yk);
\foreach \x/\o in {7.0/0, 11.6/0, 16.2/0, 20.8/1, 25.4/1, 30.0/0, 34.6/1, 39.2/0} {
  \ifnum\o=1 \node[fobjtap, minimum height=2.8mm] at (\x,\yk) {}; \else \node[fmetatap, minimum height=2.8mm] at (\x,\yk) {}; \fi}
\node[fnote, fill=envT, inner xsep=0.4mm] at (23.0,55.0) {8 decision points};
\draw[fbox=envC, fill=envT] (49.6,48.4) rectangle (\W,60.4);
\node[flbl, anchor=west] at (51.0,58.0) {\faPlug\enspace\textbf{real harnesses}};
\node[fnote, anchor=west, text=ink] at (51.0,54.6) {mini-SWE-agent};
\node[fnote, anchor=west, text=ink] at (51.0,51.2) {pi + Harbor, Codex, \ldots};
\draw[fbox=ink2, fill=white] (0,35.0) rectangle (52.4,45.4);
\node[flbl, anchor=west] at (1.4,42.6) {\textbf{combination}: \;$d \mapsto (\text{mode}_d,\ \tau_d,\ \text{backend}_d)$};
\node[fnote, anchor=west, text=ink] at (1.4,38.2) {%
  \figmki{Rep}\,\figmki{Cas}\,\figmki{Ass}\,\figmki{Gua}\,\figmki{Tri}\enspace default: the selected combination};
\draw[fbox=envC, fill=envT] (55.6,35.0) rectangle (\W,45.4);
\node[flbl, anchor=west] at (56.6,42.6) {\textbf{calibrate}};
\node[fnote, anchor=west, text=ink] at (56.6,38.2) {trajectories\,$\to\hat\tau^\star_d$};
\draw[farr] (55.6,40.2) -- (52.4,40.2);
\draw[farr] (23.0,48.4) -- (23.0,45.4);
\draw[farr] (50.8,48.4) -- (50.8,45.4);
\node[fnote, anchor=west] at (24.0,46.9) {typed question};
\draw[fbox=sOne, fill=sOneT] (0,15.6) rectangle (35.6,29.8);
\node[fs1g, anchor=west] at (1.2,27.2) {\faBolt};
\node[flbl, anchor=west] at (5.0,27.2) {\textbf{\sysone{}}, \jev{} wire API};
\node[fnote, anchor=west, text=ink] at (5.0,23.6) {\jev{} (vendor API)};
\node[fnote, anchor=west, text=ink] at (5.0,20.6) {Intern-Decision (in-process)};
\node[fnote, anchor=west, text=ink] at (5.0,17.6) {\ours{} (served)};
\draw[fbox=sTwo, fill=sTwoT] (41.4,15.6) rectangle (\W,29.8);
\node[fs2g, anchor=west] at (42.6,27.2) {\faBrain};
\node[flbl, anchor=west] at (46.6,27.2) {\textbf{\systwo{} LLM}};
\node[fnote, anchor=west, text=ink] at (46.6,23.6) {native tool calls, free text};
\node[fnote, anchor=west, text=ink] at (46.6,20.6) {deferred decisions};
\node[fnote, anchor=west, text=ink] at (46.6,17.6) {proposals a guard vets};
\draw[farr, draw=sOne] (11.0,35.0) -- (11.0,29.8);
\draw[farr, draw=sOne] (17.0,29.8) -- (17.0,35.0);
\node[fnote, anchor=east] at (10.4,32.4) {$q_d$};
\node[fnote, anchor=west] at (17.6,32.4) {$p_f$, $c$};
\draw[fdefer] (46.0,35.0) -- (46.0,29.8);
\draw[farr, draw=sTwo] (51.0,29.8) -- (51.0,35.0);
\node[fnote, anchor=east, text=sTwo!85!black] at (45.4,32.4) {defer};
\node[fnote, anchor=west, text=sTwo!85!black] at (51.6,32.4) {$a_L$};
\draw[fbox=envC!60, fill=envT] (0,6.6) rectangle (\W,11.8);
\node[fnote, anchor=west, text=ink] at (1.4,9.2) {\faDollarSign\,\faStopwatch\enspace\textbf{ledger}: every call $\to$
  decision point, tokens, \$, ms};
\draw[envC, line width=0.5pt, dash pattern=on 1.2pt off 0.9pt] (17.8,15.6) -- (17.8,11.8);
\draw[envC, line width=0.5pt, dash pattern=on 1.2pt off 0.9pt] (59.2,15.6) -- (59.2,11.8);
\end{tikzpicture}
\caption{\textbf{Architecture of \harnessname{}.} A typed question from the reference loop, or from a
hook of an existing harness, goes to the point's entry in the combination (mode, threshold, \sysone{} backend), whose mode executes, narrows,
vets or defers \sysone{}'s answer. One ledger meters every call.}
\label{fig:harness}
\end{figure}
\paragraph{Combinations, call graph and serving.}
A YAML file defines each combination of delegations (Figure~\ref{fig:harness}) by setting mode, $\tau$ and $k$ per decision point for each agent LLM, and every episode records the file's hash. Under \modeCas{}, below the tool threshold
the LLM makes the whole tool call. Otherwise each bounded or copyable argument of the accepted tool is its own
question with its own threshold, and the LLM fills the other arguments in one forced call. Any service that answers
\jev{}'s request format can be the decision model. Decision-model cache hits replay the original cost and latency,
and LLM calls are never cached.

\paragraph{Integrations.}
Table~\ref{tab:affordances} lists the modes that the adapters for mini-SWE-agent \citep{yang2024sweagent}, pi
\citep{pi2026} via Harbor \citep{harbor2026}, Codex CLI \citep{codexcli2026} and the DeepSeek Harness
\citep{deepseekharness2026} can express. \modeRep{} and \modeCas{} would need a hook that returns the next action
before the LLM is called.

\begin{table}[!htbp]
\centering\footnotesize
\setlength{\tabcolsep}{3pt}
\resizebox{\columnwidth}{!}{%
\begin{tabular}{@{}lccccc@{}}
\toprule
Mode & Own loop & pi & Codex & DSH & mini-SWE\\
\midrule
\modeRep{}/\modeCas{} tool & \checkmark & -- & -- & -- & --\\
\modeAss{} (prune tools) & \checkmark & \checkmark & -- & -- & n/a\\
\modeGua{} (approve) & \checkmark & \checkmark & deny only & \checkmark & \checkmark\\
\modeTri{} stop & \checkmark & \checkmark & \checkmark & \checkmark & \checkmark\\
\modeTri{} routing & \checkmark & \checkmark & record only & \checkmark & \checkmark\\
\modeTri{} memory/reflect & \checkmark & \checkmark & \checkmark & \checkmark & \checkmark\\
\midrule
Status & E & S, E & I & I & S, E\\
\bottomrule
\end{tabular}}
\caption{Modes each harness can express without forking. DSH: DeepSeek Harness, also serving Claude Code. --: not
expressible. n/a: mini-SWE-agent's action is one bash command. Status: I = implemented and unit-tested, S =
smoke-tested end to end, E = evaluated.}
\label{tab:affordances}
\end{table}

\subsection{The Final Harness (Exploratory)}
\label{app:harnessfinal}
Table~\ref{tab:harnessfinal} compares the final harness (\S\ref{sec:results:deploy}) with LLM-only on the same
\res{hf/n} fresh tasks, with the GPU shared by about four episodes, as in Stage~2, or dedicated to one. Cost includes
the GPU at \$1.624 per hour. With four episodes per GPU, queueing for the decision model took
\res{hf/diag/rl/queue_share}\% of the episode time.

\begin{table}[!htbp]
\centering\small
\setlength{\tabcolsep}{3.5pt}
\resizebox{\columnwidth}{!}{%
\begin{tabular}{@{}lrr@{}}
\toprule
 & shared GPU & dedicated GPU \\
\midrule
Success, LLM-only (\%) & \res{hf/success/llm_only} & \res{hf_ded/success/llm_only} \\
Success, final harness (\%) & \res{hf/success/harness_final} & \res{hf_ded/success/harness_final} \\
\quad difference [95\% CI] & \res{hf/success/delta} [\res{hf/success/delta_ci/lo}, \res{hf/success/delta_ci/hi}] & \res{hf_ded/success/delta} [\res{hf_ded/success/delta_ci/lo}, \res{hf_ded/success/delta_ci/hi}] \\
\midrule
\$ / episode, LLM-only & \res{hf/cost/llm_only} & \res{hf_ded/cost/llm_only} \\
\$ / episode, final harness & \res{hf/cost/harness_final} & \res{hf_ded/cost/harness_final} \\
\quad of which GPU & \res{hf/cost/q} & \res{hf_ded/cost/q} \\
\quad saving (\%) [95\% CI] & \res{hf/cost/saving} [\res{hf/cost/saving_ci/lo}, \res{hf/cost/saving_ci/hi}] & \res{hf_ded/cost/saving} [\res{hf_ded/cost/saving_ci/lo}, \res{hf_ded/cost/saving_ci/hi}] \\
\midrule
s / episode, LLM-only & \res{hf/wall_s/llm_only} (\res{hf/wall_s/median/llm_only}) & \res{hf_ded/wall_s/llm_only} (\res{hf_ded/wall_s/median/llm_only}) \\
s / episode, final harness & \res{hf/wall_s/harness_final} (\res{hf/wall_s/median/harness_final}) & \res{hf_ded/wall_s/harness_final} (\res{hf_ded/wall_s/median/harness_final}) \\
\quad difference [95\% CI] & \res{hf/wall_s/delta} [\res{hf/wall_s/delta_ci/lo}, \res{hf/wall_s/delta_ci/hi}] & \res{hf_ded/wall_s/delta} [\res{hf_ded/wall_s/delta_ci/lo}, \res{hf_ded/wall_s/delta_ci/hi}] \\
\bottomrule
\end{tabular}}
\caption{The final harness against LLM-only on \res{hf/n} fresh \gym{} tasks on one A10G, with the GPU shared
or dedicated (exploratory). Success in \%, its difference in points. Time: mean (median) s per episode, with the
difference of means.}
\label{tab:harnessfinal}
\end{table}
\subsection{\harnessname{} in mini-SWE-agent (Exploratory)}
\label{app:e4}
In these registered exploratory runs, \jev{} routes steps in an unmodified mini-SWE-agent (Table~\ref{tab:e4}). Before each step \jev{} is asked whether the step needs careful reasoning, planning, debugging or
error recovery. The strong LLM takes the step if \jev{} says \emph{yes} with probability at least $\tau$ =
\res{e4/tau} or gives no answer, and the fast LLM takes it otherwise.
\ifcsname jevres@e4r/jevagent/success\endcsname
On \res{e4r/n_eval} instances (the first \res{e4r/n_rep} and \res{e4r/n_ext} more in a registered extension), routing
resolved \res{e4r/jevagent/resolved} against \res{e4r/always_deep/resolved} for the strong LLM alone at
\res{e4r/jevagent/vs_deep/saving_pct}\% lower cost, without establishing non-inferiority
(Table~\ref{tab:mainresults-ci}). The new instances alone saved \res{e4r/ext/jevagent/vs_deep/saving_pct}\%.
\else
On the first \res{e4/n_eval} evaluation instances, routing resolved as many as the strong LLM alone at
\res{e4/jevagent/vs_deep/saving_pct}\% lower cost, and the fast LLM alone cost less still.
\fi
With Gemini~3.8 Flash in both roles, the fast LLM alone resolved more than both other arms.

\begin{table}[!t]
\centering\scriptsize
\setlength{\tabcolsep}{2.2pt}
\providecommand{\eFour}[3]{#1 & \res{#2/#3/n} & \res{#2/#3/resolved} & \res{#2/#3/vs_deep/wins}/\res{#2/#3/vs_deep/losses}
  & \res{#2/#3/cost_usd} & \res{#2/#3/vs_deep/saving_pct} & \res{#2/#3/wall_median_s} & \res{#2/#3/wall_mean_s}
  & \res{#2/#3/fast_share} & \res{#2/#3/limits_exceeded}\\}
\providecommand{\eFourRef}[2]{Strong alone & \res{#1/always_deep/n} & \res{#1/always_deep/resolved} & -- & \res{#1/always_deep/cost_usd}
  & -- & \res{#1/always_deep/wall_median_s} & \res{#1/always_deep/wall_mean_s} & \res{#1/always_deep/fast_share}
  & \res{#1/always_deep/limits_exceeded}\\}
\resizebox{\columnwidth}{!}{%
\begin{tabular}{@{}lrrrrrrrrr@{}}
\toprule
& & & & & Saved & \multicolumn{2}{c}{Time (s)} & Fast & \\
\cmidrule(lr){7-8}
Arm & $n$ & Res. & W/L & \$/inst. & (\%) & med. & mean & (\%) & Limit\\
\midrule
\multicolumn{10}{@{}l}{\emph{DeepSeek-V4: strong = V4-Pro with thinking, fast = V4-Flash without}}\\
\eFourRef{e4}{}
\eFour{Routing ($\tau$ \res{e4/tau})}{e4}{jevagent}
\eFour{Fast alone}{e4}{always_fast}
\midrule
\multicolumn{10}{@{}l}{\emph{Gemini 3.8 Flash: strong = thinking high, fast = low (no development run)}}\\
\eFourRef{e4b}{}
\eFour{Routing ($\tau$ 0.50)}{e4b}{jevagent}
\eFour{Fast alone$^\ast$}{e4b}{always_fast}
\bottomrule
\end{tabular}}
\caption{Step routing by \jev{} in mini-SWE-agent on the same \res{e4/n_eval} fresh SWE-bench Verified instances
(exploratory, one run per instance). Res.: resolved. W/L: resolved only by the arm / only by the strong LLM alone.
\$/inst.: API cost per instance, \jev{} included. Fast: share of LLM calls to the fast LLM. Limit: stopped by the step
or cost limit. $^\ast$First \res{e4b/always_fast/n} instances only, compared with the strong LLM on the same ones.}
\label{tab:e4}
\end{table}
\ifcsname jevres@e25/tb/jev_router/success\endcsname
\subsection{Routing in Two More Open-Source Harnesses}
\label{app:e25}
We added the routing of App.~\ref{app:e4} to pi on all \res{e25/tb/baseline/n} tasks of Terminal-Bench~2.0 and to
SWE-agent on \res{e25/swe/baseline/n} held-out SWE-bench Verified instances, with Gemini~3.5 Flash as the strong and
Flash-Lite as the fast LLM and nothing tuned (Table~\ref{tab:e25}). In both harnesses most
requests stayed with the strong LLM, so routing saved little cost and no time. It kept success in pi. In SWE-agent the
routed agent needed more steps per instance (\res{e25/swe/jev_router/steps_mean} against
\res{e25/swe/baseline/steps_mean}).

\makeatletter
\providecommand{\eTwentyFiveRow}[3]{%
  \ifcsname jevres@e25/#1/#2/n\endcsname
  #3 & \res{e25/#1/#2/n} & \res{e25/#1/#2/success}
  & \ifcsname jevres@e25/#1/#2/delta\endcsname\resm{e25/#1/#2/delta} [\resm{e25/#1/#2/delta_lo}, \resm{e25/#1/#2/delta_hi}]\else--\fi
  & \resna{e25/#1/#2/ni}
  & \res{e25/#1/#2/cost_usd} & \resna{e25/#1/#2/cost_change}
  & \res{e25/#1/#2/wall_mean_s} / \res{e25/#1/#2/wall_median_s} & \resna{e25/#1/#2/time_change}
  & \res{e25/#1/#2/taken_share}\\
  \fi}
\makeatother
\begin{table}[!htbp]
\centering\footnotesize
\setlength{\tabcolsep}{2.5pt}
\resizebox{\columnwidth}{!}{%
\begin{tabular}{@{}l r r r c r r r r r@{}}
\toprule
Arm & $n$ & Succ.\ (\%) & $\Delta$ pt [95\,\%] & NI & \$/task & Cost (\%) & Time (s) & Time (\%) & Fast (\%)\\
\midrule
\multicolumn{10}{@{}l}{\textit{pi on Terminal-Bench~2.0}}\\
\eTwentyFiveRow{tb}{baseline}{Strong LLM alone}
\eTwentyFiveRow{tb}{jev_router}{Routing by \jev{}}
\eTwentyFiveRow{tb}{always_cheap}{Fast LLM alone}
\midrule
\multicolumn{10}{@{}l}{\textit{SWE-agent on SWE-bench Verified}}\\
\eTwentyFiveRow{swe}{baseline}{Strong LLM alone}
\eTwentyFiveRow{swe}{jev_router}{Routing by \jev{}}
\eTwentyFiveRow{swe}{always_cheap}{Fast LLM alone}
\bottomrule
\end{tabular}}
\caption{Step routing in pi and SWE-agent (exploratory, one attempt per task, the same tasks in every arm).
$\Delta$: success difference to the strong LLM alone with a 95\,\% task-bootstrap interval. NI: non-inferior at a
margin of \res{e25/margin} points. Cost (\%) and Time (\%): change of the mean against the strong LLM alone. Time
(s): mean / median agent seconds per task. Fast: share of LLM requests sent to the fast LLM. Infrastructure failures
count as failures. For pi the table reports the second routed run, after a fix for Gemini's signed reasoning traces.}
\label{tab:e25}
\end{table}

\fi
\ifcsname jevres@e29/go/decision\endcsname
\paragraph{The decision layer as a proxy.}
\label{app:e29}
We then placed the decision layer in a proxy between an unmodified harness and its LLM provider. The proxy sent a step to the fast LLM only when the confidence fell in a range fitted on calibration tasks. In pi no range qualified (AUROC \res{e29/gate/tb/auroc}),
and in mini-SWE-agent the gate routed \res{e29/pilot/mswe/fast_share}\,\% of the requests on
\res{e29/pilot/mswe/n} pilot instances without changing success.
\fi
\ifcsname jevres@e29b/go/decision\endcsname\else\InputIfFileExists{generated/e29b_route}{}{}\fi
\ifcsname jevres@e29b/go/decision\endcsname
\label{app:e29b}%
A follow-up sent the steps that \jev{} predicted would only read to the fast LLM and discarded any fast-LLM response
that would write, run code or submit. \jev{}'s answers hardly separated the steps that only read from the others (AUROC
\res{e29b/gate/tb/auroc} in pi and \res{e29b/gate/mswe/auroc} in mini-SWE-agent). No fast-LLM action that would
write, run code or submit reached a task, and none of these pilots met the criterion for a full run, so none followed.
\fi
\subsection{Roles of a Decision Model in a Harness}
\label{app:coverage}
A decision model can take any recurring choice in an agent loop that has a short list of answers.
Table~\ref{tab:coverage} lists such roles, drawn from the hook points of Codex CLI, OpenHands and mini-SWE-agent
\citep{codexcli2026,wang2025openhands,yang2024sweagent} and from work on cascades, routing, asking, reflection, memory
and tree search \citep{chen2024frugalgpt,ong2025routellm,ross2025when2call,shinn2023reflexion,packer2023memgpt,zhou2024lats},
and how far we tested each.

\begin{table*}[!t]
\centering\scriptsize
\setlength{\tabcolsep}{4pt}
\begin{tabular}{@{}>{\raggedright\arraybackslash}p{2.7cm}>{\raggedright\arraybackslash}p{5.3cm}%
  >{\raggedright\arraybackslash}p{7.4cm}@{}}
\toprule
Role & What it decides & Our status\\
\midrule
Tool choice & the next tool, or a shortlist of tools for the LLM & live, registered\\
Tool arguments & the values of bounded or copied arguments & live, registered (with tool choice)\\
Ask or act & ask the user now, or act & live, registered (an option of tool choice); offline by state type\\
Stop and verify & accept a proposed finish, or send the agent back & live, registered (\gym{}); live, exploratory
  (mini-SWE-agent, pi); offline as a gate\\
Reflection & stop and rethink now, or go on & live, registered (\gym{}); live, exploratory (pi)\\
Memory write and read & store this observation; which notes to show the LLM & live, exploratory (\gym{});
  hooks implemented, not evaluated\\
Model and reasoning-depth routing & strong or cheap LLM, deep or shallow thinking, per step & live, registered
  (\gym{}); live, exploratory (mini-SWE-agent)\\
Action choice & the next action of an embodied agent & live, exploratory (ALFWorld)\\
Approval and safety gate & allow or veto a call the LLM proposed & live, exploratory (\gym{}, mini-SWE-agent, pi);
  irreversible calls offline only; harness hooks implemented, not evaluated\\
Context compaction & which messages to keep, drop or summarise & implemented (pi), not evaluated\\
Retry and backtrack & retry, undo or abandon a line of work & not covered\\
Sub-agent dispatch and planning & whether and which sub-agent to start; plan or re-plan & not covered\\
Parallel batching & which calls can run together & not covered\\
\bottomrule
\end{tabular}
\caption{Roles a decision model can play in a harness, and our evidence for each. \emph{Live, registered}: a
pre-registered task-level test, whatever its outcome. \emph{Live, exploratory}: task-level runs outside the registered
tests. \emph{Offline}: re-scored stored answers or trajectories. \emph{Implemented}: code and tests, evaluated at most in smoke
tests or pilots.}
\label{tab:coverage}
\end{table*}

\paragraph{Ask or act, and approval gates.}
At the \res{cov/ask/ask_now/occurrences} common states where the right move was to ask, \oursV{} chose correctly in
\res{cov/ask/rl/ask_now/acc}\%, \jev{} in \res{cov/ask/jev/ask_now/acc}\% and Intern-Decision-4B in
\res{cov/ask/intern/ask_now/acc}\%, and most errors were actions taken too early. As a stop guard on Flash-Lite,
\jev{} rejected \res{cov/gate/stop/lite_jev/first/vetoed_wrong} of the \res{cov/gate/stop/lite_jev/first/wrong} wrong first attempts to finish, but only \res{cov/gate/stop/lite_jev/episodes/caught_then_success} of those episodes then succeeded.
\paragraph{Who accepts the answer (exploratory).}
\label{app:e8}
On \res{e8/n} calibration decisions with Flash, the fitted confidence gate ranked right above wrong answers about as
well as the gap between the two best options or a second question to the decision model (AUROC
\res{e8/toolselect/g1tau/auroc} for tool choice and \res{e8/toolparams/g1tau/auroc} for arguments,
Table~\ref{tab:parts}).
A check by Flash cost \res{e8/toolselect/g4overdc} (tools) and \res{e8/toolparams/g4overdc} (arguments) times the price of the call it was meant to save, so the net saving was negative. Asking a second question below the threshold was the only check that raised the saving, and only for arguments.

\section{Post-Training \ours{}}
\label{app:rlsec}

\subsection{Readout, Labels and Recipes}
\seclabel{app:training}
\paragraph{Readout.}
The state, question and coded options (\texttt{A.}, \texttt{B.}, \ldots) form one prompt. One prefill gives the
logits $z_k$ of the option-code tokens, and $p = \mathrm{softmax}(z/T)$. A yes/no question is a choice between two options. A question with more than 52 options is split into groups of 52 that compete in a knockout tournament.

\paragraph{Data and labels.}
About half of the examples are decision points from harness runs on training seeds ($\ge$10{,}000, disjoint from all
other seeds) and half are When2Call training data, split by task and, from data v1 on, 8-gram decontaminated
(App.~\ref{app:hygiene}). Labels are \emph{gold} (the \gym{} answer key or dataset labels), \emph{teacher} soft labels
$y = \alpha\,\text{gold} + (1-\alpha)\,p_{\text{teacher}}$ \citep{hinton2015distilling}, or \emph{counterfactual Monte
Carlo} values $\hat{Q}(s,a) = (\text{wins}+1)/(n_{\text{MC}}+2)$ from $n_{\text{MC}}$ replays of a recorded state $s$
with $a$ forced.

\paragraph{Recipes.}
SFT minimises $\mathcal{L} = -\sum_k y_k \log p_k + \mu \sum_k (p_k - y_k)^2$ with a fresh option permutation per
example \citep{zheng2024pride}. GRPO runs the decider in the \gym{} harness with a fixed \systwo{} executor, $G$
rollouts per task, GiGPO anchor-state groups \citep{feng2025gigpo}, PPO clipping and an exact KL over the option set.
Its reward is
$\mathbf{1}[\text{success}]\,(1 - 0.5\min(1, n_{\text{calls}}/10)) - 0.2\,\mathbf{1}[\text{irreversible wrong call}]$,
with $n_{\text{calls}}$ the episode's LLM calls. OPD distils on policy \citep{agarwal2024gkd}. Temperatures are fitted
per question type and option-count bucket on held-out tasks \citep{guo2017calibration}. Only SFT, OPD and GRPO entered
the training chain.

\paragraph{Hyper-parameters and compute.}
\label{app:hparams}
AdamW, weight decay 0, 3\% warm-up, cosine decay, gradient clipping 1.0, bf16, effective batch 64, maximum length
4{,}096 tokens. LoRA on all linear layers ($r{=}32$, $\alpha{=}64$) for 4B, full fine-tuning for 0.6B. SFT: lr
$10^{-4}$, 2 epochs, Brier weight 0.5 (the Brier term was not ablated on its own). GRPO, as run for \oursV{}: 30 iterations $\times$ 8 tasks $\times$ $G{=}8$ on
strict \gym{} with the 60 distractor tools of the evaluation setting, 1 inner epoch, 64 decisions per step, clip
$\epsilon \in [0.2, 0.28]$ \citep{schulman2017ppo}, KL 0.02, GiGPO $\omega{=}1$, $\gamma{=}0.95$, lr $10^{-5}$,
Flash-Lite as executor. We kept the checkpoint of iteration 15. Every recipe fits one 48\,GB GPU.
The training round of \oursV{} ran on one \res{rl/gpu} for \res{spend/gpu/round5/hours} host-hours
(\$\res{spend/gpu/round5/usd}) plus \$\res{spend/api/training_round5} of API calls.

\subsection{The Training Chain}
\seclabel{app:rl}
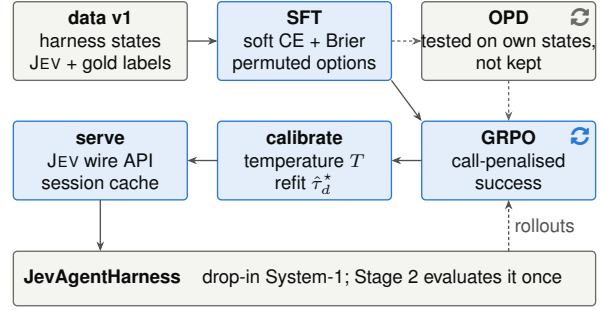
\begin{figure}[t]
\centering
\begin{tikzpicture}[jevfig]
\def\W{77.0} \def\bw{23.0} \def\bh{10.4} \def\gap{4.0}
\colorlet{envCT}{envT}
\newcommand{\stage}[8]{%
  \node[fbox=#4, fill=#4T, minimum width=\bw mm, minimum height=\bh mm, anchor=west] (#1) at (#2,#3) {};
  \node[font=\figfs\sffamily\bfseries, anchor=base] at ($(#1.center)+(0,2.2)$) {#5};
  \node[fnote, text=ink, anchor=base] at ($(#1.center)+(0,-0.8)$) {#6};
  \node[fnote, text=ink, anchor=base] at ($(#1.center)+(0,-3.7)$) {#7};
  \ifnum#8=1 \node[text=#4, font=\fontsize{7}{7}\selectfont, anchor=north east] at ($(#1.north east)+(-0.8,-0.8)$)
    {\faIcon{sync-alt}};\fi}
\pgfmathsetmacro\xa{0}
\pgfmathsetmacro\xb{\bw+\gap}
\pgfmathsetmacro\xc{2*(\bw+\gap)}
\def\ya{38.0} \def\yb{22.2}
\stage{s0}{\xa}{\ya}{envC}{data v1}{harness states}{\jev{} + gold labels}{0}
\stage{s1}{\xb}{\ya}{sOne}{SFT}{soft CE + Brier}{permuted options}{0}
\stage{s2}{\xc}{\ya}{envC}{OPD}{tested on own states,}{not kept}{1}
\stage{s3}{\xc}{\yb}{sOne}{GRPO}{call-penalised}{success}{1}
\stage{s4}{\xb}{\yb}{sOne}{calibrate}{temperature $T$}{refit $\hat\tau^\star_d$}{0}
\stage{s5}{\xa}{\yb}{sOne}{serve}{\jev{} wire API}{session cache}{0}
\draw[fbox=envC, fill=envT] (0,3.0) rectangle (\W,10.2);
\node[anchor=west, font=\figfs\sffamily\bfseries] at (1.4,6.6) {\harnesstext};
\node[fnote, text=ink, anchor=west] at (25.0,6.6) {drop-in \sysone{}; Stage 2 evaluates it once};
\draw[farr] (s0.east) -- (s1.west);
\draw[farr, dash pattern=on 1.2pt off 1pt, draw=envC] (s1.east) -- (s2.west);
\draw[farr, dash pattern=on 1.2pt off 1pt, draw=envC] (s2.south) -- (s3.north);
\draw[farr] (s1.south east) -- (s3.north west);
\draw[farr] (s3.west) -- (s4.east);
\draw[farr] (s4.west) -- (s5.east);
\draw[farr] (s5.south) -- (\xa+\bw/2,10.2);
\draw[farr, dash pattern=on 1.2pt off 0.9pt] (\xc+\bw/2,10.2) -- (s3.south);
\node[fnote, anchor=west] at (\xc+\bw/2+0.8,13.6) {rollouts};
\end{tikzpicture}
\caption{\textbf{Post-training \oursV{}: SFT followed by GRPO.} Data v1: decisions at states the
harness visits, labelled by \jev{} and gold. SFT, then GRPO on call-count-penalised success
(\faIcon{sync-alt}: rollouts in the harness, training seeds). On-policy distillation (OPD) from \jev{} at
the states the student reaches was tested and not kept by its development gate (grey). Each stage
is kept only if it passes development gates.}
\label{fig:rl}
\end{figure}
Figure~\ref{fig:rl} shows the chain, which uses development and training seeds only.

\paragraph{Why visited states.}\label{app:devdiag}%
We train on visited states because of the development-only measurements in Table~\ref{tab:devdiag-gap}.
Intern-Decision-4B is close to \jev{} on isolated items, falls behind on the states that the LLM-only agent visits, and falls further behind on the states that its own choices lead to. Our SFT model trained on recorded states (data v0) reached 0.995 accuracy on the gate set but 0.45 on the states it visited in an on-policy pilot, where \jev{} reached 0.97.

\begin{table}[!htbp]
\centering\footnotesize
\setlength{\tabcolsep}{3pt}
\begin{tabular}{@{}lcccc@{}}
\toprule
& Gate & BFCL & \multicolumn{2}{c@{}}{\gym{}}\\
Model & set & dev & shadow & \modeRep{}\\
\midrule
\jev{} & 0.971 & 0.965 & 0.933 & 0.944\\
Intern-Dec.\ 4B & 0.862 & 0.953 & 0.760 & 0.423\\
Intern-Dec.\ 2B & 0.638 & 0.906 & -- & 0.162\\
Intern-Dec.\ 0.8B & 0.571 & 0.929 & -- & 0.038\\
\bottomrule
\end{tabular}
\caption{Development diagnostic (small samples, no test): tool-selection accuracy on isolated items (210 gate-set
decisions, 85 BFCL dev items) and on a 10-task \gym{} development slice with Flash. Shadow: the states LLM-only visits.
\modeRep{}: the states the decision model reaches itself.}
\label{tab:devdiag-gap}
\end{table}

\paragraph{Data v1 and stages.}
The harness ran on training seeds with Flash-Lite as executor in three behaviour arms: \emph{shadow} (\sysone{} is
only asked), \emph{near-miss replace} (a scripted \sysone{} with injected errors reaches recovery states, as in DAgger
\citealp{ross2011dagger}) and \emph{guard} (a noisy answer key guards every tool call and the stop). The target is
$0.3\,\text{gold} + 0.7\,p_{\jev{}}$, or the gold one-hot where \jev{}'s argmax is wrong. Executed calls outside the
acceptable set become guard questions labelled \emph{no}, so that a guard does not learn to always say yes.
\emph{SFT v1} trains Qwen3-4B with LoRA on data v1 plus a 0.45$\times$ replay share, and \emph{SFT v0} is the same
recipe on recorded LLM-only states. \emph{OPD} minimises
$\mathbb{E}_{s \sim d^{\pi_\theta}}\,\mathrm{KL}\big(p_{\jev{}}(\cdot \mid s) \,\|\, p_\theta(\cdot \mid s)\big)$ over
the student's own states from SFT v1 for \res{leak/opd/iterations} iterations, and \emph{GRPO} runs for 30 iterations
from SFT v1. Each training stage was kept only if it passed registered checks on development data.

\paragraph{Chain decisions.}
The chain kept \res{rl/chain/kept}. OPD was dropped because it lowered development and recovery-state accuracy, and
GRPO passed its gates and was frozen as \oursV{} (Table~\ref{tab:recipe}). The development slice is saturated. SFT
v1 already matches \oursV{} on common-state accuracy, on task-level success (\res{stage2_rl/sft_v1/live_success}\%)
and, after temperature scaling, on ECE, so no result isolates a gain from GRPO.

\subsection{Recipe Takeaways by Stage}
\label{app:recipe}
Table~\ref{tab:paradigms} compares the ways to obtain a decision model. SFT on the agent's own states gives almost
all of the gain for about two GPU-hours, and GRPO on top added no measurable gain. Public tool-use data raises
isolated accuracy but lowers task-level success on shifted tasks.

\begin{table}[!htbp]
\centering\footnotesize
\setlength{\tabcolsep}{2.4pt}
\resizebox{\columnwidth}{!}{%
\begin{tabular}{@{}l l r r r r@{}}
\toprule
& & & \multicolumn{3}{c}{Shifted tasks (\%)}\\
\cmidrule(lr){4-6}
Option & Training cost & L1 & L2 & must-ask & L3\\
\midrule
\jev{}, off the shelf & none (API) & \res{e9/jev/acc} & \res{e21/jev/l2/all} & \res{e21/jev/l2/mustask} & \res{e21/jev/l3/success}\\
Qwen3-4B, untrained & none & \res{e9/base/acc} & \res{e21/base/l2/all} & \res{e21/base/l2/mustask} & \res{e21/base/l3/success}\\
\ifcsname jevres@e31/rlonly/l2/all\endcsname
\quad RL only (GRPO) & \res{e31/rlonly/gpu_h} GPU-h, \$\res{e31/rlonly/api_usd} & \resna{e31/rlonly/l1/acc} & \res{e31/rlonly/l2/all}
  & \res{e31/rlonly/l2/mustask} & \resna{e31/rlonly/l3/success}\\
\fi
\quad + SFT on agent states & \res{recipe/stage/sft_v1/gpu_h} GPU-h, \$\res{spend/api/training_v1_data} & \res{e9/sftv1/acc}
  & \res{e21/sft_v1/l2/all} & \res{e21/sft_v1/l2/mustask} & \res{e21/sft_v1/l3/success}\\
\quad + GRPO (\oursV{}) & +\res{recipe/stage/grpo/gpu_h} GPU-h, +\$\res{recipe/stage/grpo/api_usd} & \res{e9/rl/acc}
  & \res{e21/rl/l2/all} & \res{e21/rl/l2/mustask} & \res{e21/rl/l3/success}\\
\quad + public data (\ours{}) & +\fpeval{round(\csname jevres@e12/train/min\endcsname/60,1)} GPU-h & \res{e9/e12/acc}
  & \res{e21/e12/l2/all} & \res{e21/e12/l2/mustask} & \res{e21/e12/l3/success}\\
StartLux-4B + SFT, public data & \fpeval{round(\csname jevres@e12c/train/min\endcsname/60,1)} GPU-h & \res{e9/e12c/acc}
  & \res{e21/e12c/l2/all} & \res{e21/e12c/l2/mustask} & \res{e21/e12c/l3/success}\\
\bottomrule
\end{tabular}}
\caption{Ways to obtain a decision model: training cost (GPU-hours, and \$ for the LLM calls that collect data or run
GRPO episodes), isolated accuracy (L1, all items of Figure~\ref{fig:frontier:a}) and accuracy and success on the
shifted tasks (Table~\ref{tab:train}). GPU types differ across rows (App.~\ref{app:cost}).}
\label{tab:paradigms}
\end{table}

\begin{table*}[!t]
\centering\footnotesize
\setlength{\tabcolsep}{3pt}
\resizebox{\textwidth}{!}{%
\begin{tabular}{@{}llrrrrrrrrr@{}}
\toprule
& & & & \multicolumn{3}{c}{v1 dev} & \multicolumn{3}{c}{Common states} & \\
\cmidrule(lr){5-7}\cmidrule(lr){8-10}
Checkpoint & Data and training signal & GPU h & API \$ & acc. & recov. & ECE & acc. & off-plan & ECE & Dev\\
\midrule
\multicolumn{11}{@{}l}{\emph{Track 1: Qwen3-4B, LoRA (\oursV{})}}\\
Untrained & -- & -- & -- & \res{rl/dev/base/v1_acc} & \res{rl/dev/base/v1_recovery} & \res{rl/dev/base/v1_ece}
  & \res{stage2_rl/base/corpus_acc} & \res{stage2_rl/base/corpus_acc_offplan} & \res{stage2_rl/base/ece} & --\\
SFT v0 & \res{recipe/data/v0/rows} recorded success-only states; gold & -- & --
  & \res{rl/dev/sft_v0/v1_acc} & \res{rl/dev/sft_v0/v1_recovery} & \res{rl/dev/sft_v0/v1_ece} & -- & -- & --
  & \res{rl/dev/sft_v0/ts_replace_success}\\
SFT v1 & \res{recipe/data/v1/rows} visited states (\res{recipe/data/v1/after_error_share}\% off-plan); gold + \jev{}
  & \res{recipe/stage/sft_v1/gpu_h} & \res{spend/api/training_v1_data}
  & \res{rl/dev/sft_v1/v1_acc} & \res{rl/dev/sft_v1/v1_recovery} & \res{rl/dev/sft_v1/v1_ece}
  & \res{stage2_rl/sft_v1/corpus_acc} & \res{stage2_rl/sft_v1/corpus_acc_offplan} & \res{stage2_rl/sft_v1/ece}
  & \res{rl/dev/sft_v1/ts_replace_success}\\
\,+ OPD, it.\ 12 (dropped) & own states, forward KL to \jev{} & \res{recipe/stage/opd/gpu_h} & \res{recipe/stage/opd/api_usd}
  & \res{rl/dev/opd_i12/v1_acc} & \res{rl/dev/opd_i12/v1_recovery} & \res{rl/dev/opd_i12/v1_ece} & -- & -- & --
  & \res{rl/dev/opd_i12/ts_replace_success}\\
\,+ GRPO, it.\ 15 (\oursV{}) & episodes; call- and error-penalised success & \res{recipe/stage/grpo/gpu_h}
  & \res{recipe/stage/grpo/api_usd}
  & \res{rl/dev/grpo_i15/v1_acc} & \res{rl/dev/grpo_i15/v1_recovery} & \res{rl/dev/grpo_i15/v1_ece}
  & \res{stage2_rl/frozen/corpus_acc} & \res{stage2_rl/frozen/corpus_acc_offplan} & \res{stage2_rl/frozen/ece}
  & \res{rl/dev/grpo_i15/ts_replace_success}\\
\midrule
\multicolumn{11}{@{}l}{\emph{Track 2: StartLux-Decision-4B, full fine-tuning on Miles (exploratory)}}\\
Zero-shot & -- & -- & -- & \res{rl2/dev/zeroshot_startlux/v1_acc} & -- & \res{rl2/dev/zeroshot_startlux/v1_ece}
  & \res{stage2_s1/sl4b/corpus_acc}$^\S$ & \res{stage2_s1/sl4b/corpus_acc_offplan}$^\S$ & \res{stage2_s1/sl4b/ece}$^\S$ & --\\
SFT v2 (kept) & data v1, as SFT v1 & -- & -- & \res{rl2/dev/sft_v2/v1_acc} & \res{rl2/dev/sft_v2/v1_recovery}
  & \res{rl2/dev/sft_v2/v1_ece} & \res{stage2_v2/corpus_acc} & \res{stage2_v2/corpus_acc_offplan} & \res{stage2_v2/ece}
  & \res{rl2/stress/sft_v2/S}$^\P$\\
\,+ OPD, it.\ 12 & own states, forward KL to \jev{} & -- & -- & \res{rl2/dev/opd_i12/v1_acc} & \res{rl2/dev/opd_i12/v1_recovery}
  & \res{rl2/dev/opd_i12/v1_ece} & -- & -- & -- & \res{rl2/stress/opd_i12/S}$^\P$\\
\,+ GRPO, it.\ 15 & episodes; call- and error-penalised success & -- & -- & \res{rl2/dev/grpo_i15/v1_acc}
  & \res{rl2/dev/grpo_i15/v1_recovery} & \res{rl2/dev/grpo_i15/v1_ece} & -- & -- & -- & \res{rl2/stress/grpo_i15/S}$^\P$\\
\midrule
\jev{} (teacher) & -- & -- & -- & \res{rl/dev/jev/v1_acc} & \res{rl/dev/jev/v1_recovery} & \res{rl/dev/jev/v1_ece}
  & \res{stage2_rl/jev/corpus_acc} & \res{stage2_rl/jev/corpus_acc_offplan} & \res{stage2_rl/jev/ece}
  & \res{rl/dev/jev/ts_replace_success}\\
Intern-Decision-4B & -- & -- & -- & \res{rl/dev/intern/v1_acc} & \res{rl/dev/intern/v1_recovery} & --
  & \res{stage2_rl/intern/corpus_acc} & \res{stage2_rl/intern/corpus_acc_offplan} & \res{stage2_rl/intern/ece}
  & \res{rl/dev/intern/ts_replace_success}\\
\bottomrule
\end{tabular}}
\caption{What each training stage used and bought (\%). GPU h: host hours on one \res{rl/gpu}. API \$: building data
v1, or rollouts, teacher and development evaluations. v1 dev and common states (Stage~2, descriptive): accuracy on all
states and after an off-plan step, and ECE (raw on v1 dev). Dev: tool \modeRep{} success on the 50-task development slice.
$^\P$Success on track~2's 200-task stress slice (\jev{} \res{rl2/stress/jev/S}\%, Intern-Decision-4B
\res{rl2/stress/intern/S}\%). $^\S$Served by the vendor's server. --: not measured.}
\label{tab:recipe}
\end{table*}

In Table~\ref{tab:recipe}, adding the states after an off-plan step (\res{recipe/data/v1/after_error_share}\% of data v1) raised development accuracy on such states from
\res{rl/dev/sft_v0/v1_recovery}\% to \res{rl/dev/sft_v1/v1_recovery}\% (App.~\ref{app:e3}). OPD lowered development
accuracy in both tracks, and GRPO kept it and lowered raw ECE, but a fitted temperature of
\res{recipe/calib/sft_v1/temp_global} gave SFT v1 the same calibration.

\paragraph{Second track (exploratory).}
\label{app:rl2}
A second registered track trained StartLux-Decision-4B (non-commercial licence) with the same losses on the Miles
framework. It froze SFT v2 and makes no RL claim,
since OPD and GRPO lowered success on a harder stress slice of \res{rl2/stress/n} development tasks. In the Stage-2
combination v2 succeeded in \res{stage2_v2/profile/success_arm}\% of \res{stage2_v2/profile/n_pairs} fresh tasks
against \res{stage2_v2/profile/success_ref}\% for LLM-only, at \res{stage2_v2/profile/cost_savings}\% lower cost with
its GPU, the same picture as \oursV{}.
\subsection{Which Parts of the Recipe Matter}
\label{app:e3}
We retrained the SFT step with one ingredient changed at a time (Qwen3-4B with LoRA, one seed per arm,
Table~\ref{tab:e3}). On the common states a tenth of the training tasks stayed within
\resabs{recipe_abl/contrast/size_10_vs_anchor/c_acc_delta} points of SFT on all of them, and gold labels alone matched
the mixed labels. The model trained without off-plan states lost \resabs{recipe_abl/contrast/canon_vs_match_25/c_acc_delta} points on the common states but scored higher on its own off-plan states, so the evidence on which states to train on is
mixed.

\begin{table}[!htbp]
\centering\footnotesize
\setlength{\tabcolsep}{2.6pt}
\resizebox{\columnwidth}{!}{%
\begin{tabular}{@{}lrrrrrr@{}}
\toprule
& & & \multicolumn{2}{c}{$C$ (\%)} & \multicolumn{2}{c}{Dev.\ (\%)}\\
\cmidrule(lr){4-5}\cmidrule(lr){6-7}
Arm & Tasks & SFT min & all & off-plan & all & off-plan\\
\midrule
SFT v1 (all tasks) & -- & -- & \res{recipe_abl/sft_v1/c_acc} & \res{recipe_abl/sft_v1/c_acc_offplan}
  & \res{recipe_abl/sft_v1/v1_acc} & \res{recipe_abl/sft_v1/v1_recovery}\\
50\% of tasks & \res{recipe_abl/D50/tasks} & \res{recipe_abl/D50/sft_min} & \res{recipe_abl/D50/c_acc}
  & \res{recipe_abl/D50/c_acc_offplan} & \res{recipe_abl/D50/v1_acc} & \res{recipe_abl/D50/v1_recovery}\\
25\% of tasks & \res{recipe_abl/D25/tasks} & \res{recipe_abl/D25/sft_min} & \res{recipe_abl/D25/c_acc}
  & \res{recipe_abl/D25/c_acc_offplan} & \res{recipe_abl/D25/v1_acc} & \res{recipe_abl/D25/v1_recovery}\\
10\% of tasks & \res{recipe_abl/D10/tasks} & \res{recipe_abl/D10/sft_min} & \res{recipe_abl/D10/c_acc}
  & \res{recipe_abl/D10/c_acc_offplan} & \res{recipe_abl/D10/v1_acc} & \res{recipe_abl/D10/v1_recovery}\\
\midrule
25\%, gold labels only & \res{recipe_abl/L_gold_25/tasks} & \res{recipe_abl/L_gold_25/sft_min}
  & \res{recipe_abl/L_gold_25/c_acc} & \res{recipe_abl/L_gold_25/c_acc_offplan} & \res{recipe_abl/L_gold_25/v1_acc}
  & \res{recipe_abl/L_gold_25/v1_recovery}\\
25\%, \jev{} labels only & \res{recipe_abl/L_teach_25/tasks} & \res{recipe_abl/L_teach_25/sft_min}
  & \res{recipe_abl/L_teach_25/c_acc} & \res{recipe_abl/L_teach_25/c_acc_offplan} & \res{recipe_abl/L_teach_25/v1_acc}
  & \res{recipe_abl/L_teach_25/v1_recovery}\\
\midrule
No off-plan states & \res{recipe_abl/S_canon_25/tasks} & \res{recipe_abl/S_canon_25/sft_min}
  & \res{recipe_abl/S_canon_25/c_acc} & \res{recipe_abl/S_canon_25/c_acc_offplan} & \res{recipe_abl/S_canon_25/v1_acc}
  & \res{recipe_abl/S_canon_25/v1_recovery}\\
Random rows, same size & \res{recipe_abl/S_match_25/tasks} & \res{recipe_abl/S_match_25/sft_min}
  & \res{recipe_abl/S_match_25/c_acc} & \res{recipe_abl/S_match_25/c_acc_offplan} & \res{recipe_abl/S_match_25/v1_acc}
  & \res{recipe_abl/S_match_25/v1_recovery}\\
\midrule
\jev{} (teacher) & & & \res{recipe_abl/pub_jev/c_acc} & \res{recipe_abl/pub_jev/c_acc_offplan} & & \\
\bottomrule
\end{tabular}}
\caption{SFT ablation (exploratory, one seed per arm). $C$: tool-choice accuracy on the common states of the fresh
tasks. Dev.: held-out development decisions. Off-plan: after an off-plan step. SFT minutes on one A10G. Smaller arms
also get fewer optimizer steps.}
\label{tab:e3}
\end{table}
\ifcsname jevres@e31/rlonly/iterations\endcsname\ifcsname jevres@e31/rlonly/iterations\endcsname
\paragraph{GRPO without SFT.}
\label{app:e31}
We ran \oursV{}'s GRPO stage from the untrained Qwen3-4B, with a fresh LoRA and everything else unchanged, for
\res{e31/rlonly/iterations} iterations. The model made \res{e31/rlonly/decisions_made_share}\% of the training
decisions itself, yet it stayed at the untrained model's level on every measure (Table~\ref{tab:paradigms}), with
\res{e31/rlonly/c/acc}\% on the common states against \res{e31/ref/base/c/acc}\% before and
\res{e31/ref/sft_v1/c/acc}\% after SFT. In \res{e31/rlonly/zero_std_share}\% of its task groups all eight episodes
had the same reward and so gave no episode-level signal (\res{e31/rlonly/v1/zero_std_share}\% in \oursV{}'s first
\res{e31/rlonly/iterations} iterations).
\fi
\fi%
\subsection{A Sub-1B Decision Model with a Scoring Head}
\label{app:e14}
We gave Qwen3.5-0.8B a fresh scoring head of the Clef design~\citep{cloudflare2026clef}, which returns one logit per
allowed option for every question of a request in one pass, and trained it with LoRA and Unsloth's recipe~\citep{unsloth2026decision} on \oursV{}'s supervised data (arm A) or on Unsloth's public
mix of classification and typed-decision data (arm G)\ifcsname jevres@e14/ag/c/acc\endcsname, or on both (arm
AG)\fi. Arm A matched \oursV{} on the common states (Table~\ref{tab:e14}). Arm G matched arm A on BFCL's tool choices but scored \res{e14/a/c/vse14g/all} points lower on the common states. Arm A's confidences sat close to 1, and the registered
fit placed both thresholds within $10^{-9}$ of 1. In the final harness (App.~\ref{app:harnessfinal}) arm A solved
\res{e14/b/a/success}\% of \res{e14/b/n} tasks, as the 4B harness did, at \res{e14/b/vsllm/saving}\% lower cost than
Flash alone. Unsloth's decision-model code is AGPL-3.0, and we redistribute none of it.

\begin{table}[!htbp]
\centering\footnotesize
\setlength{\tabcolsep}{2.2pt}
\resizebox{\columnwidth}{!}{%
\begin{tabular}{@{}lrrrrrrrr@{}}
\toprule
& \multicolumn{2}{c}{Isolated (\%)} & \multicolumn{3}{c}{Common states (\%)} & Closed & \multicolumn{2}{c}{Serving}\\
\cmidrule(lr){2-3}\cmidrule(lr){4-6}\cmidrule(lr){8-9}
Model & all & tools & all & off-plan & ask & loop \% & ms & GB\\
\midrule
\oursV{} & \res{e14/ref/rl/e1acc} & \res{e14/ref/rl/e1toolchoice} & \res{e14/ref/rl/c/acc} & \res{e14/ref/rl/c/offplan}
  & \res{e14/ref/rl/c/ask} & \res{e14/ref/rl/live} & \res{e14/lat/4b/c1/p50} & \res{e14/lat/4b/mem}\\
0.8B, agent states (A) & \res{e9/e14a/acc} & \res{e14/a/e1/toolchoice} & \res{e14/a/c/acc} & \res{e14/a/c/offplan}
  & \res{e14/a/c/ask} & \res{e14/a/live/success} & \res{e14/lat/a/c1/p50} & \res{e14/lat/a/mem}\\
0.8B, generic mix (G) & \res{e9/e14g/acc} & \res{e14/g/e1/toolchoice} & \res{e14/g/c/acc} & \res{e14/g/c/offplan}
  & \res{e14/g/c/ask} & \resna{e14/g/live/success} & \res{e14/lat/g/c1/p50} & \res{e14/lat/g/mem}\\
\ifcsname jevres@e14/ag/c/acc\endcsname
0.8B, both (AG) & \res{e9/e14ag/acc} & \res{e14/ag/e1/toolchoice} & \res{e14/ag/c/acc} & \res{e14/ag/c/offplan}
  & \res{e14/ag/c/ask} & -- & -- & --\\
\fi
\bottomrule
\end{tabular}}
\caption{A 0.8B decision model with a scoring head (exploratory, one seed per arm). Isolated: all
\res{e14/ref/e1n} isolated decisions and their BFCL tool choices. Common states: all, after an off-plan step, and
must-ask. Closed loop: success when the model picks every tool on \res{e14/a/live/n} fresh \gym{} tasks. Serving:
median ms per request and GPU memory, each model alone on one A10G.}
\label{tab:e14}
\end{table}
\subsection{\gym{}-Hard}
\label{app:e7}
\gym{}-Hard keeps the frozen generator, serves \res{e7/setting/multi_customers} or
\res{e7b/setting/multi_customers} customers per task and adds a \emph{tier} rule that decides from the customer's tier
and the order amount whether each order is cancelled or shipped (Table~\ref{tab:e7}). With Stage~2's combination
\jev{} kept success at a small saving but was slower, and \oursV{}, which had never seen the tier rule, took nearly
every cancel-or-ship decision, was often wrong and lost most tasks. After refitting, the threshold left every tool choice of \oursV{} to the LLM, and \jev{}'s refitted cascade kept success at lower cost once the final answer listed each customer. One more epoch on
tier-rule states raised \oursV{}'s held-out cancel-or-ship accuracy from \res{e7d/frozen/heldout/op/acc}\% to
\res{e7d/tier/heldout/op/acc}\% and kept its common-state accuracy (\res{e7d/tier/c/acc}\% against
\res{e7d/frozen/c/acc}\%). In the agent it accepted no wrong cancel-or-ship decision, but the paired test did not
establish non-inferiority at the 3-point margin (Table~\ref{tab:mainresults-ci}). As registered, its last
\res{e7f/ext/n} tasks were added after the first \res{e7f/first/n} had been read.

\begin{table}[!htbp]
\centering\footnotesize
\setlength{\tabcolsep}{2.4pt}
\providecommand{\eSevRow}[3]{#1 & \res{#2/success} & \resm{#2/delta}\,{\scriptsize[\resm{#2/delta_ci/lo}, \resm{#2/delta_ci/hi}]}
  & \res{#2/saving} & \res{#2/op_coverage} & \res{#2/op_acc_accepted} & \res{#3/fail/wrongwrite}\\}
\resizebox{\columnwidth}{!}{%
\begin{tabular}{@{}lrrrrrr@{}}
\toprule
Arm & Succ. & $\Delta$ pt [95\% CI] & Saved \% & Took \% & Acc. \% & Wrong write\\
\midrule
\multicolumn{7}{@{}l}{\textit{\res{e7/setting/multi_customers} customers, Stage-2 combination}}\\
LLM-only & \res{e7/llm_only/success} & -- & -- & -- & -- & \res{e7/llm_only/fail/wrongwrite}\\
\eSevRow{Combination, \jev{}}{e7/jevagent_best}{e7/jevagent_best}
\eSevRow{Combination, \oursV{}}{e7/jevagent_best_rl}{e7/jevagent_best_rl}
\multicolumn{7}{@{}l}{\textit{\res{e7b/setting/multi_customers} customers, Stage-2 combination}}\\
LLM-only & \res{e7b/llm_only/success} & -- & -- & -- & -- & \res{e7b/llm_only/fail/wrongwrite}\\
\eSevRow{Combination, \jev{}}{e7b/jevagent_best}{e7b/jevagent_best}
\eSevRow{Combination, \oursV{}}{e7b/jevagent_best_rl}{e7b/jevagent_best_rl}
\multicolumn{7}{@{}l}{\textit{\res{e7c/setting/multi_customers} customers, thresholds refitted on the new rule}}\\
LLM-only & \res{e7c/llm_only/success} & -- & -- & -- & -- & \res{e7c/llm_only/fail/wrongwrite}\\
\eSevRow{Cascade, \jev{}}{e7c/jev_refit}{e7c/jev_refit}
\multicolumn{7}{@{}l}{\textit{The same, final answer given as a list per customer}}\\
LLM-only & \res{e7e/llm_only/success} & -- & -- & -- & -- & \res{e7e/llm_only/fail/wrongwrite}\\
\eSevRow{Cascade, \jev{}}{e7e/jev_refit}{e7e/jev_refit}
\multicolumn{7}{@{}l}{\textit{The same, decision model retrained on the tier rule}}\\
LLM-only & \res{e7f/rl_tier/success_ref} & -- & -- & -- & -- & \res{e7f/llm_only/fail/wrongwrite}\\
\eSevRow{Combination, retrained 4B}{e7f/rl_tier}{e7f/rl_tier}
\ifcsname jevres@e19/rl_random/success\endcsname
\eSevRow{Combination, random-rule 4B$^\ast$}{e19/rl_random}{e19/rl_random}
\fi
\ifcsname jevres@e13/luna/jev_refit/success\endcsname
\multicolumn{7}{@{}l}{\textit{List answer, thresholds refitted, agent LLM \res{e13/luna/name}}}\\
LLM-only & \res{e13/luna/llm_only/success} & -- & -- & -- & -- & \res{e13/luna/llm_only/fail/wrongwrite}\\
\eSevRow{Cascade, \jev{}}{e13/luna/jev_refit}{e13/luna/jev_refit}
\fi
\bottomrule
\end{tabular}}
\caption{\gym{}-Hard on fresh tasks (exploratory). The two Stage-2 bands and the retrained band have
\res{e7/setting/n} tasks, the bands with refitted thresholds \res{e7c/setting/n}. The agent LLM is Flash unless named,
and dollar figures compare only within a band. Took and Acc.: share of the cancel-or-ship decisions the decision
model took, and its accuracy on those. Wrong write: episodes ended by a wrong state-changing call. $^\ast$Trained on
random order rules, compared with the same LLM-only arm.}
\label{tab:e7}
\end{table}
\subsection{The Training Recipe on ALFWorld}
\label{app:e11}
We repeated the SFT step of \S\ref{sec:rl} (a fresh LoRA on Qwen3-4B) at ALFWorld's action decision point on its
training games only, with the environment's next correct action as the label, and skipped GRPO, since SFT alone matched
\oursV{} on \gym{} (App.~\ref{app:recipe}). The registered fit accepted every confidence level, so on
\res{e11/cascade/n} valid\_seen games the cascade took every action. It solved \res{e11/cascade/success}\% against
\res{e11/llm_only/success}\% for Flash alone, at \res{e11/cascade/saving}\% lower cost with the GPU charged, and was non-inferior at the 3-point margin, which we report as descriptive (Table~\ref{tab:mainresults-ci}). Valid\_seen shares its
rooms with the training games, and the model comes from a single training run.
\ifcsname jevres@e23/cascade/success\endcsname
\paragraph{Unseen rooms.}
\label{app:e23}
On ALFWorld's valid\_unseen split (\res{e23/n_games} games in \res{e23/n_scenes} rooms that occur in no training or
calibration game), the unchanged cascade took \res{e23/cascade/coverage}\% of the actions and solved
\res{e23/cascade/success}\% of the games against \res{e23/llm_only/success}\% for Flash alone, at
\res{e23/cascade/saving}\% lower cost, and non-inferiority was not shown ($p = \res{e23/cascade/ni/p}$). Its actions
were right about as often as in seen rooms, but it failed some of the simplest games that Flash alone always solved.
\fi
\ifcsname jevres@e24/e11/pooled/acceptable\endcsname\ifcsname jevres@e24/e11/pooled/acceptable\endcsname
\subsection{Agent-State Accuracy on Public Benchmarks}
\label{app:e24}
We replayed every recorded trajectory of Flash alone on ALFWorld's valid\_seen and valid\_unseen games
(App.~\ref{app:e11}), which reproduced all \res{e24/alf/n/states} states of \res{e24/alf/n/games} games, and
every model answered the same states offline (Table~\ref{tab:e24}).

\begin{table}[h]
\centering\scriptsize
\setlength{\tabcolsep}{3pt}
\begin{tabular}{@{}l r r r r r@{}}
\toprule
& \multicolumn{2}{c}{L2 ALFWorld} & Expert & L2 & Live alone\\
\cmidrule(lr){2-3}
Model & seen & unseen & action & $\tau^2$ & seen / unseen\\
\midrule
\multicolumn{6}{@{}l}{\textit{Trained on ALFWorld's training games}}\\
Qwen3-4B + SFT & \res{e24/e11/seen/acceptable} & \res{e24/e11/unseen/acceptable} & \res{e24/e11/pooled/strict}
  & \resna{e24/e11/tau2/tool/pooled} & \res{e24/live/e11/seen/replace} / \res{e24/live/e11/unseen/replace}\\
\ifcsname jevres@e24/e22/tau2/tool/pooled\endcsname
\multicolumn{6}{@{}l}{\textit{Trained on $\tau^2$-bench's training tasks}}\\
0.8B scoring head + SFT & -- & -- & -- & \res{e24/e22/tau2/tool/pooled} & --\\
\fi
\multicolumn{6}{@{}l}{\textit{Trained on \gym{} (transfer)}}\\
\ours{} & \res{e24/e12/seen/acceptable} & \res{e24/e12/unseen/acceptable} & \res{e24/e12/pooled/strict}
  & \resna{e24/e12/tau2/tool/pooled} & --\\
0.8B scoring head & \res{e24/e14a/seen/acceptable} & \res{e24/e14a/unseen/acceptable} & \res{e24/e14a/pooled/strict}
  & \resna{e24/e14a/tau2/tool/pooled} & --\\
\multicolumn{6}{@{}l}{\textit{Off the shelf}}\\
Qwen3-4B & \res{e24/qwen4b/seen/acceptable} & \res{e24/qwen4b/unseen/acceptable} & \res{e24/qwen4b/pooled/strict}
  & \resna{e24/qwen4b/tau2/tool/pooled} & --\\
StartLux-4B & \res{e24/sl4b/seen/acceptable} & \res{e24/sl4b/unseen/acceptable} & \res{e24/sl4b/pooled/strict}
  & \resna{e24/sl4b/tau2/tool/pooled} & --\\
\ifcsname jevres@e24/sl9b/pooled/acceptable\endcsname
StartLux-9B & \res{e24/sl9b/seen/acceptable} & \res{e24/sl9b/unseen/acceptable} & \res{e24/sl9b/pooled/strict}
  & \resna{e24/sl9b/tau2/tool/pooled} & --\\
\fi
\ifcsname jevres@e24/sl27b/pooled/acceptable\endcsname
StartLux-27B & \res{e24/sl27b/seen/acceptable} & \res{e24/sl27b/unseen/acceptable} & \res{e24/sl27b/pooled/strict}
  & \resna{e24/sl27b/tau2/tool/pooled} & --\\
\fi
\jev{} & \res{e24/jev/seen/acceptable} & \res{e24/jev/unseen/acceptable} & \res{e24/jev/pooled/strict}
  & \resna{e24/jev/tau2/tool/pooled} & \res{e24/live/jev/seen/replace} / --\\
OpenAI Decisions & \res{e24/oaidec/seen/acceptable} & \res{e24/oaidec/unseen/acceptable} & \res{e24/oaidec/pooled/strict}
  & \resna{e24/oaidec/tau2/tool/pooled} & --\\
\midrule
Flash's own action & \res{e24/flash/seen/acceptable} & \res{e24/flash/unseen/acceptable} & \res{e24/flash/pooled/strict}
  & \resna{e24/flash/tau2/tool/pooled} & \res{e24/live/flash/seen} / \res{e24/live/flash/unseen}\\
\bottomrule
\end{tabular}
\caption{Agent-state accuracy (\%) on the states of Flash alone (game means). L2 ALFWorld: correct when
ALFWorld's expert takes the action or it shortens the expert's path. Expert action: agreement with the expert's single
next action (a lower bound). L2 $\tau^2$: share of states where the model chooses the reference write action that is
due (task means). Live alone: success when the model takes every action (for Flash, the agent alone). --: not run.}
\label{tab:e24}
\end{table}

The model trained on ALFWorld's training games led, also in unseen rooms, though it learned from the expert that
defines the labels. The models trained on \gym{} did not transfer to ALFWorld (\ours{} was right at \res{e24/e12/pooled/acceptable}\% of states and its untrained base at \res{e24/qwen4b/pooled/acceptable}\%).
\ifcsname jevres@e24/e12/tau2/tool/pooled\endcsname
On $\tau^2$-bench the ranking did not follow ALFWorld's (Kendall rank correlation \resm{e24/kendall/alf_tau2} over
\res{e24/kendall/alf_tau2/n} models).
\fi
\fi
\fi%
\ifcsname jevres@e22/pooled/e22_final/success\endcsname\paragraph{The training recipe on \texorpdfstring{$\tau^2$}{tau2}-bench.}
\label{app:e22}
We trained the 0.8B scoring head of App.~\ref{app:e14} further on $\tau^2$-bench's retail and airline training
tasks, on states visited by Flash alone and labelled with Flash's choice when the episode succeeded or when that call
matched a pending action of the reference solution. On the test tasks the harness lost success against Flash alone at
its fitted thresholds, so non-inferiority was not shown, and at a threshold of 0.5 it saved more and lost less
(Table~\ref{tab:e22}). The result rests on one training run.

\begin{table}[!htbp]
\centering\footnotesize
\setlength{\tabcolsep}{3pt}
\resizebox{\columnwidth}{!}{%
\begin{tabular}{@{}l r r r r r r r r@{}}
\toprule
& & \multicolumn{2}{c}{Success (\%)} & & & & & \\
\cmidrule(lr){3-4}
Test tasks & $n$ & LLM & with & $\Delta$ pt [95\,\%] & NI $p$ & LLM-free (\%) & saved (\%) & time (s)\\
\midrule
\multicolumn{9}{@{}l}{\textit{Fitted thresholds}}\\
Both domains & \res{e22/pooled/e22_final/n} & \res{e22/pooled/llm_only/success} & \res{e22/pooled/e22_final/success}
  & \resm{e22/pooled/e22_final/delta} [\resm{e22/pooled/e22_final/lo}, \resm{e22/pooled/e22_final/hi}]
  & \res{e22/pooled/e22_final/p} & \res{e22/pooled/e22_final/llmfree} & \res{e22/pooled/e22_final/saving}
  & \res{e22/pooled/llm_only/wall_mean} / \res{e22/pooled/e22_final/wall_mean}\\
Retail & \res{e22/retail/e22_final/n} & \res{e22/retail/llm_only/success} & \res{e22/retail/e22_final/success}
  & \resm{e22/retail/e22_final/delta} [\resm{e22/retail/e22_final/lo}, \resm{e22/retail/e22_final/hi}]
  & \res{e22/retail/e22_final/p} & \res{e22/retail/e22_final/llmfree} & \res{e22/retail/e22_final/saving}
  & \res{e22/retail/llm_only/wall_mean} / \res{e22/retail/e22_final/wall_mean}\\
Airline & \res{e22/airline/e22_final/n} & \res{e22/airline/llm_only/success} & \res{e22/airline/e22_final/success}
  & \resm{e22/airline/e22_final/delta} [\resm{e22/airline/e22_final/lo}, \resm{e22/airline/e22_final/hi}]
  & \res{e22/airline/e22_final/p} & \res{e22/airline/e22_final/llmfree} & \res{e22/airline/e22_final/saving}
  & \res{e22/airline/llm_only/wall_mean} / \res{e22/airline/e22_final/wall_mean}\\
\multicolumn{9}{@{}l}{\textit{Threshold 0.5 at both decision points}}\\
Both domains & \res{e22/pooled/e22_tau05/n} & \res{e22/pooled/llm_only/success} & \res{e22/pooled/e22_tau05/success}
  & \resm{e22/pooled/e22_tau05/delta} [\resm{e22/pooled/e22_tau05/lo}, \resm{e22/pooled/e22_tau05/hi}]
  & \res{e22/pooled/e22_tau05/p} & \res{e22/pooled/e22_tau05/llmfree} & \res{e22/pooled/e22_tau05/saving}
  & \res{e22/pooled/llm_only/wall_mean} / \res{e22/pooled/e22_tau05/wall_mean}\\
\bottomrule
\end{tabular}}
\caption{The recipe on $\tau^2$-bench's test tasks (exploratory). $n$ test tasks, each run \res{e22/k} times per
arm, success averaged per task. $\Delta$: paired, with a task-bootstrap interval. NI $p$: certified test at the
5-point margin on each task's majority outcome. LLM-free: agent turns the decision model completed without the LLM.
Saved: cost saving with the GPU charged. Time: s per episode (LLM alone / with the model). Retail's natural-language
checks were judged by Gemini 3.5 Flash instead of $\tau^2$-bench's gpt-4.1.}
\label{tab:e22}
\end{table}
\fi%
\subsection{More Training on Public Tool-Use Data}
\label{app:e12}
We continued \oursV{}'s adapter for an epoch on \res{e12/data/new} public items from When2Call's training split and
ToolACE, mixed one to one with \res{e12/data/replay} of its own SFT states, after a leakage audit against every test
pool. The resulting checkpoint, \ours{}, raised isolated accuracy from \res{e12/frozen/e1/acc}\% to
\res{e12/e12/e1/acc}\%, was non-inferior to \oursV{} on the common states at the registered 1-point margin
(\res{e12/e12/c/acc}\% against \res{e12/frozen/c/acc}\%) and solved \res{e12/e12/live/success}\% of
\res{e12/e12/live/n} fresh tasks. The same recipe and data on StartLux-Decision-4B (CC BY-NC 4.0 weights) reached
\res{e12c/e12c/e1/acc}\% on the isolated questions, \resm{e12c/e1/vs/sl4b/all} points above its base, was non-inferior
to \oursV{} on the common states (\res{e12c/e12c/c/acc}\%) and solved \res{e12c/e12c/live/success}\% of the same tasks.

\paragraph{Reading a rule.}
Trained instead on \gym{}-Hard tasks with an order rule drawn per task (never the evaluated one), the random-rule
checkpoint was right at \res{e12b/e12b/heldout/op/acc}\% of held-out cancel-or-ship states of the unseen tier rule,
against \res{e12b/frozen/heldout/op/acc}\% for \oursV{} and \res{e12b/e7dtier/heldout/op/acc}\% for the model trained
on that rule itself (App.~\ref{app:e7}).

\subsection{Train/Evaluation Hygiene}
\label{app:hygiene}
An offline audit checked both training tracks against every evaluation set. The training seeds lie between
\res{leak/seeds/train/min} and \res{leak/seeds/train/max} and share \res{leak/seeds/overlap} seeds with the
\res{leak/seeds/eval/n} evaluation seeds, and no evaluation task repeats a training task's world or gold plan.
\res{leak/state/c/exact/rate}\% of the common-state occurrences repeat a training state word for word (early steps
whose state is little more than the instruction), and \oursV{} scores \res{leak/rescore/rl/without_exact}\% without
them and \res{leak/rescore/rl/all}\% with them. The $\tau^2$ questions have no near duplicate among the training rows.
Data v1's When2Call replay shares no 8-gram with a test question. Data v0's shares one in
\res{leak/static/v0-replay-when2call/ngram8} rows, but it trained only SFT v0, which was never scored on L1. \oursV{} was
chosen on development inputs only, and every \gym{} threshold was fitted on calibration seeds. We disclose one
exposure. Before access to Stage-1 results was blocked, the person training v1 printed the first record of one Stage-1 trajectory file to learn the trajectory format. No aggregate was opened and no pre-fixed rule changed, but a strict reader may treat the two
Stage-2 tests of \oursV{} (App.~\ref{app:stage2}) as exploratory. The code checkout of the exploratory second track contained the selection summary of the registered $\tau^2$ transfer test. Intern-Decision and StartLux-Decision were released
before any evaluation task list was frozen, but their training data are closed, so overlap with the public L1 items
cannot be excluded.

\ifcsname jevres@tpl/c/all/lookup/acc\endcsname\paragraph{Template baselines.}
\label{app:tplbase}
\gym{} builds every task from a few request templates, so the right tool may follow from the template alone. We fitted
lookups on the \res{tpl/train/rows} tool-choice rows of the supervised training data (Table~\ref{tab:tplbase}). One
tool per key would get almost every common state right, at the must-ask states too, so on most common states the
trained models' lead over off-the-shelf models reflects the generator's conventions. The trained models pull ahead on
the states whose key never occurs in training, most of which follow an off-plan step.
\ifcsname jevres@e21/rl/l2/renamed\endcsname On the shifted split of App.~\ref{app:e21}, a lookup by tool name scores
\res{tpl/shift/renamed/lookup/acc}\% on the states whose right tool was renamed, while \oursV{} scores
\res{e21/rl/l2/renamed}\%.\fi

\newcommand{\tplrow}[2]{#1 & \resna{tpl/c/all/#2/acc} & \resna{tpl/c/must_ask/#2/acc}
  & \resna{tpl/c/after_off_plan/#2/acc} & \resna{tpl/c/cov/full/covered/#2/acc} & \resna{tpl/c/cov/full/uncovered/#2/acc}\\}
\begin{table}[!t]
\centering\footnotesize
\setlength{\tabcolsep}{3pt}
\resizebox{\columnwidth}{!}{%
\begin{tabular}{@{}lccccc@{}}
\toprule
& All & Must- & After & Seen & Unseen\\
& states & ask & off-plan & key & key\\
\midrule
States $n$ & \res{tpl/c/all/n} & \res{tpl/c/must_ask/n} & \res{tpl/c/after_off_plan/n} & \res{tpl/c/cov/full/covered/n}
  & \res{tpl/c/cov/full/uncovered/n}\\
\midrule
\tplrow{(a) Prior}{prior}
\tplrow{(c) Last tool}{lookup_last}
\tplrow{(b) Tool sequence}{lookup}
\tplrow{(b+) Back-off}{lookup_backoff}
\tplrow{Best per key$^a$}{key_ceiling}
\midrule
\tplrow{\jev{}}{jev}
\tplrow{StartLux-27B}{sl27b}
\tplrow{Qwen3-4B base}{base}
\midrule
\tplrow{SFT v1$^\ast$}{sft_v1}
\tplrow{\oursV{}$^\ast$}{rl}
\tplrow{0.8B head$^\ast$}{e14a}
\bottomrule
\end{tabular}}
\caption{Template baselines and decision models on the common states, as task-weighted accuracy (\%), post hoc and
descriptive. The baselines are fitted on the training rows only: (a) the most frequent label, (b) the most frequent
label for the key (task family and tools executed so far), (c) the same keyed on the last executed tool, (b+) (b), then
(c), then (a). \emph{Seen key}: the key occurs in the training rows. $^a$A bound that picks one tool per key on the
common states themselves. $^\ast$Trained on these rows. 0.8B head: App.~\ref{app:e14}.}
\label{tab:tplbase}
\end{table}
\fi%

\section{Protocol and Statistics}
\label{app:protocol}
\paragraph{Metrics.}\seclabel{app:metrics}%
Every success number comes with dollars, latency (the critical path of model calls) and LLM calls, as in cost-aware
evaluation of agents, routers, calibration and selective prediction
\citep{kapoor2025agentsthatmatter,yao2025taubench,ong2025routellm,naeini2015bbq,geifman2019biasreduced}.
Utility is $U = W S - C - \lambda L$ for success $S$, cost $C$, latency $L$, task value $W$ and latency price
$\lambda$, and ECE is the 10-bin top-label calibration error (other metrics in \texttt{docs/\allowbreak
BENCHMARK\_SPEC.md}). Paired comparisons resample tasks and keep the trials of one task together
\citep{efron1993bootstrap,koehn2004significance}. A task that still errors in either arm after two retries leaves
that comparison.

\paragraph{Non-inferiority decision rule.}
For margin $m$ on a paired binary endpoint, $W$ and $L$ count the tasks the arm wins and loses. The outcomes
$(W, L)$ are ordered by the monotone hull of the log $p$-value of Tango's one-sided score test of
$H_0{:}\ \Delta \le -m$ \citep{tango1998equivalence}. The decisive $p$ is $\min\{1, \gamma + U\}$, where $U$ bounds
the tail probability under every boundary null whose discordance lies in a $(1-\gamma)$ Clopper--Pearson set
($\gamma = 10^{-4}$, \citealp{suissa1985exact,berger1994pvalue}). The test is valid at every $n$, with exact size at
most $0.997\alpha$ at the registered sizes. At zero discordance we report the exact upper bound $1 - \alpha^{1/n}$ on
the loss rate.

\paragraph{Other tests.}
The tool cascade, the three-tool shortlist and the stop veto must hold for both \jev{} and Intern-Decision, so their
decisive $p$ is $\max(p_{\jev{}}, p_{\text{Intern}})$ (Table~\ref{tab:l2-families}). Superiority primaries use the
exact conditional test on discordant pairs and, for the reflection trigger, the one-sided paired $t$-test of the
per-task utility difference. Sign-flip tests and the paired task bootstrap ($B = 10{,}000$) are descriptive. On
$\tau^2$ the per-task endpoint is majority pass over three attempts, and the transfer test refits $\hat\tau^\star$ on
15 held-out airline tasks, counting a decision as right if it matches the next reference action. Pilots on retired
seeds fixed the suite settings, and no pilot result is reported as a finding. The Flash-Lite pairs and the $\tau^2$
transfer test are underpowered and can end inconclusive (exact power in \texttt{results/\allowbreak reviews/\allowbreak 2026-10-02/\allowbreak ni\_power\_exact\_r8.md}).

\paragraph{Registration and claims.}\seclabel{app:stage2}%
The registration, its dated deviations and the wording of each test are in the supplementary repository
(\texttt{docs/PREREGISTRATION.md}, \texttt{docs/PREREG\_A7\_STAGE2.md} and their generated effective versions).
Stage~2 was registered before any Stage-1 output was opened, and Table~\ref{tab:tests} lists every registered test
with its result. Holm's method keeps the family-wise error of the eleven Stage-1 tests at 5\%
\citep{holm1979simple}, and the four Stage-2 tests form a second family. A registered claim holds when all of its conditions pass. Every other number is descriptive. Experiments added after Stage~1 were each written down
before they ran but lie outside both families, and a dagger (\dg{}) marks them where a display mixes both kinds
(paired intervals in Table~\ref{tab:mainresults-ci}).

\paragraph{Selection and arms.}
A script chose the combination once from the Stage-1 results, per decision-point group and \systwo{} model, among the
registered \jev{} arms at their registered settings. A candidate qualified when it delegated, was non-inferior (NI
$p \le 0.05$ at 3 points) and was cheaper (one-sided paired $t$-test, $p \le 0.05$). Among qualifiers within 2 points
of the largest net saving, the one that leaves most to the LLM won, and a group with no qualifier stayed with the LLM.
On 500 new \gym{} seeds one scheduler runs every arm on every task in random order: LLM-only, the combination with
\jev{}, its \emph{best single} delegation, the combination with \oursV{} (thresholds refitted on its own calibration
arms) and the development combination. No Flash-Lite delegation qualified. A Stage-2 test rejects when all its
conditions reject (Table~\ref{tab:stage2-clauses}), and the six secondary rows are Holm-adjusted within their table. The common states
of the accuracy test are the tool-selection questions four frozen policies met on the first 200 new seeds.

\paragraph{Training behind a barrier and a third \sysone{} family.}
A manifest fixed the training inputs, recipes, budgets, stage order and checkpoint-selection rules before evaluation
outputs reached model development (App.~\ref{app:hygiene}). An amendment registered before any Stage-1 result was
read adds StartLux-Decision (open weights, 0.8B--27B), served without tuning on our data. Its 4B runs the combination
as a secondary test.

\section{Models, Cost and Licences}
\label{app:modelscost}\seclabel{app:cost}
\paragraph{Models.}
The agent LLMs are Gemini~3.5 Flash (\texttt{gemini-3.5-flash}) and the cheaper, weaker Flash-Lite
(\texttt{gemini-3.5-flash-lite}), both with minimal thinking, temperature~0 and native tool calls. Gemma~4
(\texttt{gemma-4-26b-a4b-it}) is an L1 baseline. All were accessed September--October 2026 (\S\ref{sec:setup}). \jev{} 1.13 \citep{typesafe2026jev}, a closed typed decision API
released on 15 September 2026, is pinned as \texttt{typesafe/jev-1.13} (\S\ref{sec:background}).

\paragraph{Cost and latency.}
Episode cost is
\begin{align*}
&\textstyle\sum_{\text{LLM calls}} (n_{\text{in}}\$_{\text{in}} + n_{\text{cached}}\$_{\text{cache}} + n_{\text{out}}\$_{\text{out}})\\
&\textstyle{}+ \sum_{\text{\sysone{} calls}} n_{\text{in}}\$^{\text{S1}}_{\text{in}},
\end{align*}
with list prices on the run dates applied to measured tokens, and savings compare with a reference run on the same
tasks. \jev{} is billed through OpenRouter at \$0.042/M input tokens and nothing for output. In Stage~1,
Intern-Decision carries a GPU price amortised over the model card's unbatched throughput. In Stage~2, \oursV{} and
StartLux-Decision are charged a per-episode GPU price $q$, the host's hourly price times a development run's
wall-clock window divided by its completed episodes. For \oursV{}, $q = \$\res{stage2/S2-4/tariff/q}$ per episode
(\$\res{spend/s24_host/usd_per_h} per host-hour), and the combination with \oursV{} saves
\res{stage2/S2-4/cost_savings}\% in its Stage-2 test. The saving is
\res{stage2/S2-4/tariff/busy_throughput/cost_savings}\% at the busy-throughput price,
\res{stage2/S2-4/tariff/jev_api_price/cost_savings}\% at \jev{}'s API price and
\res{stage2/S2-4/tariff/zero/cost_savings}\% at a zero GPU price. Training and calibrating \oursV{} cost
\$\res{spend/amortisation/training_usd}, which this saving pays back after \res{spend/amortisation/episodes} episodes. Critical-path latency sums sequential calls and counts concurrent calls once, at their maximum.
The study's API spend from the run ledgers was \$\res{spend/api/total} (by track in \texttt{results/\allowbreak
spend/\allowbreak spend\_ledger.md}).

\paragraph{Licences and release.}\seclabel{app:licences}%
Table~\ref{tab:licences} lists the licences or terms of the artefacts we use. We release the benchmark code and
configurations, \harnessname{} with its adapters, the post-training recipes, seeds and evaluation code. Weights and
data derived from \jev{} or Gemini outputs are released only where the providers' terms permit. The textless icons in
Figures~\ref{fig:teaser}--\ref{fig:wherehow} were generated with OpenAI's gpt-image-2.5 from our prompts, except the
Microsoft Fluent UI Emoji in Figure~\ref{fig:teaser}. Every label and number in the figures is typeset text.

\begin{table}[!htbp]
\centering\footnotesize
\setlength{\tabcolsep}{3pt}
\begin{tabular}{@{}p{0.48\columnwidth}p{0.46\columnwidth}@{}}
\toprule
Artefact & Licence or terms\\
\midrule
When2Call & CC BY 4.0 (attribution)\\
Intern-Decision 0.8B / 2B / 4B & Apache-2.0 (Qwen licence preserved)\\
StartLux-Decision (weights; server code) & CC BY-NC 4.0; Apache-2.0\\
pi, DeepSeek Harness, mini-SWE-agent, $\tau^2$-bench, ALFWorld & MIT\\
BFCL v4 (Gorilla), Harbor, Codex CLI (hook schemas) & Apache-2.0\\
Qwen3, SWE-bench Verified, Terminal-Bench 2.0 / -Science & as stated at the source\\
LoCoMo; LongMemEval (not reported) & CC BY-NC 4.0; MIT\\
Microsoft Fluent UI Emoji (icons in Figure~\ref{fig:teaser}) & MIT\\
\jev{} via OpenRouter; Gemini API; OpenAI API (GPT-6.1-Sol row; figure icons) & provider terms\\
\bottomrule
\end{tabular}
\caption{Licences and terms of the artefacts used.}
\label{tab:licences}
\end{table}

\section{Result Tables}
\label{app:results}

\subsection{All Registered Tests}
\label{app:tests}
Table~\ref{tab:tests} lists every registered test with its outcome, and Table~\ref{tab:l2-families} splits the
primaries that must hold for both decision-model families.

\newcommand{\tsteff}[2]{\resm{#1/#2}\,{\scriptsize[\resm{#1/#2_ci/lo}, \resm{#1/#2_ci/hi}]}}
\newcommand{\tstout}[1]{\parbox[t]{4.2cm}{\raggedright\resplain{#1}\strut}}
\newcommand{\tstwhat}[1]{\parbox[t]{7.6cm}{\raggedright #1\strut}}
\newcommand{\tst}[6]{{\scriptsize #1} & \tstwhat{#2} & \tsteff{#3}{#4} & #5 & #6 & \res{#3/p_holm} & \tstout{#3/decision}\\}
\newcommand{\sv}[1]{\resm{#1}}
\newcommand{\tstna}[3]{{\scriptsize #1} & \tstwhat{#2} & -- & -- & -- & -- & \tstout{#3/decision}\\}
\newcommand{\tstgroup}[2]{\multicolumn{7}{@{}l}{\rule{0pt}{2.6ex}\textbf{#1} #2}\\}
\begin{table*}[!t]
\centering\footnotesize
\setlength{\tabcolsep}{2.6pt}
\resizebox{\textwidth}{!}{%
\begin{tabular}{@{}llrcccl@{}}
\toprule
Test & What is tested
  & \begin{tabular}[b]{@{}r@{}}Success difference,\\points [95\% CI]\end{tabular}
  & \begin{tabular}[b]{@{}c@{}}Cost\\saved (\%)\end{tabular}
  & \begin{tabular}[b]{@{}c@{}}Decided by\\the model (\%)\end{tabular}
  & \begin{tabular}[b]{@{}c@{}}Adjusted\\$p$\end{tabular} & Outcome\\
\midrule
\tstgroup{Stage 1: is each use of the decision model safe and useful?}{(11 tests; non-inferior to LLM-only on \gym{} with Flash unless stated)}
\tst{C1F}{Model picks tool and arguments if confident (\modeCas{})}{primaries/C1F}{jev/delta}{\sv{primaries/C1F/jev/readouts/cost_savings}}{\res{primaries/C1F/jev/tool_coverage}}
\tst{C1L}{the same, with the cheaper Flash-Lite}{primaries/C1L}{jev/delta}{\sv{primaries/C1L/jev/readouts/cost_savings}}{\res{primaries/C1L/jev/tool_coverage}}
\tst{C2F}{Model shortlists three tools for the LLM (\modeAss{})}{primaries/C2F}{jev/delta}{\sv{primaries/C2F/jev/readouts/cost_savings}}{--}
\tst{C2L}{the same, with Flash-Lite}{primaries/C2L}{jev/delta}{\sv{primaries/C2L/jev/readouts/cost_savings}}{--}
\tst{C3}{\jev{}'s shortlist is better than a BM25 keyword shortlist}{primaries/C3}{jev/delta}{--}{--}
\tst{C4}{Model decides when Flash-Lite hands a step to Flash (\modeTri{}); better than Flash-Lite alone}{primaries/C4}{jev/delta}{\sv{primaries/C4/jev/readouts/cost_savings}}{--}
{\scriptsize C5} & \tstwhat{Model decides when to reflect (\modeTri{}, Flash-Lite); better than an equally frequent random trigger}
  & \begin{tabular}[t]{@{}r@{}}\resm{primaries/C5/jev/du_gross}\\{\scriptsize[\resm{primaries/C5/jev/du_gross_ci/lo}, $+$\res{primaries/C5/jev/du_gross_ci/hi}]}\end{tabular} & -- & --
  & \res{primaries/C5/p_holm} & \tstout{primaries/C5/decision}\\
\tst{C6F}{Model can veto the LLM's decision to stop (\modeGua{})}{primaries/C6F}{jev/delta}{\sv{primaries/C6F/jev/readouts/cost_savings}}{--}
\tst{C6L}{the same, with Flash-Lite}{primaries/C6L}{jev/delta}{\sv{primaries/C6L/jev/readouts/cost_savings}}{--}
\tst{C7}{Cheapest non-inferior \gym{} setup, refitted, on $\tau^2$}{primaries/C7}{jev/delta}{\sv{primaries/C7/jev/readouts/cost_savings}}{--}
\tst{C8}{\jev{} against Flash on isolated BFCL tool-choice questions}{primaries/C8}{jev/delta}{--}{--}
\tstgroup{Stage 2: does the chosen combination hold on fresh tasks?}{(4 tests: three combinations non-inferior to their reference and cheaper on 500 new \gym{} tasks, and accuracy on the same recorded states)}
\tst{S2-1}{Combination with \jev{}}{stage2/S2-1}{delta}{\sv{stage2/S2-1/cost_savings}}{\res{stage2/S2-1/coverage/tool_coverage}}
\tst{S2-2}{Combination against its best single part, the tool cascade}{stage2/S2-2}{delta}{\sv{stage2/S2-2/cost_savings}}{\res{stage2/S2-2/coverage/tool_coverage}}
\tst{S2-3}{On the same states, \oursV{} beats Intern-Decision and is at most 5 points below \jev{}}{stage2/S2-3}{delta}{--}{--}
\tst{S2-4}{Combination with \oursV{} as decision model}{stage2/S2-4}{delta}{\sv{stage2/S2-4/cost_savings}}{\res{stage2/S2-4/coverage/tool_coverage}}
\tstgroup{Stage 2, secondary tests}{(6 tests; as above unless stated)}
\tst{S2-1T}{Combination with \jev{} is faster (seconds saved per episode)}{stage2/S2-1T}{delta}{--}{\res{stage2/S2-1/coverage/tool_coverage}}
\tstna{S2-1L}{Combination with Flash-Lite}{stage2/S2-1L}
\tst{S2-L3a}{Combination on $\tau^2$ telecom (5-point margin)}{stage2/S2-L3a}{delta}{\sv{stage2/S2-L3a/cost_savings}}{\res{stage2/S2-L3a/coverage/tool_coverage}}
\tstna{S2-L3b}{Combination on ALFWorld, against a ReAct agent}{stage2/S2-L3b}
\tst{S2-1X}{Combination with DeepSeek-V4-Pro as the agent LLM}{stage2/S2-1X}{delta}{\sv{stage2/S2-1X/cost_savings}}{\res{stage2/S2-1X/coverage/tool_coverage}}
\tst{S2-4S}{Combination with StartLux-4B as decision model}{stage2/S2-4S}{delta}{\sv{stage2/S2-4S/cost_savings}}{\res{stage2/S2-4S/coverage/tool_coverage}}
\bottomrule
\end{tabular}}
\caption{Every registered test. Success difference: setup minus reference in points (reflection trigger: utility in
\$ per task, \oursV{} on the same states: accuracy against Intern-Decision, speed: seconds saved per episode).
Adjusted $p$: Holm-corrected within the group, and a claim holds at $p \le 0.05$. Non-inferior: at most 3 points
(5 on $\tau^2$) below the reference. For the cascade with Flash, an exact bound puts the share of tasks \jev{}'s cascade turned into failures below \res{primaries/C1F/jev/loss_rate_ub}\%.}
\label{tab:tests}\label{tab:primaries}\label{tab:prereg}\label{tab:stage2}
\end{table*}

\newcommand{\lfamc}[2]{\resm{primaries/#1/#2/delta} & {\scriptsize[\resm{primaries/#1/#2/delta_ci/lo}, \resm{primaries/#1/#2/delta_ci/hi}]}
  & \res{primaries/#1/#2/p_decisive}}
\newcommand{\lfam}[4]{#1 & \lfamc{#2}{jev} & #3 & \resm{primaries/#2/jev/readouts/cost_savings}
  & \lfamc{#2}{intern} & #4 & \resm{primaries/#2/intern/readouts/cost_savings}
  & \res{primaries/#2/p_decisive} & \resdec{primaries/#2/decision}\\}
\begin{table*}[!t]
\centering\footnotesize
\setlength{\tabcolsep}{3.0pt}
\resizebox{\textwidth}{!}{%
\begin{tabular}{@{}l ccccc ccccc cl@{}}
\toprule
& \multicolumn{5}{c}{\jev{}} & \multicolumn{5}{c}{Intern-Decision-4B} & & \\
\cmidrule(lr){2-6}\cmidrule(lr){7-11}
Test & $\Delta$ & 95\% CI & $p$ & Cov. & Sav. & $\Delta$ & 95\% CI & $p$ & Cov. & Sav. & $p_{\max}$ & Registered label\\
\midrule
\lfam{Cascade, Flash}{C1F}{\res{primaries/C1F/jev/tool_coverage}}{\res{primaries/C1F/intern/tool_coverage}}
\lfam{Cascade, Flash-Lite}{C1L}{\res{primaries/C1L/jev/tool_coverage}}{\res{primaries/C1L/intern/tool_coverage}}
\lfam{Shortlist, Flash}{C2F}{--}{--}
\lfam{Shortlist, Flash-Lite}{C2L}{--}{--}
\lfam{Stop veto, Flash}{C6F}{--}{--}
\lfam{Stop veto, Flash-Lite}{C6L}{--}{--}
\bottomrule
\end{tabular}}
\caption{\gym{} primaries per decision-model family. $\Delta$: success minus LLM-only (points). $p$: the family's
non-inferiority $p$, and $p_{\max}$ the $p$ that enters Holm. Cov.: tool-decision coverage (\%). Sav.: net savings.
\emph{Split}: one family passes. $^\|$NI with Intern-Decision at zero coverage.}
\label{tab:l2-families}
\end{table*}

\subsection{Stage 1: The Module Map}
\label{app:modulemap}
\begin{figure}[t]
\centering
{\mmDetailtrue\mmMap{flash}}
\caption{\textbf{The module map of Figure~\ref{fig:modulemap} with every mark.} Each cell shows the change in task
success against the LLM alone (points) $|$ cost saved (\%), with \jev{} and Gemini~3.5 Flash.}
\label{fig:modulemap-full}
\end{figure}

Figure~\ref{fig:modulemap-full} repeats Figure~\ref{fig:modulemap} with every mark. A cell is green if success fell by at most 1 point, cost fell, and the decision model changed the harness's choice at 5\% or more of the steps where it could. It is pink if success fell by 3 points or more. This rule was fixed in code before any number was computed. Every Stage-1
configuration is in \texttt{results/module\_map/module\_map.md}.

\paragraph{Exploratory cells.}
\label{app:modulemap-x}
Every remaining defined cell ran after Stage~1, each registered first (diamonds in Figure~\ref{fig:modulemap-full}).
The action row comes from ALFWorld. Of these cells, only the argument shortlist and the action cascade changed the agent's choices at lower cost. The
second and third \gym{} batches ran on another route to the same model, so their costs compare only with their own
LLM-only arm.

\subsection{More Stage-1 and Stage-2 Results}
\label{app:stage1res}\label{app:static}
\begin{table*}[tp]%
\centering\small
\setlength{\tabcolsep}{5pt}
\begin{tabular}{l ccc ccc cc cc}
\toprule
& \multicolumn{3}{c}{Tool selection (BFCL)} & \multicolumn{3}{c}{Arguments (BFCL)} & \multicolumn{2}{c}{When2Call} & \multicolumn{2}{c}{Efficiency}\\
\cmidrule(lr){2-4}\cmidrule(lr){5-7}\cmidrule(lr){8-9}\cmidrule(lr){10-11}
Decider & Acc & ECE & AUROC & Acc & Acc$_{\text{cand}}$ & ECE & Acc & ECE & p50 ms & \$/1k\\
\midrule
\multicolumn{11}{@{}l}{\emph{\systwo{} LLM}}\\
Gemini 3.5 Flash, native FC & \eone{gemini_flash_fc}{tool_select/bfcl/acc} & -- & --
  & \eone{gemini_flash_fc}{tool_params/bfcl/acc} & \eone{gemini_flash_fc}{tool_params/bfcl/acc_gold_in_candidates}
  & -- & \eone{gemini_flash_fc}{tool_select/when2call/acc} & -- & \eone{gemini_flash_fc}{tool_select/bfcl/latency_p50_ms}
  & \eone{gemini_flash_fc}{tool_select/bfcl/cost_per_1k_usd}\\
\eonerow{Gemini 3.5 Flash, typed question}{gemini_flash_typed}
\midrule
\multicolumn{11}{@{}l}{\emph{Typed decision models}}\\
\eonerow{\jev{}}{jev}
\eonerowk{\oursV{}}{e1_bfcl_when2call_rl/jevagent_rl}
\midrule
\multicolumn{11}{@{}l}{\emph{Generative deciders (reference)}}\\
\eonerow{Flash-Lite, verbalised}{gemini_lite_dist}
\eonerow{Gemma 4 26B-A4B, verbalised}{gemma4_dist}
\midrule
\multicolumn{11}{@{}l}{\emph{Retrieval controls}}\\
Embedding & \eone{gemini_embed}{tool_select/bfcl/acc} & \eone{gemini_embed}{tool_select/bfcl/ece}
  & \eone{gemini_embed}{tool_select/bfcl/auroc} & -- & -- & -- & -- & -- & \eone{gemini_embed}{tool_select/bfcl/latency_p50_ms}
  & \eone{gemini_embed}{tool_select/bfcl/cost_per_1k_usd}\\
\eonerow{BM25}{bm25}
\eonerow{TF-IDF}{tfidf}
\bottomrule
\end{tabular}
\caption{L1 results: accuracy (\%), ECE, AUROC, p50 latency and \$ per 1k decisions. Argument Acc credits
\textsc{Other} (defer) only when no candidate is right, and Acc$_{\text{cand}}$ covers slots whose gold value is a candidate.}
\label{tab:static}
\end{table*}

\paragraph{Stage 1.} Table~\ref{tab:static} gives the isolated decisions. Only \res{data_stats/bfcl_e1/in_candidates}\% of the
\res{data_stats/bfcl_e1/n_param_slots} argument slots have their gold value among the candidates, hence the
in-candidate column. The Stage-1 $\tau^2$ arms are in \texttt{results/\allowbreak tau2/} and the Intern-Decision arms
in \texttt{results/\allowbreak gpu\_round4\_main/\allowbreak e2a\_intern\_report.md}. A registered check compared how
\gym{} and the 35 airline tasks rank the \res{secondary/proxy/n} configurations run on both (Kendall's $\tau_b$
$\resm{secondary/proxy/delta_success/tau_b}$ by success difference, $p = \res{secondary/proxy/delta_success/p}$).

\paragraph{Stage 2.}\label{app:stage2res}%
Table~\ref{tab:stage2-clauses} splits each Stage-2 test into its conditions, and Table~\ref{tab:scorecard} gives the
secondary tests and descriptive rows (score cards in \texttt{results/\allowbreak stage2/\allowbreak stage2.md}).
With \jev{}, the reflection trigger fired in all \res{stage2/intervention/jevagent_best/failures} failures of the
combination.

\newcommand{\scl}[4]{#1 & #3 & \resm{stage2/#2/#4/est}\,{\scriptsize[\resm{stage2/#2/#4/est_ci/lo}, \resm{stage2/#2/#4/est_ci/hi}]}
  & \resm{stage2/#2/#4/boundary} & \res{stage2/#2/#4/p} & \res{stage2/#2/#4/decision}\\}
\newcommand{\sclna}[2]{#1 & -- & -- & -- & -- & #2\\}
\begin{table*}[!t]
\centering\footnotesize
\setlength{\tabcolsep}{3pt}
\begin{tabular}{@{}>{\raggedright\arraybackslash}p{2.6cm}lccc>{\raggedright\arraybackslash}p{3.5cm}@{}}
\toprule
Test & Condition & Estimate [95\% CI] & Null & Raw $p$ & Reading\\
\midrule
\scl{With \jev{}}{S2-1}{NI on success (pt)}{ni}
\scl{}{S2-1}{cost saving (\$/ep.)}{cost}
\scl{Against cascade alone}{S2-2}{NI on success (pt)}{ni}
\scl{}{S2-2}{cost saving (\$/ep.)}{cost}
\scl{\oursV{}, same states}{S2-3}{acc.\ vs.\ Intern (pt)}{intern}
\scl{}{S2-3}{acc.\ vs.\ \jev{} (pt)}{jev}
\scl{With \oursV{}}{S2-4}{NI on success (pt)}{ni}
\scl{}{S2-4}{cost saving (\$/ep.)}{cost}
\midrule
\scl{With \jev{}}{S2-1T}{latency saving (s/ep.)}{latency}
\sclna{With Flash-Lite}{not tested: no Flash-Lite delegation qualified}
\scl{On $\tau^2$ telecom}{S2-L3a}{NI on success (pt)}{ni}
\scl{}{S2-L3a}{cost saving (\$/task)}{cost}
\sclna{On ALFWorld}{not run: ALFWorld did not run in Stage~1}
\scl{With DeepSeek-V4-Pro}{S2-1X}{NI on success (pt)}{ni}
\scl{}{S2-1X}{cost saving (\$/ep.)}{cost}
\scl{With StartLux-4B}{S2-4S}{NI on success (pt)}{ni}
\scl{}{S2-4S}{cost saving (\$/ep.)}{cost}
\bottomrule
\end{tabular}
\caption{Each Stage-2 condition with its estimate and 95\% interval, null boundary, raw one-sided $p$ and reading.
Secondary tests are below the line. The rows with \oursV{} and StartLux-4B include a GPU price per episode
(\$\res{stage2/S2-4/tariff/q} and \$\res{stage2/S2-4S/tariff/q}). At \jev{}'s API price the StartLux-4B combination
would save \resm{stage2/S2-4S/tariff/jev_api_price/cost_savings}\%, and \resm{stage2/S2-4S/tariff/zero/cost_savings}\%
at a zero GPU price.}
\label{tab:stage2-clauses}
\end{table*}
\newcommand{\ssS}[3]{#1 & \resm{stage2/#2/ni/est} & \resm{stage2/#2/cost_savings}
  & \res{stage2/#2/coverage/tool_coverage} & \res{econ/#3/wall/mean} & \krOut{stage2/#2/decision}\\}
\newcommand{\ssGroup}[1]{\multicolumn{6}{@{}l}{\rule{0pt}{2.3ex}\emph{#1}}\\}
\begin{table*}[tp]
\centering\small
\setlength{\tabcolsep}{4pt}
\begin{tabular}{@{}l r r r r c@{}}
\toprule
& Effect & Cost & By & Wall- &\\
What is compared & pt & saved \% & model \% & clock s & Outcome\\
\midrule
\ssGroup{Secondary tests: 500 fresh \gym{} tasks unless noted, Flash}
Combination with \jev{}: model-call seconds saved$^a$ & \resm{stage2/S2-1T/delta} & -- & -- & --
  & \krOut{stage2/S2-1T/decision}\\
\ssS{Combination with StartLux-4B as decision model$^b$}{S2-4S}{jevagent_best_sl4b}
\ssS{Combination with DeepSeek-V4-Pro as agent LLM$^c$}{S2-1X}{jevagent_best_x}
Combination on $\tau^2$ telecom$^d$ & \resm{stage2/S2-L3a/ni/est} & \resm{stage2/S2-L3a/cost_savings}
  & \res{stage2/S2-L3a/coverage/tool_coverage} & -- & \krOut{stage2/S2-L3a/decision}\\
\ssGroup{Descriptive, fresh \gym{} tasks}
DeepSeek-V4-Flash alone$^c$ & \resm{stage2/scorecard/llm_only_xsmall/vs_baseline/delta}
  & \resm{stage2/scorecard/llm_only_xsmall/vs_baseline/cost_savings} & -- & \res{econ/llm_only_xsmall/wall/mean} & \krDesc\\
Combination on GPT-6.1-Sol$^e$ & \resm{stage2_sol/jevagent_best_sol/delta} & \resm{stage2_sol/jevagent_best_sol/cost_savings}
  & \res{stage2_sol/intervention/jevagent_best_sol/tool_coverage} & \res{stage2_sol/arm/jevagent_best_sol/wall_s} & \krDesc\\
GPT-6-Luna alone$^e$ & \resm{stage2_sol/llm_only_solsmall/delta} & \resm{stage2_sol/llm_only_solsmall/cost_savings}
  & -- & \res{stage2_sol/arm/llm_only_solsmall/wall_s} & \krDesc\\
\bottomrule
\end{tabular}
\caption{Stage-2 secondary tests and descriptive rows. Effect: success difference to LLM-only Flash unless noted
(points). By model: share of tool decisions the decision model took. $^a$Model-call seconds per episode, negative when
slower. $^b$\res{econ/jevagent_best_sl4b/concurrency} instead of \res{econ/llm_only/concurrency}
parallel episodes. $^c$Against V4-Pro alone (\res{econ/llm_only_x/wall/mean}\,s). $^d$5-point margin.
$^e$Registered optional row, against GPT-6.1-Sol alone (\res{stage2_sol/arm/llm_only_sol/success}\%,
\res{stage2_sol/arm/llm_only_sol/wall_s}\,s).}
\label{tab:scorecard}\label{tab:s2secondary}
\end{table*}

\subsection{Results Near the Ceiling}
\label{app:ceiling}
\providecommand{\krBandRowE}[1]{\rowcolor{krBand}\multicolumn{8}{@{\hspace{1.2mm}}l}{\rule{0pt}{2.4ex}\textit{#1}}\\}
\begin{table*}[!t]
\centering\small
\setlength{\tabcolsep}{3.5pt}
\renewcommand{\arraystretch}{1.12}
\resizebox{\textwidth}{!}{%
\begin{tabular}{@{}l r r r r r r c@{}}
\toprule
& \multicolumn{2}{c}{Success} & \multicolumn{2}{c}{Cost} & By & Wall- &\\
\cmidrule(lr){2-3}\cmidrule(lr){4-5}
Arm & \% & $\Delta$ pt & \$/1k ep. & saved \% & model \% & clock s & Outcome\\
\midrule
\krBandRowE{\res{data_stats/jevgym_eval/n_tasks} fresh \gym{} tasks, agent LLM Flash}
\krMk\krMkLo LLM-only & \res{stage2/S2-1/ni/success_ref} & & \resk{stage2/S2-1/cost_ref} & & &
  \res{econ/llm_only/wall/mean} & \krRef\\
\krMk\krMkCas Tool cascade alone, \jev{} & \res{stage2/S2-2/ni/success_ref}
  & \resm{stage2/scorecard/best_single/vs_baseline/delta} & \resk{stage2/S2-2/cost_ref}
  & \krSave{\res{stage2/scorecard/best_single/vs_baseline/cost_savings}}
  & \res{stage2/intervention/best_single/tool_coverage} & \res{econ/best_single/wall/mean} & \krDesc\\
\krMk\krMkPr Combination (cascade + trigger), \jev{} & \res{stage2/S2-1/ni/success_arm}
  & \resm{stage2/S2-1/ni/est} & \resk{stage2/S2-1/cost_arm} & \krSave{\res{stage2/S2-1/cost_savings}}
  & \res{stage2/S2-1/coverage/tool_coverage} & \res{econ/jevagent_best/wall/mean} & \krOut{stage2/S2-1/decision}\\
\krMkNone{}\quad the same, vs.\ the cascade alone & & \resm{stage2/S2-2/ni/est} &
  & \krSave{\res{stage2/S2-2/cost_savings}} & & & \krOut{stage2/S2-2/decision}\\
\krMk\krMkRl Same combination, \oursV{} & \res{stage2/S2-4/ni/success_arm}
  & \resm{stage2/S2-4/ni/est} & \resk{stage2/S2-4/cost_arm} & \krSave{\res{stage2/S2-4/cost_savings}}
  & \res{stage2/S2-4/coverage/tool_coverage} & \res{econ/jevagent_best_rl/wall/mean}$^a$ & \krOut{stage2/S2-4/decision}\\
\krBandRowE{Common states (\resnum{stage2/S2-3/corpus/occurrences} states of \res{stage2/S2-3/corpus/tasks}
  fresh tasks): tool-choice accuracy instead of success}
\krMkNone{}\oursV{} vs.\ Intern-Decision
  & \res{stage2/S2-3/accuracy/rl} & \resm{stage2/S2-3/intern/est} & & & & & \multirow{2}{*}{\krOut{stage2/S2-3/decision}}\\
\krMkNone{}\quad and vs.\ \jev{} (non-inferior, 5-pt margin) & & \resm{stage2/S2-3/jev/est}
  & & & & &\\
\bottomrule
\end{tabular}}
\caption{\textbf{Near the ceiling, delegation cut cost, and whether success held depended on the decision model.}
Registered Stage-2 tests on \gym{} tasks the LLM alone nearly always solved. $\Delta$: to
LLM-only unless the row names another reference. By model: share of tool decisions the decision model took. A claim
that did not hold was not established, which does not show a loss beyond the margin.
$^a$\res{econ/jevagent_best_rl/concurrency} instead of \res{econ/llm_only/concurrency} parallel episodes.}
\label{tab:keyresults}
\end{table*}
Table~\ref{tab:keyresults} holds the registered Stage-2 rows on tasks the LLM alone nearly always solved. For the
harder settings of the main text, Table~\ref{tab:mainresults-ci} gives the paired counts, the 95\% score interval of
the success difference \citep{tango1998equivalence} and the one-sided non-inferiority $p$ (rows added after Stage~1
apply the registered rule descriptively).

\begin{table}[!htbp]
\centering\scriptsize
\setlength{\tabcolsep}{2.5pt}
\providecommand{\hnRow}[2]{#1 & \resna{hardni/#2/n} & \resna{hardni/#2/wins} & \resna{hardni/#2/losses}
  & \resmna{hardni/#2/delta} & [\resmna{hardni/#2/lo}, \resmna{hardni/#2/hi}] & \resna{hardni/#2/p}\\}
\resizebox{\columnwidth}{!}{%
\begin{tabular}{@{}l r r r r c r@{}}
\toprule
Row of Table~\ref{tab:mainresults} & $n$ & wins & losses & $\Delta$ pt & 95\% interval & $p$\\
\midrule
\hnRow{\gym{}-Hard, combination, \jev{}}{e7b_jev}
\hnRow{\gym{}-Hard, combination, \oursV{}}{e7b_rl}
\ifcsname jevres@hardni/e7f/n\endcsname\hnRow{\gym{}-Hard, \oursV{} retrained}{e7f}\fi
\ifcsname jevres@hardni/e19/n\endcsname\hnRow{\gym{}-Hard, \oursV{} trained on random rules}{e19}\fi
\hnRow{\gym{}-Hard, cascade refitted, \jev{}}{e7e_jev}
\ifcsname jevres@hardni/e13_luna/n\endcsname\hnRow{\gym{}-Hard, GPT-6-Luna, cascade refitted}{e13_luna}\fi
\hnRow{SWE-bench, DeepSeek-V4}{e4}
\hnRow{SWE-bench, Gemini 3.8 Flash (App.~\ref{app:e4})}{e4b}
\ifcsname jevres@hardni/e4r/n\endcsname\hnRow{SWE-bench, DeepSeek-V4, 100 + 200 new}{e4r}\else
\ifcsname jevres@e4r/jevagent/vs_deep/n\endcsname SWE-bench, DeepSeek-V4, 100 + 200 new & \res{e4r/jevagent/vs_deep/n}
  & \res{e4r/jevagent/vs_deep/wins} & \res{e4r/jevagent/vs_deep/losses} & \res{e4r/jevagent/vs_deep/delta}
  & [\res{e4r/jevagent/vs_deep/tango_lo}, \res{e4r/jevagent/vs_deep/tango_hi}] & \res{e4r/jevagent/vs_deep/ni_p}\\\fi\fi
\hnRow{ALFWorld, action cascade}{e6a}
\ifcsname jevres@hardni/e11/n\endcsname\hnRow{ALFWorld, 4B trained there, cascade}{e11}\fi
$\tau^2$-bench telecom (registered) & \res{stage2/S2-L3a/n_pairs} & \res{stage2/S2-L3a/ni/wins}
  & \res{stage2/S2-L3a/ni/losses} & \resm{stage2/S2-L3a/ni/est}
  & [\resm{stage2/S2-L3a/ni/est_ci/lo}, \resm{stage2/S2-L3a/ni/est_ci/hi}] & \res{stage2/S2-L3a/ni/p}\\
\bottomrule
\end{tabular}}
\caption{\textbf{Paired tests for the rows of Table~\ref{tab:mainresults}.} Wins: tasks only the delegated arm
solved. Losses: tasks only the reference solved. $p$: one-sided non-inferiority test with a 3-point margin (5 points on
$\tau^2$-bench), non-inferior when $p \le 0.05$. 100 + 200 new: the instances of the first routing run pooled with
those of its extension (App.~\ref{app:e4}).}
\label{tab:mainresults-ci}
\end{table}

\subsection{Notes on Figure~\ref{fig:frontier} and Tables~\ref{tab:mainresults}--\ref{tab:train}}
\label{app:tablenotes}
\paragraph{Figure~\ref{fig:frontier}.} In (b), isolated means BFCL tool choice and spread is best minus worst. (a) \res{e9/n} recorded BFCL and When2Call decisions. \oursS{} is StartLux-0.8B
with our recipe, and a ring marks a hosted API. Orange diamonds: Gemini~3.5 Flash and Flash-Lite asked the same
multiple-choice question, and the open diamond is Flash's own tool call.\ifcsname
jevres@e30/e12/lg\endcsname{} Open models are priced at their cheapest deployment (App.~\ref{app:e30}), APIs and LLMs
at list price.\fi{} Models below 70\,\% are in App.~Table~\ref{tab:e15}. (b) Shifted \gym{} tasks
(App.~\ref{app:e21}). Isolated: BFCL tool choice on \res{diff/e1/bfcl_tool_select/n} items. Step level:
\res{e21/corpus/occurrences} recorded states of \res{e21/corpus/tasks} tasks. Task level: \res{e21/e12/l3/n} tasks with
the model picking every tool, of which Flash alone solved \res{e21/llm_only/l3/success}\%\ifcsname
jevres@e21/sl27b/l3/n\endcsname{} (StartLux-27B: the first \res{e21/sl27b/l3/n})\fi. \ours{} saw only the unshifted
generator.

\paragraph{Table~\ref{tab:mainresults}.} The reference is the LLM alone on the same tasks, Gemini~3.5 Flash unless the
row names another LLM. The refitted-cascade and GPT-6-Luna rows use other \gym{}-Hard tasks with list answers and their
own LLM-only runs (App.~\ref{app:e7}).\ifcsname jevres@e13/luna/llm_only/cached_share\endcsname{} In the Luna row,
``$\times$'' is the cost as a multiple of the reference's, whose run had \res{e13/luna/llm_only/cached_share}\% of its
input cached (\res{e13/luna/jev_refit/cached_share}\% with the cascade) and stopped
\res{e13/luna/llm_only/fail/wrongwrite} episodes early at a wrong write.\fi{} \ifeRLive The mini-SWE-agent row pools
\res{e4r/n_rep} earlier with \res{e4r/n_ext} new instances (App.~\ref{app:e4}). \fi\ifebLive The ALFWorld 4B model is
SFT on ALFWorld's training games (App.~\ref{app:e11}). \fi Paired tests: Table~\ref{tab:mainresults-ci}.

\paragraph{Table~\ref{tab:parts}.} ``Gate'' accepts the decision model's answer, ``4B'' is \oursV{} and ``4B + public data'' is \ours{}. The
fixed-threshold row covers tool choice on the first \res{mm/tool_select/cascade/jev_ts_cascade_tau0.5/n} tasks. Gate
rows are offline, per argument decision (App.~\ref{app:e8}). The new-rule row uses other tasks (App.~\ref{app:e7}).
``Own GPU'' rows compare with a reference run in the same session, and the coding row with the strong LLM.\ifcsname
jevres@e18/final/saving_vs_llm\endcsname{} The 0.8B row learns when to defer (App.~\ref{app:e18}).\fi{} Negative time
savings mean slower.

\paragraph{Table~\ref{tab:train}.} L1: isolated accuracy (\%) on the questions of Figure~\ref{fig:frontier:a}. L2 and
L3 use \gym{}-Shift (App.~\ref{app:e21}). L2: step-level accuracy on recorded states, with ``ask'' for the states where
the agent must ask (for ALFWorld, the share of acceptable actions). L3: success on \res{e21/rl/l3/n} shifted tasks with
the model making every tool choice. Saved: cost saved when deployed on new tasks of the trained kind. Green marks a
non-inferior result and rose one that is not. --: not deployed. ``+ GRPO (Stage~2)'' is \oursV{}.

\subsection{Evaluation Levels and Reach}
\label{app:levels}
Table~\ref{tab:levels} holds the numbers behind Figure~\ref{fig:levels}. Over the five pre-specified models the
isolated and task-level rankings disagree (Kendall rank correlation \resm{stage2_s1/kendall/e1_vs_live}).

\newcommand{\levci}[1]{\resm{#1}\,{\scriptsize[\resm{#1_ci/lo}, \resm{#1_ci/hi}]}}
\newcommand{\levrow}[3]{#1 & \res{explore/startlux_live/#2/e1_acc} & \res{#3/#2/corpus_acc} & \levci{#3/#2/vs_jev/corpus_delta}
  & \res{#3/#2/corpus_acc_offplan} & \res{#3/#2/ece} & \res{explore/startlux_live/#2/own_acc}
  & \res{explore/startlux_live/#2/live_success} & \levci{explore/startlux_live/#2/live_vs_jev}
  & \res{explore/startlux_live/#2/steps} & \res{explore/startlux_live/#2/api_cost}\\}
\newcommand{\levrl}[2]{#1 & \res{stage2_rl/#2/l1_acc} & \res{stage2_rl/#2/corpus_acc} & --
  & \res{stage2_rl/#2/corpus_acc_offplan} & \res{stage2_rl/#2/ece} & \res{stage2_rl/#2/own_acc}
  & \res{stage2_rl/#2/live_success} & -- & -- & --\\}
\begin{table*}[!t]
\centering\footnotesize
\setlength{\tabcolsep}{3.2pt}
\resizebox{\textwidth}{!}{%
\begin{tabular}{@{}l r rlrr r rl rr@{}}
\toprule
& Isolated & \multicolumn{4}{c}{Step-level: common states} & Own & \multicolumn{2}{c}{Task-level, model alone} & & API\\
\cmidrule(lr){3-6}\cmidrule(lr){8-9}
Decision model & acc. & acc. & $\Delta$ vs.\ \jev{} [95\% CI] & off-plan & ECE & states & success & $\Delta$ vs.\ \jev{} [95\% CI] & Steps & \$/ep.\\
\midrule
\jev{} & \res{explore/startlux_live/jev/e1_acc} & \res{stage2_s1/jev/corpus_acc} & -- & \res{stage2_s1/jev/corpus_acc_offplan}
  & \res{stage2_s1/jev/ece} & \res{explore/startlux_live/jev/own_acc} & \res{explore/startlux_live/jev/live_success} & --
  & \res{explore/startlux_live/jev/steps} & \res{explore/startlux_live/jev/api_cost}\\
\levrow{Intern-Decision-4B}{intern}{stage2_s1}
\levrow{StartLux-0.8B$^\ddagger$}{sl0.8b}{stage2_ladder}
\levrow{StartLux-2B$^\ddagger$}{sl2b}{stage2_ladder}
\levrow{StartLux-4B}{sl4b}{stage2_s1}
\levrow{StartLux-9B$^\ddagger$}{sl9b}{stage2_ladder}
\levrow{StartLux-27B (Q8\_0)}{sl27b}{stage2_s1}
\midrule
\levrow{\oursV{}$^\ast$}{frozen}{stage2_s1}
\levrl{\quad SFT only$^{\ast\S}$}{sft_v1}
\levrl{\quad Qwen3-4B, untrained$^\S$}{base}
\bottomrule
\end{tabular}}
\caption{Evaluation levels (descriptive except the registered common-state test). Tool-choice accuracy (\%) on
\resnum{stage2_s1/jev/l1_n} isolated BFCL questions, on the \resnum{stage2/S2-3/corpus/occurrences} common states of
\res{stage2/S2-3/corpus/tasks} new tasks and on each model's own states. Task level: the model picks every tool on
\res{stage2_s1/jev/live_n} new tasks (API cost, GPU not charged). $^\ddagger$Exploratory size ladder
(App.~\ref{app:explore}). $^\ast$Trained on \gym{}'s generator. $^\S$Task level on the first
\res{stage2_rl/sft_v1/live_n} tasks.}
\label{tab:levels}\label{tab:rlmain}\label{tab:rl}\label{tab:ladder}\label{tab:ladder-live}
\end{table*}

\paragraph{External reference agreement on \texorpdfstring{$\tau^2$}{tau2}.}
A registered offline diagnostic scores agreement with $\tau^2$'s reference trajectories on the
\res{t2ref/n_questions} questions the Stage-1 $\tau^2$ arms asked \jev{}. \oursV{} agreed more often than
Intern-Decision-4B and less often than \jev{}, but always choosing the reply to the user agrees about as often
(\res{t2ref/all/tool_select/majority/agree} against \res{t2ref/all/tool_select/frozen/agree}\%). The reference marks
one next step where several can be acceptable, so this agreement says little about $\tau^2$ success.

\paragraph{StartLux size ladder at the task level (exploratory).}
\label{app:explore}\label{app:explore-e2}
The StartLux-Decision rows 0.8B, 2B and 9B of Table~\ref{tab:levels} ($^\ddagger$) use the same items, states and
tasks as the other rows. Common-state accuracy rose with every size step (Kendall correlation with size
\res{explore/startlux_live/kendall/corpus_acc}) and task-level success did not
(\res{explore/startlux_live/kendall/live_success}).

\paragraph{Cost--accuracy frontier of isolated decisions.}
\label{app:e9}
Figure~\ref{fig:frontier} re-plots the recorded answers to the \res{e9/n} isolated decisions (post hoc), with API
models at their recorded price and open models at \$\res{e9/gpu/usdh} per GPU-hour of recorded serving time, an upper
bound (App.~\ref{app:e30}). A registered check on the \res{e9/check/n}-item development prefix found \jev{} at
\res{e9/check/jev/acc}\%, Flash at \res{e9/check/flash/acc}\% and Flash-Lite at \res{e9/check/lite/acc}\%.
\subsection{More Decision Models as Baselines}
\label{app:e15}
After Stage~1 we registered a round with every public decision model we could serve, each with its vendor's defaults
on one L40S GPU: Clef-Flash and Clef~\citep{cloudflare2026clef}, OpenJev in two
forms~\citep{openjev2026,openjevhf2026}, JevK5~\citep{jevk52026}, CLM~\citep{clm2026}, Laya~\citep{laya2026},
Verdict~\citep{verdict2026}, Imajev-4B~\citep{imajev2026}, decider-4b~\citep{decider2026},
Bosun~\citep{bosun2026} and OpenAI's hosted Decisions API~\citep{openai2026decisions}, which scores only the allowed
options.

\newcommand{\efrow}[3]{#1 & #2 & \res{e9/#3/acc} & \res{e9/#3/usdk} & \res{e15/#3/e1_acc} & \res{e15/#3/corpus_acc}
  & \res{e15/#3/ece} & \res{e15/#3/livecell}\\}
\newcommand{\efrowl}[3]{#1 & #2 & \res{e9/#3/acc} & \res{e9/#3/usdk} & \res{e15/#3/e1_acc} & -- & -- & --\\}
\begin{table}[t]
\centering\small
\setlength{\tabcolsep}{3.5pt}
\resizebox{\linewidth}{!}{%
\begin{tabular}{llrrrrrr}
\toprule
& & \multicolumn{2}{c}{L1: isolated} & tool choice & \multicolumn{2}{c}{L2: common states} & L3: task level \\
\cmidrule(lr){3-4}\cmidrule(lr){6-7}
Decision model & Licence & acc.\ (\%) & \$ / 1k & isolated (\%) & acc.\ (\%) & ECE & success (\%)\\
\midrule
\efrow{OpenJev-27B}{CC BY-NC}{openjevhf}
\efrow{OpenJev}{Apache-2.0}{openjev}
\efrow{OpenAI Decisions}{API}{oaidec}
\efrow{Imajev-4B}{Apache-2.0}{imajev}
\efrow{JevK5}{Apache-2.0}{jevk5}
\efrow{decider-4b}{Apache-2.0}{decider}
\efrow{Clef-Flash}{Apache-2.0}{clefflash}
\efrow{Bosun}{Apache-2.0}{bosun}
\efrow{Verdict}{Apache-2.0}{verdict}
\efrow{CLM}{Apache-2.0}{clm}
\efrow{Laya}{Apache-2.0}{laya}
\bottomrule
\end{tabular}}
\caption{\textbf{Isolated scores do not rank the new decision models either.} L1: accuracy and US\$ per 1,000
decisions on the \res{e9/n} isolated questions, then accuracy on their BFCL tool choices. L2: accuracy and calibration
error on the common states. L3: task success when the model picks every tool on the first
\res{explore/startlux_live/jev/live_n} fresh tasks with Flash, run for the API and the two open models best on the
common states. The reference models are in Table~\ref{tab:levels}. Clef's 27B answered
the isolated questions only. The Decisions API's \res{e15/oaidec/l1/refusals} refusals count as wrong.}
\label{tab:e15}
\end{table}

Most new models come close to \jev{} on isolated tool choice, the common states separate them, and the task level
reorders them again (Table~\ref{tab:e15}). Over the \res{e15/kendall/offtheshelf/n} off-the-shelf models, the Kendall
rank correlation with task-level success is \res{e15/kendall/offtheshelf/static_live} for isolated tool choice and
\res{e15/kendall/offtheshelf/common_live} for the common states.
\ifcsname jevres@e30/meta/price_h\endcsname\ifcsname jevres@e30/meta/price_h\endcsname
\providecommand{\eTrowYes}[2]{#1 & \res{e30/#2/prec} & \res{e30/#2/acc} & \res{e30/#2/ref} & \res{e30/#2/dps}
  & \res{e30/#2/usdk} & \res{e30/#2/lat_p50} & \res{e30/#2/lat_p90}\\}
\providecommand{\eTrowNo}[2]{}
\providecommand{\eTrow}[2]{\ifcsname jevres@e30/#2/acc\endcsname\expandafter\eTrowYes\else\expandafter\eTrowNo\fi{#1}{#2}}
\paragraph{Every open decision model served the cheapest way.}
\label{app:e30}
We served every open model the same cheap way (Table~\ref{tab:e30}): vLLM with prefix caching and a full queue on
one \res{e30/meta/gpu} GPU at \$\res{e30/meta/price_h} per hour, one forward pass per question, and 8-bit weights
unless they lost more than \res{e30/meta/margin} point of accuracy. The cost of 1,000 decisions is the hourly price
over the answers per second, a lower bound.
\ifcsname jevres@e30/clefflash/acc\endcsname\def\eTheads{1}\fi\ifcsname jevres@e30/e27sp/acc\endcsname\def\eTheads{1}\fi
\ifcsname jevres@e30/clefflash/acc\endcsname Imajev-4B, which combines four reads of each question through a trained
read-out, was not served this way.\else Clef-Flash and Imajev-4B were not served this way.\fi{}
For \res{e30/meta/samefor}, accuracy stayed within \res{e30/meta/maxdiff_same} points of Figure~\ref{fig:frontier:a}.
\ours{} costs \$\res{e30/e12/usdk} per 1,000 decisions, \res{e30/e12/vse9} times less than its point in the figure.

\begin{table}[t]
\centering\small
\setlength{\tabcolsep}{3pt}
\resizebox{\columnwidth}{!}{%
\begin{tabular}{@{}llrrrrrr@{}}
\toprule
& & \multicolumn{2}{c}{Isolated acc.\ (\%)} & Answers & \$ per & \multicolumn{2}{c}{Latency (ms)}\\
\cmidrule(lr){3-4}\cmidrule(lr){7-8}
Decision model & Precision & here & Fig.~\ref{fig:frontier:a} & per s & 1,000 & median & p90\\
\midrule
\eTrow{StartLux-0.8B}{sl0.8b}
\eTrow{StartLux-2B}{sl2b}
\eTrow{StartLux-4B}{sl4b}
\eTrow{StartLux-9B}{sl9b}
\eTrow{StartLux-27B$^\ast$}{sl27b}
\eTrow{Intern-Decision-4B}{intern}
\eTrow{decider-4b}{decider}
\eTrow{JevK5 (4B)}{jevk5}
\eTrow{Clef-Flash (9B)$^\S$}{clefflash}
\eTrow{OpenJev$^\ddagger$}{openjev}
\eTrow{OpenJev-27B$^\ast$}{openjevhf}
\midrule
\eTrow{\oursS{}}{e28}
\eTrow{\ours{}}{e12}
\eTrow{\oursV{}}{rl}
\eTrow{StartLux-4B + our recipe}{e12c}
\eTrow{Qwen3.5-0.8B variant$^\S$}{e27sp}
\bottomrule
\end{tabular}}
\caption{Open decision models served by vLLM on one GPU, on the \res{e30/meta/n} isolated questions of
Figure~\ref{fig:frontier:a}. Precision: the cheaper of BF16 and FP8 within \res{e30/meta/margin} point of the
figure's accuracy. Latency: one question at a time.\ifcsname jevres@e30/sl27b/acc\endcsname{} $^\ast$4-bit file on
llama.cpp.\fi\ifdefined\eTheads{} $^\S$Scoring head, own server.\fi\ifcsname jevres@e30/openjev/acc\endcsname{}
$^\ddagger$Own server on one \res{e30/openjev/gpu} GPU (\$\res{e30/openjev/price_h} per hour).\fi}
\label{tab:e30}
\end{table}
\fi
\fi%
\paragraph{Why step-level accuracy can miss a task-level loss.}
\label{app:drift}
OpenJev-27B and StartLux-27B match \jev{} on the common states yet solve about twenty points fewer tasks
(Tables~\ref{tab:e15} and~\ref{tab:levels}). We traced this in the stored trajectories (post hoc). Each tool-choice state of a model's own episodes is hashed like the common states, which tells whether it is in the corpus or \emph{novel} (Table~\ref{tab:drift}).

\newcommand{\driftrow}[2]{#1 & \res{drift/#2/live/success} & \res{drift/#2/c/acc} & \resna{drift/#2/acc/in_c}
  & \resna{drift/#2/acc/novel} & \res{drift/#2/states/novel_share} & \res{drift/#2/first/irreversible}
  & \res{drift/#2/mustask/tasks}\,/\,\resna{drift/#2/mustask/acc}
  & \resna{drift/#2/gate/09/wrong_below}\,/\,\resna{drift/#2/gate/09/right_below}\\}
\begin{table}[!htbp]
\centering\scriptsize
\setlength{\tabcolsep}{2pt}
\resizebox{\columnwidth}{!}{%
\begin{tabular}{@{}lrrrrrrrr@{}}
\toprule
& Closed- & Com. & In C & Novel & Novel & Irrev. & Must-ask & Below 0.9\\
& loop & states & acc. & acc. & share & first & tasks\,/\,acc. & wrong\,/\,right\\
\midrule
\driftrow{\jev{}}{jev}
\driftrow{Intern-Decision-4B}{intern}
\driftrow{StartLux-0.8B}{sl0.8b}
\driftrow{StartLux-2B}{sl2b}
\driftrow{StartLux-4B}{sl4b}
\driftrow{StartLux-9B}{sl9b}
\driftrow{StartLux-27B}{sl27b}
\driftrow{OpenAI Decisions}{oaidec}
\driftrow{OpenJev}{openjev}
\driftrow{OpenJev-27B}{openjevhf}
\midrule
\driftrow{\oursV{}}{rl}
\driftrow{\ours{}}{e12}
\driftrow{StartLux-4B + our recipe}{e12c}
\driftrow{0.8B head, agent states}{e14a}
\driftrow{0.8B head, generic mix}{e14g}
\midrule
\driftrow{Qwen3-4B (untrained)$^a$}{base}
\driftrow{SFT v1$^a$}{sft_v1}
\driftrow{StartLux-4B SFT (track 2)}{v2}
\bottomrule
\end{tabular}}
\caption{\textbf{The task-level loss comes from a few irreversible tool choices, and unfamiliar states explain little
of it.} Models picking every tool on the same \res{drift/jev/live/n} fresh tasks (\%): task-level success, common-state
accuracy, own-episode accuracy on corpus and novel states, novel share, tasks whose first wrong choice was
irreversible, tasks with a must-ask state and the accuracy there, and wrong and right choices with confidence below
0.9. $^a$First \res{drift/base/live/n} tasks. --: not defined.}
\label{tab:drift}
\end{table}

Serving does not explain the gap (live and offline choices agreed on \res{drift/openjevhf/faith/agree}\% of
\res{drift/openjevhf/faith/n} common states met live by OpenJev-27B), and the novel share hardly tracks success
(Kendall \res{drift/kendall/novel_share_vs_live} over \res{drift/n_models} models). Every failed task of \jev{},
OpenJev-27B and StartLux-27B ended at the model's first wrong tool choice, a cancellation, refund or shipment where a
lookup or a question was due. For the two 27B models it always came at a must-ask state. Offline the three models are
equally poor at must-ask states (\res{drift/jev/c/mustask_acc}, \res{drift/openjevhf/c/mustask_acc} and
\res{drift/sl27b/c/mustask_acc}\%), but these are only \res{drift/c/mustask/share}\% of the corpus, so the average
hides them. Counting a task as lost at its first irreversible wrong answer ranks the off-the-shelf models better than
the common-state mean (Kendall \res{drift/pred/kendall/offtheshelf/pred} against
\res{drift/pred/kendall/offtheshelf/cmean}) but misses task-level success by \res{drift/pred/mae/offtheshelf/pred}
points on average.
\subsection{Variants of the Decision Model}
\paragraph{Voting decision models.}
\label{app:e20}
We combined stored answers (post hoc) by majority vote, confidence-weighted vote, or an agreement gate that delegates
only when $m$ of $k$ members agree. Members and $m$ were chosen on one half of the states and scored on the other. No
vote reaches \oursV{} (Table~\ref{tab:e20}). The models make the same mistakes, and every off-the-shelf model fails
\res{e20/a/corr/all/must/failall}\% of the must-ask occurrences. In a registered follow-up, fitted thresholds made the
OpenAI Decisions API, alone or gated with \jev{}, defer every tool choice, so as registered neither ran at the task
level.

\newcommand{\ensrow}[3]{#1 & \resna{e20/a/l1/#3/acc} & \resna{e20/a/c/#2/acc} & \resna{e20/a/c/#2/must} & -- & --
  & \resna{e20/a/c/#2/pred}\\}
\newcommand{\gaterow}[3]{#1 & -- & \res{e20/a/c/#2/accacc}\textsuperscript{a} & \resna{e20/a/c/#2/musterr}\textsuperscript{b}
  & \res{e20/a/c/#2/cov} & \res{e20/a/c/#2/err}\,/\,\res{e20/a/c/#2/own/#3} & \res{e20/a/c/#2/pred}\\}
\begin{table}[!htbp]
\centering\footnotesize
\setlength{\tabcolsep}{3pt}
\resizebox{\columnwidth}{!}{%
\begin{tabular}{@{}lrrrrrr@{}}
\toprule
& Isolated & Common & Must- & Delegated & Errors: gate\,/ & Pred.\\
& (\%) & states (\%) & ask (\%) & (\%) & best member & L3 (\%)\\
\midrule
\ensrow{Best single off-the-shelf model}{ots/single}{ots/single}
\ensrow{Off the shelf, majority vote}{ots/majority}{ots/majority}
\ensrow{Off the shelf, weighted vote}{ots/weighted}{ots/weighted}
\gaterow{Off the shelf, all of \res{e20/a/c/ots/gate/k} agree}{ots/gate}{sl27b}
\ensrow{Hosted (\jev{} + Decisions API), weighted vote}{hosted/weighted}{hosted/weighted}
\gaterow{Hosted, both agree}{hosted/gate}{jev}
\ensrow{Off the shelf + our trained models, weighted vote}{mixed/weighted}{mixed/weighted}
\gaterow{Off the shelf + our trained models, \res{e20/a/c/mixed/gate/m} of \res{e20/a/c/mixed/gate/k} agree}{mixed/gate}{e12c}
\midrule
\oursV{} at its threshold & \res{e20/a/l1/ref/rl/acc} & \res{e20/a/c/ref/rl/acc} & \res{e20/a/c/ref/rl/must}
  & \res{e20/a/c/ref/rl/cov} & \res{e20/a/c/ref/rl/err} & \res{e20/a/c/ref/rl/pred}\\
0.8B scoring head at its threshold & \res{e20/a/l1/ref/e14a/acc} & \res{e20/a/c/ref/e14a/acc} & \res{e20/a/c/ref/e14a/must}
  & \res{e20/a/c/ref/e14a/cov} & \res{e20/a/c/ref/e14a/err} & \res{e20/a/c/ref/e14a/gpred}\\
\bottomrule
\end{tabular}}
\caption{\textbf{Voting does not replace training.} Held-out halves. Delegated: share of common-state decisions the
gate, or the model at its fitted threshold, takes. Errors: accepted wrong tool choices of the gate, then of its
strongest member at the same coverage. \textsuperscript{a}Accuracy of the delegated decisions.
\textsuperscript{b}Accepted errors on must-ask states. Pred.\ L3: the offline score of App.~\ref{app:drift}.}
\label{tab:e20}
\end{table}
\ifcsname jevres@e16/a/e1/acc\endcsname\paragraph{More public data and a larger open base.}
\label{app:e16}
Arm A continues \oursV{}'s adapter for two epochs on \res{e16/data/new} public tool-use items, mixed one to one with
its own SFT states. Arm B applies the SFT recipe to StartLux-Decision-9B (CC~BY-NC~4.0) on the same mix. Neither arm
closed the isolated gap to StartLux-27B, and When2Call stayed the weak part (Table~\ref{tab:e16}). Both arms were
non-inferior to \oursV{} on the common states at the registered 1-point margin.

\begin{table}[h]
\centering\footnotesize
\setlength{\tabcolsep}{3pt}
\resizebox{\columnwidth}{!}{%
\begin{tabular}{@{}l r r r r r r@{}}
\toprule
& \multicolumn{4}{c}{Isolated (\%)} & Step & Task\\
\cmidrule(lr){2-5}
Arm & tool & args & When2Call & all & level & level\\
\midrule
\oursV{} & \res{e9/rl/acc/toolchoice} & \res{e9/rl/acc/args} & \res{e9/rl/acc/w2c} & \res{e9/rl/acc}
  & \res{e16/row/frozen/l2} & \res{e16/row/frozen/l3}\\
A: \oursV{} + more data & \res{e16/a/e1/acc/toolchoice} & \res{e16/a/e1/acc/args} & \res{e16/a/e1/acc/w2c}
  & \res{e16/a/e1/acc} & \res{e16/a/c/acc} & \res{e16/a/live/success}\\
\ifcsname jevres@e16/b/e1/acc\endcsname
B: StartLux-9B + recipe & \res{e16/b/e1/acc/toolchoice} & \res{e16/b/e1/acc/args} & \res{e16/b/e1/acc/w2c}
  & \res{e16/b/e1/acc} & \res{e16/b/c/acc} & \res{e16/b/live/success}\\
\fi
\midrule
Best untrained open model$^\ast$ & \res{e9/sl27b/acc/toolchoice} & \res{e9/sl27b/acc/args} & \res{e9/sl27b/acc/w2c}
  & \res{e16/best_open/l1/value} & \res{e16/best_open/l2/value} & \res{e16/best_open/l3/value}\\
\bottomrule
\end{tabular}}
\caption{Two trained open models at the three levels (exploratory). Isolated: the \res{e9/n} decisions and their
three parts. Step level: common states. Task level: the model
picks every tool on \res{e16/a/live/n} fresh tasks. $^\ast$Best open model we did not train: \res{e16/best_open/l1/model}
(isolated, step level) and \res{e16/best_open/l3/model} (task level).}
\label{tab:e16}
\end{table}
\fi%
\ifcsname jevres@e18/final/success\endcsname\paragraph{Scoring head against one-token read-out.}
\label{app:e17}
Arm (i) is the 0.8B scoring head of App.~\ref{app:e14}, whose fitted thresholds sat within $10^{-9}$ of 1. Arm (ii)
reads the next-token probability over option letters from the same backbone and \res{e14/a/train/records} records.
\ifcsname jevres@e17/gen/c/acc\endcsname
Accuracy stayed the same (Table~\ref{tab:e17}). Under the registered criterion, \res{e17/reading/ts} covers more of these near-ceiling states. Its threshold, however, accepted \res{e18/c/gen/accerr} wrong tool choices on the common states, and the head's threshold accepted none. The head is also \res{e17/lat/head/vsgen/c1} times faster and needs
\res{e17/lat/head/mem} against \res{e17/lat/gen/mem}\,GB of GPU memory.
\fi
\ifcsname jevres@e17/gen/train/epochs\endcsname\else
The generative arm did not finish on the host, so the readout reports the head's numbers only.
\fi

\begin{table}[!htbp]
\centering\footnotesize
\setlength{\tabcolsep}{3pt}
\resizebox{\columnwidth}{!}{%
\begin{tabular}{@{}lrrrrrrrrr@{}}
\toprule
& \multicolumn{2}{c}{Held-out rows} & \multicolumn{2}{c}{Common states} & \multicolumn{3}{c}{Calibration states at $\hat\tau$} & \multicolumn{2}{c}{Serving (ms)}\\
\cmidrule(lr){2-3}\cmidrule(lr){4-5}\cmidrule(lr){6-8}\cmidrule(lr){9-10}
Read-out & acc.\ (\%) & ECE & acc.\ (\%) & off-plan (\%) & tools (\%) & args (\%) & errors & $c{=}1$ & $c{=}4$\\
\midrule
(i) scoring head & \res{e17/head/v1/acc} & \res{e17/head/v1/ece} & \res{e17/head/c/acc} & \res{e17/head/c/offplan}
  & \res{e17/head/cal/ts/cov} & \res{e17/head/cal/tp/cov} & \res{e17/head/cal/ts/errors}+\res{e17/head/cal/tp/errors}
  & \resna{e17/lat/head/c1/p50} & \resna{e17/lat/head/c4/p50}\\
(ii) one-token read-out & \resna{e17/gen/v1/acc} & \resna{e17/gen/v1/ece} & \resna{e17/gen/c/acc} & \resna{e17/gen/c/offplan}
  & \resna{e17/gen/cal/ts/cov} & \resna{e17/gen/cal/tp/cov} & \resna{e17/gen/cal/ts/errors}+\resna{e17/gen/cal/tp/errors}
  & \resna{e17/lat/gen/c1/p50} & \resna{e17/lat/gen/c4/p50}\\
\midrule
\oursV{} & -- & -- & \res{e14/ref/rl/c/acc} & -- & \res{e17/fourb/cal/ts/cov} & \res{e17/fourb/cal/tp/cov}
  & \res{e17/fourb/cal/ts/errors}+\res{e17/fourb/cal/tp/errors} & \resna{e17/lat/fourb/c1/p50} & \resna{e17/lat/fourb/c4/p50}\\
\bottomrule
\end{tabular}}
\caption{One backbone, two read-outs. Held-out rows: SFT v1's development file. Common states: accuracy, all and
after an off-plan step. Calibration states: share of \oursV{}'s calibration decisions taken at each fitted threshold,
and the wrong ones. Serving: median ms per request at one and four concurrent requests on one A10G.}
\label{tab:e17}
\end{table}

\paragraph{A deferral-aware objective.}
\label{app:e18}
In the spirit of learning to defer~\citep{mozannar2020consistent}, a defer head learns when to leave a choice to the LLM. It
is trained on the \res{e18/head/ts/n} tool choices and \res{e18/head/tp/n} argument questions of \oursV{}'s
calibration arm, with target $d = 1$ where the LLM is right and the head is wrong or its two top probabilities differ
by less than the cost margin $k$ of Proposition~1 ($k = \res{e18/head/ts/k}$ for tools, $\res{e18/head/tp/k}$ for
arguments). Because there are only \res{e18/head/positives} positive examples, the defer head is the registered fallback, a logistic regression on the frozen head's features. It finds the states where the model errs but still takes most of them (Table~\ref{tab:e18}).
\ifcsname jevres@e18/final/success\endcsname
On the \res{e18/final/n} tasks of the final harness it solved \res{e18/final/success}\% (Flash alone
\res{e18/final/llm/success}\%), took \res{e18/final/toolcov}\% of the tool choices and cost
\res{e18/final/saving_vs_llm}\% less than Flash alone. The registered criterion (higher coverage, success within 3
points) \res{e18/reading/live}.
\fi
\ifcsname jevres@e18/f/train/minutes\endcsname
Arm F, which trained the target in as an extra option, almost never deferred.
\fi

\begin{table}[!htbp]
\centering\footnotesize
\setlength{\tabcolsep}{4pt}
\resizebox{\columnwidth}{!}{%
\begin{tabular}{@{}lrrrr@{}}
\toprule
Rule on the common states & coverage (\%) & accepted right (\%) & accepted errors & deferred right (\%)\\
\midrule
arm A + defer head, defer iff $P(d{=}1 \mid x) > k$ & \res{e18/c/e18/cov} & \res{e18/c/e18/accacc} & \res{e18/c/e18/accerr} & \res{e18/c/e18/defacc}\\
arm A at its fitted threshold & \res{e18/c/a/cov} & \res{e18/c/a/accacc} & \res{e18/c/a/accerr} & \res{e18/c/a/defacc}\\
arm A, every decision taken & \res{e18/c/areplace/cov} & \res{e18/c/areplace/accacc} & \res{e18/c/areplace/accerr} & --\\
\oursV{} at its fitted threshold & \res{e18/c/rl/cov} & \res{e18/c/rl/accacc} & \res{e18/c/rl/accerr} & \resna{e18/c/rl/defacc}\\
\ifcsname jevres@e18/c/e18f/cov\endcsname
arm F (the option trained in), defer iff $p_{\texttt{defer}} > k$ & \res{e18/c/e18f/cov} & \res{e18/c/e18f/accacc} & \res{e18/c/e18f/accerr} & \resna{e18/c/e18f/defacc}\\
\fi
\bottomrule
\end{tabular}}
\caption{The learned deferral rule on the \res{e18/c/e18/n} tool-choice occurrences of the common states. Arm A:
the 0.8B scoring head.}
\label{tab:e18}
\end{table}
\fi%
\ifcsname jevres@e21/rl/l3/success\endcsname\subsection{A Shifted DecisionGym Split}
\label{app:e21}
\gym{}-Shift\dg{} renames every core tool and the clarification tool, rewords the requests, the policy line and the
customer's replies, and doubles the distractor tools (\res{e21/corpus/options} options per tool choice instead of 69). It adds must-ask states and three
plan shapes the training generator never produces: asking twice, a request that becomes ambiguous only after a lookup,
and a confirmation message after shipping. The common states are \res{e21/corpus/occurrences} states of
\res{e21/corpus/tasks} shifted tasks, and live runs use \res{e21/llm_only/l3/n} further tasks.

\newcommand{\eturow}[2]{#1 & \resna{e21/#2/in/c} & \resna{e21/#2/l2/all} & \resna{e21/#2/l2/mustask}
  & \resna{e21/#2/l2/renamed} & \resna{e21/#2/in/live} & \resna{e21/#2/l3/success}\\}
\newcommand{\etxrow}[3]{#1 & \res{#2/l2in/all} & \res{#2/l2/all} & \res{#2/l2/mustask} & \res{#2/l2/renamed} & #3
  & \res{#2/l3/success}\\}
\begin{table}[t]
\centering\small
\setlength{\tabcolsep}{3.5pt}
\resizebox{\linewidth}{!}{%
\begin{tabular}{lrrrrrr}
\toprule
& \multicolumn{4}{c}{Common states, accuracy (\%)} & \multicolumn{2}{c}{Live success (\%)}\\
\cmidrule(lr){2-5}\cmidrule(lr){6-7}
Decision model & in-domain & shift & must-ask & renamed & in-domain & shift\\
\midrule
\eturow{Untrained Qwen3-4B}{base}
\eturow{+ SFT}{sft_v1}
\eturow{\oursV{}}{rl}
\eturow{\ours{} (+ public data)}{e12}
\eturow{StartLux-4B + recipe}{e12c}
\eturow{0.8B scoring head}{e14a}
\midrule
\ifcsname jevres@e28/l2/all\endcsname
\etxrow{\oursS{}}{e28}{\res{e28/l3in/success}}
\fi
\ifcsname jevres@e27/s/l2/all\endcsname
\etxrow{Qwen3.5-0.8B, renamed tools}{e27/s}{--}
\etxrow{Qwen3.5-0.8B variant}{e27/sp}{\res{e27/sel/l3in/success}}
\fi
\midrule
\eturow{\jev{}}{jev}
\eturow{OpenAI Decisions}{oaidec}
\eturow{StartLux-27B}{sl27b}
\eturow{StartLux-9B}{sl9b}
\eturow{StartLux-4B}{sl4b}
\ifcsname jevres@e28/base/l2/all\endcsname
\etxrow{StartLux-0.8B}{e28/base}{\res{e28/base/l3in/success}}
\fi
\eturow{OpenJev-27B}{openjevhf}
\eturow{Intern-Decision-4B}{intern}
\midrule
Agent LLM alone & -- & -- & -- & -- & \resna{e21/llm_only/in/live} & \resna{e21/llm_only/l3/success}\\
\bottomrule
\end{tabular}}
\caption{Decision models on the shifted split\dg{} beside their in-domain numbers. Common states: accuracy on all,
must-ask and renamed-tool states. Live: success when the model picks every tool on \res{e21/rl/l3/n} shifted tasks
(StartLux-27B: the first \res{e21/sl27b/l3/n}). The in-domain live numbers of the untrained model and of + SFT cover
the first 100 tasks. Top: trained on the generator. Second block: 0.8B models trained with renamed tools
(App.~\ref{app:e27}). Third block: off-the-shelf models. --: not run.}
\label{tab:e21}\label{tab:e27}\label{tab:e28}
\end{table}

The 4B models trained on the generator lost little and stayed ahead of every off-the-shelf model at both levels
(Table~\ref{tab:e21}). The 0.8B head did not carry over. With the old names and wording restored it answered
\res{e21/e14a/l2/nosurf}\% of the same states, so it had learned the generator's names more than the decisions.
\fi%
\ifcsname jevres@e21b/rl/success\endcsname\ifcsname jevres@e21b/rl/success\endcsname
\paragraph{The confidence gate under shift.}
\label{app:e21b}
We ran the final harness\dg{}, which takes a decision only when the model's confidence clears its threshold, on the
same \res{e21b/rl/n} shifted tasks with each model's in-domain thresholds (Table~\ref{tab:e21b}). The two 4B models
stayed confident and kept taking most tool choices, so success fell. Only the 0.8B head's learned deferral handed
choices back, and non-inferiority was shown for none of the three.

\begin{table}[!htbp]
\centering\footnotesize
\setlength{\tabcolsep}{2.5pt}
\resizebox{\columnwidth}{!}{%
\begin{tabular}{@{}lrrrrrrr@{}}
\toprule
& & & \multicolumn{2}{c}{Taken (\%)} & Wrong & Saved & Every\\
\cmidrule(lr){4-5}
Decision model & Succ. & $\Delta$ & tools & args & taken & (\%) & tool\\
\midrule
\oursV{} & \res{e21b/rl/success} & \res{e21b/rl/delta} & \res{e21b/rl/toolcov} (\res{e21b/rl/in/toolcov})
  & \res{e21b/rl/argcov} (\res{e21b/rl/in/argcov}) & \res{e21b/rl/acc_wrong_tool} & \res{e21b/rl/saving} & \res{e21b/rl/replace/success}\\
\ours{} & \res{e21b/e12/success} & \res{e21b/e12/delta} & \res{e21b/e12/toolcov} (\res{e21b/e12/in/toolcov})
  & \res{e21b/e12/argcov} (\res{e21b/e12/in/argcov}) & \res{e21b/e12/acc_wrong_tool} & \res{e21b/e12/saving} & \res{e21b/e12/replace/success}\\
0.8B head + deferral & \res{e21b/e18/success} & \res{e21b/e18/delta} & \res{e21b/e18/toolcov} (\res{e21b/e18/in/toolcov})
  & \res{e21b/e18/argcov} (\res{e21b/e18/in/argcov}) & \res{e21b/e18/acc_wrong_tool} & \res{e21b/e18/saving} & \res{e21b/e18/replace/success}\\
\midrule
LLM alone & \res{e21b/ref/success} & -- & -- & -- & -- & -- & --\\
\bottomrule
\end{tabular}}
\caption{The final harness\dg{} on \res{e21b/rl/n} shifted tasks at in-domain thresholds: success (\%), difference to
the LLM alone (points), share of decisions taken (in-domain in parentheses), wrong tool choices accepted, cost saved
with the GPU, and success when the model makes every tool choice (Table~\ref{tab:e21}).}
\label{tab:e21b}
\end{table}
\fi
\fi%
\ifcsname jevres@e27/sel/l3/success\endcsname\ifcsname jevres@e27/s/l2/all\endcsname
\paragraph{A 0.8B model trained on renamed tools.}
\label{app:e27}
The 0.8B scoring head of App.~\ref{app:e14} failed on the shifted tasks. We retrained it\dg{} for one epoch on agent
states whose surface we varied. In a random half of them the core tools got new names, the text was reworded, and
near-copies of the renamed tools were added as distractors. A variant kept by a rule fixed before training also mixed
in public tool-use data (the Qwen3.5-0.8B variant). Renaming in training closed the step-level gap, while the new task
shapes still cost task-level success (Table~\ref{tab:e27}). The registered claim had two conditions under shift. Step-level accuracy stayed within 3 points of \ours{}, but the task-level gap to \ours{} (\res{e27/sp/l3/vs_e12} points, lower confidence bound \res{e27/sp/l3/vs_e12_lo}) missed the $-5$-point bound.
\ifcsname jevres@e27/sel/final_in/success\endcsname
In the final harness on \res{e27/sel/final_in/n} in-domain tasks the variant took \res{e27/sel/final_in/toolcov}\% of
the tool choices without a wrong one at \res{e27/sel/final_in/saving}\% lower cost, but non-inferiority was not shown
($p = \res{e27/sel/final_in/ni_p}$).
\fi
\ifcsname jevres@e27/sel/final_shift/success\endcsname
On shifted tasks it handed \res{e27/sel/final_shift/handed}\% of the tool choices back, every must-ask state among
them, and the agent stayed non-inferior to the LLM alone at a \res{e27/sel/final_shift/saving}\% saving.
\fi
\fi
\fi%
\ifcsname jevres@e28/l3/success\endcsname\ifcsname jevres@e28/l2/all\endcsname
\paragraph{Our recipe on a decision-tuned 0.8B base.}
\label{app:e28}
\oursS{} is StartLux-Decision-0.8B\dg{} trained with the recipe of App.~\ref{app:e12} on \res{e28/data/agentrows}
agent states and \res{e28/data/publicrows} public tool-use items, with the tools renamed in a random half of the
states. Its CC BY-NC 4.0 licence carries over. Training improved the base model at every level (Table~\ref{tab:e28})\ifcsname jevres@e28/l1/usdk\endcsname, and the trained model costs \$\res{e28/l1/usdk} per 1{,}000 decisions on
the base model's own engine\fi. As with the Qwen3.5-0.8B variant, step-level accuracy held under shift but task-level
success did not (\res{e28/l3/genkinds}\% of the shifted tasks of the generator's kinds, \res{e28/l3/newkinds}\% of the
new shapes).
\ifcsname jevres@e28/final_in/success\endcsname
In the final harness on \res{e28/final_in/n} in-domain tasks it took \res{e28/final_in/toolcov}\% of the tool choices
without a wrong one and was non-inferior to the LLM alone (\res{e28/final_in/success}\% against
\res{e28/final_in/ref}\%, $p = \res{e28/final_in/ni_p}$). It saved \res{e28/final_in/saving_api}\% of the API cost,
but our slow serving stack made the run \resabs{e28/final_in/saving}\% more expensive once GPU time is counted.
\fi
\ifcsname jevres@e28/final_shift/success\endcsname
On the shifted tasks its gate did not hand control back. It took \res{e28/final_shift/toolcov}\% of the tool choices,
\res{e28/final_shift/toolwrong} of them wrong, and non-inferiority was not shown (\res{e28/final_shift/success}\%
against \res{e28/final_shift/ref}\%).
\fi
\fi
\fi%

\subsection{Hard Tasks (Exploratory)}
\label{app:hard}
\newcommand{\ehard}[3]{#1 & \res{#2/#3/success} & \res{#2/#3/vs_baseline/delta}\,{\scriptsize(\res{#2/#3/vs_baseline/delta_lb90})}
  & \res{#2/#3/vs_baseline/wins}/\res{#2/#3/vs_baseline/losses}
  & \res{#2/#3/wall_p50_s} & \res{#2/#3/wall_p90_s} & \res{#2/#3/cost_usd} & \res{#2/#3/steps}
  & \res{#2/#3/screen} & \res{#2/#3/faster}\\}
\newcommand{\ehardref}[3]{#1 & \res{#2/#3/success} & -- & --
  & \res{#2/#3/wall_p50_s} & \res{#2/#3/wall_p90_s} & \res{#2/#3/cost_usd} & \res{#2/#3/steps} & -- & --\\}
\begin{table*}[!t]
\centering\footnotesize
\setlength{\tabcolsep}{4pt}
\begin{tabular}{@{}lccccccccc@{}}
\toprule
Configuration & Succ. & $\Delta$ (LB90) & W/L & p50\,s & p90\,s & \$/task & Steps & Screen & Faster\\
\midrule
\multicolumn{10}{@{}l}{\emph{SWE-bench Verified (\res{swe_verified_100/n_tasks} instances, $k{=}1$), mini-SWE-agent}}\\
\ehardref{Unmodified harness}{swe_verified_100}{baseline}
\ehard{+ command \modeGua{}}{swe_verified_100}{action_guard}
\ehard{+ submit \modeGua{}}{swe_verified_100}{stop_guard}
\ehard{+ routing \modeTri{}}{swe_verified_100}{route}
\multicolumn{10}{@{}l}{\emph{Terminal-Bench-Science (\res{tbsci/n_tasks} hard tasks, $k{=}\res{tbsci/k}$), pi + Harbor}}\\
\ehardref{Unmodified harness}{tbsci}{baseline}
\ehard{+ stop \modeGua{} \& refl.\ \modeTri{}$^\ast$}{tbsci}{jev_stop_reflect}
\ehard{+ tool \& stop \modeGua{}}{tbsci}{jev_gate_stop}
\multicolumn{10}{@{}l}{\emph{Terminal-Bench 2.0 (89 tasks): not run (budget)}}\\
\bottomrule
\end{tabular}
\caption{Hard tasks (exploratory) with Gemini~3.5 Flash: success (\%), paired $\Delta$ to the unmodified harness
(points, one-sided 90\% lower bound), task wins/losses, agent time (s), cost per task (\$) and steps. Screen: lower
bound above $-5$ points and at least as many wins as losses. Faster: less agent time on jointly solved tasks (sign
test, $p<0.05$), n.e.\ when no task was solved by both. Terminal-Bench-Science sits at a floor (\res{tbsci/baseline/solved} of \res{tbsci/baseline/n}
attempts solved), so its screen passes trivially. $^\ast$Primary arm, fixed in advance.}
\label{tab:hard}
\end{table*}

SWE-bench Verified allows at most 250 steps and \$3 per attempt, and Terminal-Bench-Science \citep{tbscience2026}
runs pi in Harbor sandboxes for at most 1\,h per attempt. No arm reaches the registered verdict. The later
step-routing runs (App.~\ref{app:e4}) supersede the routing row.

\section{Extended Related Work}
\label{app:related}
\paragraph{Decision models.}
Beyond \S\ref{sec:related}, \jev{} has served as \modeRep{} for action-only and memory control
\citep{ma2026jevstar,jiang2026jevmem} and as a judge or monitor \citep{guo2026justaskjev,abbasi2026controlharness}.
Studies examine its calibration and robustness \citep{deusser2026benchmarkingjev,wu2026decisionhijacking}. Closest
to us, JEV-as-a-Judge fits an acceptance threshold on about 100 local labels \citep{li2026jevjudge}, and REFLEX
escalates agent actions below one global threshold, an action-level \modeCas{} \citep{wu2026reflex}. We compare decision points and modes under one cost ledger, set thresholds by price, and test
the chosen combination and a post-trained model on new tasks.

\paragraph{Routers, cascades and deferral.}
Cascade routing \citep{dekoninck2025cascaderouting}, agentic routers \citep{liu2026agenticrouting} and fast--slow
controllers \citep{yang2026cogrouter,zhang2025adaptthink} decide which model or how much thinking a step gets.
\jev{} and LLM judges err on the same items, so a cascade between them mainly saves cost
\citep{rao2026jevrubricjudges}. Agents also ask for help through uncertainty-aware deferral
\citep{piatrashyn2026redact} or conformal prediction \citep{ren2023knowno}, and finite-sample risk control
\citep{angelopoulos2024crc} could replace our plug-in thresholds, and serving systems such as vLLM
\citep{kwon2023vllm} lower the price of each call. Tool retrieval \citep{patil2024gorilla} is
\modeAss{} with a similarity scorer, as in our BM25 control \citep{robertson2009bm25}, and reflection
\citep{shinn2023reflexion} is a \modeTri{} decision in our terms.

\paragraph{Agent benchmarks and step-level evaluation.}
Static parity that hides in-loop differences was reported for browser agents \citep{webjev2026} and legal documents
\citep{zhang2026samescores}, and a public decision index reports mixed interactive results for \jev{} alone
\citep{decisionindex2026}. Paired replay of recorded states has precedent in search stopping
\citep{deepsearcher2026jevstopping}, and process reward models score agent steps \citep{xi2025agentprm}.
\benchname{} measures the gap directly by scoring the same models on isolated questions, recorded states and live
tasks.

\paragraph{Training small models on agent states.}
Agentic RL trains the generating LLM on multi-turn trajectories \citep{zhou2024archer,song2024eto,wang2025ragen}, tool
use \citep{qian2025toolrl,jin2025searchr1} or tool-call efficiency \citep{wang2025actingless}. We train the decision
model. Open decision models trained on a target loop have beaten \jev{} there \citep{webjev2026} or on a static set
without transferring \citep{cheng2026thisthatmodel}, and small models are argued to suffice for most agentic calls
\citep{belcak2025slmagentic}. We compare on-policy distillation \citep{agarwal2024gkd,gu2024minillm} with
distillation on fixed states \citep{hinton2015distilling,kang2025agentdistill} and use group-relative RL
\citep{shao2024deepseekmath,feng2025gigpo} with calibration-aware rewards \citep{damani2025rlcr}.

\par\providecommand{\defres}[2]{\expandafter\def\csname jevres@#1\endcsname{#2}}
\defres{cases/alf/height}{60.80}%
\defres{cases/alf/n_rows}{4}%
\defres{cases/alf/n_tasks}{140}%
\defres{cases/alf/rank}{1}%
\defres{cases/alf/ref/api}{0.0036}%
\defres{cases/alf/ref/cost}{0.0036}%
\defres{cases/alf/ref/count/args}{0}%
\defres{cases/alf/ref/count/args_deferred}{0}%
\defres{cases/alf/ref/count/args_dm}{0}%
\defres{cases/alf/ref/count/free}{0}%
\defres{cases/alf/ref/count/tools}{4}%
\defres{cases/alf/ref/count/tools_dm}{0}%
\defres{cases/alf/ref/gpu}{0.0000}%
\defres{cases/alf/ref/llm_calls}{4}%
\defres{cases/alf/ref/n}{4}%
\defres{cases/alf/ref/s/1/alt}{}%
\defres{cases/alf/ref/s/1/altp}{}%
\defres{cases/alf/ref/s/1/by}{llm}%
\defres{cases/alf/ref/s/1/cnum}{-1}%
\defres{cases/alf/ref/s/1/conf}{}%
\defres{cases/alf/ref/s/1/irr}{0}%
\defres{cases/alf/ref/s/1/llm}{1}%
\defres{cases/alf/ref/s/1/must}{0}%
\defres{cases/alf/ref/s/1/nl}{0}%
\defres{cases/alf/ref/s/1/nm}{0}%
\defres{cases/alf/ref/s/1/obs}{You arrive at desk 1. On the desk 1, you see a alarmclock 2, a book 1, a bowl 1, a cd 2, a desklamp 1, a mug 2, a…}%
\defres{cases/alf/ref/s/1/obsk}{env}%
\defres{cases/alf/ref/s/1/obsn}{2}%
\defres{cases/alf/ref/s/1/ok}{}%
\defres{cases/alf/ref/s/1/tool}{go to desk 1}%
\defres{cases/alf/ref/s/2/alt}{}%
\defres{cases/alf/ref/s/2/altp}{}%
\defres{cases/alf/ref/s/2/by}{llm}%
\defres{cases/alf/ref/s/2/cnum}{-1}%
\defres{cases/alf/ref/s/2/conf}{}%
\defres{cases/alf/ref/s/2/irr}{0}%
\defres{cases/alf/ref/s/2/llm}{1}%
\defres{cases/alf/ref/s/2/must}{0}%
\defres{cases/alf/ref/s/2/nl}{0}%
\defres{cases/alf/ref/s/2/nm}{0}%
\defres{cases/alf/ref/s/2/obs}{You turn on the desklamp 1.}%
\defres{cases/alf/ref/s/2/obsk}{env}%
\defres{cases/alf/ref/s/2/obsn}{1}%
\defres{cases/alf/ref/s/2/ok}{}%
\defres{cases/alf/ref/s/2/tool}{use desklamp 1}%
\defres{cases/alf/ref/s/3/alt}{}%
\defres{cases/alf/ref/s/3/altp}{}%
\defres{cases/alf/ref/s/3/by}{llm}%
\defres{cases/alf/ref/s/3/cnum}{-1}%
\defres{cases/alf/ref/s/3/conf}{}%
\defres{cases/alf/ref/s/3/irr}{0}%
\defres{cases/alf/ref/s/3/llm}{1}%
\defres{cases/alf/ref/s/3/must}{0}%
\defres{cases/alf/ref/s/3/nl}{0}%
\defres{cases/alf/ref/s/3/nm}{0}%
\defres{cases/alf/ref/s/3/obs}{On the desk 1, you see a alarmclock 2, a book 1, a bowl 1, a cd 2, a desklamp 1, a mug 2, a mug 1, a pen 2, and a…}%
\defres{cases/alf/ref/s/3/obsk}{env}%
\defres{cases/alf/ref/s/3/obsn}{2}%
\defres{cases/alf/ref/s/3/ok}{}%
\defres{cases/alf/ref/s/3/tool}{examine desk 1}%
\defres{cases/alf/ref/s/4/alt}{}%
\defres{cases/alf/ref/s/4/altp}{}%
\defres{cases/alf/ref/s/4/by}{llm}%
\defres{cases/alf/ref/s/4/cnum}{-1}%
\defres{cases/alf/ref/s/4/conf}{}%
\defres{cases/alf/ref/s/4/irr}{0}%
\defres{cases/alf/ref/s/4/llm}{1}%
\defres{cases/alf/ref/s/4/must}{0}%
\defres{cases/alf/ref/s/4/nl}{0}%
\defres{cases/alf/ref/s/4/nm}{0}%
\defres{cases/alf/ref/s/4/obs}{You pick up the alarmclock 2 from the desk 1.}%
\defres{cases/alf/ref/s/4/obsk}{env}%
\defres{cases/alf/ref/s/4/obsn}{1}%
\defres{cases/alf/ref/s/4/ok}{}%
\defres{cases/alf/ref/s/4/tool}{take alarmclock 2 from desk 1}%
\defres{cases/alf/ref/steps}{4}%
\defres{cases/alf/ref/success}{1}%
\defres{cases/alf/ref/wall}{24.1}%
\defres{cases/alf/row/1/h}{17.00}%
\defres{cases/alf/row/1/obs}{2}%
\defres{cases/alf/row/1/y}{-0.00}%
\defres{cases/alf/row/2/h}{13.40}%
\defres{cases/alf/row/2/obs}{1}%
\defres{cases/alf/row/2/y}{-17.00}%
\defres{cases/alf/row/3/h}{17.00}%
\defres{cases/alf/row/3/obs}{2}%
\defres{cases/alf/row/3/y}{-30.40}%
\defres{cases/alf/row/4/h}{13.40}%
\defres{cases/alf/row/4/obs}{1}%
\defres{cases/alf/row/4/y}{-47.40}%
\defres{cases/alf/run/api}{0.0000}%
\defres{cases/alf/run/cost}{0.0021}%
\defres{cases/alf/run/count/args}{0}%
\defres{cases/alf/run/count/args_deferred}{0}%
\defres{cases/alf/run/count/args_dm}{0}%
\defres{cases/alf/run/count/free}{0}%
\defres{cases/alf/run/count/tools}{4}%
\defres{cases/alf/run/count/tools_dm}{4}%
\defres{cases/alf/run/gpu}{0.0021}%
\defres{cases/alf/run/llm_calls}{0}%
\defres{cases/alf/run/n}{4}%
\defres{cases/alf/run/s/1/alt}{go to sidetable 2}%
\defres{cases/alf/run/s/1/altp}{0.29}%
\defres{cases/alf/run/s/1/by}{dm}%
\defres{cases/alf/run/s/1/cnum}{0.3536}%
\defres{cases/alf/run/s/1/conf}{0.35}%
\defres{cases/alf/run/s/1/irr}{0}%
\defres{cases/alf/run/s/1/llm}{0}%
\defres{cases/alf/run/s/1/must}{0}%
\defres{cases/alf/run/s/1/nl}{0}%
\defres{cases/alf/run/s/1/nm}{0}%
\defres{cases/alf/run/s/1/obs}{You arrive at sidetable 1. On the sidetable 1, you see a alarmclock 3, a cd 3, a cellphone 2, and a pen 3.}%
\defres{cases/alf/run/s/1/obsk}{env}%
\defres{cases/alf/run/s/1/obsn}{2}%
\defres{cases/alf/run/s/1/ok}{}%
\defres{cases/alf/run/s/1/tool}{go to sidetable 1}%
\defres{cases/alf/run/s/2/alt}{look}%
\defres{cases/alf/run/s/2/altp}{$<$0.01}%
\defres{cases/alf/run/s/2/by}{dm}%
\defres{cases/alf/run/s/2/cnum}{0.9857}%
\defres{cases/alf/run/s/2/conf}{0.99}%
\defres{cases/alf/run/s/2/irr}{0}%
\defres{cases/alf/run/s/2/llm}{0}%
\defres{cases/alf/run/s/2/must}{0}%
\defres{cases/alf/run/s/2/nl}{0}%
\defres{cases/alf/run/s/2/nm}{0}%
\defres{cases/alf/run/s/2/obs}{You pick up the alarmclock 3 from the sidetable 1.}%
\defres{cases/alf/run/s/2/obsk}{env}%
\defres{cases/alf/run/s/2/obsn}{1}%
\defres{cases/alf/run/s/2/ok}{}%
\defres{cases/alf/run/s/2/tool}{take alarmclock 3 from sidetable 1}%
\defres{cases/alf/run/s/3/alt}{go to safe 1}%
\defres{cases/alf/run/s/3/altp}{0.23}%
\defres{cases/alf/run/s/3/by}{dm}%
\defres{cases/alf/run/s/3/cnum}{0.4152}%
\defres{cases/alf/run/s/3/conf}{0.42}%
\defres{cases/alf/run/s/3/irr}{0}%
\defres{cases/alf/run/s/3/llm}{0}%
\defres{cases/alf/run/s/3/must}{0}%
\defres{cases/alf/run/s/3/nl}{0}%
\defres{cases/alf/run/s/3/nm}{0}%
\defres{cases/alf/run/s/3/obs}{You arrive at desk 1. On the desk 1, you see a alarmclock 2, a book 1, a bowl 1, a cd 2, a desklamp 1, a mug 2, a…}%
\defres{cases/alf/run/s/3/obsk}{env}%
\defres{cases/alf/run/s/3/obsn}{2}%
\defres{cases/alf/run/s/3/ok}{}%
\defres{cases/alf/run/s/3/tool}{go to desk 1}%
\defres{cases/alf/run/s/4/alt}{go to sidetable 1}%
\defres{cases/alf/run/s/4/altp}{0.01}%
\defres{cases/alf/run/s/4/by}{dm}%
\defres{cases/alf/run/s/4/cnum}{0.9769}%
\defres{cases/alf/run/s/4/conf}{0.98}%
\defres{cases/alf/run/s/4/irr}{0}%
\defres{cases/alf/run/s/4/llm}{0}%
\defres{cases/alf/run/s/4/must}{0}%
\defres{cases/alf/run/s/4/nl}{0}%
\defres{cases/alf/run/s/4/nm}{0}%
\defres{cases/alf/run/s/4/obs}{You turn on the desklamp 1.}%
\defres{cases/alf/run/s/4/obsk}{env}%
\defres{cases/alf/run/s/4/obsn}{1}%
\defres{cases/alf/run/s/4/ok}{}%
\defres{cases/alf/run/s/4/tool}{use desklamp 1}%
\defres{cases/alf/run/steps}{4}%
\defres{cases/alf/run/success}{1}%
\defres{cases/alf/run/wall}{14.2}%
\defres{cases/alf/task}{look at alarmclock under the desklamp.}%
\defres{cases/alf/tau}{0.12}%
\defres{cases/alf/tau_num}{0.1208}%
\defres{cases/ask/first_wrong}{3}%
\defres{cases/ask/first_wrong_conf}{0.98}%
\defres{cases/ask/height}{67.00}%
\defres{cases/ask/jev/api}{0.0034}%
\defres{cases/ask/jev/cost}{0.0034}%
\defres{cases/ask/jev/count/args}{6}%
\defres{cases/ask/jev/count/args_deferred}{0}%
\defres{cases/ask/jev/count/args_dm}{0}%
\defres{cases/ask/jev/count/free}{0}%
\defres{cases/ask/jev/count/tools}{3}%
\defres{cases/ask/jev/count/tools_dm}{3}%
\defres{cases/ask/jev/gpu}{0.0000}%
\defres{cases/ask/jev/llm_calls}{3}%
\defres{cases/ask/jev/n}{3}%
\defres{cases/ask/jev/s/1/alt}{ask\_user}%
\defres{cases/ask/jev/s/1/altp}{0.41}%
\defres{cases/ask/jev/s/1/by}{dm}%
\defres{cases/ask/jev/s/1/cnum}{0.5500}%
\defres{cases/ask/jev/s/1/conf}{0.55}%
\defres{cases/ask/jev/s/1/irr}{0}%
\defres{cases/ask/jev/s/1/l/1/c}{}%
\defres{cases/ask/jev/s/1/l/1/free}{0}%
\defres{cases/ask/jev/s/1/l/1/k}{name}%
\defres{cases/ask/jev/s/1/l/1/v}{Mateo Smith}%
\defres{cases/ask/jev/s/1/llm}{1}%
\defres{cases/ask/jev/s/1/must}{0}%
\defres{cases/ask/jev/s/1/nl}{1}%
\defres{cases/ask/jev/s/1/nm}{0}%
\defres{cases/ask/jev/s/1/obs}{}%
\defres{cases/ask/jev/s/1/obsk}{}%
\defres{cases/ask/jev/s/1/obsn}{0}%
\defres{cases/ask/jev/s/1/ok}{1}%
\defres{cases/ask/jev/s/1/tool}{find\_user}%
\defres{cases/ask/jev/s/2/alt}{notify\_warehouse}%
\defres{cases/ask/jev/s/2/altp}{$<$0.01}%
\defres{cases/ask/jev/s/2/by}{dm}%
\defres{cases/ask/jev/s/2/cnum}{0.9900}%
\defres{cases/ask/jev/s/2/conf}{0.99}%
\defres{cases/ask/jev/s/2/irr}{0}%
\defres{cases/ask/jev/s/2/l/1/c}{}%
\defres{cases/ask/jev/s/2/l/1/free}{0}%
\defres{cases/ask/jev/s/2/l/1/k}{user\_id}%
\defres{cases/ask/jev/s/2/l/1/v}{U4937}%
\defres{cases/ask/jev/s/2/l/2/c}{}%
\defres{cases/ask/jev/s/2/l/2/free}{0}%
\defres{cases/ask/jev/s/2/l/2/k}{status}%
\defres{cases/ask/jev/s/2/l/2/v}{delivered}%
\defres{cases/ask/jev/s/2/llm}{1}%
\defres{cases/ask/jev/s/2/must}{0}%
\defres{cases/ask/jev/s/2/nl}{2}%
\defres{cases/ask/jev/s/2/nm}{0}%
\defres{cases/ask/jev/s/2/obs}{}%
\defres{cases/ask/jev/s/2/obsk}{}%
\defres{cases/ask/jev/s/2/obsn}{0}%
\defres{cases/ask/jev/s/2/ok}{1}%
\defres{cases/ask/jev/s/2/tool}{get\_orders}%
\defres{cases/ask/jev/s/3/alt}{get\_order}%
\defres{cases/ask/jev/s/3/altp}{0.01}%
\defres{cases/ask/jev/s/3/by}{dm}%
\defres{cases/ask/jev/s/3/cnum}{0.9800}%
\defres{cases/ask/jev/s/3/conf}{0.98}%
\defres{cases/ask/jev/s/3/irr}{1}%
\defres{cases/ask/jev/s/3/l/1/c}{}%
\defres{cases/ask/jev/s/3/l/1/free}{0}%
\defres{cases/ask/jev/s/3/l/1/k}{order\_id}%
\defres{cases/ask/jev/s/3/l/1/v}{ORD-3726}%
\defres{cases/ask/jev/s/3/l/2/c}{}%
\defres{cases/ask/jev/s/3/l/2/free}{0}%
\defres{cases/ask/jev/s/3/l/2/k}{method}%
\defres{cases/ask/jev/s/3/l/2/v}{original\_payment}%
\defres{cases/ask/jev/s/3/l/3/c}{}%
\defres{cases/ask/jev/s/3/l/3/free}{0}%
\defres{cases/ask/jev/s/3/l/3/k}{amount}%
\defres{cases/ask/jev/s/3/l/3/v}{291.29}%
\defres{cases/ask/jev/s/3/llm}{1}%
\defres{cases/ask/jev/s/3/must}{1}%
\defres{cases/ask/jev/s/3/nl}{3}%
\defres{cases/ask/jev/s/3/nm}{0}%
\defres{cases/ask/jev/s/3/obs}{Irreversible error: refund(…) was applied incorrectly.}%
\defres{cases/ask/jev/s/3/obsk}{err}%
\defres{cases/ask/jev/s/3/obsn}{1}%
\defres{cases/ask/jev/s/3/ok}{0}%
\defres{cases/ask/jev/s/3/tool}{refund}%
\defres{cases/ask/jev/steps}{3}%
\defres{cases/ask/jev/success}{0}%
\defres{cases/ask/jev/wall}{2.5}%
\defres{cases/ask/n_rows}{5}%
\defres{cases/ask/n_tasks}{200}%
\defres{cases/ask/ours/api}{0.0061}%
\defres{cases/ask/ours/cost}{0.0061}%
\defres{cases/ask/ours/count/args}{7}%
\defres{cases/ask/ours/count/args_deferred}{0}%
\defres{cases/ask/ours/count/args_dm}{0}%
\defres{cases/ask/ours/count/free}{2}%
\defres{cases/ask/ours/count/tools}{5}%
\defres{cases/ask/ours/count/tools_dm}{5}%
\defres{cases/ask/ours/gpu}{0.0000}%
\defres{cases/ask/ours/llm_calls}{5}%
\defres{cases/ask/ours/n}{5}%
\defres{cases/ask/ours/s/1/alt}{find\_user}%
\defres{cases/ask/ours/s/1/altp}{0.10}%
\defres{cases/ask/ours/s/1/by}{dm}%
\defres{cases/ask/ours/s/1/cnum}{0.8925}%
\defres{cases/ask/ours/s/1/conf}{0.89}%
\defres{cases/ask/ours/s/1/irr}{0}%
\defres{cases/ask/ours/s/1/l/1/c}{}%
\defres{cases/ask/ours/s/1/l/1/free}{1}%
\defres{cases/ask/ours/s/1/l/1/k}{question}%
\defres{cases/ask/ours/s/1/l/1/v}{Would you like…}%
\defres{cases/ask/ours/s/1/l/2/c}{}%
\defres{cases/ask/ours/s/1/l/2/free}{0}%
\defres{cases/ask/ours/s/1/l/2/k}{missing}%
\defres{cases/ask/ours/s/1/l/2/v}{refund\_method}%
\defres{cases/ask/ours/s/1/llm}{1}%
\defres{cases/ask/ours/s/1/must}{0}%
\defres{cases/ask/ours/s/1/nl}{2}%
\defres{cases/ask/ours/s/1/nm}{0}%
\defres{cases/ask/ours/s/1/obs}{Please refund it to my original payment method.}%
\defres{cases/ask/ours/s/1/obsk}{user}%
\defres{cases/ask/ours/s/1/obsn}{1}%
\defres{cases/ask/ours/s/1/ok}{1}%
\defres{cases/ask/ours/s/1/tool}{ask\_user}%
\defres{cases/ask/ours/s/2/alt}{get\_orders}%
\defres{cases/ask/ours/s/2/altp}{0.15}%
\defres{cases/ask/ours/s/2/by}{dm}%
\defres{cases/ask/ours/s/2/cnum}{0.7950}%
\defres{cases/ask/ours/s/2/conf}{0.80}%
\defres{cases/ask/ours/s/2/irr}{0}%
\defres{cases/ask/ours/s/2/l/1/c}{}%
\defres{cases/ask/ours/s/2/l/1/free}{0}%
\defres{cases/ask/ours/s/2/l/1/k}{name}%
\defres{cases/ask/ours/s/2/l/1/v}{Mateo Smith}%
\defres{cases/ask/ours/s/2/llm}{1}%
\defres{cases/ask/ours/s/2/must}{0}%
\defres{cases/ask/ours/s/2/nl}{1}%
\defres{cases/ask/ours/s/2/nm}{0}%
\defres{cases/ask/ours/s/2/obs}{}%
\defres{cases/ask/ours/s/2/obsk}{}%
\defres{cases/ask/ours/s/2/obsn}{0}%
\defres{cases/ask/ours/s/2/ok}{1}%
\defres{cases/ask/ours/s/2/tool}{find\_user}%
\defres{cases/ask/ours/s/3/alt}{refund}%
\defres{cases/ask/ours/s/3/altp}{0.01}%
\defres{cases/ask/ours/s/3/by}{dm}%
\defres{cases/ask/ours/s/3/cnum}{0.9874}%
\defres{cases/ask/ours/s/3/conf}{0.99}%
\defres{cases/ask/ours/s/3/irr}{0}%
\defres{cases/ask/ours/s/3/l/1/c}{}%
\defres{cases/ask/ours/s/3/l/1/free}{0}%
\defres{cases/ask/ours/s/3/l/1/k}{user\_id}%
\defres{cases/ask/ours/s/3/l/1/v}{U4937}%
\defres{cases/ask/ours/s/3/l/2/c}{}%
\defres{cases/ask/ours/s/3/l/2/free}{0}%
\defres{cases/ask/ours/s/3/l/2/k}{status}%
\defres{cases/ask/ours/s/3/l/2/v}{delivered}%
\defres{cases/ask/ours/s/3/llm}{1}%
\defres{cases/ask/ours/s/3/must}{0}%
\defres{cases/ask/ours/s/3/nl}{2}%
\defres{cases/ask/ours/s/3/nm}{0}%
\defres{cases/ask/ours/s/3/obs}{}%
\defres{cases/ask/ours/s/3/obsk}{}%
\defres{cases/ask/ours/s/3/obsn}{0}%
\defres{cases/ask/ours/s/3/ok}{1}%
\defres{cases/ask/ours/s/3/tool}{get\_orders}%
\defres{cases/ask/ours/s/4/alt}{get\_orders}%
\defres{cases/ask/ours/s/4/altp}{$<$0.01}%
\defres{cases/ask/ours/s/4/by}{dm}%
\defres{cases/ask/ours/s/4/cnum}{0.9911}%
\defres{cases/ask/ours/s/4/conf}{0.99}%
\defres{cases/ask/ours/s/4/irr}{1}%
\defres{cases/ask/ours/s/4/l/1/c}{}%
\defres{cases/ask/ours/s/4/l/1/free}{0}%
\defres{cases/ask/ours/s/4/l/1/k}{order\_id}%
\defres{cases/ask/ours/s/4/l/1/v}{ORD-3726}%
\defres{cases/ask/ours/s/4/l/2/c}{}%
\defres{cases/ask/ours/s/4/l/2/free}{0}%
\defres{cases/ask/ours/s/4/l/2/k}{amount}%
\defres{cases/ask/ours/s/4/l/2/v}{291.29}%
\defres{cases/ask/ours/s/4/l/3/c}{}%
\defres{cases/ask/ours/s/4/l/3/free}{0}%
\defres{cases/ask/ours/s/4/l/3/k}{method}%
\defres{cases/ask/ours/s/4/l/3/v}{original\_payment}%
\defres{cases/ask/ours/s/4/llm}{1}%
\defres{cases/ask/ours/s/4/must}{0}%
\defres{cases/ask/ours/s/4/nl}{3}%
\defres{cases/ask/ours/s/4/nm}{0}%
\defres{cases/ask/ours/s/4/obs}{Refunded 291.29 via original\_payment.}%
\defres{cases/ask/ours/s/4/obsk}{env}%
\defres{cases/ask/ours/s/4/obsn}{1}%
\defres{cases/ask/ours/s/4/ok}{1}%
\defres{cases/ask/ours/s/4/tool}{refund}%
\defres{cases/ask/ours/s/5/alt}{send\_message}%
\defres{cases/ask/ours/s/5/altp}{0.01}%
\defres{cases/ask/ours/s/5/by}{dm}%
\defres{cases/ask/ours/s/5/cnum}{0.9915}%
\defres{cases/ask/ours/s/5/conf}{0.99}%
\defres{cases/ask/ours/s/5/irr}{0}%
\defres{cases/ask/ours/s/5/l/1/c}{}%
\defres{cases/ask/ours/s/5/l/1/free}{1}%
\defres{cases/ask/ours/s/5/l/1/k}{answer}%
\defres{cases/ask/ours/s/5/l/1/v}{I have…}%
\defres{cases/ask/ours/s/5/llm}{1}%
\defres{cases/ask/ours/s/5/must}{0}%
\defres{cases/ask/ours/s/5/nl}{1}%
\defres{cases/ask/ours/s/5/nm}{0}%
\defres{cases/ask/ours/s/5/obs}{}%
\defres{cases/ask/ours/s/5/obsk}{}%
\defres{cases/ask/ours/s/5/obsn}{0}%
\defres{cases/ask/ours/s/5/ok}{1}%
\defres{cases/ask/ours/s/5/tool}{finish}%
\defres{cases/ask/ours/steps}{5}%
\defres{cases/ask/ours/success}{1}%
\defres{cases/ask/ours/wall}{45.0}%
\defres{cases/ask/rank}{27}%
\defres{cases/ask/ref/api}{0.0304}%
\defres{cases/ask/ref/cost}{0.0304}%
\defres{cases/ask/ref/first_ask}{3}%
\defres{cases/ask/ref/gpu}{0.0000}%
\defres{cases/ask/ref/llm_calls}{5}%
\defres{cases/ask/ref/steps}{5}%
\defres{cases/ask/ref/success}{1}%
\defres{cases/ask/ref/wall}{3.5}%
\defres{cases/ask/row/1/h}{13.40}%
\defres{cases/ask/row/1/obs}{1}%
\defres{cases/ask/row/1/y}{-0.00}%
\defres{cases/ask/row/2/h}{9.80}%
\defres{cases/ask/row/2/obs}{0}%
\defres{cases/ask/row/2/y}{-13.40}%
\defres{cases/ask/row/3/h}{17.00}%
\defres{cases/ask/row/3/obs}{1}%
\defres{cases/ask/row/3/y}{-23.20}%
\defres{cases/ask/row/4/h}{17.00}%
\defres{cases/ask/row/4/obs}{1}%
\defres{cases/ask/row/4/y}{-40.20}%
\defres{cases/ask/row/5/h}{9.80}%
\defres{cases/ask/row/5/obs}{0}%
\defres{cases/ask/row/5/y}{-57.20}%
\defres{cases/ask/seed}{300026}%
\defres{cases/ask/task}{Refund the full amount of Mateo Smith's most recent delivered order (highest order id).}%
\defres{cases/take/height}{59.80}%
\defres{cases/take/n_rows}{5}%
\defres{cases/take/n_tasks}{200}%
\defres{cases/take/rank}{1}%
\defres{cases/take/ref/api}{0.0301}%
\defres{cases/take/ref/cost}{0.0301}%
\defres{cases/take/ref/gpu}{0.0000}%
\defres{cases/take/ref/llm_calls}{5}%
\defres{cases/take/ref/steps}{5}%
\defres{cases/take/ref/success}{1}%
\defres{cases/take/ref/wall}{4.4}%
\defres{cases/take/row/1/h}{13.40}%
\defres{cases/take/row/1/obs}{1}%
\defres{cases/take/row/1/y}{-0.00}%
\defres{cases/take/row/2/h}{9.80}%
\defres{cases/take/row/2/obs}{0}%
\defres{cases/take/row/2/y}{-13.40}%
\defres{cases/take/row/3/h}{13.40}%
\defres{cases/take/row/3/obs}{0}%
\defres{cases/take/row/3/y}{-23.20}%
\defres{cases/take/row/4/h}{17.00}%
\defres{cases/take/row/4/obs}{1}%
\defres{cases/take/row/4/y}{-36.60}%
\defres{cases/take/row/5/h}{6.20}%
\defres{cases/take/row/5/obs}{0}%
\defres{cases/take/row/5/y}{-53.60}%
\defres{cases/take/run/api}{0.0025}%
\defres{cases/take/run/cost}{0.0042}%
\defres{cases/take/run/count/args}{6}%
\defres{cases/take/run/count/args_deferred}{1}%
\defres{cases/take/run/count/args_dm}{5}%
\defres{cases/take/run/count/free}{2}%
\defres{cases/take/run/count/tools}{5}%
\defres{cases/take/run/count/tools_dm}{5}%
\defres{cases/take/run/gpu}{0.0017}%
\defres{cases/take/run/llm_calls}{2}%
\defres{cases/take/run/n}{5}%
\defres{cases/take/run/s/1/alt}{find\_user}%
\defres{cases/take/run/s/1/altp}{0.03}%
\defres{cases/take/run/s/1/by}{dm}%
\defres{cases/take/run/s/1/cnum}{0.9600}%
\defres{cases/take/run/s/1/conf}{0.96}%
\defres{cases/take/run/s/1/irr}{0}%
\defres{cases/take/run/s/1/l/1/c}{0.993}%
\defres{cases/take/run/s/1/l/1/free}{0}%
\defres{cases/take/run/s/1/l/1/k}{missing}%
\defres{cases/take/run/s/1/l/1/v}{shipping\_speed}%
\defres{cases/take/run/s/1/l/2/c}{}%
\defres{cases/take/run/s/1/l/2/free}{1}%
\defres{cases/take/run/s/1/l/2/k}{question}%
\defres{cases/take/run/s/1/l/2/v}{Could you please specify the…}%
\defres{cases/take/run/s/1/llm}{1}%
\defres{cases/take/run/s/1/must}{0}%
\defres{cases/take/run/s/1/nl}{2}%
\defres{cases/take/run/s/1/nm}{0}%
\defres{cases/take/run/s/1/obs}{Express shipping, please.}%
\defres{cases/take/run/s/1/obsk}{user}%
\defres{cases/take/run/s/1/obsn}{1}%
\defres{cases/take/run/s/1/ok}{1}%
\defres{cases/take/run/s/1/tool}{ask\_user}%
\defres{cases/take/run/s/2/alt}{ship\_order}%
\defres{cases/take/run/s/2/altp}{0.12}%
\defres{cases/take/run/s/2/by}{dm}%
\defres{cases/take/run/s/2/cnum}{0.8406}%
\defres{cases/take/run/s/2/conf}{0.84}%
\defres{cases/take/run/s/2/irr}{0}%
\defres{cases/take/run/s/2/llm}{0}%
\defres{cases/take/run/s/2/m/1/c}{1.000}%
\defres{cases/take/run/s/2/m/1/k}{name}%
\defres{cases/take/run/s/2/m/1/v}{Alice Smith}%
\defres{cases/take/run/s/2/must}{0}%
\defres{cases/take/run/s/2/nl}{0}%
\defres{cases/take/run/s/2/nm}{1}%
\defres{cases/take/run/s/2/obs}{}%
\defres{cases/take/run/s/2/obsk}{}%
\defres{cases/take/run/s/2/obsn}{0}%
\defres{cases/take/run/s/2/ok}{1}%
\defres{cases/take/run/s/2/tool}{find\_user}%
\defres{cases/take/run/s/3/alt}{ship\_order}%
\defres{cases/take/run/s/3/altp}{0.05}%
\defres{cases/take/run/s/3/by}{dm}%
\defres{cases/take/run/s/3/cnum}{0.9414}%
\defres{cases/take/run/s/3/conf}{0.94}%
\defres{cases/take/run/s/3/irr}{0}%
\defres{cases/take/run/s/3/llm}{0}%
\defres{cases/take/run/s/3/m/1/c}{1.000}%
\defres{cases/take/run/s/3/m/1/k}{user\_id}%
\defres{cases/take/run/s/3/m/1/v}{U4140}%
\defres{cases/take/run/s/3/m/2/c}{1.000}%
\defres{cases/take/run/s/3/m/2/k}{status}%
\defres{cases/take/run/s/3/m/2/v}{pending}%
\defres{cases/take/run/s/3/must}{0}%
\defres{cases/take/run/s/3/nl}{0}%
\defres{cases/take/run/s/3/nm}{2}%
\defres{cases/take/run/s/3/obs}{}%
\defres{cases/take/run/s/3/obsk}{}%
\defres{cases/take/run/s/3/obsn}{0}%
\defres{cases/take/run/s/3/ok}{1}%
\defres{cases/take/run/s/3/tool}{get\_orders}%
\defres{cases/take/run/s/4/alt}{get\_shipping\_quote}%
\defres{cases/take/run/s/4/altp}{0.03}%
\defres{cases/take/run/s/4/by}{dm}%
\defres{cases/take/run/s/4/cnum}{0.9719}%
\defres{cases/take/run/s/4/conf}{0.97}%
\defres{cases/take/run/s/4/irr}{1}%
\defres{cases/take/run/s/4/llm}{0}%
\defres{cases/take/run/s/4/m/1/c}{1.000}%
\defres{cases/take/run/s/4/m/1/k}{order\_id}%
\defres{cases/take/run/s/4/m/1/v}{ORD-3381}%
\defres{cases/take/run/s/4/m/2/c}{1.000}%
\defres{cases/take/run/s/4/m/2/k}{speed}%
\defres{cases/take/run/s/4/m/2/v}{express}%
\defres{cases/take/run/s/4/must}{0}%
\defres{cases/take/run/s/4/nl}{0}%
\defres{cases/take/run/s/4/nm}{2}%
\defres{cases/take/run/s/4/obs}{Order ORD-3381 shipped (express).}%
\defres{cases/take/run/s/4/obsk}{env}%
\defres{cases/take/run/s/4/obsn}{1}%
\defres{cases/take/run/s/4/ok}{1}%
\defres{cases/take/run/s/4/tool}{ship\_order}%
\defres{cases/take/run/s/5/alt}{send\_message}%
\defres{cases/take/run/s/5/altp}{0.01}%
\defres{cases/take/run/s/5/by}{dm}%
\defres{cases/take/run/s/5/cnum}{0.9911}%
\defres{cases/take/run/s/5/conf}{0.99}%
\defres{cases/take/run/s/5/irr}{0}%
\defres{cases/take/run/s/5/l/1/c}{}%
\defres{cases/take/run/s/5/l/1/free}{1}%
\defres{cases/take/run/s/5/l/1/k}{answer}%
\defres{cases/take/run/s/5/l/1/v}{Alice Smith's pending order…}%
\defres{cases/take/run/s/5/llm}{1}%
\defres{cases/take/run/s/5/must}{0}%
\defres{cases/take/run/s/5/nl}{1}%
\defres{cases/take/run/s/5/nm}{0}%
\defres{cases/take/run/s/5/obs}{}%
\defres{cases/take/run/s/5/obsk}{}%
\defres{cases/take/run/s/5/obsn}{0}%
\defres{cases/take/run/s/5/ok}{1}%
\defres{cases/take/run/s/5/tool}{finish}%
\defres{cases/take/run/steps}{5}%
\defres{cases/take/run/success}{1}%
\defres{cases/take/run/wall}{3.7}%
\defres{cases/take/saved_calls}{3}%
\defres{cases/take/seed}{300500}%
\defres{cases/take/tariff}{1.624}%
\defres{cases/take/task}{Ship Alice Smith's pending order.}%
\defres{cases/take/tau_arg}{0.996}%
\defres{cases/take/tau_tool}{0.45}%
\defres{cases/take/tau_tool_num}{0.4450}%
\definecolor{caseRed}{HTML}{B5483B}%
\definecolor{caseRedT}{HTML}{F8E6E2}%
\definecolor{caseOk}{HTML}{1B7A4C}%
\tikzset{
  csbox/.style={rounded corners=0.8pt, minimum height=3.9mm, inner xsep=0.9mm, inner ysep=0pt, anchor=west,
    font=\ttfamily\figfs, text=white, align=left},
  cschip/.style={rounded corners=0.8pt, minimum height=3.0mm, inner xsep=0.8mm, inner ysep=0pt, anchor=west,
    font=\figfs\sffamily, line width=0.35pt},
  cstxt/.style={anchor=west, font=\figfs\sffamily},
  csobs/.style={anchor=north west, font=\figfs\sffamily\itshape, align=left, fill=white, rounded corners=0.8pt,
    line width=0.3pt, inner xsep=0.8mm, inner ysep=0.45mm},
  cslane/.style={rounded corners=1.5pt, draw=none},
}%
\def\csPitch{3.6}\def\csTool{4.6}\def\csPad{1.6}%
\def\csC#1{\res{cases/\csCase/#1}}%
\def\csR#1{\res{cases/\csCase/\csRun/#1}}%
\def\csLineY#1{\pgfmathsetmacro\csY{\csTop-\csTool-(#1-2)*\csPitch-\csPitch/2}}%

\newcommand{\csBar}[6]{%
  \fill[envT] (#1,#2-0.8) rectangle ++(#3,1.6);
  \pgfmathsetmacro\csBw{#3*#4}%
  \fill[#6] (#1,#2-0.8) rectangle ++(\csBw,1.6);
  \draw[envC!55, line width=0.25pt] (#1,#2-0.8) rectangle ++(#3,1.6);
  \pgfmathparse{#5>-0.5 ? 1 : 0}%
  \ifnum\pgfmathresult=1 \draw[white, line width=1.3pt] ({#1+#3*#5},#2-0.8) -- ({#1+#3*#5},#2+0.8);
    \draw[ink, line width=0.5pt] ({#1+#3*#5},#2-1.45) -- ({#1+#3*#5},#2+1.45);\fi}%

\newcommand{\csDrawRun}{%
  \pgfmathtruncatemacro\csN{\csR{n}}%
  \pgfmathsetmacro\csXl{\csX+\csWd+(\csWd>0 ? 1 : 0)}%
  \pgfmathsetmacro\csWb{\csXl+\csWl-\csX}%
  \foreach \i in {1,...,\csN} {%
    \pgfmathsetmacro\csTop{\csC{row/\i/y}}%
    \pgfmathsetmacro\csH{\csC{row/\i/h}}%
    \pgfmathsetmacro\csYa{\csTop-\csTool/2}%
    \ifnum\pdfstrcmp{\csR{s/\i/ok}}{0}=0 \def\csCol{caseRed}%
      \fill[caseRedT, rounded corners=1pt] (\csX,\csTop+0.55) rectangle ++(\csWb,-\csH+0.5);%
    \else\def\csCol{sOne}\fi
    \ifnum\pdfstrcmp{\csR{s/\i/irr}}{1}=0 \def\csIrr{\hfill\faExclamationTriangle}\else\def\csIrr{}\fi
    \ifnum\csBoxNat=1 \tikzset{csboxw/.style={minimum width=10mm}}%
    \else\tikzset{csboxw/.style={text width={(\csBoxW-1.8)*1mm}}}\fi
    \ifnum\pdfstrcmp{\csR{s/\i/by}}{dm}=0
      \node[csbox, csboxw, fill=\csCol] at (\csX+1,\csYa) {\csR{s/\i/tool}\csIrr};
      \pgfmathsetmacro\csBx{\csX+1+\csBoxW+1.6}%
      \csBar{\csBx}{\csYa}{\csBarL}{\csR{s/\i/cnum}}{\csTau}{\csCol}
      \node[cstxt, text=\csCol] at (\csBx+\csBarL+0.9,\csYa) {\csR{s/\i/conf}};
      \pgfmathtruncatemacro\csK{2}%
      \ifnum\pdfstrcmp{\csR{s/\i/alt}}{}=0 \else
        \ifnum\csAltOwn=1 \csLineY{\csK}\pgfmathtruncatemacro\csK{\csK+1}%
          \node[cstxt, text=ink2] at (\csX+1.3,\csY) {2nd: \texttt{\csR{s/\i/alt}} \csR{s/\i/altp}};
        \else
          \node[cstxt, text=ink2] at (\csBx+\csBarL+7.2,\csYa) {2nd: \texttt{\csR{s/\i/alt}} \csR{s/\i/altp}};
        \fi
      \fi
      \ifnum\pdfstrcmp{\csR{s/\i/must}}{1}=0 \csLineY{\csK}\pgfmathtruncatemacro\csK{\csK+1}%
        \node[cstxt, text=caseRed, font=\figfs\sffamily\bfseries] at (\csX+1.3,\csY)
          {only acceptable: \texttt{\mdseries ask\_user}};\fi
      \pgfmathtruncatemacro\csNm{\csR{s/\i/nm}}%
      \ifnum\csNm>0 \foreach \j in {1,...,\csNm} {%
        \csLineY{\csK+\j-1}%
        \node[cschip, fill=sOneT, draw=sOne!70] (csc) at (\csX+1,\csY)
          {\textcolor{ink2}{\csR{s/\i/m/\j/k}} \texttt{\csR{s/\i/m/\j/v}}};
        \node[cstxt, text=sOne] at ([xshift=0.8mm]csc.east) {\csR{s/\i/m/\j/c}};}\fi
      \pgfmathtruncatemacro\csKl{1}%
      \pgfmathtruncatemacro\csUsed{\csK-1+\csNm}%
    \else
      \node[csbox, csboxw, fill=sTwo] at (\csXl+1,\csYa) {\csR{s/\i/tool}\csIrr};
      \pgfmathtruncatemacro\csKl{2}%
      \pgfmathtruncatemacro\csUsed{1}%
    \fi
    \pgfmathtruncatemacro\csNl{\csR{s/\i/nl}}%
    \ifnum\csNl>0 \foreach \j in {1,...,\csNl} {%
      \pgfmathtruncatemacro\csKj{\csKl+\j-1}%
      \ifnum\csKj=1 \pgfmathsetmacro\csY{\csYa}\else\csLineY{\csKj}\fi
      \ifnum\pdfstrcmp{\csR{s/\i/l/\j/free}}{1}=0
        \node[cschip, fill=sTwoT, draw=sTwo!70] at (\csXl+1,\csY)
          {\textcolor{ink2}{\csR{s/\i/l/\j/k}} ``\csR{s/\i/l/\j/v}''};
      \else
        \node[cschip, fill=sTwoT, draw=sTwo!70] (csc) at (\csXl+1,\csY)
          {\textcolor{ink2}{\csR{s/\i/l/\j/k}} \texttt{\csR{s/\i/l/\j/v}}};
        \ifnum\pdfstrcmp{\csR{s/\i/l/\j/c}}{}=0 \else
          \node[cstxt, text=sTwo] at ([xshift=0.8mm]csc.east) {decision model \csR{s/\i/l/\j/c} $<\tau$};\fi
      \fi}\fi
    \pgfmathtruncatemacro\csUsed{max(\csUsed,\csKl-1+\csNl)}%
    \pgfmathtruncatemacro\csNoLlm{(\csWl>0) && (\csNl==0) && (\csR{s/\i/llm}==0) ? 1 : 0}%
    \ifnum\csNoLlm=1 \node[cstxt, text=sTwo!70!envC, font=\figfs\sffamily\itshape] at (\csXl+1.3,\csYa)
      {no LLM call};\fi
    \ifnum\pdfstrcmp{\csR{s/\i/obs}}{}=0 \else
      \def\csOc{ink2}\def\csOd{envC!45}\def\csOg{$\hookrightarrow$}%
      \ifnum\pdfstrcmp{\csR{s/\i/obsk}}{err}=0 \def\csOc{caseRed}\def\csOd{caseRed!55}\fi
      \ifnum\pdfstrcmp{\csR{s/\i/obsk}}{user}=0 \def\csOg{\textup{\faUser}}\fi
      \pgfmathsetmacro\csOy{\csTop-\csTool-(\csUsed-1)*\csPitch-0.2}%
      \ifnum\csR{s/\i/obsn}>1 \tikzset{csobsw/.style={text width={(\csWb-4.4)*1mm}}}%
      \else\tikzset{csobsw/.style={}}\fi
      \node[csobs, csobsw, text=\csOc, draw=\csOd] at (\csX+1,\csOy) {\csOg\,\csR{s/\i/obs}};\fi
  }%
  \pgfmathtruncatemacro\csNr{\csC{n_rows}}%
  \ifnum\csN<\csNr \pgfmathtruncatemacro\csNn{\csN+1}\pgfmathsetmacro\csYo{\csC{row/\csNn/y}-\csTool/2}%
  \else \pgfmathsetmacro\csYo{-\csC{height}-1.5}\fi
  \ifnum\pdfstrcmp{\csR{success}}{1}=0
    \node[cstxt, text=caseOk, font=\figfs\sffamily\bfseries] at (\csX+1.3,\csYo) {\faCheckCircle\ task solved};
  \else
    \node[cstxt, text=caseRed, font=\figfs\sffamily\bfseries] at (\csX+1.3,\csYo) {\faTimesCircle\ task failed};
  \fi}%

\newcommand{\csFrame}[3]{%
  \pgfmathsetmacro\csFb{-\csC{height}-3.6}%
  \ifdim #2pt>0pt \fill[sOneT!55, cslane] (#1,2.0) rectangle ++(#2,\csFb-2.0);\fi
  \pgfmathsetmacro\csFx{#1+#2+(#2>0 ? 1 : 0)}%
  \ifdim #3pt>0pt \fill[sTwoT!55, cslane] (\csFx,2.0) rectangle ++(#3,\csFb-2.0);\fi}%

\newcommand{\csSpine}{%
  \pgfmathtruncatemacro\csNr{\csC{n_rows}}%
  \foreach \i in {1,...,\csNr} {%
    \pgfmathsetmacro\csTop{\csC{row/\i/y}}%
    \node[circle, fill=envT, draw=envC!70, line width=0.35pt, minimum size=3.3mm, inner sep=0pt,
      font=\figfs\sffamily, text=ink2] at (3.4,\csTop-\csTool/2) {\i};
    \ifnum\i>1 \draw[envC!35, line width=0.3pt, dash pattern=on 0.6pt off 0.8pt] (7.5,\csTop+0.8) -- (160,\csTop+0.8);\fi}
  \node[cstxt, text=ink2, anchor=south] at (3.4,2.4) {step};}%

\newcommand{\csKeyStart}[2]{\coordinate (cskn) at (#1,#2);}%
\newcommand{\csKeyNext}[2]{\path (cskn.east) ++(3.0,0) coordinate (cskp);
  \begin{scope}[shift={(cskp)}] #1 \end{scope}
  \node[cstxt, text=ink2] (cskn) at ([xshift=1.9mm]cskp) {#2};}%
\def\csSwBox#1{\fill[#1, rounded corners=0.5pt] (-1.1,-0.9) rectangle (1.1,0.9);}%
\def\csSwConf{\fill[envT] (-1.5,-0.6) rectangle (1.5,0.6); \fill[sOne] (-1.5,-0.6) rectangle (0.6,0.6);}%
\def\csSwTau{\csSwConf \draw[white, line width=1.3pt] (0,-0.6) -- (0,0.6); \draw[ink, line width=0.5pt] (0,-1.3) -- (0,1.3);}%
\def\csSwIrr{\node[text=ink2, font=\figfs] at (0,0) {\faExclamationTriangle};}%
\def\csSwObs{\draw[envC!55, fill=white, line width=0.3pt, rounded corners=0.6pt] (-1.4,-0.95) rectangle (1.4,0.95);}%

\newcommand{\csFoot}[4]{\def\csFy{#1}%
  \fill[envT, rounded corners=1.5pt] (0,\csFy+1.9) rectangle (160,\csFy-8.1);
  \node[cstxt, text=ink, font=\figfs\sffamily\bfseries] at (1.2,\csFy) {#2};
  \node[cstxt, text=sOne] at (1.2,\csFy-3.3) {#3};
  \node[cstxt, text=sTwo] at (1.2,\csFy-6.0) {#4};}%
\newcommand{\csMini}[4]{%
  \node[cstxt, text=ink, font=\figfs\sffamily\bfseries] at (#1,\csFy) {#2};
  \pgfmathsetmacro\csMm{max(#3,#4)}%
  \pgfmathsetmacro\csMa{22*#3/\csMm}\pgfmathsetmacro\csMb{22*#4/\csMm}%
  \fill[sOne] (#1,\csFy-4.15) rectangle ++(\csMa,1.7);
  \node[cstxt, text=sOne] at (#1+\csMa+0.9,\csFy-3.3) {#3};
  \fill[sTwo] (#1,\csFy-6.85) rectangle ++(\csMb,1.7);
  \node[cstxt, text=sTwo] at (#1+\csMb+0.9,\csFy-6.0) {#4};}%

\section{Case Studies}
\label{app:cases}
The three figures below show recorded trajectories step by step, one row per step from top to bottom. Blue marks the
choices the decision model made and orange those the LLM made, so each figure shows who decided at every step. Each
case is the first task, in a fixed order, that meets the
rule stated in its caption.

\paragraph{Where the decision model makes the choice.}
On the \gym{} task of Figure~\ref{fig:case-take}, \ours{} chose all \res{cases/take/run/count/tools} tools and
\res{cases/take/run/count/args_dm} of the \res{cases/take/run/count/args} arguments that are not free text. The LLM
wrote the question to the customer and the final message, and it set the one argument whose confidence fell just
below its threshold. The task took \res{cases/take/run/llm_calls} LLM calls instead of \res{cases/take/ref/llm_calls}.

\paragraph{A confident irreversible choice.}
In Figure~\ref{fig:case-ask}, \jev{} and \ours{} each picked every tool of a task that does not say how the customer
wants to be refunded. \jev{} looked up the customer and the order and then chose the refund with confidence
\res{cases/ask/first_wrong_conf}, where asking was the only acceptable choice, and the LLM filled in a refund method
the customer had never given. \ours{} asked first and solved the task, as did Gemini 3.5 Flash alone, which asked at
step~\res{cases/ask/ref/first_ask}. Every failed task of \jev{} ends at its first wrong tool choice, an
irreversible action like this one (App.~\ref{app:drift}).

\paragraph{ALFWorld.}
In Figure~\ref{fig:case-alf}, the 4B model trained on ALFWorld states (App.~\ref{app:e11}) chose every action of a
household game, because its fitted threshold of \res{cases/alf/tau} let every confidence through. It solved the game
in \res{cases/alf/run/steps} steps without an LLM call, as many steps as Gemini 3.5 Flash alone took.

\begin{figure*}[!t]
\centering
\begin{tikzpicture}[jevfig]
\def\csCase{take}
\node[cstxt, text=ink] at (0,15.0) {\textbf{Task}\enspace\textit{\csC{task}}};
\csKeyStart{7.0}{10.4}
\csKeyNext{\csSwBox{sOne}}{decision model chose}
\csKeyNext{\csSwBox{sTwo}}{LLM chose or wrote}
\csKeyNext{\csSwTau}{confidence and threshold $\tau$}
\csKeyNext{\csSwIrr}{changes the world}
\csKeyNext{\csSwObs}{environment's reply}
\csFrame{8.5}{88.5}{62}
\node[cstxt] at (9.8,4.6) {\textcolor{sOne}{\faBolt\,\textbf{decision model}}\enspace\ours{}\enspace
  \textcolor{ink2}{$\tau$ = \csC{tau_tool} for tools, \csC{tau_arg} for arguments}};
\node[cstxt] at (99.8,4.6) {\textcolor{sTwo}{\faBrain\,\textbf{LLM}}\enspace Gemini 3.5 Flash};
\csSpine
\def\csRun{run}\def\csX{8.5}\def\csWd{88.5}\def\csWl{62}\def\csBoxW{21}\def\csBoxNat{0}\def\csBarL{20}
\def\csTau{\csC{tau_tool_num}}\def\csAltOwn{0}
\csDrawRun
\pgfmathsetmacro\csFyy{-\csC{height}-8.4}
\csFoot{\csFyy}{Same task}{with \ours{}}{Flash alone}
\csMini{36}{LLM calls}{\csC{run/llm_calls}}{\csC{ref/llm_calls}}
\csMini{78}{cost per task (\$, GPU included)}{\csC{run/cost}}{\csC{ref/cost}}
\csMini{120}{time (s)}{\csC{run/wall}}{\csC{ref/wall}}
\end{tikzpicture}
\caption{\textbf{\ours{} took every tool choice of this \gym{} task, and the LLM only wrote free text and set one
argument below its threshold.} Bars show the decision model's confidence in its choice against the fitted threshold
$\tau$, next to the runner-up option and its probability. Chips are arguments. The footer compares the same task with
Gemini 3.5 Flash alone, with the GPU charged at \$\res{cases/take/tariff} per hour. The task is the first of the \res{cases/take/n_tasks} new tasks, in seed order, that both runs solved and in which at least one choice was left to the LLM. Long text is abridged (\dots).}
\label{fig:case-take}
\end{figure*}
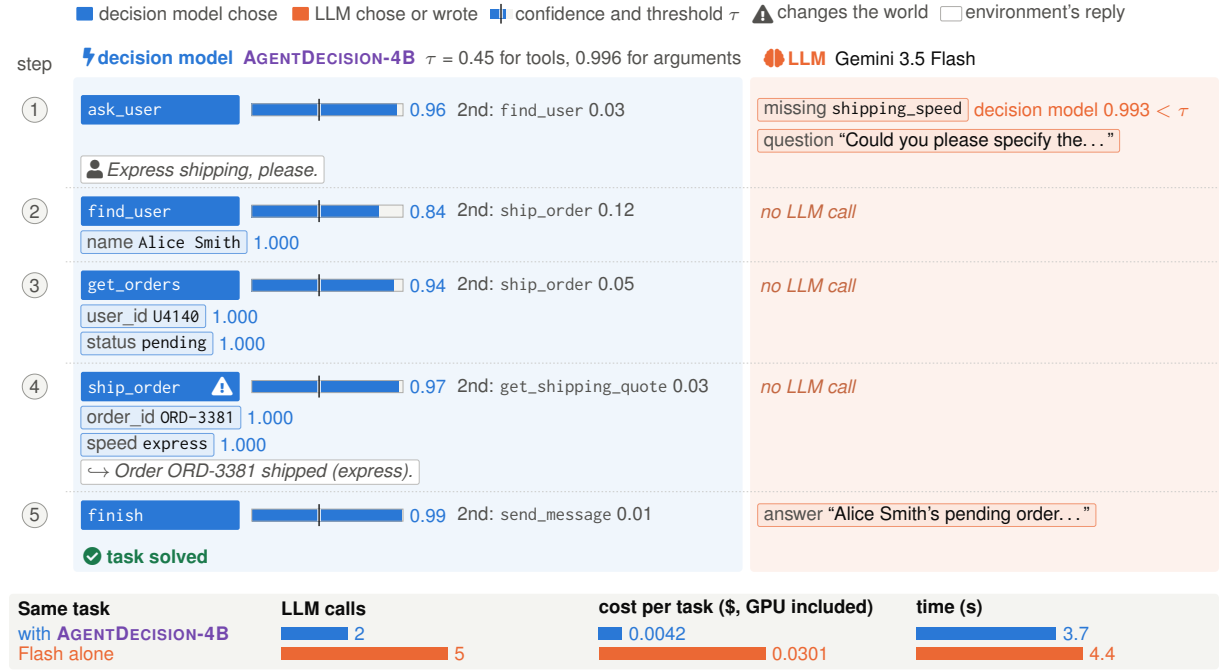

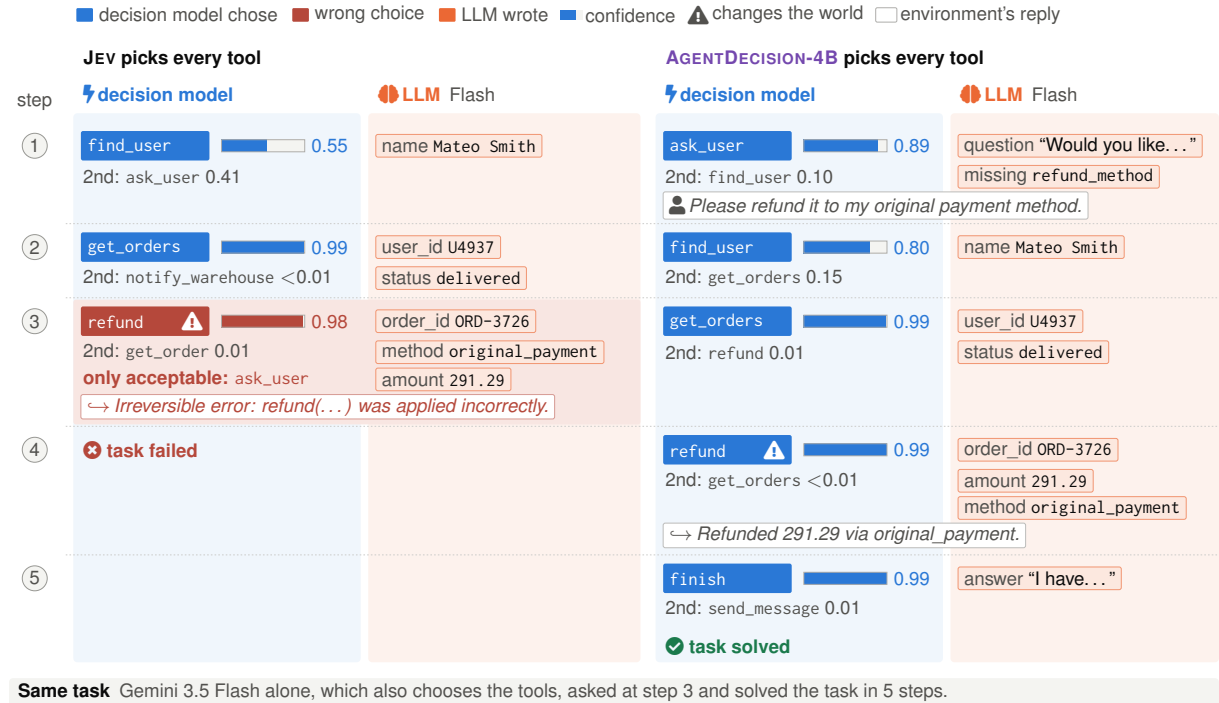
\begin{figure*}[!t]
\centering
\begin{tikzpicture}[jevfig]
\def\csCase{ask}
\node[cstxt, text=ink] at (0,19.6) {\textbf{Task}\enspace\textit{\csC{task}}};
\csKeyStart{7.0}{15.0}
\csKeyNext{\csSwBox{sOne}}{decision model chose}
\csKeyNext{\csSwBox{caseRed}}{wrong choice}
\csKeyNext{\csSwBox{sTwo}}{LLM wrote}
\csKeyNext{\csSwConf}{confidence}
\csKeyNext{\csSwIrr}{changes the world}
\csKeyNext{\csSwObs}{environment's reply}
\csFrame{8.5}{38}{36}
\csFrame{85.5}{38}{35.5}
\node[cstxt, font=\figfs\sffamily\bfseries] at (9.8,9.2) {\jev{} picks every tool};
\node[cstxt, font=\figfs\sffamily\bfseries] at (86.8,9.2) {\ours{} picks every tool};
\node[cstxt, text=sOne] at (9.8,4.6) {\faBolt\,\textbf{decision model}};
\node[cstxt, text=sTwo] at (48.8,4.6) {\faBrain\,\textbf{LLM}\enspace\textcolor{ink2}{Flash}};
\node[cstxt, text=sOne] at (86.8,4.6) {\faBolt\,\textbf{decision model}};
\node[cstxt, text=sTwo] at (125.8,4.6) {\faBrain\,\textbf{LLM}\enspace\textcolor{ink2}{Flash}};
\csSpine
\def\csWd{38}\def\csBoxW{17}\def\csBoxNat{0}\def\csBarL{11}\def\csTau{-1}\def\csAltOwn{1}
\def\csRun{jev}\def\csX{8.5}\def\csWl{36}\csDrawRun
\def\csRun{ours}\def\csX{85.5}\def\csWl{35.5}\csDrawRun
\pgfmathsetmacro\csFyy{-\csC{height}-7.6}
\fill[envT, rounded corners=1.5pt] (0,\csFyy+1.9) rectangle (160,\csFyy-1.9);
\node[cstxt, text=ink2] at (1.2,\csFyy) {\textbf{\textcolor{ink}{Same task}}\enspace Gemini 3.5 Flash alone, which also
  chooses the tools, asked at step \csC{ref/first_ask} and solved the task in \csC{ref/steps} steps.};
\end{tikzpicture}
\caption{\textbf{\jev{} refunded with high confidence where it had to ask, while \ours{} asked first.} Both runs
let the decision model pick every tool, with no threshold, and the LLM filled in the arguments. The red note names
the only acceptable tool in the task's answer key. The task is the first of the \res{cases/ask/n_tasks} fresh tasks,
by seed, on which \jev{}'s first wrong tool choice was an irreversible action at a state where only asking was
acceptable, \jev{} failed and \ours{} solved the task. Long text is abridged (\dots).}
\label{fig:case-ask}
\end{figure*}

\begin{figure*}[!t]
\centering
\begin{tikzpicture}[jevfig]
\def\csCase{alf}
\node[cstxt, text=ink] at (0,15.0) {\textbf{Task}\enspace\textit{\csC{task}}};
\csKeyStart{7.0}{10.4}
\csKeyNext{\csSwBox{sOne}}{decision model chose}
\csKeyNext{\csSwBox{sTwo}}{LLM chose}
\csKeyNext{\csSwTau}{confidence and threshold $\tau$ = \csC{tau}}
\csKeyNext{\csSwObs}{environment's reply}
\csFrame{8.5}{75}{0}
\csFrame{85.5}{0}{74.5}
\node[cstxt, text=sOne] at (9.8,4.6) {\faBolt\,\textbf{decision model}\enspace\textcolor{ink}{4B trained on ALFWorld
  states}};
\node[cstxt, text=sTwo] at (86.8,4.6) {\faBrain\,\textbf{LLM}\enspace\textcolor{ink}{Gemini 3.5 Flash alone}};
\csSpine
\def\csBoxW{44}\def\csBoxNat{1}\def\csBarL{14}\def\csAltOwn{1}
\def\csRun{run}\def\csX{8.5}\def\csWd{75}\def\csWl{0}\def\csTau{\csC{tau_num}}\csDrawRun
\def\csRun{ref}\def\csX{85.5}\def\csWd{0}\def\csWl{74.5}\def\csTau{-1}\csDrawRun
\pgfmathsetmacro\csFyy{-\csC{height}-8.4}
\csFoot{\csFyy}{Same game}{trained 4B}{Flash alone}
\csMini{36}{LLM calls}{\csC{run/llm_calls}}{\csC{ref/llm_calls}}
\csMini{78}{cost per game (\$, GPU included)}{\csC{run/cost}}{\csC{ref/cost}}
\csMini{120}{time (s)}{\csC{run/wall}}{\csC{ref/wall}}
\end{tikzpicture}
\caption{\textbf{On ALFWorld the trained 4B model took every action and solved the game without the LLM.} Its fitted
threshold let every confidence through, so even its least confident move ran without an LLM call. The footer
compares the same game with Gemini 3.5 Flash alone, with the GPU charged per episode as in App.~\ref{app:e11}. The
game is the first of the \res{cases/alf/n_tasks} valid\_seen games, by game id, that both runs solved. Long text is
abridged (\dots).}
\label{fig:case-alf}
\end{figure*}

\end{document}